\documentclass{applemlr}

\usepackage{amsmath}
\usepackage{enumerate}
\usepackage{amsfonts}
\usepackage{amsthm}
\usepackage{diagbox}
\usepackage{colortbl}
\usepackage{amssymb}
\usepackage{xspace}
\usepackage{wrapfig}
\usepackage{adjustbox}
\usepackage{tabularx}
\usepackage{mathtools}
\usepackage{tikz}
\usepackage{enumitem}
\usepackage{silence}
\usepackage{dsfont}
\usepackage{makecell}
\usepackage{xfakebold}
\usepackage{enumitem}
\usepackage{amsmath,amsfonts,bm}

\def\eqref#1{equation~\ref{#1}}
\def\Eqref#1{Equation~\ref{#1}}

\def\1{\bm{1}}

\DeclareMathAlphabet{\mathsfit}{\encodingdefault}{\sfdefault}{m}{sl}
\SetMathAlphabet{\mathsfit}{bold}{\encodingdefault}{\sfdefault}{bx}{n}

\newcommand{\E}{\mathbb{E}}

\newcommand{\R}{\mathbb{R}}

\newcommand{\softmax}{\mathrm{softmax}}

\DeclareMathOperator*{\argmin}{arg\,min}

\newcommand{\src}{\mathrm{src}}
\newcommand{\tgt}{\mathrm{tgt}}

\definecolor{textgray}{HTML}{6E6E73}
\usetikzlibrary{positioning, calc}
\usetikzlibrary{decorations.pathmorphing}

\makeatletter
\patchcmd{\wrong@fontshape}{\@gobbletwo}{}{}{}
\makeatother
\numberwithin{equation}{section}
\makeatletter
\AtBeginDocument{
  \urlstyle{sf}
  
}
\makeatother

\definecolor{light}{RGB}{125, 125, 125}
\crefname{tcb@cnt@pbox}{code}{code}
\Crefname{tcb@cnt@pbox}{Code}{Code}
\crefname{assumption}{assumption}{assumption}
\Crefname{assumption}{Assumption}{Assumptions}

\newtcolorbox[auto counter]{pbox}[2][]{
  colback=white,
  title=Code~\thetcbcounter: #2,
  #1,fonttitle=\sffamily,
  fontupper=\sffamily,
  arc=2pt,
  colframe=bgcolor,
  coltitle=fgcolor,
  colbacktitle=bgcolor,
  toptitle=0.25cm,
  bottomtitle=0.125cm
}

\makeatletter
\newcommand\applefootnote[1]{%
  \begingroup
  \renewcommand\thefootnote{}%
  \renewcommand\@makefntext[1]{\noindent##1}%
  \footnote{#1}%
  \addtocounter{footnote}{-1}%
  \endgroup
}
\makeatother

\definecolor{cverbbg}{gray}{0.90}

\usepackage{url}

\graphicspath{{figures/}}

\theoremstyle{remark}
\newtheorem{remark}{Remark}

\title{KV-Lingo: Learning KV-Cache Translators with Distillation}

\author[*]{Valérie Castin}
\author[*]{Keitaro Sakamoto}
\author{Anastasiia Filippova}
\author{João Monteiro}
\author{Marco Cuturi}
\author{Pierre Ablin}

\affiliation{Apple}

\abstract{

Large language models represent context with a key-value (KV) cache.
Caches are model-specific: for the same text, models with different architectures or weights produce incompatible representations.
This makes it costly to switch models over a shared context: although the context has already been processed by one model, the incoming model must process it again to build its own cache.
We introduce KV-Lingo, a method for translating the KV cache of a source model into one that can be read by a target model. 
KV-Lingo consists of a collection of linear maps, typically one per layer of the target model, that are applied independently on all tokens' key and value representations.
We train these maps using distillation, minimising the divergence between the target model’s predictions from its native cache and those from the translated cache.
We consider several model pairs spanning multiple sizes and architectures, training one translator per pair on a generic text corpus.
The resulting translators preserve strong downstream performance in both small-to-large and large-to-small transfers.
Since a switch then costs a linear map and a single decoding step instead of a prefill, replacing re-prefill with cache translation reduces the time to first token after a model switch by 9.6× already on a 64-token prompt for Qwen models on an Apple M3 Ultra, and by up to 29× at 32k context length on an H100.
These gains make KV-Lingo particularly useful for dynamic model routing: a context can be processed by one model and handed off to another only when needed, without re-prefilling the shared prefix.
We finally show that KV-Lingo can be used for seamless model switching, staying close to re-prefill across repeated switches in our multi-turn evaluations.
}

\metadata[Correspondence]{\sffamily valerie.castin@ens.psl.eu, sakakei-1999@g.ecc.u-tokyo.ac.jp,\\ \{a\_filippova, jmonteiro2, cuturi, p\_ablin\}@apple.com}
\metadata[Note]{\sffamily $^{*}$ equal contribution. Work done while VC and KS were interns at Apple.}
\date{\sffamily September 26, 2026}

\usepackage{comment}
\excludecomment{iclrblock}
\includecomment{arxivblock}

\newcommand{\vary}[2]{#1}
\newcommand{\arxivonly}[1]{#1}
\newcommand{\iclronly}[1]{\ignorespaces}

\begin{document}
\maketitle

\section{Introduction}
\label{sec:intro}

Autoregressive transformers~\citep{vaswani2017attention} generate one token at a time, attending over all previous tokens.
The key-value (KV) cache stores the past keys and values so that each decode step costs $\mathcal{O}(n)$ rather than $\mathcal{O}(n^2)$ in the context length $n$; at long context, it dominates the memory cost of serving~\citep{kwon2023efficient, hooper2024kvquant}.
The cache is tied to the model that built it: its shape follows the architecture and its coefficients the parameters, so handing it to another model, even one of matching shape, severely degrades generation~\citep{liu2024droidspeak}.

 In many settings, several models run over one context: routers and cascades send harder queries to a stronger model~\citep{chen2024frugalgpt, ong2025routellm}, token-level routers defer selected tokens to a larger one~\citep{shen2024learning, fu2025r2r}, and multi-agent pipelines share a working context between specialists~\citep{wu2024autogen}.
In these cases, the KV-cache incompatibility across models has a cost: systems must either maintain a cache for every model that may be called, or compute the target model’s cache when it is first invoked. This means re-encoding the entire prefix, duplicating much of the computation spent processing the same context, at a cost that grows with the context length.

In order to fix this incompatibility, we propose training \emph{cache translators}, which transform the cache of a source model into a cache readable by a target model.
The target model can then start generating right away, without paying the prefilling cost.
We call our approach KV-Lingo. Its translators are simple and cheap to apply: each target layer uses separate token-wise linear maps for keys and values, reading from a small set of source layers. 
We train these maps with distillation, which yields strong downstream performance despite this restricted architecture (Figure~\ref{fig:main}).
The fact that translators as simple as linear maps perform so well is remarkable; we show in Appendix~\ref{app:cka_linearity} that the KV caches of different models typically have a very similar geometry, and are approximately linearly related, which empirically supports the above observation.
Our contributions are as follows:

\textbf{KV-Lingo, a 
simple translator model and training procedure}
    (\S\ref{sec:method}): a per-target-layer linear map,
    split across keys and values, fit in closed form by minimising the MSE, and then refined by
    self-distillation against the target's own next-token distribution. A central finding is that cache reconstruction is not enough: translators that closely match target activations can generate poorly, whereas self-distillation turns the same simple linear family into effective drop-in caches.
    We justify each architectural choice empirically: linearity, self-distillation, head mixing, pre- vs post-RoPE. We also adapt KV-Lingo so that it can translate caches across models using different tokenizers.

\textbf{A systematic evaluation of KV-Lingo} (\S\ref{sec:exp}) spanning
    multiple model pairs, from the families Qwen3~\citep{yang2025qwen3}, Gemma-3~\citep{team2025gemma}, Gemma-4~\citep{team2026gemma} and Mistral-v0.3~\citep{jiang2023mistral7b}, scales ranging from 0.6B to a 30B mixture of experts,
    evaluated on a range of tasks from language modelling and multi-turn chat, to long-context understanding. On most tasks, the translated cache performs as well as or better than the smaller model. We also show that KV-Lingo translators are robust to several model switches in multi-turn settings.

\textbf{An efficiency study} (Figure~\ref{fig:cost}): model switching with KV-Lingo costs a linear translation and
a single decoding step of the target instead of a full prefill. 
We quantify the speed-up this yields on an Apple M3 Ultra and on H100 GPUs, reporting between $4.8\times$ and $29\times$ faster time to first token (TTFT) depending on model and prefill length.

The closest prior work, like ours, avoids prefilling the shared context with the target model. \citet{chen2026see} introduce a per-head nonlinear adapter, while \citet{horton2024kv} jointly train an auxiliary model and per-layer linear maps. Concurrent works instead use closed-form linear fits between source and target caches~\citep{heo2026cross, qu2026cachebridge, li2026universal}; this corresponds to the first stage of our training procedure (\S\ref{sec:method:training}). Appendix~\ref{app:related} discusses the broader literature.

\begin{figure}[t]
\centering
\includegraphics[width=\linewidth]{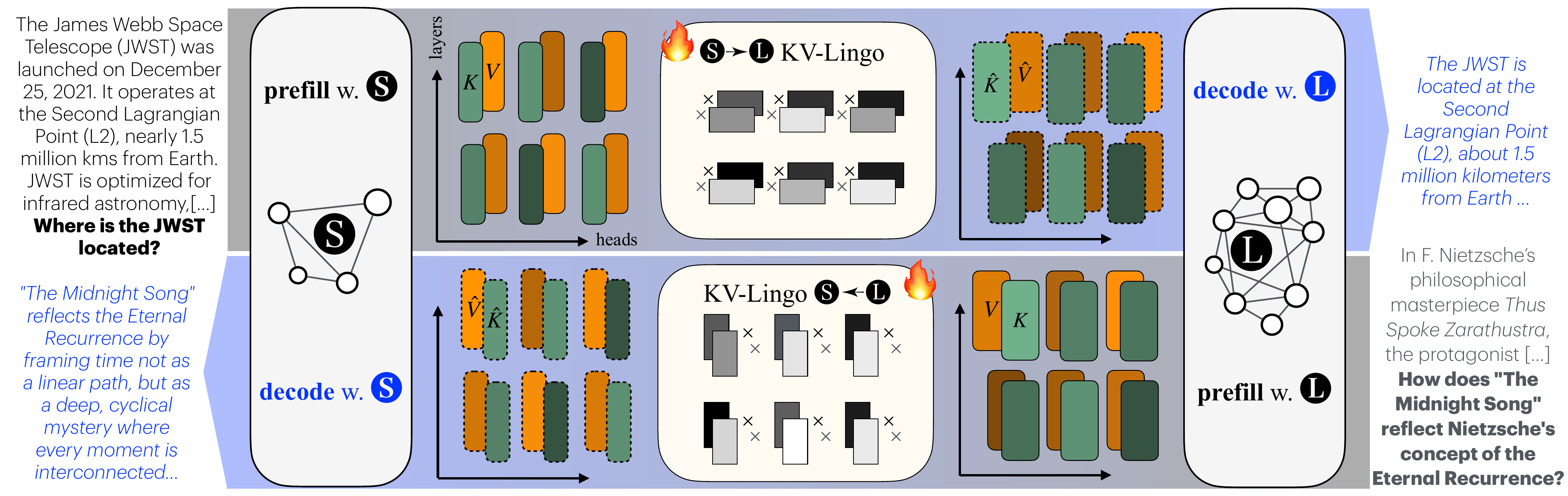}
\caption{\textbf{KV-Lingo translates a KV cache between two frozen models}, so a
conversation can switch model without re-prefilling its context. Top: the small model $S$
prefills the prompt, and the large model $L$ decodes the answer from the translated cache
$(\widehat{K}, \widehat{V})$, dashed. Bottom: the same from $L$ to $S$, with a second
translator. Each translator (flame, the only trained part) is a set of linear maps, one per
target layer for keys and one for values, applied token by token, so its cost grows
linearly with the prefix length where re-prefilling grows quadratically.}
\label{fig:main}
\end{figure}

\section{Problem setting}
\label{sec:setting}

\textbf{Notation.} We call \emph{context} the tokens initially supplied to the models, and \emph{prefix} the full sequence preceding the model switch: the context, possibly followed by a continuation generated by a model.
Let $\mathcal{V}$ be the vocabulary and $x_{1:n} \in \mathcal{V}^n$ the prefix. A model with $N$ layers, $H$ key/value
heads and head dimension $d_h$ computes, at each layer, the keys and values of every
prefix token. Collecting them layerwise, the \emph{key-value (KV) cache} is the collection $\mathcal{C} = (K^{\ell},V^{\ell})_{\ell=1}^N$, with $K^\ell, V^\ell \in \R^{d \times n}$ and $d = H \cdot d_h$,
with one column per prefix token and the $H$ heads stacked along the $d$-dimensional axis.
RoPE rotates keys by their token's position~\citep{su2024roformer}, and some
models also normalise keys per head beforehand;
we consider translating both pre-RoPE and post-RoPE keys.
A model fills the cache in a single \emph{prefill} pass over the prefix, then \emph{decodes} a continuation one token at a time: each step reads the whole cache and appends the new token's keys and values to the cache.

An LLM models the probability of a sequence as $p(y_{1:m}\mid x_{1:n})
=
\prod_{t=1}^{m}
p(y_t \mid x_{1:n}, y_{<t}).$
Conditioning on the prefix $x_{1:n}$ only requires the cache
$\mathcal{C}(x_{1:n-1})$ of the past tokens together with the current token $x_n$:
a forward pass on $x_n$ attends to the cache and produces the next-token distribution.
We therefore write
$p\bigl(y_{1:m} \mid \mathcal{C}(x_{1:n-1}),\, x_n\bigr)
=
\prod_{t=1}^{m}
p\bigl(y_t \mid \mathcal{C}(x_{1:n-1}),\, x_n,\, y_{<t}\bigr).$

\subsection{Cache translation}
\label{sec:setting:translation}

We consider two
models: a \emph{source}, which produces the cache of a prefix $x_{1:n}$ it may have partly generated, and a \emph{target}, which must continue generating.
We write $\mathcal{C}_\src(x_{1:n})$ and $\mathcal{C}_\tgt(x_{1:n})$ for the caches they produce on
the same prefix.

Decoding from the target's cache,
$p_\tgt\bigl(\,\cdot \mid \mathcal{C}_\tgt(x_{1:n-1}),\, x_n\bigr)$, is \emph{native}
decoding. A \emph{translator} is a map $T$ from caches in the source's format to caches in
the target's format such that
\begin{equation}
    p_\tgt\bigl(\,\cdot \mid T(\mathcal{C}_\src(x_{1:n-1})),\, x_n\bigr) \approx
p_\tgt\bigl(\,\cdot \mid \mathcal{C}_\tgt(x_{1:n-1}),\, x_n\bigr), 
\label{eq:desiderata_translation}
\end{equation}
as distributions over
continuations $y_{1:m}$, for prefixes drawn from the contexts the translator is deployed on. 
In typical deployment, $x_n$ is the last token of the generation prompt and
carries no content.

Two remarks shape the problem. First, $T$ need not reproduce $\mathcal{C}_\tgt$ entrywise, only the target's next-token distributions; cache reconstruction error and continuation quality can in fact disagree sharply (\S\ref{sec:exp}). Second, translation pays off only if applying $T$ costs less than re-prefilling; since a $100$-token cache of Qwen3-4B already holds $7.4$ million values, this forces structured sparsity on $T$ (\S\ref{sec:method:family}).

We study the \emph{post-hoc} setting where source and target are frozen; only $T$ is trained, so no access to either model's
pretraining pipeline or resources is needed. Training the two models jointly for compatible caches
is an interesting direction, outside the scope of this paper.

\subsection{Deployment scenarios}
\label{sec:setting:metrics}

\textbf{Model switching.} We focus on settings involving two models of different capability: a \emph{large} model
$L$ (better but slower) and a \emph{small} model $S$ (less capable but faster).
Currently, when switching models, the target re-prefills the whole prefix before
decoding its first token~\citep{liu2024droidspeak,
geng2026relaycachingacceleratingllmcollaboration}. KV-lingo bypasses this. The two directions of transfer serve different purposes:
\begin{itemize}[noitemsep,nolistsep,leftmargin=*]
    \item In $S \to L$ transfer, $S$ processes the prefix cheaply and hands it off to
    $L$ only when the stronger model is needed. Translation reduces the latency of
    this hand-off by avoiding the $L$ prefill, at the cost of a possible degradation
    relative to $L$ decoding from its native cache.
    \item In $L \to S$ transfer, the representation computed by $L$ can be handed
    off to the cheaper model $S$, allowing subsequent generation to benefit from the wealth in latent representation of
    $L$'s processing while decoding with $S$. This is useful, for example, in 
    agentic pipelines where a large model processes a shared context or produces
    a plan before delegating subsequent work to smaller models.
    Another usecase is privacy, where $S$ and $L$ models differ in data-access policies. A large model can process a non-sensitive context, whose translated cache is handed to a restricted model that subsequently receives private queries. This avoids exposing the private input to the large model while avoiding a second prefill of the shared context. In realistic scenarios where $S$ runs on-device, this strategy would be assessed by comparing the gain in performance and the added communication cost to transfer the cache, against the Watts cost of prefilling with the small model.
\end{itemize}

\textbf{Prefilling for a mixture of experts.}
In memory-bound inference, such as
on-device generation, a mixture-of-experts (MoE)
model~\citep{shazeer2017outrageously} decodes about as fast as
a dense model of its active size, reading only the experts each token is routed to, but
prefills more slowly, as the tokens of a prompt together reach most of its experts. A dense
$S$ can prefill the prompt, and an MoE $L$ decode from its translated cache
(\S\ref{sec:exp:moe}). The MoE, having more capacity, then decodes with better quality.

\begin{remark}
    In a memory-bound case, a smaller model $S$ could be used to prefill the context once and keep it as its own, smaller cache, directly used by $L$ for decoding, by merging the translators with queries.
    This method, which requires post-RoPE and sparse translators, in only tested as a proof-of-concept (Appendix~\ref{app:memory})
\end{remark}

\section{KV-Lingo: linear translators trained with self-distillation}
\label{sec:method}

\arxivonly{%
\begin{figure}[t]
\centering
\includegraphics[width=\linewidth]{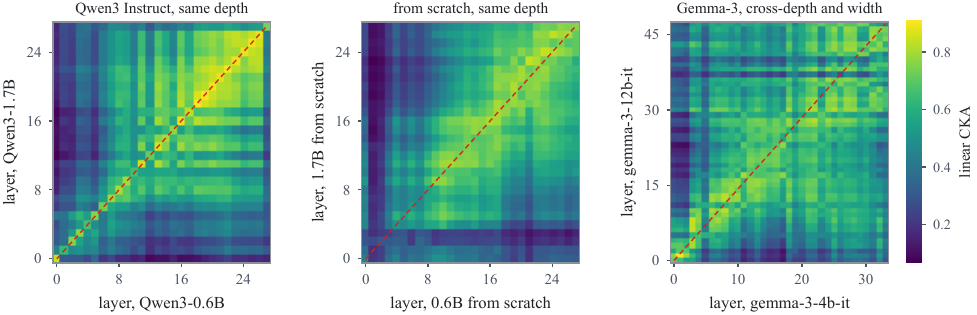}
\caption{\textbf{Layer-to-layer cache correspondence.} Linear CKA between
every layer of the large (rows) and small (columns) model, on post-$k$-norm keys over
$16$k tokens; shared colour scale, dashed line at equal relative depth. \emph{Left:}
pretrained Qwen3 Instruct pair of equal depth. \emph{Middle:} $1.7$B/$0.6$B pair
pretrained from scratch, sharing no checkpoint. \emph{Right:} pretrained Gemma-3 pair
differing in depth and key/value width. Appendix~\ref{app:cka} adds every other pair we
measured, including two randomly initialised controls.}
\label{fig:cka}
\end{figure}
}

\subsection{Linear translator family}
\label{sec:method:family}

Given a prefix $x_{1:n}$, let $\mathcal{C}_\src(x_{1:n}) = (K_\src^{\ell},V_\src^{\ell})_{\ell=1}^{N_\src}$ be the source cache, where
$K_{\src, i}^{\ell}, V_{\src,i}^{\ell}\in \R^{d_\src}$ 
stack every head of token $i$.
KV-Lingo translates the cache of target layer $\ell$ from a small subset
$\mathcal{E}_\ell$ of source layers, simply $\{\ell\}$ when both models have the same
number of layers:
\begin{equation}
  \label{eq:map}
  \widehat{K}^{\ell}_i \;=\; \sum_{\ell'\in \mathcal{E}_\ell} T_K^{\ell'\to\ell}\, K_{\src,i}^{\ell'},
  \qquad
  \widehat{V}^{\ell}_i \;=\; \sum_{\ell'\in \mathcal{E}_\ell} T_V^{\ell'\to \ell}\, V_{\src,i}^{\ell'},
  \qquad i = 1,\dots,n-1,
\end{equation}
with $T_K^{\ell'\to\ell},\, T_V^{\ell'\to\ell} \in \R^{\,d_\tgt \times d_\src}$
(the token-wise structure is motivated in Appendix~\ref{app:cka_linearity}). We write
$T_K^\ell \in \R^{\,d_\tgt \times |\mathcal{E}_\ell|\, d_\src}$ for their concatenation over
$\ell'\in\mathcal{E}_\ell$, and likewise $T_V^\ell$. The keys and values of token $n$ come
instead from a forward pass of the target on the last prefix token, attending to the
translated cache.
We write $T(\mathcal{C}_\src(x_{1:n}))$ for the resulting cache.
Sharing the token index $i$ across both sides presumes the two models have the same tokenizer; \S\ref{sec:exp:xtok} treats a pair that does not, where the correspondence
is built between token \emph{groups} covering the same characters.
\Eqref{eq:map} leaves several choices open: whether heads mix, where in the target's stack
the map acts, and how to set $\mathcal{E}_\ell$.

\textbf{The power of linearity.}
We choose linear maps first for speed. Translating a prefix is, like prefilling, compute-bound, so its
wall-clock cost is set by the translator's FLOPs, and a linear map translates faster than
an MLP while nearly matching its performance (Figures~\ref{fig:qwen3small}(a)
and~\ref{fig:ttft}). Linearity brings two further benefits: the first training stage has
a closed-form solution (\S\ref{sec:method:training}), and choosing the source layers from
data costs only ridge solves (greedy match, below).

\textbf{Keys and values apart.} $T_K^\ell$ acts only on keys, and $T_V^\ell$ only on values: this halves the parameter count against a joint map over $[K;V]$ at negligible cost on performance (Appendix~\ref{app:splitjoint}).

\textbf{Head mixing.}
We propose two variants of KV-Lingo: a \emph{head-wise} version where target head $h$ reads only
source head $h$, so $T_K^{\ell'\to\ell}$ and $T_V^{\ell'\to\ell}$ are block-diagonal; and a \emph{head-mixing} one which leaves the matrices unconstrained.
Head-mixing
stores $2\,|\mathcal{E}_\ell|\,d_\tgt d_\src$ parameters per target layer; head-wise has $H$ times fewer, making it faster but less expressive.

\textbf{Rotation and normalisation.}
Translators can act \emph{post-RoPE} (on rotated keys), \emph{post-norm} (after key
normalisation, before RoPE) or \emph{pre-norm} (before both); in the last two cases the
translated keys are normalised (pre-norm only) and rotated before being stored. Post-norm
is worse than pre-norm on long context (Table~\ref{app:tab:knorm}). Pre-norm and
post-RoPE match on short prefixes ($\lesssim 2$k tokens), but pre-norm is slightly better on
longer ones (Table~\ref{tab:pre_vs_post_rope}) and extrapolates better beyond training
lengths (Figure~\ref{fig:pre_vs_post_rope_extrapolation}). We therefore use pre-norm in most
experiments, although post-RoPE translation has a significant advantage in the memory-bound
setting of Appendix~\ref{app:memory}.

\textbf{Source-layer selection.}
We call a pair \emph{shape-preserving} when its two caches have the same number of layers, where each target layer reads its counterpart,
$\mathcal{E}_\ell = \{\ell\}$, and \emph{cross-layer} when depth differs, so that an
assignment $\mathcal{E}_\ell$ must be chosen. Every target layer then reads the same number
$\nu$ of source layers.
\emph{Relative-depth match} takes the band of $\nu$ consecutive source layers centred on the
relative-depth counterpart $c(\ell) \coloneqq 1 + (\ell-1) (N_\src-1)/(N_\tgt-1)$ of target
layer $\ell \in \{1,\ldots, N_\tgt\}$.
The window
is shifted at the boundaries, so $|\mathcal{E}_{\ell}| = \nu$ for every
target layer.
Figure~\ref{fig:xdepth_assignment} illustrates this method.
\emph{Greedy match} instead picks $\mathcal{E}_\ell$ from data, separately for each target
layer, by forward selection over source layers on the residual of a ridge fit to that
layer, and the \emph{$R^2$ assignment} keeps the $\nu$ source layers whose single-layer fits
score best.
A larger $\nu$ trades speed for quality (Appendix~\ref{app:xdepth}).

\subsection{Training the translator}
\label{sec:method:training}

Training data are texts split into a prefix $x_{1:n}$, whose cache is translated, and a
continuation $y_{1:m}$. Only the matrices $\{T_K^\ell, T_V^\ell\}_\ell$ are trained, in two
stages. For ease of exposition we consider $\mathcal{E}_\ell = \{\ell\}$, the extension to more involved assignments is straightforward.

\textbf{Stage 1: closed-form initialisation on the prefixes.}
Each map is initialised to the minimiser of the cache-reconstruction Frobenius error
\begin{equation}
  \label{eq:hat}
  T_K^{\ell} \;=\; \argmin_{M}\ \E_{x_{1:n}}\,\bigl\| M K_\src^{\ell}(x_{1:n}) - K_\tgt^{\ell}(x_{1:n}) \bigr\|_F^2 ,
\end{equation}
and likewise for $T_V^\ell$. It is solved in closed form on the fitting corpus, without
gradient steps: the normal equations need only the second moments $\E[K_\src K_\src^\top]$
and $\E[K_\src K_\tgt^\top]$ per layer, and are solved through a pseudo-inverse for
numerical stability.

\textbf{Stage 2: self-distillation on the continuation.}
As stated in \eqref{eq:desiderata_translation}, the ultimate goal of translation is to match the distribution of continuations one would obtain with the native cache.
This directly translates into a distillation objective:
\begin{equation}
  \label{eq:kl}
  \mathcal{L}(T) \;=\; \E_{(x_{1:n},\,y_{1:m})}\ \frac{1}{m}\sum_{j=1}^{m}
  \mathrm{KL}\Bigl(\,
    p_\tgt\bigl(\cdot \mid \mathcal{C}_\tgt(x_{1:n}),\, y_{<j}\bigr)
    \ \big\|\
    p_\tgt\bigl(\cdot \mid T(\mathcal{C}_\src(x_{1:n})),\, y_{<j}\bigr)
  \Bigr),
\end{equation}
the divergence between the target's predictions from its \emph{own} cache and from the
translated one. 
We minimise $\mathcal{L}$ using Adam, starting from the stage-1 initialisation. As usual in distillation, the supervision comes from the target; the
examples only fix the positions at which the two distributions are compared.
\iclronly{As an ablation, we replace it with a cross-entropy loss, which we find more brittle at long context
(Appendix~\ref{app:objective}).}
\vary{%
The next paragraph gives a cheaper alternative to this recipe: a cache norm better aligned with downstream performance than the Frobenius norm.
}{%
\citet{heo2026cross} and \citet{qu2026cachebridge} use a Frobenius loss similar to \eqref{eq:hat} to train their translators, with architectures different from KV-Lingo's and without stage 2 (Appendix~\ref{app:related}). In our experiments, we find that stage 2 training yields vastly superior performance.
In Appendix~\ref{app:norms} we find a reweighted norm in \eqref{eq:hat} which empirically gives better initialisations for stage 2, but these gains vanish when training stage 2 long enough.

}

\paragraph{Closed-form translators with other norms.}

Stage 1 of our training recipe measures the reconstruction residual of~\eqref{eq:hat} in the
Frobenius norm, weighting every position of the prefix equally. 
However, this choice of norm is not always aligned with downstream performance, as Table~\ref{app:tab:cachedist} shows.
We propose the \emph{attention-weighted norm}, which replaces the uniform weighting of the Frobenius norm by the average attention each prefix position receives from the continuation's queries.
For $\Delta \in \R^{d\times n}$ with columns $\delta_1, \ldots, \delta_n$, we define
\begin{equation}
  \|\Delta\|_{\mathrm{attn}}^2 \;=\; \sum_{i=1}^{n} \bar a_i\,\|\delta_i\|_2^2
    \;=\; \bigl\|\Delta \operatorname{diag}(\bar a)^{1/2}\bigr\|_F^2, \quad
  \bar a_i \;=\; \frac{1}{m}\sum_{j=1}^{m} a_{j,i},
  \label{eq:attn-norm}
\end{equation}
where $a_{j,i} \;=\; \softmax_i\!\bigl(q_j^\top k_i / \sqrt{d_h}\bigr)$ is the attention
that the $j$-th continuation query $q_j$ places on prefix position $i$.
The norm follows from a first-order expansion of the attention output.
The target consumes the translated cache only through attention, which at a continuation query $q$ outputs $\sum_{i=1}^{n} a_i v_i$ at each head. 
Perturbing the cache by $(\delta k_i, \delta v_i)$ changes that output, to first order, by
$\sum_{i} a_i\,\delta v_i
\;+\; \sum_{i} a_i\Bigl(\delta s_i - \sum_{j} a_j\,\delta s_j\Bigr) v_i$,
with $\delta s_i = q^\top \delta k_i / \sqrt{d_h}$. 
Both the key and the value residual contribute to the output, weighted by that position's attention weight $a_i$, and averaging over the continuation's queries gives the weights $\bar a$ of~\eqref{eq:attn-norm}. 
A prefix token the continuation barely attends to therefore contributes little to the output, regardless of how badly it is reconstructed, whereas the Frobenius norm charges it as much as, for example, an attention sink.
 
When fitting a closed-form translator (stage 1 of training) with the attention norm instead of the Frobenius norm, we obtain a significantly better off-the-shelf validation loss and downstream performance (Table~\ref{app:tab:norms}).
However, when running self-distillation (i.e., stage 2) from that improved initialisation then absorbs most of that advantage: both initialisations lead to a similar validation loss (Figure~\ref{app:fig:norms}).
For this reason, we keep the uniform fit as our default initialisation, as it is cheaper to compute; 
while the closed-form translator fitted with the attention-weighted norm can be used without subsequent training
in computationally-bound settings.
\vary{%
Appendix~\ref{app:norms} provides more details on this experiment.
}{%
The rest of this appendix reports the comparison in full.
}

\section{Experiments}
\label{sec:exp}

\subsection{Setup and baselines}
\label{sec:exp:setup}

We train one translator per model pair on a single multi-turn chat dataset,
Nemotron-SFT-Instruction-Following-Chat-v2~\citep{nvidia2025nemotron3nanoopen},
disjoint from every evaluation benchmark we consider below. Each sample is a (prefix, continuation) pair split at
the start of an assistant turn, where a switch would happen, with reasoning-on and
reasoning-off samples alternating; prefixes range from $n=10$ to $16$k tokens and
continuations are capped at $m=2048$. The \emph{same} translator serves every task and language we
evaluate. Appendix~\ref{app:data} details the training mixture, Appendix~\ref{app:protocol} the
training procedure.

\textbf{Evaluation datasets.}
We cover six
abilities: \emph{language modelling}, by decoding on held-out training conversations;
\emph{knowledge and reasoning}, with MMLU-Pro~\citep{wang2024mmlupro},
ARC-Challenge~\citep{clark2018think} and GSM8K~\citep{cobbe2021training};
\emph{instruction following}, with IFEval~\citep{zhou2023instruction}; \emph{context-grounded QA}, with RepLiQA~\citep{monteiro2024repliqa},
whose fictional documents rule out answers from parametric memory, CoQA~\citep{reddy2019coqa}
for the multi-turn case and, in $11$ languages, XQuAD~\citep{artetxe2020cross};
\emph{multi-turn chat}, judged on MT-Bench-101~\citep{bai2024mt}; and \emph{long context},
with RULER~\citep{hsieh2024ruler}, LongMemEval~\citep{wu2025longmemeval} and
LongBench-v2~\citep{bai2025longbench}. Judged benchmarks use a local Qwen3.6-27B judge, and
several are run both reasoning-on and reasoning-off.
Table~\ref{tab:evalsets} gives the prefix length range for these tasks, from a few hundred tokens to $32$k.

\textbf{Baselines.}
We compare KV-Lingo to three baselines.

\emph{Head-wise linear closed-form translator}~\citep{heo2026cross}: the head-wise stage-1 solution
of~\eqref{eq:hat} used on its own, without stage 2. 

\emph{Head-wise MLP translator}: the architecture of~\citet{chen2026see}, a head-wise
two-layer MLP, trained with our recipe (Appendix~\ref{app:protocol}): stage 1 with Adam, as
no closed form exists, and stage 2 by self-distillation, which we find stronger than the
cross-entropy they use (Appendix~\ref{app:objective}).

\emph{DroidSpeak}~\citep{liu2024droidspeak}: the source cache is
reused unchanged wherever the two models agree closely enough, and re-prefilled at the
layers where they diverge most. It requires same-shape caches, and its
cost grows with the fraction of layers it recomputes. Its results are in
Appendix~\ref{app:droidspeak}.

We compare only against methods whose translator costs less than the re-prefill it replaces (\S\ref{sec:setting:translation}).
This excludes cache-fusion methods such as Cache-to-Cache~\citep{fu2026cachetocache} and the transformer adapters of~\citet{dery2026latent}, where translating is no cheaper than prefilling.

\subsection{Shape-preserving cache translation}
\label{sec:exp:shape}

\begin{figure}[]
\centering
\includegraphics[width=\linewidth]{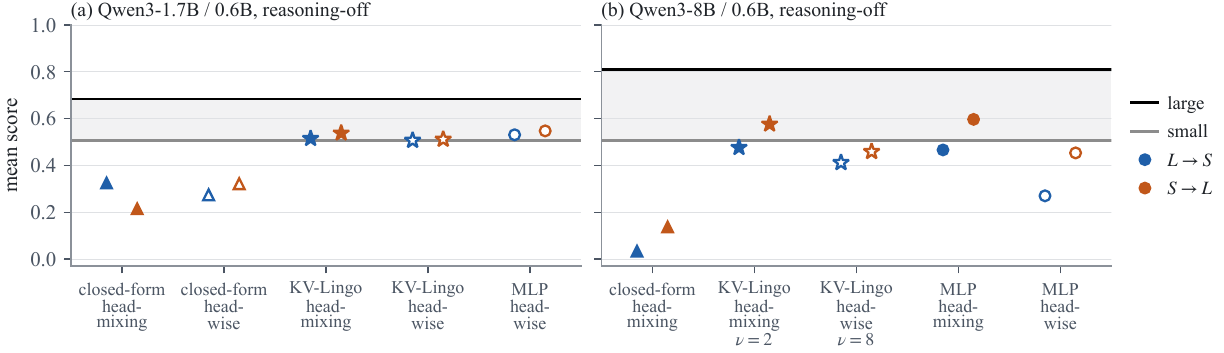}
\caption{\textbf{Cache translation scores}, reasoning-off, mean of MMLU-Pro, \mbox{ARC-C},
RULER, LongMemEval, CoQA, IFEval, XQuAD, RepLiQA and MT-Bench-101. \emph{(a)
Shape-preserving}, Qwen3-0.6B/1.7B: a distilled head-wise linear translator suffices.
\emph{(b) Cross-layer}, Qwen3-0.6B/8B; head-mixing KV-Lingo at $\nu = 2$ uses the $R^2$
assignment. $\nu=1$ for MLPs.
}
\label{fig:qwen3small}
\end{figure}

We first evaluate cache translation in the setting where the source and the target have the same cache shape.
This is the case for two Qwen3 model pairs, both with $H = 8$ and $d_h = 128$: the $0.6$B/$1.7$B pair at $N = 28$ layers and the $4$B/$8$B pair at $N = 36$.
Figure~\ref{fig:qwen3small}(a) reports the score of KV-Lingo translators and baselines when translating for the $0.6$B/$1.7$B pair, averaged over nine benchmarks, and evaluated reasoning-off.
We observe that distilled translators significantly outperform closed-form ones, and that our linear KV-Lingo translators match the score of the head-wise MLP of~\citet{chen2026see}, reaching the level of the small model in both directions.
Reasoning-on, the score of $S \to L$ translation rises from the level of the small model to close to that of the large model (Figure~\ref{fig:suite_small_on_mean}).
Moreover, in $L \to S$, when the task is easy enough that source and target have similar scores, translation performs at the level of the large model (Figure~\ref{fig:suite_off}), which makes KV-Lingo translators suited to the agentic hand-off setting described in \S\ref{sec:setting:metrics}. 
Replacing the head-wise MLP with a head-mixing one does not change performance (Figure~\ref{fig:head_mixing_mlp_shape}).
Head-wise translators suffice here because the heads of these released pairs are aligned one to one, which is specific to Qwen3's training recipe (Appendix~\ref{app:head_alignment}).
In addition, the left plot of Figure~\ref{fig:ttft} shows that KV-Lingo translators are significantly more efficient than re-prefilling for long prefixes, and faster than MLPs, while almost preserving the performance of the target model on these shape-preserving pairs.
Appendix~\ref{appsec:qwen3_shape} reports a more complete set of evaluations.

\textbf{The impact of reasoning.}
We evaluate our KV-Lingo translators both reasoning-off and reasoning-on.
In $L \to S$, adding reasoning barely changes the score gap between KV-Lingo and the target model, while in $S \to L$ reasoning can significantly increase the score of translation (Figure~\ref{app:fig:reasoningboost}).
Hence, the score drop caused by handing a slightly degraded cache to a large model can be mitigated by making the model reason.
Moreover, the performance of closed-form translators is significantly degraded reasoning-on, contrary to KV-Lingo translators (Figure~\ref{fig:suite_small_on_mean}).

\textbf{Ablations.}
We test two other evaluation frameworks: (i) only translating the context, while the prompt is prefilled by the target model, showing in Appendix~\ref{app:handoffpoint} that KV-Lingo translators are robust to that change of framework (Table~\ref{tab:ctxonly}), and (ii) translating the prefix and the reasoning trace of the large model before handing it to the small model, which significantly boosts the performance of the small model (Table~\ref{tab:cothandoff}).
We also demonstrate the robustness of KV-Lingo translators to out-of-distribution languages, showing in Figure~\ref{fig:xquad} that performance is preserved across all the languages of XQuAD~\citep{artetxe2020cross}.
Finally, we test cache translation on a model pair pretrained from scratch with the architectures of Qwen3-0.6B and 1.7B, showing that the performance of KV-Lingo is preserved in that setting (Appendix~\ref{app:scratch}).

\subsection{Cross-layer cache translation}
\label{sec:exp:xdepth}

\begin{figure}[]
\centering
\includegraphics[width=\linewidth]{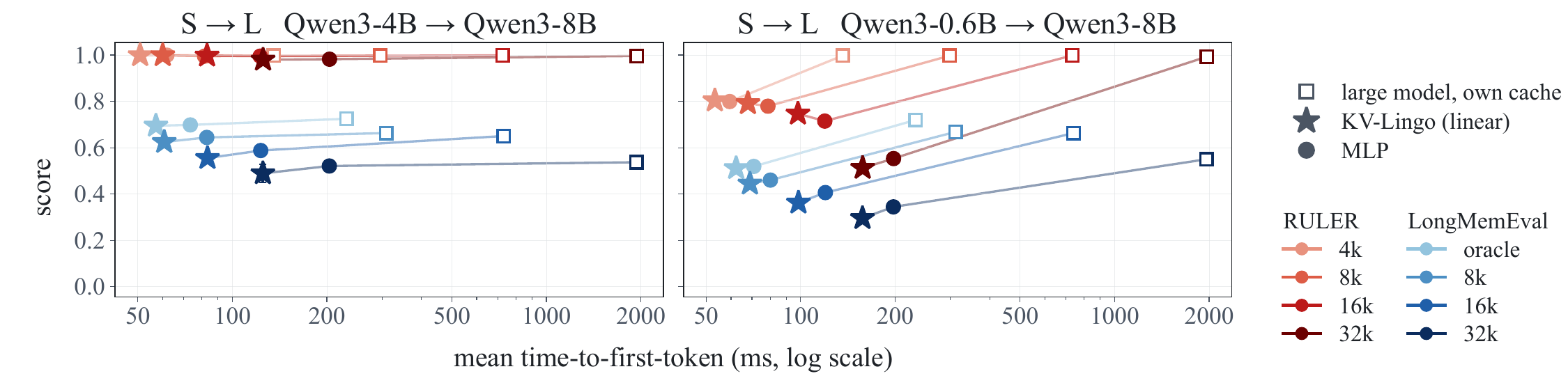}
\caption{\textbf{Score against time to first token, long context, $S \to L$.} Each colour
is one long-context evaluation task. Open squares:
Qwen3-8B decoding from its own cache, timed as one full prefill. Filled markers: Qwen3-8B
reading a translated cache, timed as the translation of an already computed source cache
plus one decoding step (Appendix~\ref{app:cost}).
\emph{Left}: shape-preserving pair.
\emph{Right}: cross-layer pair at $\nu = 2$.
Throughput is the same for all methods.}
\label{fig:ttft}
\end{figure}

\textbf{Qwen3 pairs.}
We translate between Qwen3-0.6B/8B and Qwen3-1.7B/4B (28 and 36 layers) in both directions, with the relative-depth, greedy and $R^2$ assignments of \S\ref{sec:method:family}.
On the 0.6B/8B pair, reasoning-off and averaged over nine benchmarks (Figure~\ref{fig:qwen3small}(b)), head-mixing KV-Lingo with 2 source layers and the $R^2$ assignment is almost on par with a head-mixing MLP in both directions ($0.476$ against $0.466$ in $L \to S$, $0.577$ against $0.597$ in $S \to L$), while translating faster (Figure~\ref{fig:ttft}, right).
Both stay just below the small model's $0.506$ in $L \to S$ and clear it by $0.07$--$0.09$ in $S \to L$, whereas the head-wise MLP falls well below it in $L \to S$ ($0.270$) and stays below it in $S \to L$ ($0.454$).
Both MLPs read $\nu=1$ source layer per target layer, with the greedy assignment for the head-mixing MLP and the relative-depth one for the head-wise MLP.
Mixing heads and adding source layers both improve the scores (Figure~\ref{fig:head_wise_vs_head_mixing}), notably long-context retrieval (Figure~\ref{fig:n_src_cross_depth}), at a cost proportional to $\nu$ (Figure~\ref{fig:nsrc_ttft}).
The three assignments perform similarly (Figures~\ref{fig:assignment_comparison_1} and~\ref{fig:assignment_comparison_2}).
On the unbalanced 0.6B/8B pair, translation can land well above the small model: reasoning-on, $S\to L$, on LongMemEval, the $8$B decoding from a translated $0.6$B cache scores $0.531$ ($R^2$ assignment) against the $0.6$B's own $0.383$ on the same items.
Appendix~\ref{app:xdepth} provides additional evaluations.

\textbf{Gemma pairs.}
We also test two Gemma pairs that differ in depth and in the number of key-value heads, with head dimension 256 throughout.
Gemma-3-4B-it and Gemma-3-12B-it~\citep{team2025gemma} have 34 and 48 layers, one global attention layer every 5 local ones, and 4 and 8 K/V heads, so that no head-wise method applies to either pair, and KV-Lingo translators are rectangular matrices throughout.
Gemma-4-E2B-it and Gemma-4-E4B-it~\citep{team2026gemma} have 35 and 42 layers, 4:1 and 5:1 local/global ratios, and 1 and 2 K/V heads; their last 20 and 18 layers reuse the caches of earlier layers.
Source layers are always mapped to target layers of the same attention type (local or global), with $\nu=1$ for Gemma-4 and $\nu\in\{1,2,4\}$ for Gemma-3.
On the benchmark means of Figures~\ref{fig:gemma-3} and~\ref{fig:gemma-4}, KV-Lingo outperforms the small model in $L \to S$ (for Gemma-3, with greedy match or at $\nu = 4$) but in $S \to L$ it only reaches the small model's level (within one standard error, for Gemma-3 at $\nu = 4$) rather than the large model's; Tables~\ref{tab:gemma3}, \ref{tab:gemma3nu} and~\ref{tab:gemma4} give the per-benchmark scores.

\subsection{Cross-tokenizer cache translation}
\label{sec:exp:xtok}

Everything so far assumed the two models segment the prefix identically. 
We drop that assumption with the model pair Qwen3-8B $\leftrightarrow$ Mistral-7B-Instruct-v0.3~\citep{jiang2023mistral7b},
where the source and the target use different tokenizers, so that their representations of a given prefix can have a different number
of tokens.
The models also have different depths ($36$ layers for Qwen3-8B, against $32$ layers for Mistral-7B):
we use the relative-depth assignment to determine the source-target layer mapping, with 3 source layers per target layer.
We describe in Appendix~\ref{app:subsec:xtok} our procedure to apply KV-Lingo to that setting.
Table~\ref{tab:xtok} reports both directions on seven benchmarks; the average scores are in Figure~\ref{fig:xtok_pair}. Even with a change of
tokenizer, KV-Lingo translators outperform the small model on two benchmarks in $L \to S$ and
on four in $S \to L$.

\subsection{MLA cache translation}
\label{sec:exp:mla}

We test KV-Lingo with multi-head latent attention (MLA), applying it as-is to \emph{latent cache translation} instead of KV-cache translation.
As a proof of concept, and since existing MLA pairs that share a tokenizer have $100$B--$1$T parameters~\citep{kimilinear, kimik2, deepseekv2lite, deepseekv2}, we convert the GQA models Qwen3-4B and Qwen3-8B into MLA models of similar architecture, with the conversion recipe CARE~\citep{zhou2026care} (see Appendix~\ref{app:mla} for the experimental protocol).
KV-Lingo works off-the-shelf in that setting:
the source and target latent caches have the same shape: $36$ layers $\times$ $1024$ latent dimensions, and each source layer $\ell$ is mapped to target layer $\ell$, in parallel over tokens, with one translator per layer $\ell$, separately for the latent vectors $c_{KV}$ and the decoupled RoPE vectors $k_R$: $\hat c_{KV,i}^{\tgt, \ell} = T_{KV}^\ell\, c_{KV,i}^{\src, \ell}$ and $ \hat k_{R,i}^{\tgt, \ell} = T_R^\ell \, k_{R,i}^{\src, \ell}$, for $i=1,\dots,n-1$.
Figure~\ref{fig:mla_pair} and Table~\ref{tab:mla} report the results.
Averaged over six benchmarks, KV-Lingo lands about $0.04$ below both native models ($0.588$ and $0.587$ against $0.629$ and $0.633$): latent caches translate as-is, at some cost in quality.

\subsection{Dense-to-MoE cache translation}
\label{sec:exp:moe}

We test the scenario of \S\ref{sec:setting:metrics} by prefilling with a dense Qwen3-4B, translating its cache, and
generating with Qwen3-30B-A3B~\citep{yang2025qwen3}, an MoE that activates $3.3$B of its
$30.5$B parameters per token. Each of the MoE's $48$ layers reads $\nu = 3$ of the dense model's
$36$, assigned by relative depth.
This scenario is of particular interest when serving at batch size 1, for instance on device.
The added capacity of the MoE is costly at prefill time since it requires fetching most or even all the experts' weights in parallel. Concretely, on an Apple M3 Ultra
(\texttt{bf16}, batch size $1$), the MoE decodes as fast as the dense model but takes
$152$\,ms to prefill a $64$-token prompt, against $68$\,ms, and translation adds only
$1.8$\,ms (see Figure~\ref{fig:cost}). KV-Lingo thus acts as a prefill accelerator for
the MoE.
Figure~\ref{app:fig:moe} reports quality per task category (full results in
Table~\ref{app:tab:moe}). On the two judged short-context benchmarks (RepLiQA and
MT-Bench-101), the MoE decoding from the translated cache comes within about $0.01$ of its own
cache, with reasoning on or off. On the six other short-context benchmarks it lands between
the two models with reasoning on, and below the dense model with reasoning off. Translation
fails on long context, mostly on RULER; on the dense cross-layer pairs, more source layers
per target layer buy back part of this recall (\S\ref{sec:exp:xdepth}). In that case, non-linearity does not help: an MLP in
place of the linear map reaches a higher KL loss (Table~\ref{app:tab:moemlp}).

\subsection{Multi-turn cache translation}
\label{sec:exp:multiturn}

\begin{figure}[t]
\centering
\includegraphics[width=0.95\linewidth]{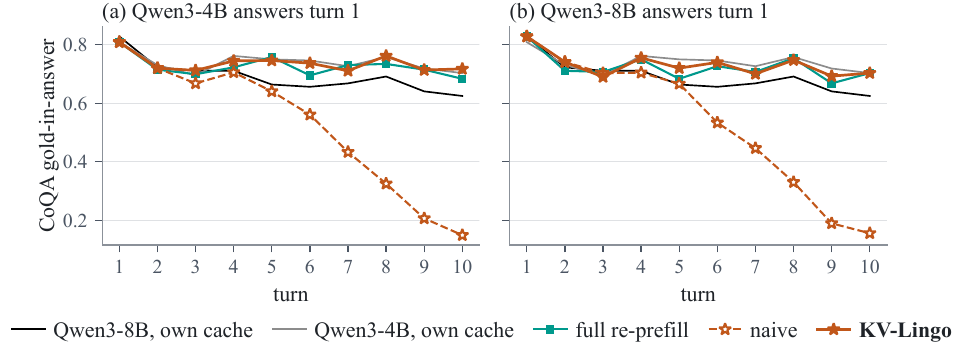}
\caption{\textbf{Multi-turn model switching on CoQA.} Qwen3-$4$B $\leftrightarrow$ Qwen3-$8$B alternate over 10 CoQA turns,
reasoning off; panels differ in which model answers turn $1$. Both strategies use the same translators. The naive strategy collapses; KV-Lingo stays level with full re-prefill through turn $10$. Translated conditions: mean over three seeds; error bars: standard error over evaluation items (Appendix~\ref{app:seeds}).}
\label{fig:multiturn_coqa}
\end{figure}

The experiments of \S\ref{sec:exp:shape} and \S\ref{sec:exp:xdepth} involve a single model switch.
For multiple switches, we propose to use KV-Lingo with \emph{alternated translation and generation}: each model keeps its own cache of the prefix, mixing native and translated
    segments (Figure~\ref{fig:multiturn}). At a switch, only the tokens generated since the previous switch are translated.
We compare this method with a naive baseline
which, at every switch, translates the full prefix: the context and earlier generations are then
translated back and forth, once per switch, and the cost of each switch grows with the prefix length.

On CoQA~\citep{reddy2019coqa}, QuAC~\citep{choi2018quac} and MultiChallenge~\citep{sirdeshmukh2025multichallengerealisticmultiturnconversation},
the naive method holds for a few turns and then drops sharply, while alternated translation and generation stays close to re-prefilling across all turns (within $0.05$ on QuAC) (Figures~\ref{fig:multiturn_coqa},~\ref{app:fig:multiturn_quac} and~\ref{app:fig:multiturn_multichallenge}).

\subsection{Computational gains of cache translation}
\label{sec:cost}

At a switch, re-prefilling is a forward pass of the target on the $n$ prefix tokens. Translation
replaces it with two much cheaper operations. The translator is a
token-wise linear map: on our shape-preserving Qwen3 pairs it costs $7.5$ to $92$ times
fewer FLOPs per token than applying the target's own weight matrices, and it computes no quadratic attention.
The target then runs a single decoding step over the current token, whose cost depends on
$n$ only mildly, through reading the cache. The time to first token thus goes from
$t_{\mathrm{prefill}}(n)$ to $t_T(n) + t_{\mathrm{step}}(n)$. 

\begin{figure}[t]
\centering
\includegraphics[width=\linewidth]{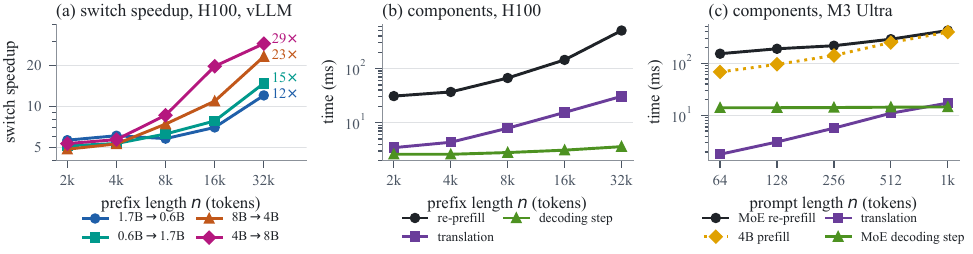}
\caption{\textbf{Switch cost of KV-lingo.} (a) Speedup of the switch on an H100,
the target served by vLLM: its time to first token after re-prefilling, divided by that
after translation plus one decoding step. (b) The terms of that ratio for the switch into
Qwen3-1.7B. (c) The same terms on an Apple M3 Ultra (MLX), from Qwen3-4B into
Qwen3-30B-A3B, plus the 4B's own prefill.}
\label{fig:cost}
\end{figure}

Figure~\ref{fig:cost} times both re-prefilling,  translation and one decoding step;
Appendix~\ref{app:cost} details the protocol and reports every timing. On an H100 with the
target served by vLLM~\citep{kwon2023efficient}, the switch reaches the first token $4.8$ to
$5.6\times$ sooner already at $2$k tokens, and $12$ to $29\times$ sooner at $32$k
(Figure~\ref{fig:cost}a). Panel (b) shows why the gap widens with $n$: re-prefill grows
quadratically, translation linearly, and the decoding step, which only reads the cache, stays
at $2.6$ to $3.6$\,ms. At long context, translation is thus most of the switch's cost. Since it
depends only on the cache width, which all four Qwen3 models share, while the prefill grows
with the target's size, larger targets gain more: $12.0\times$ into Qwen3-0.6B against
$28.9\times$ into Qwen3-8B at $32$k. On a device that decodes one request at a time, the gain
does not even need long contexts, since an MoE's prefill must load most of its experts. On an
Apple M3 Ultra with MLX~\citep{mlx2023}, Qwen3-30B-A3B prefills a $64$-token prompt in
$152$\,ms, whereas translating the cache of Qwen3-4B and one decoding step of the MoE take
$15.8$\,ms, so the MoE answers $9.6\times$ sooner (Figure~\ref{fig:cost}c, \S\ref{sec:exp:moe}).
After the switch, the target decodes at its native speed, since the translated cache has the
target's own shape.

\section{Conclusion}
KV-Lingo shows that KV caches can be translated between frozen language models using simple token-wise linear maps trained with self-distillation. The translated caches retain strong downstream performance while avoiding the target model’s re-prefill, substantially reducing the latency of model switching. Cache translation therefore provides a simple primitive for efficient dynamic routing and hand-off between language models.

\bibliographystyle{plainnat}
\bibliography{refs}

@article{vaswani2017attention,
  title   = {Attention is all you need},
  author  = {Vaswani, Ashish and Shazeer, Noam and Parmar, Niki and Uszkoreit, Jakob and Jones, Llion and Gomez, Aidan N and Kaiser, {\L}ukasz and Polosukhin, Illia},
  journal = {Advances in neural information processing systems},
  volume  = {30},
  year    = {2017}
}

@article{shazeer2017outrageously,
  title   = {Outrageously large neural networks: The sparsely-gated mixture-of-experts layer},
  author  = {Shazeer, Noam and Mirhoseini, Azalia and Maziarz, Krzysztof and Davis, Andy and Le, Quoc and Hinton, Geoffrey and Dean, Jeff},
  journal = {arXiv preprint arXiv:1701.06538},
  year    = {2017}
}

@article{su2024roformer,
  title     = {Roformer: Enhanced transformer with rotary position embedding},
  author    = {Su, Jianlin and Ahmed, Murtadha and Lu, Yu and Pan, Shengfeng and Bo, Wen and Liu, Yunfeng},
  journal   = {Neurocomputing},
  volume    = {568},
  pages     = {127063},
  year      = {2024},
  publisher = {Elsevier}
}

@article{yang2025qwen3,
  title   = {Qwen3 technical report},
  author  = {Yang, An and Li, Anfeng and Yang, Baosong and Zhang, Beichen and Hui, Binyuan and Zheng, Bo and Yu, Bowen and Gao, Chang and Huang, Chengen and Lv, Chenxu and others},
  journal = {arXiv preprint arXiv:2505.09388},
  year    = {2025}
}

@inproceedings{su2025nemotron,
  title     = {Nemotron-cc: Transforming common crawl into a refined long-horizon pretraining dataset},
  author    = {Su, Dan and Kong, Kezhi and Lin, Ying and Jennings, Joseph and Norick, Brandon and Kliegl, Markus and Patwary, Mostofa and Shoeybi, Mohammad and Catanzaro, Bryan},
  booktitle = {Proceedings of the 63rd Annual Meeting of the Association for Computational Linguistics (Volume 1: Long Papers)},
  pages     = {2459--2475},
  year      = {2025}
}

@misc{nvidia2025nemotron3nanoopen,
      title={Nemotron 3 Nano: Open, Efficient Mixture-of-Experts Hybrid Mamba-Transformer Model for Agentic Reasoning}, 
      author={NVIDIA and : and Aaron Blakeman and Aaron Grattafiori and Aarti Basant and Abhibha Gupta and Abhinav Khattar and Adi Renduchintala and Aditya Vavre and Akanksha Shukla and Akhiad Bercovich and Aleksander Ficek and Aleksandr Shaposhnikov and Alex Kondratenko and Alexander Bukharin and Alexandre Milesi and Ali Taghibakhshi and Alisa Liu and Amelia Barton and Ameya Sunil Mahabaleshwarkar and Amir Klein and Amit Zuker and Amnon Geifman and Amy Shen and Anahita Bhiwandiwalla and Andrew Tao and Ann Guan and Anubhav Mandarwal and Arham Mehta and Ashwath Aithal and Ashwin Poojary and Asif Ahamed and Asma Kuriparambil Thekkumpate and Ayush Dattagupta and Banghua Zhu and Bardiya Sadeghi and Barnaby Simkin and Ben Lanir and Benedikt Schifferer and Besmira Nushi and Bilal Kartal and Bita Darvish Rouhani and Boris Ginsburg and Brandon Norick and Brandon Soubasis and Branislav Kisacanin and Brian Yu and Bryan Catanzaro and Carlo del Mundo and Chantal Hwang and Charles Wang and Cheng-Ping Hsieh and Chenghao Zhang and Chenhan Yu and Chetan Mungekar and Chintan Patel and Chris Alexiuk and Christopher Parisien and Collin Neale and Damon Mosk-Aoyama and Dan Su and Dane Corneil and Daniel Afrimi and Daniel Rohrer and Daniel Serebrenik and Daria Gitman and Daria Levy and Darko Stosic and David Mosallanezhad and Deepak Narayanan and Dhruv Nathawani and Dima Rekesh and Dina Yared and Divyanshu Kakwani and Dong Ahn and Duncan Riach and Dusan Stosic and Edgar Minasyan and Edward Lin and Eileen Long and Eileen Peters Long and Elena Lantz and Ellie Evans and Elliott Ning and Eric Chung and Eric Harper and Eric Tramel and Erick Galinkin and Erik Pounds and Evan Briones and Evelina Bakhturina and Faisal Ladhak and Fay Wang and Fei Jia and Felipe Soares and Feng Chen and Ferenc Galko and Frankie Siino and Gal Hubara Agam and Ganesh Ajjanagadde and Gantavya Bhatt and Gargi Prasad and George Armstrong and Gerald Shen and Gorkem Batmaz and Grigor Nalbandyan and Haifeng Qian and Harsh Sharma and Hayley Ross and Helen Ngo and Herman Sahota and Hexin Wang and Himanshu Soni and Hiren Upadhyay and Huizi Mao and Huy C Nguyen and Huy Q Nguyen and Iain Cunningham and Ido Shahaf and Igor Gitman and Ilya Loshchilov and Ivan Moshkov and Izzy Putterman and Jan Kautz and Jane Polak Scowcroft and Jared Casper and Jatin Mitra and Jeffrey Glick and Jenny Chen and Jesse Oliver and Jian Zhang and Jiaqi Zeng and Jie Lou and Jimmy Zhang and Jining Huang and Joey Conway and Joey Guman and John Kamalu and Johnny Greco and Jonathan Cohen and Joseph Jennings and Joyjit Daw and Julien Veron Vialard and Junkeun Yi and Jupinder Parmar and Kai Xu and Kan Zhu and Kari Briski and Katherine Cheung and Katherine Luna and Keshav Santhanam and Kevin Shih and Kezhi Kong and Khushi Bhardwaj and Krishna C. Puvvada and Krzysztof Pawelec and Kumar Anik and Lawrence McAfee and Laya Sleiman and Leon Derczynski and Li Ding and Lucas Liebenwein and Luis Vega and Maanu Grover and Maarten Van Segbroeck and Maer Rodrigues de Melo and Makesh Narsimhan Sreedhar and Manoj Kilaru and Maor Ashkenazi and Marc Romeijn and Mark Cai and Markus Kliegl and Maryam Moosaei and Matvei Novikov and Mehrzad Samadi and Melissa Corpuz and Mengru Wang and Meredith Price and Michael Boone and Michael Evans and Miguel Martinez and Mike Chrzanowski and Mohammad Shoeybi and Mostofa Patwary and Nabin Mulepati and Natalie Hereth and Nave Assaf and Negar Habibi and Neta Zmora and Netanel Haber and Nicola Sessions and Nidhi Bhatia and Nikhil Jukar and Nikki Pope and Nikolai Ludwig and Nima Tajbakhsh and Nirmal Juluru and Oleksii Hrinchuk and Oleksii Kuchaiev and Olivier Delalleau and Oluwatobi Olabiyi and Omer Ullman Argov and Ouye Xie and Parth Chadha and Pasha Shamis and Pavlo Molchanov and Pawel Morkisz and Peter Dykas and Peter Jin and Pinky Xu and Piotr Januszewski and Pranav Prashant Thombre and Prasoon Varshney and Pritam Gundecha and Qing Miao and Rabeeh Karimi Mahabadi and Ran El-Yaniv and Ran Zilberstein and Rasoul Shafipour and Rich Harang and Rick Izzo and Rima Shahbazyan and Rishabh Garg and Ritika Borkar and Ritu Gala and Riyad Islam and Roger Waleffe and Rohit Watve and Roi Koren and Ruoxi Zhang and Russell J. Hewett and Ryan Prenger and Ryan Timbrook and Sadegh Mahdavi and Sahil Modi and Samuel Kriman and Sanjay Kariyappa and Sanjeev Satheesh and Saori Kaji and Satish Pasumarthi and Sean Narentharen and Sean Narenthiran and Seonmyeong Bak and Sergey Kashirsky and Seth Poulos and Shahar Mor and Shanmugam Ramasamy and Shantanu Acharya and Shaona Ghosh and Sharath Turuvekere Sreenivas and Shelby Thomas and Shiqing Fan and Shreya Gopal and Shrimai Prabhumoye and Shubham Pachori and Shubham Toshniwal and Shuoyang Ding and Siddharth Singh and Simeng Sun and Smita Ithape and Somshubra Majumdar and Soumye Singhal and Stefania Alborghetti and Stephen Ge and Sugam Dipak Devare and Sumeet Kumar Barua and Suseella Panguluri and Suyog Gupta and Sweta Priyadarshi and Syeda Nahida Akter and Tan Bui and Teodor-Dumitru Ene and Terry Kong and Thanh Do and Tijmen Blankevoort and Tom Balough and Tomer Asida and Tomer Bar Natan and Tugrul Konuk and Twinkle Vashishth and Udi Karpas and Ushnish De and Vahid Noorozi and Vahid Noroozi and Venkat Srinivasan and Venmugil Elango and Vijay Korthikanti and Vitaly Kurin and Vitaly Lavrukhin and Wanli Jiang and Wasi Uddin Ahmad and Wei Du and Wei Ping and Wenfei Zhou and Will Jennings and William Zhang and Wojciech Prazuch and Xiaowei Ren and Yashaswi Karnati and Yejin Choi and Yev Meyer and Yi-Fu Wu and Yian Zhang and Ying Lin and Yonatan Geifman and Yonggan Fu and Yoshi Subara and Yoshi Suhara and Yubo Gao and Zach Moshe and Zhen Dong and Zihan Liu and Zijia Chen and Zijie Yan},
      year={2025},
      eprint={2512.20848},
      archivePrefix={arXiv},
      primaryClass={cs.CL},
      url={https://arxiv.org/abs/2512.20848}, 
}

@misc{merity2016pointer,
  title         = {Pointer Sentinel Mixture Models},
  author        = {Stephen Merity and Caiming Xiong and James Bradbury and Richard Socher},
  year          = {2016},
  eprint        = {1609.07843},
  archiveprefix = {arXiv},
  primaryclass  = {cs.CL}
}

@inproceedings{clark2019boolq,
  title     = {Boolq: Exploring the surprising difficulty of natural yes/no questions},
  author    = {Clark, Christopher and Lee, Kenton and Chang, Ming-Wei and Kwiatkowski, Tom and Collins, Michael and Toutanova, Kristina},
  booktitle = {Proceedings of the 2019 Conference of the North American Chapter of the Association for Computational Linguistics: Human Language Technologies, Volume 1 (Long and Short Papers)},
  pages     = {2924--2936},
  year      = {2019}
}

@inproceedings{monteiro2024repliqa,
  title     = {RepLi{QA}: A Question-Answering Dataset for Benchmarking {LLM}s on Unseen Reference Content},
  author    = {Joao Monteiro and Pierre-Andre Noel and {\'E}tienne Marcotte and Sai Rajeswar and Valentina Zantedeschi and David Vazquez and Nicolas Chapados and Christopher Pal and Perouz Taslakian},
  booktitle = {The Thirty-eight Conference on Neural Information Processing Systems Datasets and Benchmarks Track},
  year      = {2024},
  url       = {https://openreview.net/forum?id=4diKTLmg2y}
}

@article{hsieh2024ruler,
  title   = {RULER: What's the real context size of your long-context language models?},
  author  = {Hsieh, Cheng-Ping and Sun, Simeng and Kriman, Samuel and Acharya, Shantanu and Rekesh, Dima and Jia, Fei and Zhang, Yang and Ginsburg, Boris},
  journal = {arXiv preprint arXiv:2404.06654},
  year    = {2024}
}

@inproceedings{bai2024mt,
  title     = {Mt-bench-101: A fine-grained benchmark for evaluating large language models in multi-turn dialogues},
  author    = {Bai, Ge and Liu, Jie and Bu, Xingyuan and He, Yancheng and Liu, Jiaheng and Zhou, Zhanhui and Lin, Zhuoran and Su, Wenbo and Ge, Tiezheng and Zheng, Bo and others},
  booktitle = {Proceedings of the 62nd Annual Meeting of the Association for Computational Linguistics (Volume 1: Long Papers)},
  pages     = {7421--7454},
  year      = {2024}
}

@article{clark2018think,
  title   = {Think you have solved question answering? try arc, the ai2 reasoning challenge},
  author  = {Clark, Peter and Cowhey, Isaac and Etzioni, Oren and Khot, Tushar and Sabharwal, Ashish and Schoenick, Carissa and Tafjord, Oyvind},
  journal = {arXiv preprint arXiv:1803.05457},
  year    = {2018}
}

@inproceedings{zellers2019hellaswag,
  title     = {Hellaswag: Can a machine really finish your sentence?},
  author    = {Zellers, Rowan and Holtzman, Ari and Bisk, Yonatan and Farhadi, Ali and Choi, Yejin},
  booktitle = {Proceedings of the 57th annual meeting of the association for computational linguistics},
  pages     = {4791--4800},
  year      = {2019}
}

@article{sakaguchi2021winogrande,
  title     = {Winogrande: An adversarial winograd schema challenge at scale},
  author    = {Sakaguchi, Keisuke and Bras, Ronan Le and Bhagavatula, Chandra and Choi, Yejin},
  journal   = {Communications of the ACM},
  volume    = {64},
  number    = {9},
  pages     = {99--106},
  year      = {2021},
  publisher = {ACM New York, NY, USA}
}

@inproceedings{bisk2020piqa,
  title     = {Piqa: Reasoning about physical commonsense in natural language},
  author    = {Bisk, Yonatan and Zellers, Rowan and Gao, Jianfeng and Choi, Yejin and others},
  booktitle = {Proceedings of the AAAI conference on artificial intelligence},
  volume    = {34},
  number    = {05},
  pages     = {7432--7439},
  year      = {2020}
}

@article{cobbe2021training,
  title   = {Training verifiers to solve math word problems},
  author  = {Cobbe, Karl and Kosaraju, Vineet and Bavarian, Mohammad and Chen, Mark and Jun, Heewoo and Kaiser, Lukasz and Plappert, Matthias and Tworek, Jerry and Hilton, Jacob and Nakano, Reiichiro and others},
  journal = {arXiv preprint arXiv:2110.14168},
  year    = {2021}
}

@article{team2026gemma,
  title   = {Gemma 4 technical report},
  author  = {Team, Gemma and Abd, Sherif El and Aggarwal, Vaibhav and Algayres, Robin and Andreev, Alek and Bachem, Olivier and Ballantyne, Ian and Brick, Cormac and C{\u{a}}rbune, Victor and Casbon, Michelle and others},
  journal = {arXiv preprint arXiv:2607.02770},
  year    = {2026}
}

@inproceedings{kornblith2019similarity,
  title        = {Similarity of neural network representations revisited},
  author       = {Kornblith, Simon and Norouzi, Mohammad and Lee, Honglak and Hinton, Geoffrey},
  booktitle    = {International conference on machine learning},
  pages        = {3519--3529},
  year         = {2019},
  organization = {PMlR}
}

@inproceedings{nguyen2021do,
  title     = {Do Wide and Deep Networks Learn the Same Things? Uncovering How Neural Network Representations Vary with Width and Depth},
  author    = {Thao Nguyen and Maithra Raghu and Simon Kornblith},
  booktitle = {International Conference on Learning Representations},
  year      = {2021},
  url       = {https://openreview.net/forum?id=KJNcAkY8tY4}
}

@inproceedings{lenc2015understanding,
  title     = {Understanding image representations by measuring their equivariance and equivalence},
  author    = {Lenc, Karel and Vedaldi, Andrea},
  booktitle = {Proceedings of the IEEE conference on computer vision and pattern recognition},
  pages     = {991--999},
  year      = {2015}
}

@article{bansal2021revisiting,
  title   = {Revisiting model stitching to compare neural representations},
  author  = {Bansal, Yamini and Nakkiran, Preetum and Barak, Boaz},
  journal = {Advances in neural information processing systems},
  volume  = {34},
  pages   = {225--236},
  year    = {2021}
}

@article{csiszarik2021similarity,
  title   = {Similarity and matching of neural network representations},
  author  = {Csisz{\'a}rik, Adri{\'a}n and K{\H{o}}r{\"o}si-Szab{\'o}, P{\'e}ter and Matszangosz, Akos and Papp, Gergely and Varga, D{\'a}niel},
  journal = {Advances in Neural Information Processing Systems},
  volume  = {34},
  pages   = {5656--5668},
  year    = {2021}
}

@inproceedings{moschella2023relative,
  title     = {Relative representations enable zero-shot latent space communication},
  author    = {Luca Moschella and Valentino Maiorca and Marco Fumero and Antonio Norelli and Francesco Locatello and Emanuele Rodol{\`a}},
  booktitle = {The Eleventh International Conference on Learning Representations },
  year      = {2023},
  url       = {https://openreview.net/forum?id=SrC-nwieGJ}
}

@article{li2015convergent,
  title   = {Convergent learning: Do different neural networks learn the same representations?},
  author  = {Li, Yixuan and Yosinski, Jason and Clune, Jeff and Lipson, Hod and Hopcroft, John},
  journal = {arXiv preprint arXiv:1511.07543},
  year    = {2015}
}

@misc{olmo2025olmo3,
  title         = {Olmo 3},
  author        = {Team Olmo and Allyson Ettinger and Amanda Bertsch and Bailey Kuehl and David Graham and David Heineman and Dirk Groeneveld and Faeze Brahman and Finbarr Timbers and Hamish Ivison and Jacob Morrison and Jake Poznanski and Kyle Lo and Luca Soldaini and Matt Jordan and Mayee Chen and Michael Noukhovitch and Nathan Lambert and Pete Walsh and Pradeep Dasigi and Robert Berry and Saumya Malik and Saurabh Shah and Scott Geng and Shane Arora and Shashank Gupta and Taira Anderson and Teng Xiao and Tyler Murray and Tyler Romero and Victoria Graf and Akari Asai and Akshita Bhagia and Alexander Wettig and Alisa Liu and Aman Rangapur and Chloe Anastasiades and Costa Huang and Dustin Schwenk and Harsh Trivedi and Ian Magnusson and Jaron Lochner and Jiacheng Liu and Lester James V. Miranda and Maarten Sap and Malia Morgan and Michael Schmitz and Michal Guerquin and Michael Wilson and Regan Huff and Ronan Le Bras and Rui Xin and Rulin Shao and Sam Skjonsberg and Shannon Zejiang Shen and Shuyue Stella Li and Tucker Wilde and Valentina Pyatkin and Will Merrill and Yapei Chang and Yuling Gu and Zhiyuan Zeng and Ashish Sabharwal and Luke Zettlemoyer and Pang Wei Koh and Ali Farhadi and Noah A. Smith and Hannaneh Hajishirzi},
  year          = {2025},
  eprint        = {2512.13961},
  archiveprefix = {arXiv},
  primaryclass  = {cs.CL},
  url           = {https://arxiv.org/abs/2512.13961}
}

@misc{sirdeshmukh2025multichallengerealisticmultiturnconversation,
  title         = {MultiChallenge: A Realistic Multi-Turn Conversation Evaluation Benchmark Challenging to Frontier LLMs},
  author        = {Ved Sirdeshmukh and Kaustubh Deshpande and Johannes Mols and Lifeng Jin and Ed-Yeremai Cardona and Dean Lee and Jeremy Kritz and Willow Primack and Summer Yue and Chen Xing},
  year          = {2025},
  eprint        = {2501.17399},
  archiveprefix = {arXiv},
  primaryclass  = {cs.CL},
  url           = {https://arxiv.org/abs/2501.17399}
}

@article{choi2018quac,
  author     = {Eunsol Choi and
                He He and
                Mohit Iyyer and
                Mark Yatskar and
                Wen{-}tau Yih and
                Yejin Choi and
                Percy Liang and
                Luke Zettlemoyer},
  title      = {QuAC : Question Answering in Context},
  journal    = {CoRR},
  volume     = {abs/1808.07036},
  year       = {2018},
  url        = {http://arxiv.org/abs/1808.07036},
  eprinttype = {arXiv},
  eprint     = {1808.07036},
  bibsource  = {dblp computer science bibliography, https://dblp.org}
}

@misc{geng2026relaycachingacceleratingllmcollaboration,
  title         = {RelayCaching: Accelerating LLM Collaboration via Decoding KV Cache Reuse},
  author        = {Yingsheng Geng and Yuchong Gao and Weihong Wu and Guyue Liu and Jiang Liu},
  year          = {2026},
  eprint        = {2603.13289},
  archiveprefix = {arXiv},
  primaryclass  = {cs.LG},
  url           = {https://arxiv.org/abs/2603.13289}
}

@inproceedings{huh2024position,
  title     = {Position: The Platonic Representation Hypothesis},
  author    = {Minyoung Huh and Brian Cheung and Tongzhou Wang and Phillip Isola},
  booktitle = {Forty-first International Conference on Machine Learning},
  year      = {2024},
  url       = {https://openreview.net/forum?id=BH8TYy0r6u}
}

@inproceedings{lee2025shared,
  title     = {Shared Global and Local Geometry of Language Model Embeddings},
  author    = {Andrew Lee and Melanie Weber and Fernanda Vi{\'e}gas and Martin Wattenberg},
  booktitle = {Second Conference on Language Modeling},
  year      = {2025},
  url       = {https://openreview.net/forum?id=aJDykpJAYF}
}

@inproceedings{chen2025transferring,
  title     = {Transferring Linear Features Across Language Models With Model Stitching},
  author    = {Alan Chen and Jack Merullo and Alessandro Stolfo and Ellie Pavlick},
  booktitle = {The Thirty-ninth Annual Conference on Neural Information Processing Systems},
  year      = {2025},
  url       = {https://openreview.net/forum?id=Qvvy0X63Fv}
}

@article{zhang2023h2o,
  title   = {H2o: Heavy-hitter oracle for efficient generative inference of large language models},
  author  = {Zhang, Zhenyu and Sheng, Ying and Zhou, Tianyi and Chen, Tianlong and Zheng, Lianmin and Cai, Ruisi and Song, Zhao and Tian, Yuandong and R{\'e}, Christopher and Barrett, Clark and others},
  journal = {Advances in neural information processing systems},
  volume  = {36},
  pages   = {34661--34710},
  year    = {2023}
}

@article{li2024snapkv,
  title   = {Snapkv: Llm knows what you are looking for before generation},
  author  = {Li, Yuhong and Huang, Yingbing and Yang, Bowen and Venkitesh, Bharat and Locatelli, Acyr and Ye, Hanchen and Cai, Tianle and Lewis, Patrick and Chen, Deming},
  journal = {Advances in Neural Information Processing Systems},
  volume  = {37},
  pages   = {22947--22970},
  year    = {2024}
}

@inproceedings{zhao2025smallkv,
  title     = {Small{KV}: Small Model Assisted Compensation of {KV} Cache Compression for Efficient {LLM} Inference},
  author    = {Yi Zhao and Yajuan Peng and Nguyen Cam-Tu and Zuchao Li and Wang Xiaoliang and hai zhao and Xiaoming Fu},
  booktitle = {The Thirty-ninth Annual Conference on Neural Information Processing Systems},
  year      = {2025},
  url       = {https://openreview.net/forum?id=0BVrpXMr5Y}
}

@article{heo2026cross,
  title   = {Cross-Model KV Cache Transfer in LLM Families: A Closed-Form Linear Mapping for Prefill Reuse},
  author  = {Heo, Taekyung and Shafipour, Rasoul and Zhao, Ritchie and Golub, Maximilian and Kamani, Mohammad Mahdi and Borkar, Ritika and Chandran, Makesh Tarun and Zardoshti, Pantea and Rouhani, Bita Darvish},
  journal = {arXiv preprint arXiv:2608.03893},
  year    = {2026}
}

@article{li2026universal,
  title   = {A Universal Context-Reuse Layer for Cross-Model KV Sharing},
  author  = {Li, Yi and Jiang, Dongming and Zhao, Yi and Li, Bingzhe},
  journal = {arXiv preprint arXiv:2608.30963},
  year    = {2026}
}

@article{qu2026cachebridge,
  title   = {CacheBridge: Efficient Cross-Model KV Cache Transfer},
  author  = {Qu, Xingyu and Lu, Siyuan and Chen, Zhiyu and Wang, Sheng and Lin, Tao},
  journal = {arXiv preprint arXiv:2609.00891},
  year    = {2026}
}

@inproceedings{liu2024kivi,
  title     = {{KIVI}: A Tuning-Free Asymmetric 2bit Quantization for {KV} Cache},
  author    = {Zirui Liu and Jiayi Yuan and Hongye Jin and Shaochen Zhong and Zhaozhuo Xu and Vladimir Braverman and Beidi Chen and Xia Hu},
  booktitle = {Forty-first International Conference on Machine Learning},
  year      = {2024},
  url       = {https://openreview.net/forum?id=L057s2Rq8O}
}

@article{hooper2024kvquant,
  title   = {Kvquant: Towards 10 million context length llm inference with kv cache quantization},
  author  = {Hooper, Coleman and Kim, Sehoon and Mohammadzadeh, Hiva and Mahoney, Michael W and Shao, Yakun S and Keutzer, Kurt and Gholami, Amir},
  journal = {Advances in Neural Information Processing Systems},
  volume  = {37},
  pages   = {1270--1303},
  year    = {2024}
}

@article{yang2024no,
  title   = {No token left behind: Reliable kv cache compression via importance-aware mixed precision quantization},
  author  = {Yang, June Yong and Kim, Byeongwook and Bae, Jeongin and Kwon, Beomseok and Park, Gunho and Yang, Eunho and Kwon, Se Jung and Lee, Dongsoo},
  journal = {arXiv preprint arXiv:2402.18096},
  year    = {2024}
}

@article{gim2024prompt,
  title   = {Prompt cache: Modular attention reuse for low-latency inference},
  author  = {Gim, In and Chen, Guojun and Lee, Seung-seob and Sarda, Nikhil and Khandelwal, Anurag and Zhong, Lin},
  journal = {Proceedings of Machine Learning and Systems},
  volume  = {6},
  pages   = {325--338},
  year    = {2024}
}

@inproceedings{yao2025cacheblend,
  title     = {Cacheblend: Fast large language model serving for rag with cached knowledge fusion},
  author    = {Yao, Jiayi and Li, Hanchen and Liu, Yuhan and Ray, Siddhant and Cheng, Yihua and Zhang, Qizheng and Du, Kuntai and Lu, Shan and Jiang, Junchen},
  booktitle = {Proceedings of the twentieth European conference on computer systems},
  pages     = {94--109},
  year      = {2025}
}

@inproceedings{yang2025kvlink,
  title     = {{KVL}ink: Accelerating Large Language Models via Efficient {KV} Cache Reuse},
  author    = {Jingbo Yang and Bairu Hou and Wei Wei and Yujia Bao and Shiyu Chang},
  booktitle = {The Thirty-ninth Annual Conference on Neural Information Processing Systems},
  year      = {2025},
  url       = {https://openreview.net/forum?id=oDcAGSXZZP}
}

@article{liu2024droidspeak,
  title   = {Droidspeak: Kv cache sharing for cross-LLM communication and multi-LLM serving},
  author  = {Liu, Yuhan and Huang, Yuyang and Yao, Jiayi and Feng, Shaoting and Gu, Zhuohan and Du, Kuntai and Li, Hanchen and Cheng, Yihua and Jiang, Junchen and Lu, Shan and others},
  journal = {arXiv preprint arXiv:2411.02820},
  year    = {2024}
}

@inproceedings{woo2026icarus,
  title     = {{IC}aRus: Identical Cache Reuse for Efficient Multi-Model Inference},
  author    = {Sunghyeon Woo and Jaeeun Kil and Hoseung Kim and Minsub Kim and Joonghoon Kim and Ahreum Seo and Sungjae Lee and Minjung Jo and Jiwon Ryu and Baeseong park and Se Jung Kwon and Dongsoo Lee},
  booktitle = {The Fourteenth International Conference on Learning Representations},
  year      = {2026},
  url       = {https://openreview.net/forum?id=qrMo6R7lOS}
}

@article{yang2024kvsharer,
  title   = {Kvsharer: Efficient inference via layer-wise dissimilar kv cache sharing},
  author  = {Yang, Yifei and Cao, Zouying and Chen, Qiguang and Qin, Libo and Yang, Dongjie and Zhao, Hai and Chen, Zhi},
  journal = {arXiv preprint arXiv:2410.18517},
  year    = {2024}
}

@inproceedings{brandon2024reducing,
  title     = {Reducing Transformer Key-Value Cache Size with Cross-Layer Attention},
  author    = {William Brandon and Mayank Mishra and Aniruddha Nrusimha and Rameswar Panda and Jonathan Ragan-Kelley},
  booktitle = {The Thirty-eighth Annual Conference on Neural Information Processing Systems},
  year      = {2024},
  url       = {https://openreview.net/forum?id=M2UzLRoqic}
}

@inproceedings{wu2025improving,
  title     = {Improving Model Representation and Reducing {KV} Cache via Skip Connections with First Value Heads},
  author    = {Zhoutong Wu and Yuan Zhang and Yiming Dong and Chenheng Zhang and Cong Fang and Kun Yuan and Zhouchen Lin},
  booktitle = {The Thirty-ninth Annual Conference on Neural Information Processing Systems},
  year      = {2025},
  url       = {https://openreview.net/forum?id=bjV8Y38aFF}
}

@inproceedings{wu2025systematic,
  title     = {A systematic study of cross-layer kv sharing for efficient llm inference},
  author    = {Wu, You and Wu, Haoyi and Tu, Kewei},
  booktitle = {Proceedings of the 2025 Conference of the Nations of the Americas Chapter of the Association for Computational Linguistics: Human Language Technologies (Volume 2: Short Papers)},
  pages     = {396--403},
  year      = {2025}
}

@article{filippova2026stochastic,
  title   = {Stochastic kv routing: Enabling adaptive depth-wise cache sharing},
  author  = {Filippova, Anastasiia and Grangier, David and Cuturi, Marco and Monteiro, Joao},
  journal = {arXiv preprint arXiv:2604.22782},
  year    = {2026}
}

@article{gelberg2026training,
  title   = {Training Transformers for KV Cache Compressibility},
  author  = {Gelberg, Yoav and Eitan, Yam and Bronstein, Michael and Gal, Yarin and Maron, Haggai},
  journal = {arXiv preprint arXiv:2605.05971},
  year    = {2026}
}

@inproceedings{lin2025matryoshkakv,
  title     = {Matryoshkakv: Adaptive kv compression via trainable orthogonal projection},
  author    = {Lin, Bokai and Zeng, Zihao and Xiao, Zipeng and Kou, Siqi and Hou, Tianqi and Gao, Xiaofeng and Zhang, Hao and Deng, Zhijie},
  booktitle = {International Conference on Learning Representations},
  volume    = {2025},
  pages     = {86669--86690},
  year      = {2025}
}

@inproceedings{kim2025lexico,
  title     = {Lexico: Extreme {KV} Cache Compression via Sparse Coding over Universal Dictionaries},
  author    = {Junhyuck Kim and Jongho Park and Jaewoong Cho and Dimitris Papailiopoulos},
  booktitle = {Forty-second International Conference on Machine Learning},
  year      = {2025},
  url       = {https://openreview.net/forum?id=Yh9vxlxnjA}
}

@inproceedings{kim2025kvzip,
  title     = {{KV}zip: Query-Agnostic {KV} Cache Compression with Context Reconstruction},
  author    = {Jang-Hyun Kim and Jinuk Kim and Sangwoo Kwon and Jae W. Lee and Sangdoo Yun and Hyun Oh Song},
  booktitle = {The Thirty-ninth Annual Conference on Neural Information Processing Systems},
  year      = {2025},
  url       = {https://openreview.net/forum?id=JFygzwx8SJ}
}

@inproceedings{eyuboglu2026cartridges,
  title     = {Cartridges: Lightweight and general-purpose long context representations via self-study},
  author    = {Sabri Eyuboglu and Ryan Saul Ehrlich and Simran Arora and Neel Guha and Dylan Zinsley and Emily Ruoyu Liu and Atri Rudra and James Zou and Azalia Mirhoseini and Christopher Re},
  booktitle = {The Fourteenth International Conference on Learning Representations},
  year      = {2026},
  url       = {https://openreview.net/forum?id=0k5w8O0SNg}
}

@inproceedings{zweiger2026fast,
  title     = {Fast {KV} Compaction via Attention Matching},
  author    = {Adam Zweiger and Xinghong Fu and Han Guo and Yoon Kim},
  booktitle = {Forty-third International Conference on Machine Learning},
  year      = {2026},
  url       = {https://openreview.net/forum?id=t01SKTj8pP}
}

@article{monteiro2026nectar,
  title   = {Nectar: Neural Estimation of Cached-Token Attention via Regression},
  author  = {Monteiro, Jo{\~a}o and Klein, Michal and Ablin, Pierre and Cuturi, Marco},
  journal = {arXiv preprint arXiv:2605.09778},
  year    = {2026}
}

@article{chen2024frugalgpt,
  title   = {Frugal{GPT}: How to Use Large Language Models While Reducing Cost and Improving Performance},
  author  = {Lingjiao Chen and Matei Zaharia and James Zou},
  journal = {Transactions on Machine Learning Research},
  issn    = {2835-8856},
  year    = {2024},
  url     = {https://openreview.net/forum?id=cSimKw5p6R},
  note    = {Featured Certification}
}

@inproceedings{ong2025routellm,
  title     = {Route{LLM}: Learning to Route {LLM}s from Preference Data},
  author    = {Isaac Ong and Amjad Almahairi and Vincent Wu and Wei-Lin Chiang and Tianhao Wu and Joseph E. Gonzalez and M Waleed Kadous and Ion Stoica},
  booktitle = {The Thirteenth International Conference on Learning Representations},
  year      = {2025},
  url       = {https://openreview.net/forum?id=8sSqNntaMr}
}

@inproceedings{shen2024learning,
  title     = {Learning to decode collaboratively with multiple language models},
  author    = {Shen, Zejiang and Lang, Hunter and Wang, Bailin and Kim, Yoon and Sontag, David},
  booktitle = {Proceedings of the 62nd Annual Meeting of the Association for Computational Linguistics (Volume 1: Long Papers)},
  pages     = {12974--12990},
  year      = {2024}
}

@article{fu2025r2r,
  title   = {R2r: Efficiently navigating divergent reasoning paths with small-large model token routing},
  author  = {Fu, Tianyu and Ge, Yi and You, Yichen and Liu, Enshu and Yuan, Zhihang and Dai, Guohao and Yan, Shengen and Yang, Huazhong and Wang, Yu},
  journal = {Advances in Neural Information Processing Systems},
  volume  = {38},
  pages   = {124108--124145},
  year    = {2025}
}

@inproceedings{wu2024autogen,
  title     = {AutoGen: Enabling Next-Gen {LLM} Applications via Multi-Agent Conversations},
  author    = {Qingyun Wu and Gagan Bansal and Jieyu Zhang and Yiran Wu and Beibin Li and Erkang Zhu and Li Jiang and Xiaoyun Zhang and Shaokun Zhang and Jiale Liu and Ahmed Hassan Awadallah and Ryen W White and Doug Burger and Chi Wang},
  booktitle = {First Conference on Language Modeling},
  year      = {2024},
  url       = {https://openreview.net/forum?id=BAakY1hNKS}
}

@inproceedings{fu2026cachetocache,
  title     = {Cache-to-Cache: Direct Semantic Communication Between Large Language Models},
  author    = {Tianyu Fu and Zihan Min and Hanling Zhang and Jichao Yan and Guohao Dai and Wanli Ouyang and Yu Wang},
  booktitle = {The Fourteenth International Conference on Learning Representations},
  year      = {2026},
  url       = {https://openreview.net/forum?id=LeatkxrBCi}
}

@article{horton2024kv,
  title   = {KV prediction for improved time to first token},
  author  = {Horton, Maxwell and Cao, Qingqing and Sun, Chenfan and Jin, Yanzi and Mehta, Sachin and Rastegari, Mohammad and Nabi, Moin},
  journal = {arXiv preprint arXiv:2410.08391},
  year    = {2024}
}

@article{dery2026latent,
  title   = {Latent space communication via kv cache alignment},
  author  = {Dery, Lucio M and Yahav, Zohar and Prior, Henry and Feng, Qixuan and Shen, Jiajun and Szlam, Arthur},
  journal = {arXiv preprint arXiv:2601.06123},
  year    = {2026}
}

@misc{dey2026dontlazycompletepenables,
  title         = {Don't be lazy: CompleteP enables compute-efficient deep transformers},
  author        = {Nolan Dey and Bin Claire Zhang and Lorenzo Noci and Mufan Li and Blake Bordelon and Shane Bergsma and Cengiz Pehlevan and Boris Hanin and Joel Hestness},
  year          = {2026},
  eprint        = {2505.01618},
  archiveprefix = {arXiv},
  primaryclass  = {cs.LG},
  url           = {https://arxiv.org/abs/2505.01618}
}

@inproceedings{kwon2023efficient,
  title     = {Efficient memory management for large language model serving with pagedattention},
  author    = {Kwon, Woosuk and Li, Zhuohan and Zhuang, Siyuan and Sheng, Ying and Zheng, Lianmin and Yu, Cody Hao and Gonzalez, Joseph and Zhang, Hao and Stoica, Ion},
  booktitle = {Proceedings of the 29th symposium on operating systems principles},
  pages     = {611--626},
  year      = {2023}
}

@article{zheng2024sglang,
  title   = {Sglang: Efficient execution of structured language model programs},
  author  = {Zheng, Lianmin and Yin, Liangsheng and Xie, Zhiqiang and Sun, Chuyue and Huang, Jeff and Yu, Cody H and Cao, Shiyi and Kozyrakis, Christos and Stoica, Ion and Gonzalez, Joseph E and others},
  journal = {Advances in neural information processing systems},
  volume  = {37},
  pages   = {62557--62583},
  year    = {2024}
}

@article{chen2026see,
  title   = {See What I See, Know What I Think: Dense Latent Communication Across Heterogeneous Agents},
  author  = {Chen, Siyi and Zhang, Xiaoyan and Wu, Meng and Tremblay, Jonathan and Blukis, Valts and Birchfield, Stan and Vidal, Rene and Velasquez, Alvaro and Liu, Sijia and Qu, Qing},
  journal = {arXiv preprint arXiv:2606.13594},
  year    = {2026}
}

@inproceedings{shi2026kvcomm,
  title     = {{KVC}omm: Enabling Efficient {LLM} Communication through Selective {KV} Sharing},
  author    = {Xiangyu Shi and Marco Chiesa and Gerald Q. Maguire Jr. and Dejan Kostic},
  booktitle = {The Fourteenth International Conference on Learning Representations},
  year      = {2026},
  url       = {https://openreview.net/forum?id=F7rUng23nw}
}

@inproceedings{wang2024mmlupro,
  title     = {{MMLU}-Pro: A More Robust and Challenging Multi-Task Language Understanding Benchmark},
  author    = {Yubo Wang and Xueguang Ma and Ge Zhang and Yuansheng Ni and Abhranil Chandra and Shiguang Guo and Weiming Ren and Aaran Arulraj and Xuan He and Ziyan Jiang and Tianle Li and Max Ku and Kai Wang and Alex Zhuang and Rongqi Fan and Xiang Yue and Wenhu Chen},
  booktitle = {The Thirty-eight Conference on Neural Information Processing Systems Datasets and Benchmarks Track},
  year      = {2024},
  url       = {https://openreview.net/forum?id=y10DM6R2r3}
}

@inproceedings{bai2025longbench,
  title     = {Longbench v2: Towards deeper understanding and reasoning on realistic long-context multitasks},
  author    = {Bai, Yushi and Tu, Shangqing and Zhang, Jiajie and Peng, Hao and Wang, Xiaozhi and Lv, Xin and Cao, Shulin and Xu, Jiazheng and Hou, Lei and Dong, Yuxiao and others},
  booktitle = {Proceedings of the 63rd Annual Meeting of the Association for Computational Linguistics (Volume 1: Long Papers)},
  pages     = {3639--3664},
  year      = {2025}
}

@inproceedings{wu2025longmemeval,
  title     = {LongMemEval: Benchmarking Chat Assistants on Long-Term Interactive Memory},
  author    = {Di Wu and Hongwei Wang and Wenhao Yu and Yuwei Zhang and Kai-Wei Chang and Dong Yu},
  booktitle = {The Thirteenth International Conference on Learning Representations},
  year      = {2025},
  url       = {https://openreview.net/forum?id=pZiyCaVuti}
}

@article{zhou2023instruction,
  title   = {Instruction-following evaluation for large language models},
  author  = {Zhou, Jeffrey and Lu, Tianjian and Mishra, Swaroop and Brahma, Siddhartha and Basu, Sujoy and Luan, Yi and Zhou, Denny and Hou, Le},
  journal = {arXiv preprint arXiv:2311.07911},
  year    = {2023}
}

@article{reddy2019coqa,
  title     = {Coqa: A conversational question answering challenge},
  author    = {Reddy, Siva and Chen, Danqi and Manning, Christopher D},
  journal   = {Transactions of the Association for Computational Linguistics},
  volume    = {7},
  pages     = {249--266},
  year      = {2019},
  publisher = {MIT Press One Rogers Street, Cambridge, MA 02142-1209, USA journals-info~…}
}

@article{deepseekv2,
  title   = {{DeepSeek-V2}: A Strong, Economical, and Efficient Mixture-of-Experts Language Model},
  author  = {{DeepSeek-AI}},
  journal = {arXiv preprint arXiv:2405.04434},
  year    = {2024},
  url     = {https://arxiv.org/abs/2405.04434},
  note    = {DeepSeek-V2-Lite is described in Appendix B}
}

@misc{deepseekv2lite,
  title        = {{DeepSeek-V2-Lite}},
  author       = {{DeepSeek-AI}},
  year         = {2024},
  howpublished = {\url{https://huggingface.co/deepseek-ai/DeepSeek-V2-Lite}},
  note         = {Model card, Hugging Face}
}

@article{kimilinear,
  title   = {{Kimi Linear}: An Expressive, Efficient Attention Architecture},
  author  = {{Kimi Team}},
  journal = {arXiv preprint arXiv:2510.26692},
  year    = {2025},
  url     = {https://arxiv.org/abs/2510.26692}
}

@article{kimik2,
  title   = {{Kimi K2}: Open Agentic Intelligence},
  author  = {{Kimi Team}},
  journal = {arXiv preprint arXiv:2507.20534},
  year    = {2025},
  url     = {https://arxiv.org/abs/2507.20534}
}

@inproceedings{artetxe2020cross,
  title     = {On the cross-lingual transferability of monolingual representations},
  author    = {Artetxe, Mikel and Ruder, Sebastian and Yogatama, Dani},
  booktitle = {Proceedings of the 58th annual meeting of the association for computational linguistics},
  pages     = {4623--4637},
  year      = {2020}
}

@inproceedings{zhou2026care,
  title     = {{CARE}: Covariance-Aware and Rank-Enhanced Decomposition for Enabling Multi-Head Latent Attention},
  author    = {Zhongzhu Zhou and Fengxiang Bie and Ziyan Chen and Zhenyu Zhang and Yibo Yang and Junxiong Wang and Ben Athiwaratkun and Xiaoxia Wu and Shuaiwen Leon Song},
  booktitle = {The Fourteenth International Conference on Learning Representations},
  year      = {2026},
  url       = {https://openreview.net/forum?id=DVurf4kGag}
}

@misc{jiang2023mistral7b,
  title         = {Mistral 7B},
  author        = {Albert Q. Jiang and Alexandre Sablayrolles and Arthur Mensch and Chris Bamford and Devendra Singh Chaplot and Diego de las Casas and Florian Bressand and Gianna Lengyel and Guillaume Lample and Lucile Saulnier and Lélio Renard Lavaud and Marie-Anne Lachaux and Pierre Stock and Teven Le Scao and Thibaut Lavril and Thomas Wang and Timothée Lacroix and William El Sayed},
  year          = {2023},
  eprint        = {2310.06825},
  archiveprefix = {arXiv},
  primaryclass  = {cs.CL},
  url           = {https://arxiv.org/abs/2310.06825}
}

@article{team2025gemma,
  title   = {Gemma 3 technical report},
  author  = {Team, Gemma and Kamath, Aishwarya and Ferret, Johan and Pathak, Shreya and Vieillard, Nino and Merhej, Ramona and Perrin, Sarah and Matejovicova, Tatiana and Ram{\'e}, Alexandre and Rivi{\`e}re, Morgane and others},
  journal = {arXiv preprint arXiv:2503.19786},
  year    = {2025}
}

@inproceedings{roy2007effective,
  title        = {The effective rank: A measure of effective dimensionality},
  author       = {Roy, Olivier and Vetterli, Martin},
  booktitle    = {2007 15th European signal processing conference},
  pages        = {606--610},
  year         = {2007},
  organization = {IEEE}
}

@article{hoffmann2022training,
  title   = {Training compute-optimal large language models},
  author  = {Hoffmann, Jordan and Borgeaud, Sebastian and Mensch, Arthur and Buchatskaya, Elena and Cai, Trevor and Rutherford, Eliza and Casas, Diego de Las and Hendricks, Lisa Anne and Welbl, Johannes and Clark, Aidan and others},
  journal = {arXiv preprint arXiv:2203.15556},
  year    = {2022}
}

@article{raecompressive2019,
  author  = {Rae, Jack W and Potapenko, Anna and Jayakumar, Siddhant M and
             Hillier, Chloe and Lillicrap, Timothy P},
  title   = {Compressive Transformers for Long-Range Sequence Modelling},
  journal = {arXiv preprint},
  url     = {https://arxiv.org/abs/1911.05507},
  year    = {2019}
}

@software{mlx2023,
  author  = {Awni Hannun and Jagrit Digani and Angelos Katharopoulos and Ronan Collobert},
  title   = {{MLX}: Efficient and flexible machine learning on Apple silicon},
  url     = {https://github.com/ml-explore},
  version = {0.0},
  year    = {2023}
}

@inproceedings{xiao2024efficient,
  title     = {Efficient streaming language models with attention sinks},
  author    = {Xiao, Guangxuan and Tian, Yuandong and Chen, Beidi and Han, Song and Lewis, Mike},
  booktitle = {International Conference on Learning Representations},
  volume    = {2024},
  pages     = {21875--21895},
  year      = {2024}
}

@inproceedings{nawrot2024dynamic,
  title     = {Dynamic Memory Compression: Retrofitting {LLM}s for Accelerated Inference},
  author    = {Piotr Nawrot and Adrian {\L}a{\'n}cucki and Marcin Chochowski and David Tarjan and Edoardo Ponti},
  booktitle = {Forty-first International Conference on Machine Learning},
  year      = {2024},
  url       = {https://openreview.net/forum?id=tDRYrAkOB7}
}

\appendix
\clearpage
\section{Related work}
\label{app:related}

\paragraph{Cross-model cache transfer.}
Work on making one cache serve two models spans four regimes. \textbf{First,}
the cache is handed over as it is, recomputing the layers
where two variants diverge~\citep{liu2024droidspeak}, sharing the cache-writing layers
across siblings that differ only in a LoRA adapter above
them~\citep{woo2026icarus}, or transmitting only the layers that carry the most
attention~\citep{shi2026kvcomm}. Reuse of this kind needs the two models to share an
architecture, and even then concedes recomputation, restricted model differences, or
partial transfer. \textbf{Second,} one of the two models is trained for the purpose:
\citet{horton2024kv} fit per-layer linear maps from a small auxiliary model's cache into
a base model's, fine-tuning the auxiliary through the frozen base to cut time to first
token. The construction is close to ours; we differ in keeping the source off the shelf,
in crossing families and tokenizers, and in fitting at lengths where a prefill is worth
avoiding, where their prompts stay well under a thousand tokens. \textbf{Third,} a cache is combined with the target's own rather than replacing it:
Cache-to-Cache~\citep{fu2026cachetocache} fuses the two, which beats either model alone
but presumes the target has \emph{already} prefilled the context, so the pair pays two
prefills where a switch should pay one. \textbf{Fourth,} our setting, both models stay frozen and only a translator between
them is fitted. \citet{dery2026latent} map every model's cache into one shared latent
space, a channel that also carries learned skills between models; each adapter is a
transformer as deep as the model it serves, cross-attending over the whole cache, so it
reads the prefix much as a prefill would.
The instance closest to us is a token-wise
two-layer MLP per target layer and key/value group, trained to reconstruct the
target's cache and then with cross-entropy on generation~\citep{chen2026see}, which we
reimplement and match to our translator in parameters and in FLOPs
(Appendix~\ref{app:scratch}). Its two phases mirror our two stages, but the first is
trained rather than solved, the second fits the data rather than the target's own
predictions, and the map is non-linear and confined to matching key/value groups where ours
is a single matrix per layer and side that mixes heads freely. Neither targets the cost of a switch: one map is as expensive as a prefill, and
the other is measured against communicating in text, on prompts of about $180$ tokens,
where the cost constraint that shapes our translator does not bite.

\paragraph{Concurrent work on cross-model cache transfer.}
\citet{heo2026cross} independently study cross-model KV transfer for prefill reuse
within a model family, from the same starting point as ours: the cross-model cache
relation is largely linear, a ridge fit on a small calibration set recovers much of it,
and stripping RoPE before fitting makes the map position-free. \vary{Three}{Four} things separate the
studies. \textbf{(i)} Their method is a closed-form fit, corresponding to our stage
1~\eqref{eq:hat}, so their results confirm the initialiser is sound and our contribution
is what the behavioural stage adds on top.
\textbf{(ii)} Where a closed-form map degrades, they recover it with a nonlinear MLP; we
measure the nonlinear headroom in the fit to be small (Table~\ref{tab:linearity}) and
locate the missing quality in the objective instead, which is why our translator stays
linear. 
\iclronly{\textbf{(iii)} Their pairs all sit inside one family, unequal depth included, where
ours also cross a tokenizer boundary (\S\ref{sec:exp:xtok}), latent caches
(Appendix~\ref{app:mla}) and a pair pretrained from scratch. Their map is affine, where ours is strictly linear;
Appendix~\ref{app:linearity} finds a bias term buys nothing.}
\vary{\textbf{(iii)}}{\textbf{(iv)}} The two
studies count the cost differently. They compare their map's latency against the
target's transformer body over the prefix, ignoring the forward pass the target must
still run over the current token before it can emit anything; we charge that step on
both sides and measure end to end (\S\ref{sec:cost}). Their maps also read $k = 8$ to all
$26$ source layers, since one layer leaves too much of the fit unexplained. That is paid
at every switch: the source's cache must be kept for $k$ layers rather than one, each map
is $k$ times wider, and the target's first token waits on $k$ matrix products per
layer. Ours reads a single source layer on every shape-preserving pair\vary{.}{, and three only
at a tokenizer boundary (\S\ref{sec:exp:xtok}).}

Two follow-ups stay inside the closed-form regime. \citet{qu2026cachebridge} weight the
residual by causal attention, match each target head to one source head, and invert
the source's rotation before fitting. We sweep the first two: attention
weighting cuts the off-the-shelf KL from $0.175$ to $0.064$ but leaves $0.0195$ against
$0.0186$ after stage 2\iclronly{~(Appendix~\ref{app:norms})}, and head matching is our
block-diagonal member (\S\ref{sec:method:family}). Their gains are in building the map, not in
applying it. Building happens once per pair and can take as long as it needs; applying
it happens at every switch, and that is the cost our requirement constrains: cheap
enough to beat a re-prefill, without giving up the target's quality. \citet{li2026universal} frame the
problem as we do but never state what their map is, how it is fitted, or where it acts,
so we cite it rather than compare against it.

\paragraph{Representation alignment and stitching.}
Centered kernel alignment measures how far independently trained networks agree on
their representations~\citep{kornblith2019similarity, nguyen2021do}, and what relates
them is often a simple transformation, in some cases a linear
one~\citep{li2015convergent, huh2024position}. Model stitching turns that measurement
into an intervention: a learned map, classically linear, joins one network's lower
layers to another's upper layers, and the hybrid's accuracy scores their
compatibility~\citep{lenc2015understanding, bansal2021revisiting, csiszarik2021similarity}.
Encoding each point by its similarity to shared anchors avoids fitting the map at
all~\citep{moschella2023relative}. In language models, affine maps between residual
streams transfer trained components across scales~\citep{chen2025transferring}, and a
linear map on token embeddings carries steering vectors between models of different
width~\citep{lee2025shared}. Our translator is fitted in the same spirit but is not a
stitch: its output is not injected once into a forward pass for the later layers to
absorb, it is the context the target attends over, at every layer and every decoded
position.

\paragraph{Cache eviction, compression, and reuse in a single model.}
A large literature reduces or reuses the cache of a \emph{single} model. Some methods
keep less of it, dropping entries during the prompt or after
it~\citep{zhang2023h2o, xiao2024efficient, li2024snapkv, zhao2025smallkv}, or compacting them into
a smaller cache, or a small network, that reproduces the model's attention~\citep{kim2025kvzip, eyuboglu2026cartridges, zweiger2026fast, monteiro2026nectar}. Others store them more cheaply, by
quantisation~\citep{liu2024kivi, hooper2024kvquant, yang2024no} or by learning the
compression itself~\citep{nawrot2024dynamic, lin2025matryoshkakv, kim2025lexico}, or
share them across layers~\citep{brandon2024reducing, yang2024kvsharer, wu2025systematic, wu2025improving, filippova2026stochastic}. A third group reuses what is already stored, across
requests~\citep{kwon2023efficient, zheng2024sglang} or beyond a shared
prefix~\citep{gim2024prompt, yao2025cacheblend, yang2025kvlink}. The model itself can also be
trained so that these methods work better on it~\citep{gelberg2026training}. All of this is
orthogonal to translating between two \emph{different} models, and composes with it
in principle.

\clearpage
\section{Motivating measurements: layer correspondence, linearity and rank}
\label{app:cka_linearity}

Applying a translator $T$ should cost less than the re-prefill it replaces
(\S\ref{sec:setting:translation}), which enforces some structure on the translator map; the
simplest candidate is a single matrix per target layer, applied to a well-chosen source
layer, in parallel over tokens. This appendix reports the measurements behind that choice:
how well that structure is suited to the actual geometry of caches across models.
Fix a layer of the source and of the target, and a prefix token $i$: we denote
$K_{\src,i} \in \R^{d_\src}$ the source's key and $K_{\tgt,i} \in \R^{d_\tgt}$ the target's
key. We ask to what extent a single translator map can predict every $K_{\tgt,i}$ from
$K_{\src,i}$, which source layer to choose for each target layer, and whether the
dependency between those vectors is approximately linear. Layer correspondence answers the
second question (\S\ref{app:cka}); linearity and effective rank answer the other two
(\S\ref{app:linearity}).

\subsection{Layer correspondence across every pair}
\label{app:cka}

\begin{figure}[htbp]
\centering
\includegraphics[width=0.66\linewidth]{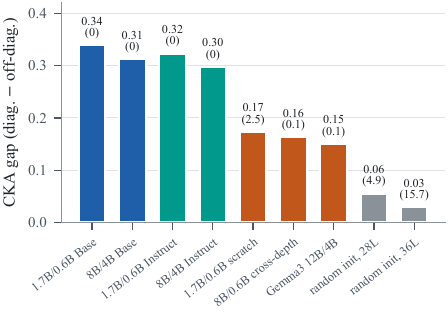}
\caption{\textbf{Diagonal contrast of the per-layer CKA grid, every pair.}
Unweighted linear CKA between source layer $\ell$ and target layer $\ell'$,
summarised as the gap between the diagonal ($\ell=\ell'$) and the off-diagonal mean;
the parenthesised number is the mean offset of the per-row argmax from the diagonal,
in layers. The released equal-depth Qwen3 pairs, Base and Instruct alike, have a
strong on-diagonal peak at offset $0$, which licenses the identity assignment. The
from-scratch, cross-layer and Gemma-3 pairs keep about half that contrast and, except
for the from-scratch pair, sit near the relative-depth diagonal; they are the pairs
for which the source layer is selected rather than assumed. Random initialisation
gives essentially no structure.}
\label{app:fig:cka}
\end{figure}

We measure layer correspondence between the source and the target with linear centred
kernel alignment (CKA)~\citep{kornblith2019similarity}, computed between every source and
target layer pair. On caches centred over tokens it is
$\mathrm{CKA}(X,Y) = \lVert X^\top Y \rVert_F^2 / (\lVert X^\top X \rVert_F\,
\lVert Y^\top Y \rVert_F)$.
Figure~\ref{fig:cka} shows three of the per-layer CKA grids in full.
Figure~\ref{app:fig:cka} summarises all of them, each grid reduced to two numbers:
the gap between its diagonal and its off-diagonal mean, and how far the per-row
argmax sits from the diagonal.

That summary splits the pairs into three regimes. The released shape-preserving Qwen3 pairs are
sharply peaked on the diagonal, exceeding the off-diagonal mean by $0.30$--$0.34$ with the
per-row argmax exactly on it, on Base and Instruct checkpoints alike. The pairs that share
no lineage or no shape keep about half that contrast: a $0.6$B/$1.7$B pair we pretrained
from scratch, sharing no checkpoint and no initialisation, sits at $0.17$, the Gemma-3
pair~\citep{team2025gemma} at $0.15$ and an unequal-depth Qwen3 pair at $0.16$; only the
last two are close to registered, the from-scratch pair's argmax drifting $2.5$ layers.
The randomly initialised controls, built on the same two architectures, reach $0.03$ and
$0.06$ at offsets of $15.7$ and $4.9$ layers, which is what an absent correspondence looks
like under this measurement.

A correspondence of layers by relative depth therefore appears during training, and is not
limited to the Qwen3 architecture and training recipe. It is what licenses the identity
layer assignment on the equal-depth pairs and a relative-depth window elsewhere
(\S\ref{sec:method:family}).

\iclronly{%
\begin{figure}[htbp]
\centering
\includegraphics[width=\linewidth]{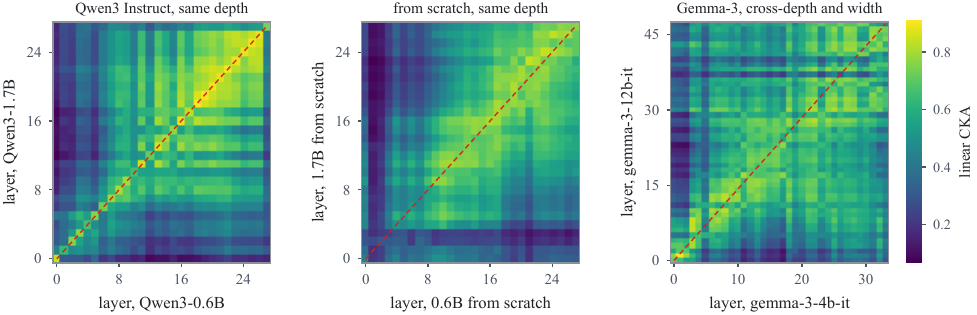}
\caption{\textbf{Layer-to-layer cache correspondence.} Linear CKA between
every layer of the large (rows) and small (columns) model, on post-$k$-norm keys over
$16$k tokens; shared colour scale, dashed line at equal relative depth. \emph{Left:}
pretrained Qwen3 Instruct pair of equal depth. \emph{Middle:} $1.7$B/$0.6$B pair
pretrained from scratch, sharing no checkpoint. \emph{Right:} pretrained Gemma-3 pair
differing in depth and key/value width. Figure~\ref{app:fig:cka} summarises every other
pair we measured, including two randomly initialised controls.}
\label{fig:cka}
\end{figure}
}

\subsection{Cache linearity and rank, per layer}
\label{app:linearity}

Linear CKA says which layers have similar geometries: two point clouds have a linear CKA
close to $1$ if they approximately differ by an orthogonal transformation, which is a
stronger condition than linear dependency. To measure how linearly related source and
target layers are, we fit a no-bias least-squares map $K_{\src,i} \mapsto T_K^{\ell}
K_{\src,i}$ per layer on held-out tokens and compute the fraction $R^2_{\mathrm{lin}}$ of
the target's cache variance it explains, against the $R^2_{\mathrm{nl}}$ of a two-layer MLP
on the same data. Table~\ref{tab:linearity} reports both; Table~\ref{tab:affine} refits the
linear map with an intercept, which is how the nonlinear headroom of
Table~\ref{tab:linearity} is attributed to curvature rather than to a mean shift; and
Table~\ref{tab:erank} reports how much of its nominal width each cache actually occupies.

The linear map reaches $0.90$--$0.99$ of what the MLP does
($\rho = R^2_{\mathrm{lin}}/R^2_{\mathrm{nl}}$, the MLP adding at most $0.073$), so a
nonlinear translator should buy little, which \S\ref{sec:exp} confirms downstream. The
unexplained residual is substantial: $1 - R^2_{\mathrm{nl}} \approx 0.2$--$0.4$, with values
harder than keys and the from-scratch pair harder than the released checkpoints, orderings
that recur downstream.

Caches also have low per-layer effective rank~\citep{roy2007effective}: $13$--$177$ against
$d = 1024$ (Table~\ref{tab:erank}), with a single principal direction carrying up to half
the variance in the first layers. Randomly initialised models of the same architecture
behave very differently: effective ranks of $365$--$830$, near-isotropic, top direction
below $4\%$. Training therefore concentrates the cache onto a low-dimensional, strongly
anisotropic subspace, which can reduce the number of samples needed to estimate the
translator accurately along the directions that dominate the cache distribution.

\begin{table}[htbp]
\centering
\footnotesize
\setlength{\tabcolsep}{5pt}
\caption{\textbf{Ablation: the impact of translating with an MLP instead of a linear map.} Per-layer linear fit of the target's cache from the source's, averaged over
layers, on held-out tokens ($L \to S$; pre-normalisation capture point, which leaves
values unchanged; $d=1024$ per side). $R^2_{\mathrm{lin}}$ is the no-bias least-squares
fit, the estimator of the closed-form translator; $R^2_{\mathrm{nl}}$ a two-layer MLP on the same
data;
$\rho = R^2_{\mathrm{lin}}/R^2_{\mathrm{nl}}$ the fraction of recoverable signal the linear map
already captures.
The $S \to L$ direction agrees to within $0.018$ on keys and
$0.037$ on values.}
\label{tab:linearity}
\begin{tabular}{llrrrr}
\toprule
Pair & side & $R^2_{\mathrm{lin}}$ & $R^2_{\mathrm{nl}}$ & nonlin.\ extra & $\rho$  \\
\midrule
\multirow{2}{*}{Qwen3 8B/4B}        & K & $0.815$ & $0.825$ & $0.010$ & $0.989$ \\
                                    & V & $0.693$ & $0.708$ & $0.015$ & $0.980$ \\
\midrule
\multirow{2}{*}{Qwen3 1.7B/0.6B}    & K & $0.766$ & $0.794$ & $0.028$ & $0.965$  \\
                                    & V & $0.658$ & $0.697$ & $0.039$ & $0.945$ \\
\midrule
\multirow{2}{*}{from scratch 1.7B/0.6B} & K & $0.752$ & $0.805$ & $0.053$ & $0.935$ \\
                                    & V & $0.651$ & $0.724$ & $0.073$ & $0.903$ \\
\bottomrule
\end{tabular}
\end{table}

\begin{table}[htbp]
\centering
\footnotesize
\setlength{\tabcolsep}{5pt}
\caption{\textbf{Ablation: the impact of adding an intercept to the translators.} We compare the per-layer fit of linear translators with and without bias, at the pre-normalisation capture point. The two $R^2$ columns are layer averages on held-out tokens.
On both Qwen3 pairs the bias is worth at most $0.001$ of $R^2$ at any
single layer, an order of magnitude below what the MLP finds (Table~\ref{tab:linearity}). The from-scratch pair has more room for an offset:
its $S \to L$ keys gain $0.012$ on average.}
\label{tab:affine}
\begin{tabular}{@{}lllrr@{}}
\toprule
 & & & \multicolumn{2}{c}{$R^2$, layer average}  \\
\cmidrule(lr){4-5} 
Pair & Dir. & side & no bias & affine \\
\midrule
\multirow{4}{*}{Qwen3 8B/4B} & \multirow{2}{*}{$L \to S$} & K & $0.815$ & $0.815$ \\
 &  & V & $0.693$ & $0.693$\\
\cmidrule(l){2-5}
 & \multirow{2}{*}{$S \to L$} & K & $0.808$ & $0.808$ \\
 &  & V & $0.676$ & $0.676$ \\
\midrule
\multirow{4}{*}{Qwen3 1.7B/0.6B} & \multirow{2}{*}{$L \to S$} & K & $0.766$ & $0.766$  \\
 &  & V & $0.658$ & $0.658$  \\
\cmidrule(l){2-5}
 & \multirow{2}{*}{$S \to L$} & K & $0.753$ & $0.754$  \\
 &  & V & $0.623$ & $0.623$  \\
\midrule
\multirow{4}{*}{from scratch 1.7B/0.6B} & \multirow{2}{*}{$L \to S$} & K & $0.752$ & $0.755$ \\
 &  & V & $0.651$ & $0.651$  \\
\cmidrule(l){2-5}
 & \multirow{2}{*}{$S \to L$} & K & $0.757$ & $0.769$  \\
 &  & V & $0.674$ & $0.675$ \\
\bottomrule
\end{tabular}
\end{table}

\begin{table}[htbp]
\centering
\footnotesize
\setlength{\tabcolsep}{7pt}
\caption{Effective rank of the per-layer cache covariance (of $d=1024$), at the first
layer, mid-stack and last layer, with the variance fraction carried by the top
direction. Trained caches are low-rank and anisotropic; randomly initialised models of
the same architecture are neither, which is the null for Figure~\ref{fig:cka}.}
\label{tab:erank}
\begin{tabular}{lrr}
\toprule
Model & effective rank (first $\to$ mid $\to$ last) & top-1 variance fraction (first $\to$ last) \\
\midrule
Qwen3-1.7B, trained            & $59 \to 15 \to 177$   & $0.31 \to 0.10$ \\
Qwen3-1.7B, \emph{random init} & $365 \to 583 \to 558$ & $0.04 \to 0.01$ \\
Qwen3-8B, trained              & $24 \to 48 \to 135$   & $0.52 \to 0.10$ \\
Qwen3-8B, \emph{random init}   & $401 \to 821 \to 830$ & $0.04 \to 0.00$ \\
from-scratch 1.7B, trained     & $13 \to 113 \to 157$  & $0.51 \to 0.10$ \\
\bottomrule
\end{tabular}
\end{table}

\paragraph{Protocol.}
For a fixed pair, direction and capture point we forward both models over a shared
Nemotron token set ($512$ documents of $256$ tokens, so $131$k aligned positions)
and collect, for every layer and every prefix token, the aligned pair $(x,y)$ = (source
cache, target cache). The split is by \emph{document}, $80/20$ into fit and test, so no
scored position shares a document with a fitted one; the MLP's early-stopping set is a
further tenth of the fitting tokens.
$R^2_{\mathrm{lin}}$ is the held-out fit of the no-bias least-squares map
of~\eqref{eq:hat}, with the per-dimension training mean as the
$\mathrm{SS}_{\mathrm{tot}}$ baseline. It differs from the deployed closed-form
initialisation in its conditioning: the diagnostic adds a small ridge to the normal
equations ($10^{-3}$ times the mean diagonal of the source Gram) where the deployed
solve truncates the pseudo-inverse instead. Both act on the tail of the source spectrum
and neither disturbs the leading directions, but they are different regularisers, so
$R^2_{\mathrm{lin}}$ is the fit quality of that estimator rather than of the shipped
matrix exactly.
$R^2_{\mathrm{nl}}$ is an MLP on the same split, early-stopped on validation with the
linear solution as its initial best checkpoint, so $R^2_{\mathrm{nl}} \ge R^2_{\mathrm{lin}}$ by
construction and is a \emph{lower} bound on what is recoverable from $x$. The variance
splits into linear-explained $R^2_{\mathrm{lin}}$, nonlinear-extra
$R^2_{\mathrm{nl}}-R^2_{\mathrm{lin}}$, and irreducible $1-R^2_{\mathrm{nl}}$.

\paragraph{What the per-layer curves show.}
The layer averages of Table~\ref{tab:linearity} hide a shape common to all three pairs,
though not the monotone one might expect. $R^2$ peaks early (at layer $0$ or $1$ on
both Qwen3 pairs, at layer $4$ on the from-scratch pair, which rises before it turns
over), then drifts down and falls off sharply at the last layer, which is the worst
layer in nine of the twelve pair $\times$ side $\times$ direction cells ($8$B/$4$B
values run $0.828$ at layer $1$ to $0.515$ at layer $35$). How much of that is trend
rather than the final-layer cliff depends on the pair: Spearman$(\text{layer}, R^2)$ is
null on $1.7$B/$0.6$B ($-0.04$ to $-0.26$, $p \ge 0.19$), moderate on $8$B/$4$B
($-0.34$ to $-0.42$, $p \le 0.04$) and strong on the from-scratch pair ($-0.51$ to
$-0.79$, $p \le 0.005$). Values sit below keys at every layer of both Qwen3 pairs and in
both directions; the from-scratch pair is the exception, with values above keys at layer
$0$ in $L \to S$ (by $0.068$) and at three early layers in $S \to L$. The
nonlinear-extra term is small but not uniform: it concentrates in the first layers on
every pair, peaking at $0.074$ against a layer average of $0.015$ for $8$B/$4$B values
and at $0.175$ against $0.073$ for from-scratch values, so what nonlinearity there is to
buy sits at the bottom of the stack. That is also where the bias term of
Table~\ref{tab:affine} lives on the from-scratch pair: its largest single-layer gains are
at layers $2$ and $5$, inside the same early band, so on that pair part of the early
nonlinear headroom is an offset rather than curvature. Both Qwen3 pairs show no such band
at any layer.

\paragraph{Depth $\times$ position.}
Refitting within position buckets on long documents (PG-19~\citep{raecompressive2019}, $256 \to 16$k) gives, for
Qwen3 $8$B/$4$B pre-normalisation keys, layer-averaged $R^2$ of $0.806$, $0.815$,
$0.814$, $0.816$ over buckets $[0,256)$, $[256,1\text{k})$, $[1\text{k},4\text{k})$,
$[4\text{k},16\text{k})$: flat to slightly rising, with the paired sign test across
layers null everywhere ($p = 0.19$--$0.87$). The depth slope, however, steepens with
position: Spearman$(\text{layer}, R^2)$ moves from $-0.29$ in the first bucket to
$-0.62$ in the last. Splitting the stack in thirds, the position effect is $-0.031$
(early), $-0.010$ (mid), $+0.032$ (late) for values. Long context helps the first two
thirds and hurts the last third, and the largest single-layer losses are all deep and
in values. That deep loss is irreducible: at $16$k the MLP adds $0.000$ over the linear
fit. Finally, the isolated low-CKA ``dead'' layers that appear in pre-normalisation
keys at short context (layers $13$, $15$--$18$ for $8$B/$4$B) largely dissolve past
${\sim}1$k tokens (their $R^2$ dip shrinks from $+0.102$ to $+0.026$ and their CKA dip
from $+0.351$ to $+0.067$), confirming by an independent route that they are a
pre-normalisation scaling artefact rather than a genuine cross-model misalignment.

\clearpage
\section{Training data and training recipe}
\label{app:data}
\suppressfloats[t]

\begin{figure}[t]
\centering
\includegraphics[width=\linewidth, trim=3.3pt 25.8pt 26.6pt 2.6pt, clip]%
  {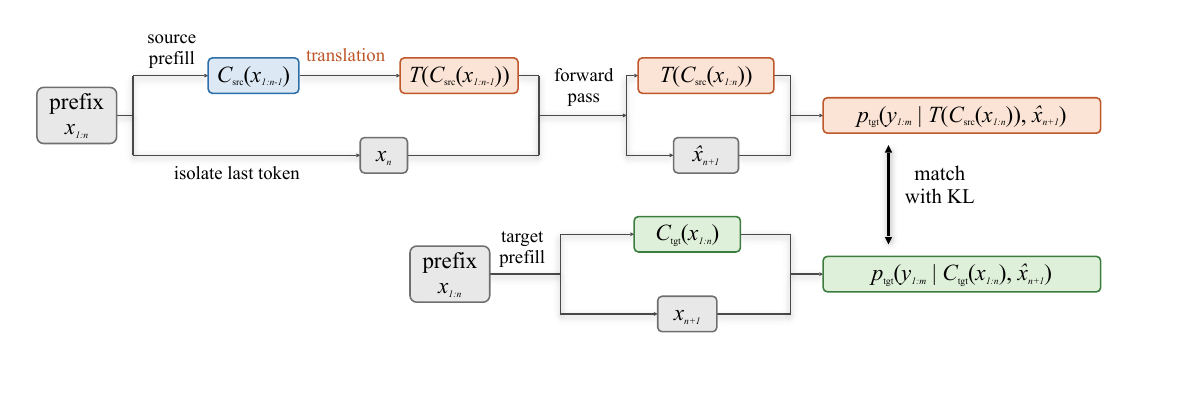}
\caption{\textbf{Architecture and training of KV-Lingo translators.} Both rows start
from the same prefix $x_{1:n}$ and end at a distribution over the same continuation
$y_{1:m}$; Stage~2 of \S\ref{sec:method:training} asks the two to agree. Top: the source
prefills $x_{1:n-1}$, the translator $T$ maps that cache into the target's format, and one
forward pass of the target on the held-back token $x_n$ completes the cache and emits
$\hat{x}_{n+1}$. Bottom: the target prefills the whole prefix itself. The cache is the only
thing that differs between the two rows. Source and target are frozen; the KL is
backpropagated through the target into $T$, which carries all the trainable parameters.}
\label{fig:kvlingo_computation}
\end{figure}

Figure~\ref{fig:kvlingo_computation} summarises how a KV-Lingo translator is applied and
trained; this section details the data it is trained on.

\paragraph{The training stream.}
We train all our translators using the same general-knowledge multi-turn conversation dataset:
Nemotron-SFT-Instruction-Following-Chat-v2
~\citep{nvidia2025nemotron3nanoopen}.
In our experiments, the training dataset is disjoint from every downstream benchmark:
our translators are universal for text data, in the sense that we train one translator per model pair once and for all, and this translator generalises to several domains and languages.

Our training samples all take the form of a (prefix, continuation) pair.
The prefix is the beginning of a conversation between a human and an assistant, containing one or several turns, and ending at the start of an assistant turn (after the last boilerplate token of the template), just before the first token of the reply.
We alternate between reasoning-on and reasoning-off samples, structured as follows.
For \emph{reasoning-on} samples, the reasoning trace is part of the continuation. For \emph{reasoning-off} samples, the continuation contains no reasoning trace, and an empty reasoning marker closes the prefix instead.

\paragraph{Prefix lengths.}
To train the translator on prefixes of varied and controlled lengths, we build a modified training dataset from Nemotron-SFT-Instruction-Following-Chat-v2, whose conversations are too short on their own to train long-context translators.
It is made of:
\begin{itemize}
    \item $50\%$ natural samples drawn from Nemotron-SFT-Instruction-Following-Chat-v2, half of which have a reasoning-off continuation and the other half a reasoning-on one (reasoning traces are removed from the prefix). The prefix length of these samples ranges from 10 to 10k tokens, and is approximately uniformly distributed over that range.
    \item $50\%$ samples obtained by concatenating natural samples, from which one question and its answer are removed at random. The question is placed at the end of the prefix, together with the boilerplate tokens that open the assistant turn, while the answer to the question is placed in the continuation. The concatenations are built so as to have prefix lengths between 12k and 16k tokens.
\end{itemize}
Figure~\ref{fig:prefixhist} shows an estimation of the resulting distribution of prefix lengths.
Continuation lengths are capped at 2048 tokens, with an average length of $750$ tokens reasoning-off and $1$k tokens reasoning-on.

\paragraph{How a sample is built.}
A training record is a multi-turn conversation. We pick one assistant turn $t$ at random,
render the conversation with the target family's chat template up to and including the
header that opens turn $t$, and drop every later turn. That render is the prefix; turn
$t$'s own answer is the continuation. The prefix therefore always ends inside the
template's boilerplate rather than at a content token,
which is exactly where a served conversation
sits when the model is about to speak, and where a switch of model would happen.

\paragraph{Where the reasoning trace goes.}
Whether a sample reasons is a property of the record, not a setting: the loader detects a
\verb!<think>! block in turn $t$ and renders the sample accordingly. If turn $t$ reasons,
the prefix ends at the bare assistant header and the continuation opens with the real
\verb!<think>! block, so the trace is scored. If it does not, Qwen3's template emits
an empty \verb!<think>\n\n</think>\n\n! placeholder, and we keep that placeholder at the
end of the \emph{prefix}, where inference puts it, prefilled by whoever holds the cache,
rather than at the start of the continuation, where it would be scored as four fully
predictable tokens. Independently of $t$, the Qwen3 template strips the reasoning block
from every assistant turn before the last, so earlier turns appear as plain chat in the
prefix. In the packed half of the mixture above the geometry is the same, with the
evaluated conversation's final user query lifted to the very end of the prefix and its answer
kept as the continuation.

\paragraph{Template tokens.}
The boilerplate of the three families that appear in the paper is as follows; 
in Qwen3 and Gemma a newline follows each delimiter and each generation prompt, and in Mistral a single space follows the opening delimiter.
\begin{itemize}
  \item \textbf{Qwen3.} Turns are wrapped in \verb!<|im_start|>{role}! $\dots$
    \verb!<|im_end|>!, with roles \texttt{user} / \texttt{assistant}; the generation
    prompt is \verb!<|im_start|>assistant!, followed on a reasoning-off sample by the
    placeholder \verb!<think>\n\n</think>\n\n!.
  \item \textbf{Gemma.} Turns are wrapped in \verb!<start_of_turn>{role}! $\dots$
    \verb!<end_of_turn>!, with roles \texttt{user} / \texttt{model}, and the generation
    prompt is \verb!<start_of_turn>model!. There is no thinking placeholder.
  \item \textbf{Mistral.} User turns are wrapped in \verb![INST]! $\dots$ \verb![/INST]!,
    and \verb![/INST]! is itself the generation prompt. There is no thinking placeholder.
\end{itemize}
Only Qwen3 exposes a thinking placeholder; Gemma's template has no such flag and does not
strip reasoning from earlier turns, and it refuses two consecutive turns in the same role.
\begin{iclrblock}
For the cross-tokenizer pair of \S\ref{sec:exp:xtok}, each model renders the conversation
with its \emph{own} template and only content tokens are put in correspondence:
\verb!<|im_start|>assistant! and \verb![/INST]! have no shared referent across the two
families.
\end{iclrblock}

\paragraph{A rendered sample.}
Below is the shape of one reasoning-off Qwen3 sample. The control tokens are verbatim; the
conversation content is abridged and stands in for a real record.

\begin{quote}\small
\emph{Prefix} (translated by $T$, never decoded):
\begin{verbatim}
<|im_start|>user
Give me three uses for baking soda.<|im_end|>
<|im_start|>assistant
Cleaning, deodorising, and leavening.<|im_end|>
<|im_start|>user
Which of those needs an acid?<|im_end|>
<|im_start|>assistant
<think>

</think>

\end{verbatim}
\emph{Continuation} (decoded by the target, scored against the target's own
distribution):
\begin{verbatim}
Leavening: it needs an acid to release CO2.<|im_end|>
\end{verbatim}
\end{quote}

On a reasoning-on sample the prefix stops one line earlier, after
\verb!<|im_start|>assistant\n!, and the continuation begins with the record's own
\verb!<think>! block.

\clearpage
\section{Evaluation data}
\label{app:evaldata}
\suppressfloats[t]

Table~\ref{tab:evalsets} lists the evaluation suite: the ability each benchmark probes, the translated prefix length, the number of scored items, the reasoning mode and generation budget, and the metric.
MMLU-Pro and ARC-Challenge are scored by generate-and-parse, and MT-Bench-101 on one
randomly chosen turn per conversation, with the earlier turns as context.
Each translator is evaluated on a subset of this evaluation suite.

\paragraph{Decoding and judging.}
Every model decodes greedily, in both reasoning modes, up to the generation budget of
Table~\ref{tab:evalsets}. The judge, run locally, is Qwen3.6-27B. It also decodes
greedily, with reasoning off, and sees the answer with any reasoning trace removed.
MT-Bench-101 uses the benchmark's
own per-task prompts and $1$--$10$ scale~\citep{bai2024mt}; since we judge a single turn per
conversation, we report the mean over conversations of that turn's rating, not the
benchmark's minimum over turns. LongMemEval uses the benchmark's own answer-check
templates~\citep{wu2025longmemeval}, with the closing \emph{Answer yes or no only} replaced
by a request for a bracketed yes-or-no verdict, so that a malformed reply cannot be read as
a wrong answer. RepLiQA answers are judged against the reference answer on the answerable
questions, under one of two rubrics: binary correctness, or the fraction of the reference's
key facts that the answer conveys, the judge first splitting the reference into those facts.
The shape-preserving Qwen3 results report the fact coverage, and the cross-tokenizer
results the binary correctness.

\begin{table}[!htbp]
\centering
\footnotesize
\setlength{\tabcolsep}{3.5pt}
\caption{The evaluation suite. \emph{Reasoning} is the mode the reported numbers are
scored in, with the generation budget (max new tokens) per mode in parentheses;
teacher-forced rows need none. \emph{Prefix} is the translated prefix length in tokens:
a range or ladder where the benchmark fixes it, a cap where it does not.
The language-modelling row is scored on held-out conversations from the
training mixture.}
\label{tab:evalsets}
\begin{tabular}{llllll}
\toprule
Benchmark & Ability & Prefix & $n_\textbf{samples}$ & Reasoning (budget) & Metric \\
\midrule
Nemotron chat  & LM              & $10$--$16$k          & held-out          & off/on ($-$)        & cont.\ PPL, KL \\
MMLU-Pro       & knowledge MC    & $\le 3$k             & $1{,}400$         & off/on ($1$k/$8$k)  & accuracy (macro) \\
ARC-Challenge  & science MC      & $\le 3$k             & $1{,}172$         & off/on ($1$k/$8$k)  & accuracy \\
GSM8K          & arithmetic      & $\le 1$k             & $256$             & on ($2$k)           & EM flexible \\
IFEval         & constraints     & $\le 2$k             & $541$             & off ($1$k)          & prompt/instr.\ acc. \\
CoQA           & grounded QA     & $\le 1$k             & $256$             & off ($128$)         & F1, gold-in-ans. \\
QuAC           & dialogue QA     & $0.7$--$1.1$k        & $256$ ($8$ turns) & off ($128$)         & F1 \\
RepLiQA        & unseen-doc QA   & $\le 3$k             & $512$ ($358$a)    & off/on ($512$/$2$k) & judge correct. \\
XQuAD          & QA, $11$ langs  & $\le 2$k             & $2{,}816$         & off ($128$)         & token-F1, DiD \\
MT-Bench-101   & multi-turn chat & $103$ (p50)          & $256$             & off/on ($2$k)       & judge, $1$--$10$ \\
MultiChallenge & multi-turn instr. & $1$--$5$k          & $273$             & off ($2$k)          & judge pass rate \\
RULER (NIAH)   & NIAH retrieval  & $4$/$8$/$16$/$32$k   & $400$/len.        & off/on ($2$k)    & recall \\
LongMemEval    & session memory  & oracle, $8$--$32$k   & $389$ common      & off/on ($2$k)       & judge accuracy \\
LongBench-v2   & long documents  & $\le 32$k            & $116$/$395$       & off/on ($512$/$4$k) & accuracy \\
\bottomrule
\end{tabular}
\\[5pt]
\begin{minipage}{\linewidth}
\scriptsize

\emph{Relevant model pairs for each dataset.} LongBench-v2 is not reported on Qwen3-0.6B$\leftrightarrow$1.7B
(either reasoning mode) nor on Gemma-4-E2B$\leftrightarrow$E4B reasoning-off: the larger model of
the pair is near or below the $25\%$ chance floor ($0.319$/$0.250$ and $0.208$
respectively), so no converted-vs-native delta is readable there. RULER reasoning-off on
Qwen3-4B$\leftrightarrow$8B is at ceiling at $4$/$8$/$16$k (native recall
$0.998$--$1.000$); only the $32$k rung separates conditions on that pair.

\end{minipage}
\end{table}

\clearpage
\section{Training protocol}
\label{app:protocol}
\suppressfloats[t]

\paragraph{Models.}
We use the following released Hugging Face checkpoints:
\begin{itemize}
  \item \texttt{Qwen/Qwen3-0.6B}, \texttt{Qwen/Qwen3-1.7B}, \texttt{Qwen/Qwen3-4B} and
    \texttt{Qwen/Qwen3-8B}~\citep{yang2025qwen3};
  \item \texttt{Qwen/Qwen3-0.6B-Base} and \texttt{Qwen/Qwen3-1.7B-Base}, in
    Appendix~\ref{app:scratch} only;
  \item \texttt{google/gemma-3-4b-it} and \texttt{google/gemma-3-12b-it}~\citep{team2025gemma};
  \item \texttt{google/gemma-4-E2B-it} and \texttt{google/gemma-4-E4B-it}~\citep{team2026gemma};
  \item \texttt{mistralai/Mistral-7B-Instruct-v0.3}~\citep{jiang2023mistral7b}.
\end{itemize}
The four Qwen3 models are the hybrid-reasoning releases. The MoE pair of Appendix~\ref{app:moe} adds Qwen3-30B-A3B, the MLA models of
Appendix~\ref{app:mla} are converted by us from Qwen3-4B and Qwen3-8B as described there,
and the from-scratch pair of Appendix~\ref{app:scratch} is pretrained by us with the recipe
given there.

\paragraph{Hardware and precision.}
Every translator is trained on NVIDIA H100 80GB GPUs. Source and target stay frozen in
\texttt{bf16} throughout; the translator's parameters are held in \texttt{fp32}. Most
translators are trained on a single GPU, accumulating gradients over the $8$ conversations
of a step; the cross-layer head-mixing translators with $\nu \ge 2$ and all head-wise
cross-layer translators use $8$-way data parallelism instead, one conversation per GPU,
at the same effective batch of $8$.

\paragraph{The training procedure.}
Both stages of the translator fitting procedure (\S\ref{sec:method:training}) read the same stream of (prefix, continuation)
pairs, one conversation at a time: prefix lengths span three orders of magnitude, so
samples are never padded together into a batch.

\paragraph{Stage 1.}
The second-moment statistics of~\eqref{eq:hat} are accumulated in \texttt{fp64} over $400$
conversations of the training stream (${\approx}2.8$M prefix positions), and the
per-layer normal equations are then solved once, offline, by symmetric eigendecomposition, with
eigenvalues below a relative tolerance of $10^{-8}$ treated as zero. 
Note that during this phase, the continuations are not used.
The resulting maps are both the initialisation of stage 2 and the closed-form
baseline we carry through every table.

\paragraph{Stage 2.}
Starting from those maps, we optimise~\eqref{eq:kl} for $5000$ steps using AdamW with an effective batch of $8$ conversations,
cosine schedule with $5\%$ warmup, no weight decay, gradient clipping at $1.0$.
Prefixes are capped at $16{,}384$ tokens and continuations at $2048$ (Appendix~\ref{app:data}).
For the Base pairs of Appendix~\ref{app:scratch}, every sample is instead a $256$-token
prefix followed by a $256$-token continuation.
The learning rate is the only tuned hyperparameter. For each (pair, direction, translator),
we train at five learning rates spaced half a decade apart, centred on the best rate of an
earlier run, with one seed, and extend the window whenever the best rate sits on its edge,
so that every selected learning rate is interior to the rates tried; over all cells the
rates tried span $3{\times}10^{-7}$ to $3{\times}10^{-2}$. We select the rate with the lowest
final validation KL on $64$ held-out conversations drawn from the same dataset and disjoint
from training, and train the other seeds at that rate. Table~\ref{tab:lrs} lists the
selected rates.

\begin{table}[!htbp]
\centering
\caption{\textbf{Selected stage-2 learning rates} of the linear translators, by pair,
direction and layer assignment. On the shape-preserving pairs every target layer reads its
counterpart, so the two $\nu = 1$ columns share one rate; on the cross-layer pairs $\nu$ is
the number of source layers per target layer. \emph{Depth}: relative-depth match.
\emph{Selected}: the $R^2$ assignment on Qwen3, which at $\nu = 1$ coincides with greedy
match up to the ridge penalty, and greedy match on Gemma. The head-mixing columns at
$\nu \ge 2$ use the $R^2$ assignment and the head-wise columns greedy match, as in
Appendix~\ref{app:xdepth}. The head-mixing rates at $\nu \ge 2$ were selected at seed $56$,
all others at seed $42$. The MLP rates are given in the text, and those of the MoE pair in
Appendix~\ref{app:moe}.}
\label{tab:lrs}
\scriptsize
\setlength{\tabcolsep}{3pt}
\begin{tabular}{@{}ll cccc cccc@{}}
\toprule
 & & \multicolumn{4}{c}{head-mixing} & \multicolumn{4}{c}{head-wise} \\
\cmidrule(lr){3-6} \cmidrule(lr){7-10}
Pair & Direction & $\nu{=}1$, depth & $\nu{=}1$, selected & $\nu{=}2$ & $\nu{=}4$
 & $\nu{=}1$ & $\nu{=}2$ & $\nu{=}4$ & $\nu{=}8$ \\
\midrule
Qwen3-0.6B/1.7B & $L \to S$ & \multicolumn{2}{c}{$3{\times}10^{-5}$} & & & $10^{-4}$ & & & \\
Qwen3-0.6B/1.7B & $S \to L$ & \multicolumn{2}{c}{$10^{-4}$} & & & $10^{-3}$ & & & \\
Qwen3-4B/8B & $L \to S$ & \multicolumn{2}{c}{$3{\times}10^{-5}$} & & & $10^{-4}$ & & & \\
Qwen3-4B/8B & $S \to L$ & \multicolumn{2}{c}{$3{\times}10^{-5}$} & & & $3{\times}10^{-4}$ & & & \\
\midrule
Qwen3-0.6B/8B & $L \to S$ & $10^{-4}$ & $3{\times}10^{-4}$ & $10^{-4}$ & $3{\times}10^{-5}$
 & $3{\times}10^{-3}$ & $10^{-3}$ & $3{\times}10^{-4}$ & $3{\times}10^{-4}$ \\
Qwen3-0.6B/8B & $S \to L$ & $3{\times}10^{-5}$ & $3{\times}10^{-5}$ & $10^{-5}$ & $3{\times}10^{-6}$
 & $3{\times}10^{-4}$ & $10^{-4}$ & $10^{-4}$ & $10^{-5}$ \\
Qwen3-1.7B/4B & $L \to S$ & $3{\times}10^{-4}$ & $3{\times}10^{-4}$ & $10^{-4}$ & $10^{-4}$
 & $3{\times}10^{-3}$ & $10^{-3}$ & $10^{-3}$ & $3{\times}10^{-4}$ \\
Qwen3-1.7B/4B & $S \to L$ & $10^{-5}$ & $10^{-5}$ & $3{\times}10^{-6}$ & $3{\times}10^{-6}$
 & $10^{-4}$ & $10^{-4}$ & $3{\times}10^{-5}$ & $10^{-5}$ \\
\midrule
Gemma-3-4B/12B & both & $3{\times}10^{-5}$ & $3{\times}10^{-5}$ & & & & & & \\
Gemma-4-E2B/E4B & both & $10^{-4}$ & $10^{-4}$ & & & & & & \\
\bottomrule
\end{tabular}
\end{table}

\paragraph{MLP translators.}
The MLP baselines read $\nu = 1$ source layer per target layer and replace each linear
block by a two-layer MLP $z \mapsto W_2\,\mathrm{GELU}(W_1 z + b_1) + b_2$, with no skip
connection. The head-wise MLP of~\citet{chen2026see} has one such MLP per target layer, KV
head and side (keys or values), $128 \to 2048 \to 128$ on Qwen3, and scales its output by a
learned sigmoid gate per layer and head, initialised at $\sigma(4) \approx 0.98$. The
head-mixing MLP of the cross-layer pairs has one gated MLP per target layer and side,
spanning all heads, $1024 \to 2048 \to 1024$, which matches the head-wise MLP in parameters
and FLOPs; the one on Qwen3-4B/8B is ungated and $1024 \to 4096 \to 1024$.
With no closed-form solution available, stage 1 trains the MLP from random initialisation
with AdamW on the cache-reconstruction error of~\eqref{eq:hat}, for $5000$ steps with the
batch and schedule of stage 2; stage 2 is then the recipe above.
For the head-wise MLP, the stage-1 learning rate is swept over
$\{10^{-3}, 3{\times}10^{-3}, 10^{-2}, 3{\times}10^{-2}\}$ (extended when the best sits on
an edge) and selected on held-out reconstruction error, and the stage-2 one over five
half-decade values around the best of an earlier short-context run. The head-mixing MLPs
are not swept: the cross-layer ones reuse the head-wise MLP's learning rates for the same
cell, and the Qwen3-4B/8B one uses $10^{-3}$ in stage 1 and the best of
$\{3{\times}10^{-6}, 10^{-5}, 3{\times}10^{-5}\}$ in stage 2. Head-wise MLPs on the
shape-preserving pairs are trained with three seeds, every other MLP with one.

Figure~\ref{fig:prefixhist} shows the prefix lengths of the long-context training mixture
of Appendix~\ref{app:data}.

Appendix~\ref{app:evaldata} gives the evaluation protocol: decoding, judging and
scoring. Appendix~\ref{app:seeds} defines the error bars and the seeds.

\begin{figure}[!htbp]
\centering
\includegraphics[width=0.55\linewidth]{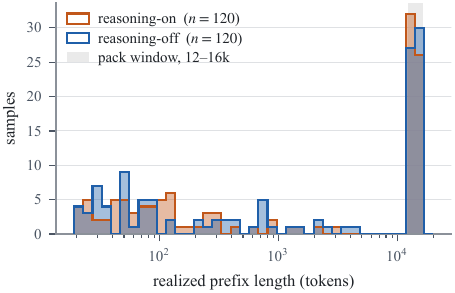}
\caption{\textbf{Realised prefix length of the long-context instruct training mixture.}
Histogram over $240$ probe samples, split by reasoning status, on log-spaced bins. Both
statuses span the short unpacked prompts and the packed window alike, so a downstream
reasoning-on/off comparison is not also a comparison of prefix lengths.}
\label{fig:prefixhist}
\end{figure}

\subsection{Seeds and error bars}
\label{app:seeds}

\paragraph{Error bars.}
Every $\pm$ in the tables and every error bar in the figures is the standard error of the
benchmark score over its evaluation items: for a score that averages the per-item scores
$s_1, \dots, s_n$, it is their sample standard deviation divided by $\sqrt{n}$. It measures
how far the score would move on a fresh draw of evaluation items, with the translator held
fixed. A score averaged over disjoint item sets (MMLU-Pro over its $14$ categories, RULER
over its rungs, a mean over benchmarks) combines the standard errors of its parts as
$\sqrt{\sum_j w_j^2\,\mathrm{se}_j^2}$. LongMemEval asks the same questions at every rung,
so its mean over rungs is clustered by question instead. A three-seed mean averages each
item over the three seeds before taking the standard error, and a judge rating rescaled to
$[0, 1]$ rescales its standard error with it. The from-scratch and released Base pairs
stored only aggregate accuracies, so theirs is the binomial $\sqrt{p(1-p)/n}$; perplexities
and validation KLs carry none.

\paragraph{Seeds.}
On the two shape-preserving Qwen3 pairs, $0.6$B/$1.7$B and $4$B/$8$B, every trained translator is
trained with seeds $42$, $43$ and $44$ and reported as their mean. Every other translator is
reported at a single seed: seed $42$ for the families that were also trained at three seeds
(the cross-layer pairs at $\nu = 1$, Gemma-3 and Gemma-4), and the only seed for the wider
maps and the MLP baselines, which were trained once. The rest of this subsection shows
what the other two seeds change.

Table~\ref{tab:seedsd} compares the two sources of variation benchmark by benchmark. On the
shape-preserving pairs the spread over seeds is below the error bar of a single seed on every
benchmark but XQuAD, whose token-F1 over eleven languages has a small standard error. On the
cross-layer pairs it is as large as the error bar or larger on MMLU-Pro, ARC-C, CoQA, XQuAD,
RepLiQA and RULER, and on Gemma-3 on all of those but RULER: there, a single-seed number
carries training noise that its $\pm$ does not show. In $54$ of the $513$ three-seed cells
behind the table, seed $42$ lies more than two of its own standard errors from the three-seed mean.

\begin{table}[htbp]
\centering
\caption{\textbf{Seed spread against evaluation noise.} For each family trained at three
seeds, \emph{seed sd} is the standard deviation of a cell's three per-seed scores and
\emph{eval se} the standard error over evaluation items of one seed's score, each the
median over the family's cells (translators, directions and reasoning modes) on that
benchmark. RULER pools its per-rung cells, LongMemEval is the mean over its four rungs, and
judge ratings are rescaled to $[0, 1]$. ---: not evaluated for that family.}
\label{tab:seedsd}
\footnotesize
\setlength{\tabcolsep}{4pt}
\begin{tabular}{@{}l rr rr rr rr@{}}
\toprule
 & \multicolumn{2}{c}{shape-preserving Qwen3} & \multicolumn{2}{c}{cross-layer Qwen3}
 & \multicolumn{2}{c}{Gemma-3} & \multicolumn{2}{c}{Gemma-4} \\
\cmidrule(lr){2-3} \cmidrule(lr){4-5} \cmidrule(lr){6-7} \cmidrule(lr){8-9}
Benchmark & seed sd & eval se & seed sd & eval se & seed sd & eval se & seed sd & eval se \\
\midrule
MMLU-Pro & $0.010$ & $0.012$ & $0.016$ & $0.012$ & $0.016$ & $0.013$ & $0.011$ & $0.013$ \\
ARC-C & $0.006$ & $0.009$ & $0.014$ & $0.012$ & $0.023$ & $0.012$ & $0.005$ & $0.010$ \\
GSM8K & $0.012$ & $0.026$ & $0.021$ & $0.026$ & $0.012$ & $0.022$ & --- & --- \\
IFEval & $0.008$ & $0.020$ & $0.012$ & $0.021$ & $0.010$ & $0.021$ & --- & --- \\
CoQA & $0.013$ & $0.025$ & $0.036$ & $0.024$ & $0.055$ & $0.027$ & --- & --- \\
XQuAD & $0.014$ & $0.007$ & $0.013$ & $0.006$ & $0.024$ & $0.008$ & $0.035$ & $0.010$ \\
RepLiQA & $0.009$ & $0.013$ & $0.017$ & $0.018$ & $0.031$ & $0.018$ & --- & --- \\
MT-Bench-101 & $0.010$ & $0.021$ & $0.014$ & $0.025$ & $0.009$ & $0.018$ & $0.005$ & $0.017$ \\
RULER & $0.004$ & $0.009$ & $0.021$ & $0.017$ & $0.009$ & $0.014$ & $0.004$ & $0.007$ \\
LongMemEval & $0.011$ & $0.020$ & $0.013$ & $0.019$ & --- & --- & --- & --- \\
LongBench-v2 & $0.021$ & $0.045$ & $0.030$ & $0.042$ & --- & --- & $0.016$ & $0.022$ \\
\bottomrule
\end{tabular}
\end{table}

Table~\ref{tab:seedsets} follows the families that the paper reports at seed $42$ through
one benchmark-set mean per translator. Seed $42$ lies within two standard errors of the
three-seed mean on $11$ of the $16$ rows. The largest departure comes from a collapsed seed:
on $0.6$B/$8$B in $L \to S$ with the greedy assignment, seed $42$ scores $0.018$ on XQuAD
against $0.190$ and $0.235$ for the other two seeds, and $0.208$ on MMLU-Pro against $0.337$
and $0.314$, which puts its four-benchmark mean at $0.360$ against a three-seed mean of
$0.408$. Collapses of this kind, where one seed loses a third or more of its score on one
benchmark, occur five times, all on the cross-layer pairs and all reasoning-off. The other
three hit seed $43$: on the same cell with the depth assignment ($0.308$ on ARC-C against
$0.650$ and $0.676$), and on $1.7$B/$4$B in $S \to L$ with the depth assignment ($0.296$ on
CoQA against $0.581$ and $0.616$, and $0.197$ on XQuAD against $0.387$ and $0.392$).

\begin{table}[htbp]
\centering
\caption{\textbf{Seed $42$ against the three-seed mean, for the families reported at seed
$42$.} One benchmark-set mean per translator, reasoning-off: the four-benchmark mean of the
cross-layer figures (MMLU-Pro, ARC-C, RULER, LongMemEval), the six-benchmark mean of
Figure~\ref{fig:gemma-4}, and the mean over the eleven rows of Table~\ref{tab:gemma3}, with
judge ratings rescaled to $[0, 1]$. \emph{Seed 42} is the score the paper reports, $\pm$ its
standard error over evaluation items; \emph{3-seed mean} is $\pm$ the seed standard
deviation.}
\label{tab:seedsets}
\footnotesize
\setlength{\tabcolsep}{4pt}
\begin{tabular}{@{}llc rr rr@{}}
\toprule
Pair & Direction & Assignment & seed $42$ & 3-seed mean & seed $43$ & seed $44$ \\
\midrule
Qwen3-0.6B/8B & $L \to S$ & depth & $0.371_{\pm 0.007}$ & $0.351_{\pm 0.061}$ & $0.282$ & $0.400$ \\
Qwen3-0.6B/8B & $L \to S$ & greedy & $0.360_{\pm 0.007}$ & $0.408_{\pm 0.042}$ & $0.439$ & $0.424$ \\
Qwen3-0.6B/8B & $S \to L$ & depth & $0.408_{\pm 0.007}$ & $0.410_{\pm 0.003}$ & $0.411$ & $0.413$ \\
Qwen3-0.6B/8B & $S \to L$ & greedy & $0.435_{\pm 0.007}$ & $0.441_{\pm 0.005}$ & $0.442$ & $0.446$ \\
Qwen3-1.7B/4B & $L \to S$ & depth & $0.458_{\pm 0.007}$ & $0.461_{\pm 0.013}$ & $0.476$ & $0.451$ \\
Qwen3-1.7B/4B & $L \to S$ & greedy & $0.478_{\pm 0.007}$ & $0.492_{\pm 0.013}$ & $0.504$ & $0.493$ \\
Qwen3-1.7B/4B & $S \to L$ & depth & $0.536_{\pm 0.007}$ & $0.530_{\pm 0.007}$ & $0.522$ & $0.531$ \\
Qwen3-1.7B/4B & $S \to L$ & greedy & $0.518_{\pm 0.007}$ & $0.519_{\pm 0.007}$ & $0.513$ & $0.526$ \\
Gemma-4-E2B/E4B & $L \to S$ & depth & $0.668_{\pm 0.006}$ & $0.678_{\pm 0.009}$ & $0.680$ & $0.685$ \\
Gemma-4-E2B/E4B & $L \to S$ & greedy & $0.665_{\pm 0.006}$ & $0.675_{\pm 0.009}$ & $0.677$ & $0.684$ \\
Gemma-4-E2B/E4B & $S \to L$ & depth & $0.657_{\pm 0.006}$ & $0.655_{\pm 0.003}$ & $0.651$ & $0.655$ \\
Gemma-4-E2B/E4B & $S \to L$ & greedy & $0.664_{\pm 0.006}$ & $0.658_{\pm 0.007}$ & $0.650$ & $0.659$ \\
Gemma-3-4B/12B & $L \to S$ & depth & $0.712_{\pm 0.005}$ & $0.725_{\pm 0.012}$ & $0.733$ & $0.730$ \\
Gemma-3-4B/12B & $L \to S$ & greedy & $0.747_{\pm 0.005}$ & $0.744_{\pm 0.007}$ & $0.736$ & $0.749$ \\
Gemma-3-4B/12B & $S \to L$ & depth & $0.692_{\pm 0.005}$ & $0.685_{\pm 0.007}$ & $0.683$ & $0.679$ \\
Gemma-3-4B/12B & $S \to L$ & greedy & $0.706_{\pm 0.005}$ & $0.694_{\pm 0.012}$ & $0.694$ & $0.683$ \\
\bottomrule
\end{tabular}
\end{table}

\clearpage
\section{Qwen3-0.6B/1.7B and 4B/8B, pretrained}
\label{appsec:qwen3_shape}
\suppressfloats[t]

\begin{figure}[!htbp]
\centering
\includegraphics[width=\linewidth]{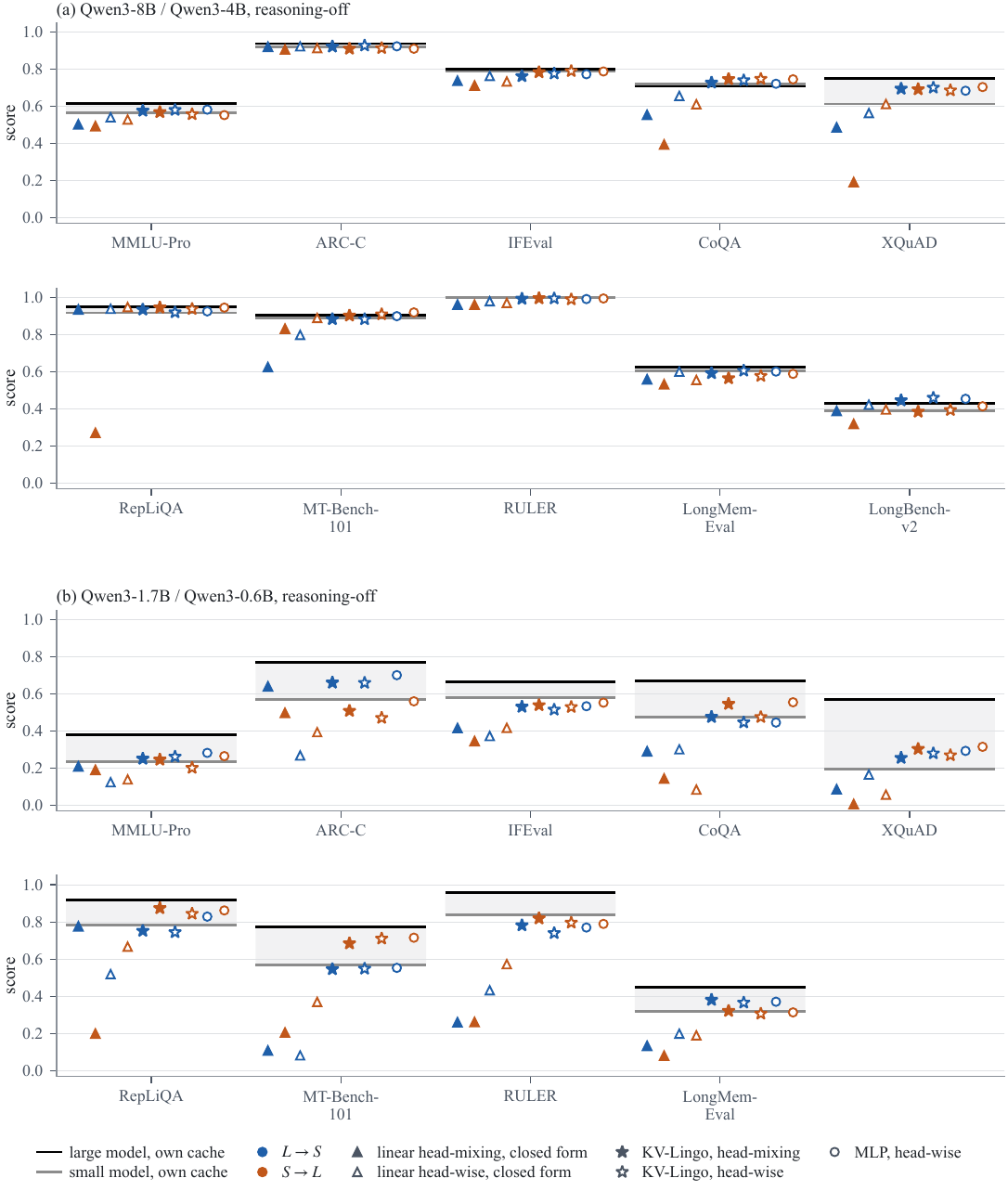}
\caption{\textbf{Shape-preserving Qwen3 pairs, reasoning-off, both
directions.} Trained translators are the mean over three seeds; error bars: standard error over evaluation items (Appendix~\ref{app:seeds}).}
\label{fig:suite_off}
\end{figure}

\begin{figure}[tbp]
\centering
\includegraphics[width=\linewidth]{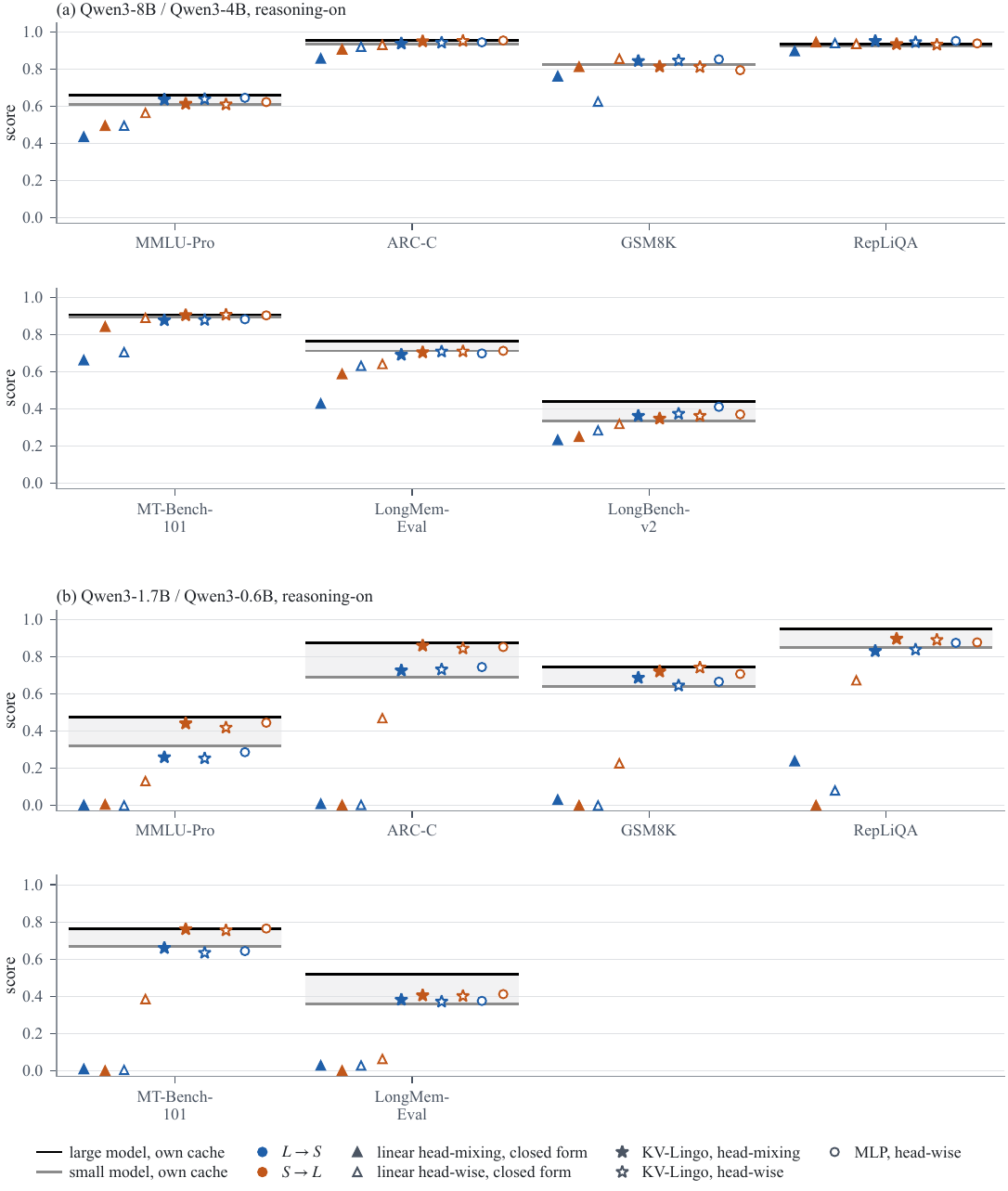}
\caption{\textbf{Shape-preserving Qwen3 pairs, reasoning-on, both directions.} Trained translators are the mean over three seeds; error bars: standard error over evaluation items (Appendix~\ref{app:seeds}).}
\label{fig:suite_on}
\end{figure}

\begin{figure}[tbp]
\centering
\includegraphics[width=\linewidth]{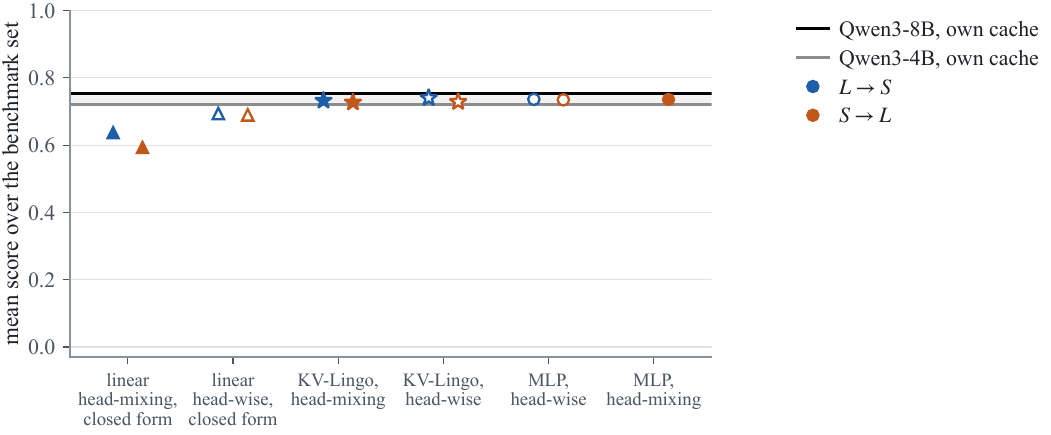}
\caption{Qwen3-4B/8B, reasoning-off, average scores of MMLU-Pro, ARC-C, IFEval, CoQA, XQuAD, MT-Bench-101, RULER, LongMemEval, LongBench-v2. The head-mixing MLP, computed only in the small to large setting, performs on par with the head-wise MLP. The head-mixing MLP is a single seed (seed 42), the other trained translators the mean over three seeds; error bars: standard error over evaluation items (Appendix~\ref{app:seeds}).}
\label{fig:head_mixing_mlp_shape}
\end{figure}

This section presents the full set of evaluations realised with the pretrained model pairs Qwen3-0.6B/1.7B and 4B/8B.
Tables~\ref{tab:qwen3ls} and~\ref{tab:qwen3sl} report both pairs over the whole evaluation
suite, one table per translation direction: the two native anchors against
the closed-form fit, the three trained translators of \S\ref{sec:exp:setup} and the
head-wise MLP adapter of~\citet{chen2026see}. Both are scored reasoning-off throughout.
Figures~\ref{fig:suite_off} and~\ref{fig:suite_on} draw the same measurement, one figure
per reasoning mode.
Figure~\ref{fig:head_mixing_mlp_shape} includes a head-mixing MLP as a baseline, in $S \to L$ with the model pair Qwen3-4B/8B, showing that it performs on par with the head-wise MLP baseline.

\begin{table}[tbp]
\centering
\footnotesize
\setlength{\tabcolsep}{3pt}
\caption{\textbf{Large to small evaluation suite, Qwen3 pairs with the same shape, reasoning-off.} The large model
prefills, the small one decodes from the translated cache. Trained translators are the mean over three seeds; the
closed-form fit is a single solve. $\pm$: standard error over evaluation items (Appendix~\ref{app:seeds}). \underline{Underline}: the translated
cache beats the small model. \textbf{Bold}: the
best translator on that row. The head-wise MLP adapter of~\citet{chen2026see} is shown at
the width that matches our map in parameters and FLOPs (hidden dimension is $4\times$ the input dimension).
Pre-normalisation capture throughout.
Metrics follow Table~\ref{tab:evalsets}: accuracy for MMLU-Pro (macro over $14$
categories), ARC-C and LongBench-v2 (untruncated subset),
prompt-level strict accuracy for IFEval, a $1$--$10$ judge score for
MT-Bench-101, token-F1 for CoQA and for XQuAD (pooled at equal $n$ over its eleven
languages), graded fact coverage on the answerable split for RepLiQA, recall averaged over
$4$k--$32$k for RULER, and judged accuracy averaged over the four dilution rungs for
LongMemEval. H-M means head-mixing. The mixed loss is an ablation, presented in Appendix~\ref{app:objective}. ---: not run.}
\label{tab:qwen3ls}
\resizebox{\ifdim\width>\linewidth\linewidth\else\width\fi}{!}{%
\begin{tabular}{@{}clrrrrrrr@{}}
\toprule
 & & \multicolumn{2}{c}{native} & \multicolumn{5}{c}{translated} \\
\cmidrule(lr){3-4} \cmidrule(lr){5-9}
 &  &  &  & closed & KV-Lingo & KV-Lingo & KV-Lingo & MLP \\
 & Benchmark & small & large & form & H-M, KL & H-M, mixed loss & head-wise & head-wise \\
\midrule
\multirow{10}{*}{\rotatebox[origin=c]{90}{Qwen3-0.6B/1.7B}} & MMLU-Pro & $0.234_{\pm 0.011}$ & $0.379_{\pm 0.013}$ & $0.210_{\pm 0.011}$ & $\underline{0.251}_{\pm 0.010}$ & $\underline{0.279}_{\pm 0.010}$ & $\underline{0.262}_{\pm 0.010}$ & $\underline{\mathbf{0.283}}_{\pm 0.010}$ \\
 & ARC-C & $0.568_{\pm 0.014}$ & $0.770_{\pm 0.012}$ & $\underline{0.640}_{\pm 0.014}$ & $\underline{0.660}_{\pm 0.012}$ & $\underline{0.694}_{\pm 0.012}$ & $\underline{0.659}_{\pm 0.012}$ & $\underline{\mathbf{0.701}}_{\pm 0.012}$ \\
 & IFEval & $0.580_{\pm 0.021}$ & $0.662_{\pm 0.020}$ & $0.416_{\pm 0.021}$ & $0.530_{\pm 0.019}$ & $0.476_{\pm 0.019}$ & $0.515_{\pm 0.019}$ & $\mathbf{0.534}_{\pm 0.020}$ \\
 & CoQA & $0.475_{\pm 0.026}$ & $0.672_{\pm 0.026}$ & $0.291_{\pm 0.022}$ & $\underline{0.476}_{\pm 0.024}$ & $\underline{\mathbf{0.516}}_{\pm 0.023}$ & $0.446_{\pm 0.023}$ & $0.446_{\pm 0.024}$ \\
 & RepLiQA & $0.782_{\pm 0.019}$ & $0.916_{\pm 0.012}$ & $0.776_{\pm 0.020}$ & $0.752_{\pm 0.017}$ & $\underline{\mathbf{0.841}}_{\pm 0.014}$ & $0.745_{\pm 0.017}$ & $\underline{0.829}_{\pm 0.015}$ \\
 & XQuAD & $0.194_{\pm 0.004}$ & $0.571_{\pm 0.008}$ & $0.087_{\pm 0.004}$ & $\underline{0.256}_{\pm 0.006}$ & $\underline{\mathbf{0.362}}_{\pm 0.007}$ & $\underline{0.280}_{\pm 0.006}$ & $\underline{0.294}_{\pm 0.006}$ \\
 & MT-Bench-101 & $6.11_{\pm 0.26}$ & $7.95_{\pm 0.22}$ & $1.97_{\pm 0.16}$ & $5.92_{\pm 0.25}$ & $5.44_{\pm 0.25}$ & $5.94_{\pm 0.26}$ & $\mathbf{5.98}_{\pm 0.28}$ \\
 & RULER & $0.839_{\pm 0.006}$ & $0.958_{\pm 0.004}$ & $0.261_{\pm 0.008}$ & $0.782_{\pm 0.005}$ & $\mathbf{0.799}_{\pm 0.004}$ & $0.740_{\pm 0.005}$ & $0.770_{\pm 0.005}$ \\
 & LongMemEval & $0.331_{\pm 0.019}$ & $0.470_{\pm 0.020}$ & $0.134_{\pm 0.011}$ & $\underline{\mathbf{0.403}}_{\pm 0.020}$ & $\underline{0.402}_{\pm 0.020}$ & $\underline{0.380}_{\pm 0.019}$ & $\underline{0.385}_{\pm 0.020}$ \\
 & LongBench-v2 & $0.276_{\pm 0.042}$ & $0.319_{\pm 0.043}$ & $\underline{\mathbf{0.284}}_{\pm 0.042}$ & $0.261_{\pm 0.030}$ & $0.270_{\pm 0.035}$ & $0.250_{\pm 0.035}$ & --- \\
\midrule
\multirow{10}{*}{\rotatebox[origin=c]{90}{Qwen3-4B/8B}} & MMLU-Pro & $0.564_{\pm 0.013}$ & $0.615_{\pm 0.013}$ & $0.503_{\pm 0.013}$ & $\underline{0.576}_{\pm 0.012}$ & $\underline{0.572}_{\pm 0.012}$ & $\underline{0.580}_{\pm 0.012}$ & $\underline{\mathbf{0.583}}_{\pm 0.012}$ \\
 & ARC-C & $0.920_{\pm 0.008}$ & $0.937_{\pm 0.007}$ & $0.920_{\pm 0.008}$ & $\underline{0.921}_{\pm 0.007}$ & $0.914_{\pm 0.007}$ & $\underline{\mathbf{0.927}}_{\pm 0.007}$ & $\underline{0.923}_{\pm 0.007}$ \\
 & IFEval & $0.786_{\pm 0.018}$ & $0.802_{\pm 0.017}$ & $0.738_{\pm 0.019}$ & $0.762_{\pm 0.016}$ & $0.754_{\pm 0.016}$ & $\mathbf{0.776}_{\pm 0.016}$ & $0.773_{\pm 0.016}$ \\
 & CoQA & $0.721_{\pm 0.024}$ & $0.712_{\pm 0.023}$ & $0.554_{\pm 0.024}$ & $\underline{0.727}_{\pm 0.023}$ & $\underline{0.725}_{\pm 0.023}$ & $\underline{\mathbf{0.740}}_{\pm 0.023}$ & $\underline{0.721}_{\pm 0.023}$ \\
 & RepLiQA & $0.916_{\pm 0.012}$ & $0.951_{\pm 0.009}$ & $\underline{0.934}_{\pm 0.011}$ & $\underline{0.934}_{\pm 0.009}$ & $\underline{\mathbf{0.944}}_{\pm 0.009}$ & $\underline{0.919}_{\pm 0.011}$ & $\underline{0.924}_{\pm 0.010}$ \\
 & XQuAD & $0.613_{\pm 0.007}$ & $0.752_{\pm 0.006}$ & $0.487_{\pm 0.007}$ & $\underline{0.695}_{\pm 0.007}$ & $\underline{0.699}_{\pm 0.006}$ & $\underline{\mathbf{0.700}}_{\pm 0.007}$ & $\underline{0.684}_{\pm 0.007}$ \\
 & MT-Bench-101 & $9.00_{\pm 0.19}$ & $9.15_{\pm 0.16}$ & $6.62_{\pm 0.26}$ & $8.94_{\pm 0.15}$ & $8.77_{\pm 0.15}$ & $8.93_{\pm 0.15}$ & $\underline{\mathbf{9.08}}_{\pm 0.13}$ \\
 & RULER & $0.998_{\pm 0.001}$ & $0.998_{\pm 0.001}$ & $0.960_{\pm 0.004}$ & $0.991_{\pm 0.002}$ & $0.992_{\pm 0.002}$ & $\mathbf{0.993}_{\pm 0.001}$ & $0.990_{\pm 0.002}$ \\
 & LongMemEval & $0.633_{\pm 0.019}$ & $0.643_{\pm 0.020}$ & $0.583_{\pm 0.020}$ & $0.621_{\pm 0.019}$ & $0.615_{\pm 0.019}$ & $\mathbf{0.632}_{\pm 0.019}$ & $0.627_{\pm 0.020}$ \\
 & LongBench-v2 & $0.388_{\pm 0.045}$ & $0.431_{\pm 0.046}$ & $0.388_{\pm 0.045}$ & $\underline{0.445}_{\pm 0.044}$ & $\underline{0.448}_{\pm 0.043}$ & $\underline{\mathbf{0.460}}_{\pm 0.044}$ & $\underline{0.454}_{\pm 0.045}$ \\
\bottomrule
\end{tabular}}
\end{table}

\begin{table}[tbp]
\centering
\footnotesize
\setlength{\tabcolsep}{3pt}
\caption{\textbf{Small to large evaluation suite, Qwen3 pairs with the same shape, reasoning-off.} The small model
prefills, the large one decodes from the translated cache. Trained translators are the mean over three seeds; the
closed-form fit is a single solve. $\pm$: standard error over evaluation items (Appendix~\ref{app:seeds}). \underline{Underline}: the translated
cache beats the small model by more than one standard error. \textbf{Bold}: the
best translator on that row. The head-wise MLP adapter of~\citet{chen2026see} is shown at
the width that matches our map in parameters and FLOPs (hidden dimension is $4\times$ the input dimension). 
Pre-normalisation capture throughout.
Metrics follow Table~\ref{tab:evalsets}: accuracy for MMLU-Pro (macro over $14$
categories), ARC-C and LongBench-v2 (untruncated subset),
prompt-level strict accuracy for IFEval, a $1$--$10$ judge score for
MT-Bench-101, token-F1 for CoQA and for XQuAD (pooled at equal $n$ over its eleven
languages), graded fact coverage on the answerable split for RepLiQA, recall averaged over
$4$k--$32$k for RULER, and judged accuracy averaged over the four dilution rungs for
LongMemEval. H-M means head-mixing. The mixed loss is an ablation, presented in Appendix~\ref{app:objective}.
On the 0.6B/1.7B pair the head-wise MLP's MT-Bench-101 cell is the mean of two seeds: the third
returns a score outside the judge's $1$--$10$ scale, a judge-parser bug that read ``1996'' as a rating. ---: not run.}
\label{tab:qwen3sl}
\resizebox{\ifdim\width>\linewidth\linewidth\else\width\fi}{!}{%
\begin{tabular}{@{}clrrrrrrr@{}}
\toprule
 & & \multicolumn{2}{c}{native} & \multicolumn{5}{c}{translated} \\
\cmidrule(lr){3-4} \cmidrule(lr){5-9}
 &  &  &  & closed & KV-Lingo & KV-Lingo & KV-Lingo & MLP \\
 & Benchmark & small & large & form & H-M, KL & H-M, mixed loss & head-wise & head-wise \\
\midrule
\multirow{10}{*}{\rotatebox[origin=c]{90}{Qwen3-0.6B/1.7B}} & MMLU-Pro & $0.234_{\pm 0.011}$ & $0.379_{\pm 0.013}$ & $0.191_{\pm 0.010}$ & $\underline{0.246}_{\pm 0.009}$ & $0.234_{\pm 0.009}$ & $0.202_{\pm 0.008}$ & $\underline{\mathbf{0.265}}_{\pm 0.010}$ \\
 & ARC-C & $0.568_{\pm 0.014}$ & $0.770_{\pm 0.012}$ & $0.497_{\pm 0.015}$ & $0.509_{\pm 0.012}$ & $0.488_{\pm 0.012}$ & $0.471_{\pm 0.012}$ & $\mathbf{0.560}_{\pm 0.013}$ \\
 & IFEval & $0.575_{\pm 0.021}$ & $0.664_{\pm 0.020}$ & $0.346_{\pm 0.020}$ & $0.539_{\pm 0.019}$ & $0.502_{\pm 0.019}$ & $0.529_{\pm 0.018}$ & $\mathbf{0.553}_{\pm 0.020}$ \\
 & CoQA & $0.475_{\pm 0.026}$ & $0.672_{\pm 0.026}$ & $0.144_{\pm 0.018}$ & $\underline{0.546}_{\pm 0.024}$ & $\underline{\mathbf{0.568}}_{\pm 0.024}$ & $0.475_{\pm 0.022}$ & $\underline{0.555}_{\pm 0.025}$ \\
 & RepLiQA & $0.782_{\pm 0.019}$ & $0.916_{\pm 0.012}$ & $0.200_{\pm 0.020}$ & $\underline{\mathbf{0.874}}_{\pm 0.013}$ & $\underline{0.854}_{\pm 0.013}$ & $\underline{0.844}_{\pm 0.013}$ & $\underline{0.862}_{\pm 0.013}$ \\
 & XQuAD & $0.194_{\pm 0.004}$ & $0.571_{\pm 0.008}$ & $0.007_{\pm 0.001}$ & $\underline{0.304}_{\pm 0.006}$ & $\underline{\mathbf{0.366}}_{\pm 0.007}$ & $\underline{0.270}_{\pm 0.006}$ & $\underline{0.316}_{\pm 0.007}$ \\
 & MT-Bench-101 & $6.11_{\pm 0.26}$ & $7.95_{\pm 0.22}$ & $2.85_{\pm 0.21}$ & $\underline{7.16}_{\pm 0.19}$ & $\underline{7.14}_{\pm 0.19}$ & $\underline{7.39}_{\pm 0.19}$ & $\underline{\mathbf{7.44}}_{\pm 0.20}$ \\
 & RULER & $0.839_{\pm 0.006}$ & $0.958_{\pm 0.004}$ & $0.262_{\pm 0.009}$ & $\mathbf{0.819}_{\pm 0.004}$ & $0.815_{\pm 0.004}$ & $0.796_{\pm 0.004}$ & $0.789_{\pm 0.005}$ \\
 & LongMemEval & $0.331_{\pm 0.019}$ & $0.470_{\pm 0.020}$ & $0.087_{\pm 0.010}$ & $\mathbf{0.335}_{\pm 0.017}$ & $0.333_{\pm 0.018}$ & $0.318_{\pm 0.017}$ & $0.328_{\pm 0.019}$ \\
 & LongBench-v2 & $0.276_{\pm 0.042}$ & $0.319_{\pm 0.043}$ & $0.302_{\pm 0.043}$ & $0.296_{\pm 0.029}$ & $\underline{\mathbf{0.322}}_{\pm 0.035}$ & $0.270_{\pm 0.031}$ & --- \\
\midrule
\multirow{10}{*}{\rotatebox[origin=c]{90}{Qwen3-4B/8B}} & MMLU-Pro & $0.564_{\pm 0.013}$ & $0.615_{\pm 0.013}$ & $0.494_{\pm 0.013}$ & $\mathbf{0.569}_{\pm 0.012}$ & $0.558_{\pm 0.012}$ & $0.558_{\pm 0.012}$ & $0.553_{\pm 0.012}$ \\
 & ARC-C & $0.920_{\pm 0.008}$ & $0.937_{\pm 0.007}$ & $0.905_{\pm 0.009}$ & $0.910_{\pm 0.008}$ & $\mathbf{0.914}_{\pm 0.007}$ & $0.913_{\pm 0.008}$ & $0.910_{\pm 0.008}$ \\
 & IFEval & $0.789_{\pm 0.018}$ & $0.799_{\pm 0.017}$ & $0.712_{\pm 0.019}$ & $0.783_{\pm 0.016}$ & $0.766_{\pm 0.016}$ & $\mathbf{0.789}_{\pm 0.016}$ & $0.787_{\pm 0.016}$ \\
 & CoQA & $0.721_{\pm 0.024}$ & $0.712_{\pm 0.023}$ & $0.394_{\pm 0.022}$ & $\underline{0.746}_{\pm 0.022}$ & $\underline{0.747}_{\pm 0.023}$ & $\underline{\mathbf{0.747}}_{\pm 0.022}$ & $\underline{0.745}_{\pm 0.022}$ \\
 & RepLiQA & $0.916_{\pm 0.012}$ & $0.951_{\pm 0.009}$ & $0.271_{\pm 0.023}$ & $\underline{0.944}_{\pm 0.009}$ & $\underline{0.944}_{\pm 0.009}$ & $\underline{0.937}_{\pm 0.009}$ & $\underline{\mathbf{0.944}}_{\pm 0.009}$ \\
 & XQuAD & $0.613_{\pm 0.007}$ & $0.752_{\pm 0.006}$ & $0.193_{\pm 0.006}$ & $\underline{0.692}_{\pm 0.007}$ & $\underline{0.677}_{\pm 0.007}$ & $\underline{0.686}_{\pm 0.007}$ & $\underline{\mathbf{0.704}}_{\pm 0.007}$ \\
 & MT-Bench-101 & $9.00_{\pm 0.19}$ & $9.15_{\pm 0.16}$ & $8.47_{\pm 0.21}$ & $9.11_{\pm 0.14}$ & $\underline{9.18}_{\pm 0.14}$ & $\underline{9.17}_{\pm 0.15}$ & $\underline{\mathbf{9.26}}_{\pm 0.14}$ \\
 & RULER & $0.998_{\pm 0.001}$ & $0.998_{\pm 0.001}$ & $0.960_{\pm 0.003}$ & $\mathbf{0.994}_{\pm 0.002}$ & $0.992_{\pm 0.002}$ & $0.988_{\pm 0.002}$ & $0.993_{\pm 0.002}$ \\
 & LongMemEval & $0.633_{\pm 0.019}$ & $0.643_{\pm 0.020}$ & $0.563_{\pm 0.020}$ & $0.591_{\pm 0.019}$ & $0.603_{\pm 0.019}$ & $0.604_{\pm 0.019}$ & $\mathbf{0.613}_{\pm 0.019}$ \\
 & LongBench-v2 & $0.388_{\pm 0.045}$ & $0.431_{\pm 0.046}$ & $0.319_{\pm 0.043}$ & $0.385_{\pm 0.041}$ & $0.399_{\pm 0.043}$ & $0.394_{\pm 0.042}$ & $\mathbf{0.414}_{\pm 0.044}$ \\
\bottomrule
\end{tabular}}
\end{table}

\begin{figure}[tbp]
\centering
\includegraphics[width=\linewidth]{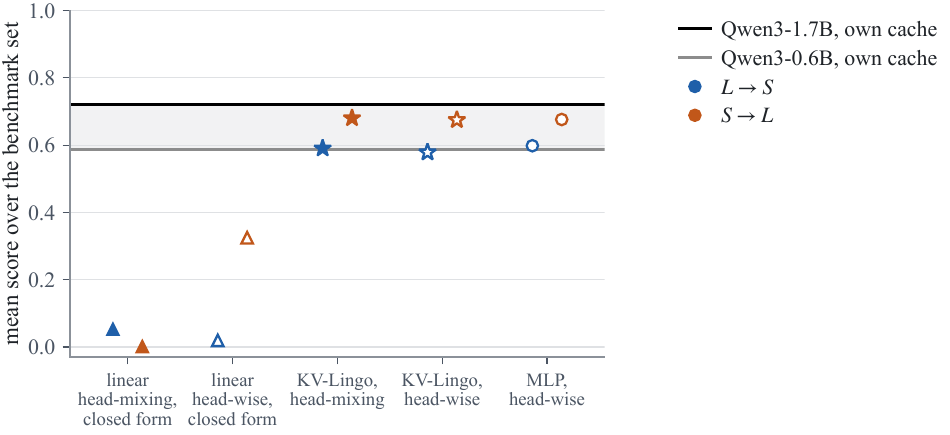}
\caption{\textbf{Shape-preserving small Qwen3 pair, reasoning-on, both directions.} Average score over MMLU-Pro, ARC-C, GSM8K, RepLiQA, MT-Bench-101 and LongMemEval.}
\label{fig:suite_small_on_mean}
\end{figure}

\section{Head alignment in released Qwen3 pairs}
\label{app:head_alignment}
\suppressfloats[t]

\begin{figure}[htbp]
\centering
\includegraphics[width=\linewidth]{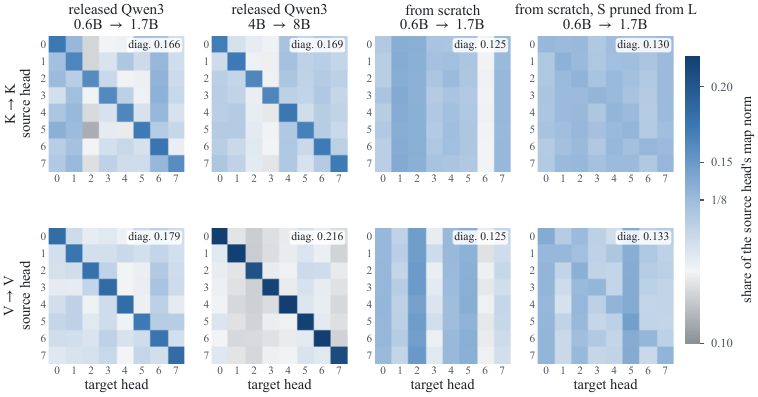}
\caption{\textbf{Per-head structure of the closed-form map, $S \to L$.} Share of each
source head's map norm sent to each target head, for key-to-key (top) and value-to-value
(bottom) blocks, averaged over layers; rows sum to one, and white is chance ($1/8$).
\emph{diag.}: mean of the diagonal. Released Qwen3-0.6B/1.7B and 4B/8B against a
0.6B/1.7B pair we pretrained from scratch (Appendix~\ref{app:scratch}), from independent
initialisations or with the small model sliced out of the large one.}
\label{fig:head_alignment}
\end{figure}

On the two shape-preserving Qwen3 pairs, head-wise translators match head-mixing ones
(\S\ref{sec:exp:shape}), whereas on the pair pretrained from scratch they fall behind
(Table~\ref{tab:scratch}). Figure~\ref{fig:head_alignment} explains the difference: the
heads of the released models correspond one to one, while those of the from-scratch
models do not.

To measure this without imposing any head structure, we take the closed-form joint map
of Appendix~\ref{app:splitjoint}, fitted by \eqref{eq:hat} with no constraint on head
mixing. We cut its key-to-key and value-to-value blocks into $8 \times 8$ blocks of size
$d_h \times d_h$. Entry $(i, j)$ of a grid is the Frobenius norm of the block that sends
source head $i$ to target head $j$, averaged over layers and normalised so that each row
sums to one. Chance is therefore $1/8$. We compare the mean diagonal against the same
statistic under $10^5$ random permutations of the target heads.

On both released pairs, in $S \to L$, every source head sends its largest block to the
target head with the same index. The mean diagonal reaches $0.166$--$0.216$, $33$--$73\%$
above its permutation-null mean. This is the block that a head-wise translator keeps,
and the off-diagonal blocks it drops are smaller, each below $0.135$. From 8B to 4B the
diagonal is stronger still ($0.196$ on keys and $0.249$ on values). The exception is
1.7B~$\to$~0.6B, whose grids sit at chance, so for that direction the closed-form map
does not by itself account for the head-wise result.

The from-scratch pair has no such structure: its diagonal is at chance ($0.125$ on both
keys and values), and each grid is nearly uniform. The only structure is by column: a
few target heads receive more from every source head. The alignment therefore comes from
how the released checkpoints were produced, not from the architecture. A shared
initialisation recovers only a small part of it. We initialise the small model by slicing
the first heads and channels out of the large one and pretrain it for the same $35$B
tokens (last column). This brings the identity back, but weakly: the diagonal is $0.130$
and $0.133$, $4\%$ and $6\%$ above chance.

\clearpage
\section{Qwen3-0.6B/8B and Qwen3-1.7B/4B, pretrained}
\label{app:xdepth}
\suppressfloats[t]

\begin{figure}[tbp]
    \centering
    \includegraphics[width=\linewidth]{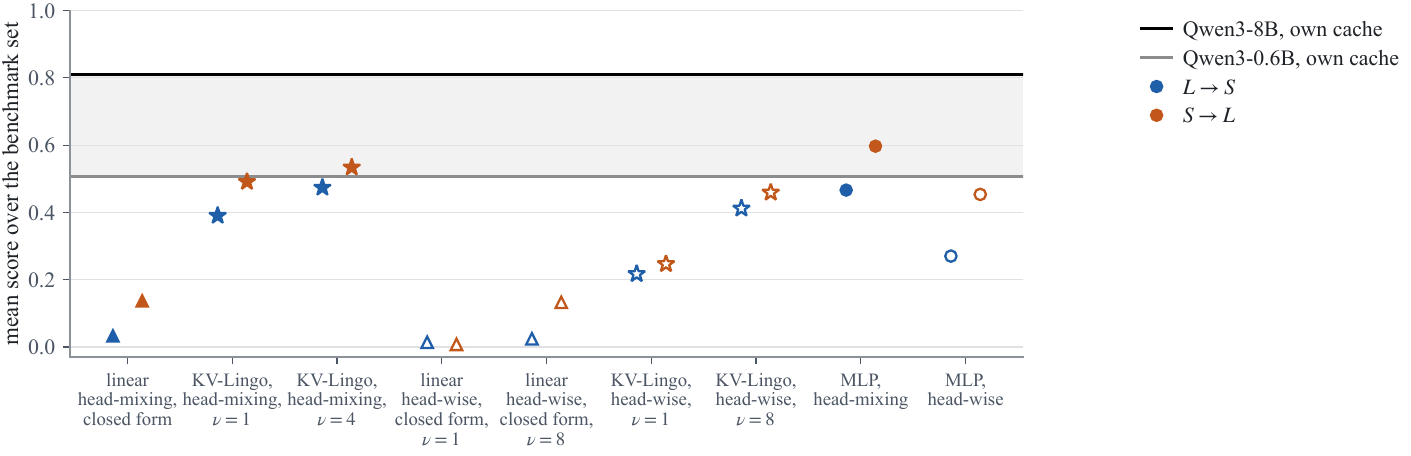}
    \caption{\textbf{Every translator on Qwen3-8B/0.6B}, both directions, reasoning-off.
    Each marker is the mean score over MMLU-Pro, ARC-C, RULER, LongMemEval, CoQA, IFEval,
    XQuAD, RepLiQA and MT-Bench-101, weighted equally, with RULER's four context
    budgets averaged first and judge scores rescaled to $[0, 1]$; LongBench-v2 is left out,
    as it was run in $S \to L$ only. Black and grey lines are Qwen3-8B and Qwen3-0.6B on
    their own cache. Triangles are closed-form fits, stars KV-Lingo after
    self-distillation, circles the two MLP translators of Appendix~\ref{app:protocol};
    filled markers mix heads, open ones are head-wise. The head-mixing linear translators
    use the $R^2$ assignment, their closed form at $\nu = 1$; the head-wise linear
    translators and the head-mixing MLP use greedy match, the head-wise MLP relative depth.
    Head-mixing KV-Lingo at $\nu = 4$ matches the two MLPs in parameters; head-wise
    KV-Lingo would need $\nu = 32$, so $\nu = 8$ stands in at a quarter of their count.
    Head-mixing KV-Lingo at $\nu = 1$ is the mean over three seeds, at $\nu = 4$ seed 56,
    and every other trained translator seed 42; no error bars.}
    \label{fig:xdepth_all_1}
\end{figure}

\begin{figure}[tbp]
    \centering
    \includegraphics[width=\linewidth]{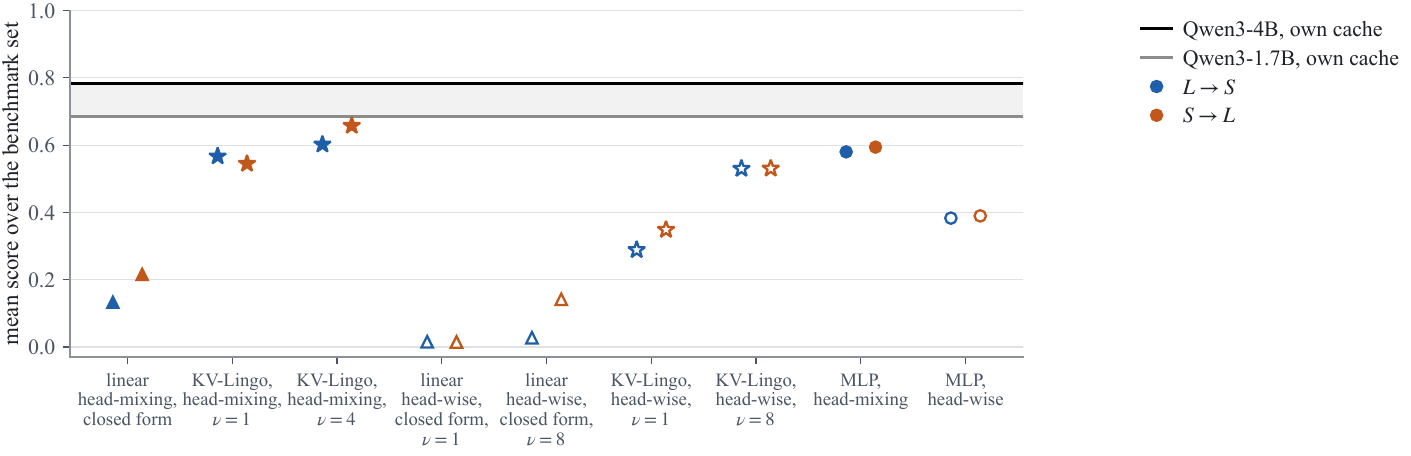}
    \caption{\textbf{Every translator on Qwen3-4B/1.7B}, both directions, reasoning-off.
    Everything as in Figure~\ref{fig:xdepth_all_1}, for the other cross-layer pair.}
    \label{fig:xdepth_all_2}
\end{figure}

\begin{figure}[htbp]
    \centering
    \includegraphics[width=\linewidth]{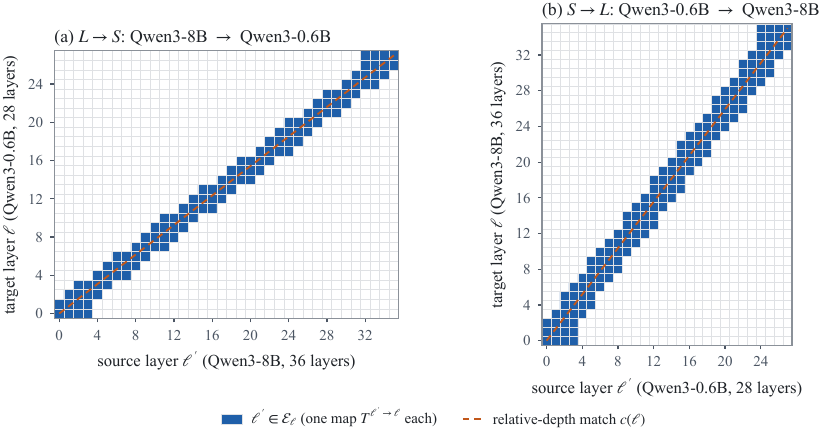}
    \caption{\textbf{The relative-depth assignment $\mathcal{E}_\ell$} of
    \S\ref{sec:method:family}, drawn for the unequal-depth Qwen3-0.6B/8B pair in both
    directions at $\nu = 4$ source layers per target layer. A filled cell $(\ell',\ell)$
    means that target layer $\ell$ reads source layer $\ell'$, through its own matrices
    $T_K^{\ell'\to\ell},\, T_V^{\ell'\to\ell}$; each row therefore carries exactly $\nu$
    filled cells, and each panel $2\nu N_\tgt$ matrices in total. The dashed line is the
    exact relative-depth counterpart, $c(\ell) = \ell (N_\src-1)/(N_\tgt-1)$ with layers indexed from $0$ as in the plot, on which the
    window is centred everywhere except at the two ends of the stack, where it shifts
    inward rather than truncating so that every target layer still reads $\nu$ distinct
    source layers. Going up in depth ($S \to L$, right) the band is steeper than the
    diagonal and consecutive target layers reuse source layers; going down ($L \to S$,
    left) it is shallower and consecutive windows slide faster than one layer at a time.}
    \label{fig:xdepth_assignment}
\end{figure}

\begin{figure}[tbp]
    \centering
    \includegraphics[width=0.9\linewidth]{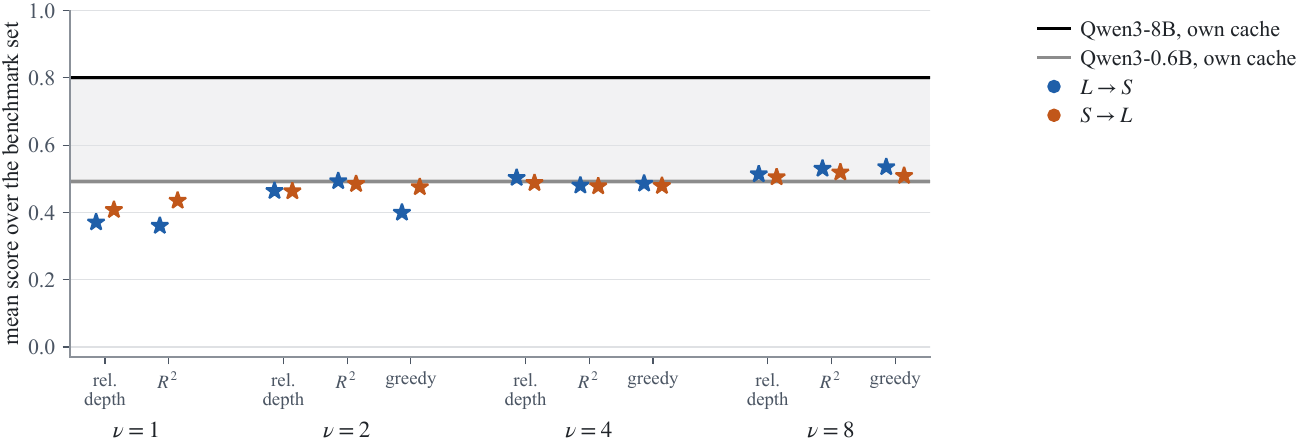}
    \caption{\textbf{Layer assignment against $\nu$, Qwen3-8B/0.6B}, both directions,
    head-mixing KV-Lingo, reasoning-off. Markers and native lines as in
    Figure~\ref{fig:head_wise_vs_head_mixing}; each marker is the mean score over MMLU-Pro,
    ARC-C, RULER and LongMemEval, weighted equally, with RULER's four context budgets
    averaged first. Within each group of $\nu$, the ticks are
    relative-depth match, the $R^2$ assignment (the $\nu$ source layers with the highest
    single-layer $R^2$, chosen separately for keys and values) and greedy match. At $\nu = 1$ greedy match reduces to the $R^2$ assignment up to the ridge penalty,
    so it has no separate translator there. Every point is a single seed, seed 42 at $\nu = 1$
    and seed 56 at $\nu \geq 2$; error bars: standard error over evaluation items (Appendix~\ref{app:seeds}).}
    \label{fig:assignment_comparison_1}
\end{figure}

\begin{figure}[tbp]
    \centering
    \includegraphics[width=0.9\linewidth]{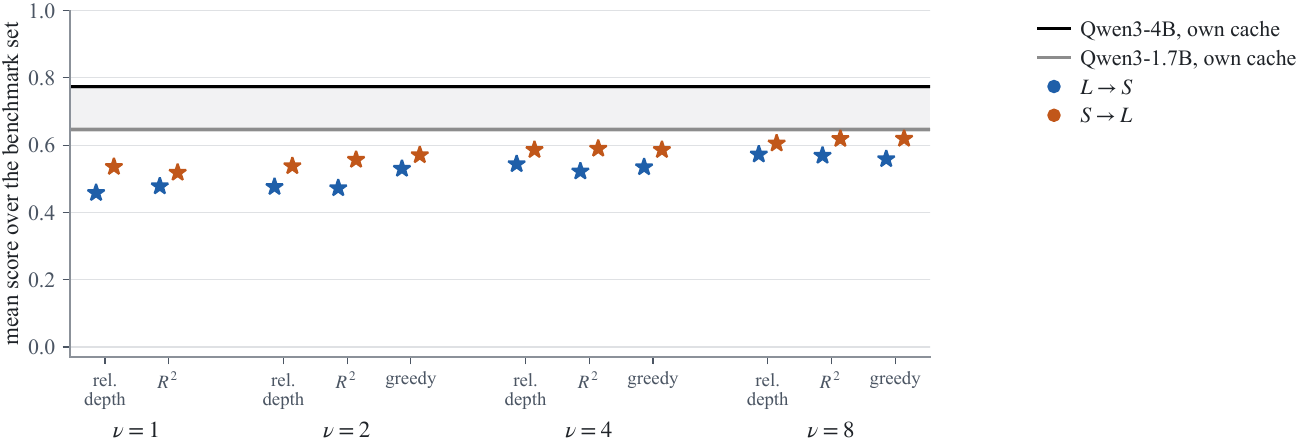}
    \caption{\textbf{Layer assignment against $\nu$, Qwen3-4B/1.7B}, both directions,
    head-mixing KV-Lingo, reasoning-off. Everything as in
    Figure~\ref{fig:assignment_comparison_1}, for the other cross-layer pair.}
    \label{fig:assignment_comparison_2}
\end{figure}

\begin{figure}[tbp]
    \centering
    \includegraphics[width=0.9\linewidth]{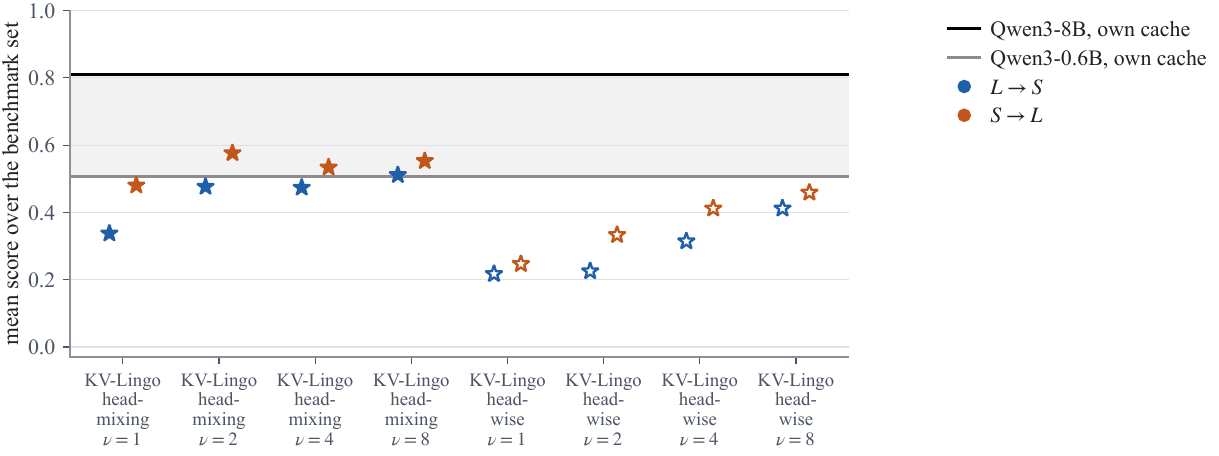}
    \caption{\textbf{Head-mixing against head-wise KV-Lingo over $\nu$}, Qwen3-8B/0.6B, both
    directions, reasoning-off. Each marker is the mean score over the nine benchmarks of
    Figure~\ref{fig:xdepth_all_1}, weighted equally. Black
    and grey lines are Qwen3-8B and Qwen3-0.6B on their own cache. Filled stars are the
    head-mixing translator with the $R^2$ assignment, open stars the head-wise one with
    greedy match. Every marker is a single seed: head-mixing at seed 42 for $\nu = 1$ and
    seed 56 for $\nu \geq 2$, head-wise at seed 42.}
    \label{fig:head_wise_vs_head_mixing}
\end{figure}

This section complements \S\ref{sec:exp:xdepth} on the two Qwen3 pairs whose depths
differ, $28$ layers for Qwen3-0.6B and Qwen3-1.7B against $36$ for Qwen3-4B and Qwen3-8B, each
translated in both directions, which makes four cross-layer cells. It first defines the
layer assignments (Figure~\ref{fig:xdepth_assignment}, then greedy match and the $R^2$
assignment below), then compares
them: at one source layer per target layer in Tables~\ref{tab:xdepth},
\ref{app:tab:xdepthruler} and~\ref{app:tab:xdepthlme}, and across the number $\nu$ of source
layers per target layer in Figures~\ref{fig:assignment_comparison_1}
and~\ref{fig:assignment_comparison_2}. Figure~\ref{fig:head_wise_vs_head_mixing} compares the
head-mixing and head-wise translators over $\nu$; the last paragraph reads what a larger
$\nu$ buys and what it costs. Figures~\ref{fig:xdepth_all_1} and~\ref{fig:xdepth_all_2}
set every translator evaluated on the two pairs side by side, closed-form, KV-Lingo and MLP,
head-mixing and head-wise, on the mean over nine reasoning-off benchmarks.

\paragraph{Greedy match.}
Greedy match (\S\ref{sec:method:family}) chooses $\mathcal{E}_\ell$ from data, independently
for each target layer. Fix a target layer $\ell$ and a calibration set of $n_{\mathrm{cal}}$
tokens; let $Y \in \R^{n_{\mathrm{cal}} \times d_\tgt}$ collect its features and
$X_s \in \R^{n_{\mathrm{cal}} \times d_\src}$ those of source layer $s$ on the same tokens. For a
support $S \subseteq \{1,\dots,N_\src\}$, let
$X_S \coloneqq [\,X_s\,]_{s \in S} \in \R^{n_{\mathrm{cal}} \times |S|d_\src}$ be the column-wise
concatenation, and define the ridge block map and its relative residual
\begin{equation}
    \hat{B}_S \coloneqq \argmin_{B \in \R^{|S|d_\src \times d_\tgt}}
        \bigl\lVert Y - X_S B \bigr\rVert_F^2 + \lambda \lVert B \rVert_F^2 ,
    \qquad
    r(S) \coloneqq \frac{\bigl\lVert Y - X_S \hat{B}_S \bigr\rVert_F^2}{\lVert Y \rVert_F^2} ,
\end{equation}
with the convention $r(\varnothing) = 1$. Since $|S|d_\src$ may exceed $n_{\mathrm{cal}}$, the
penalty $\lambda > 0$ is what guarantees that $\hat{B}_S$ is unique. The support is then
grown one source layer at a time, re-solving the full block map at every step: starting
from $S^{(0)} = \varnothing$, for $j = 1,\dots,\nu$,
\begin{equation}
    s^{(j)} \in \argmin_{s \,\notin\, S^{(j-1)}} \; r\bigl(S^{(j-1)} \cup \{s\}\bigr),
    \qquad
    S^{(j)} = S^{(j-1)} \cup \bigl\{s^{(j)}\bigr\},
\end{equation}
and we set $\mathcal{E}_{\ell} \coloneqq S^{(\nu)}$, so that $|\mathcal{E}_{\ell}| = \nu$. Each
step evaluates every remaining candidate exactly, at a cost of $O(\nu\,N_\src)$ ridge solves
per target layer. At $\nu = 1$ the single step picks the source
layer whose single-layer fit has the highest $R^2$, which is, up to the ridge penalty, the
$R^2$ assignment below.

\paragraph{The $R^2$ assignment.}
The $R^2$ assignment ranks source layers by their single-layer fit alone, without the
forward selection. For each target layer $\ell$ and source layer $s$, it scores
$R^2_\ell(s) \coloneqq 1 - r(\{s\})$, the fraction of the target layer's features explained
by the least-squares map from source layer $s$ alone, and sets $\mathcal{E}_\ell$ to the $\nu$
source layers with the highest $R^2_\ell(s)$. Keys and values are ranked separately, so a
target layer may read different source layers for its keys and for its values. The scores
are read off the Gram matrices the closed-form fit already accumulates, so the assignment
costs no extra pass over the data. Unlike greedy match, it does not account for
redundancy between the selected layers: two adjacent source layers that explain the same
part of the target are both kept.

\begin{figure}
    \centering
    \includegraphics[width=0.7\linewidth]{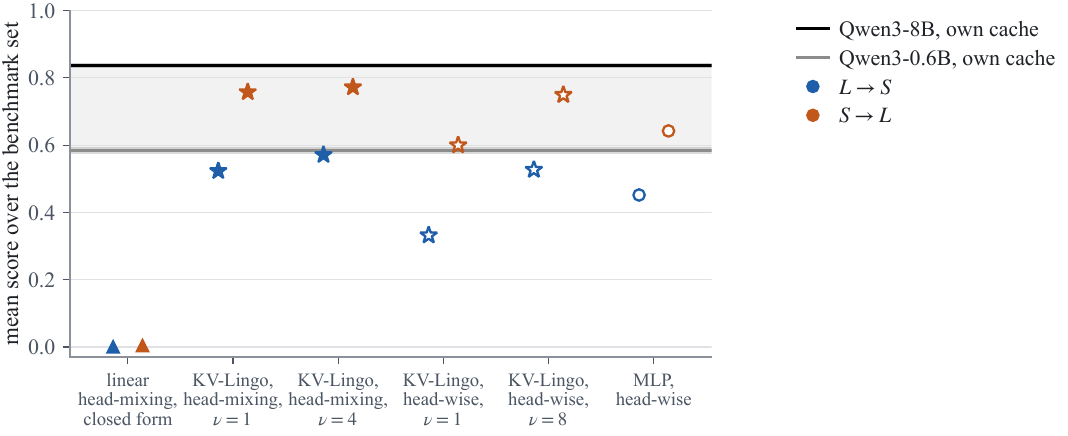}
    \caption{\textbf{Cache translation with Qwen3-0.6B/8B, reasoning-on.} Average over MMLU-Pro, ARC-C, GSM8K, RepLiQA, MT-Bench-101 and LongMemEval. Every point is a single seed, seed 42 except the head-mixing translator at $\nu = 4$ (seed 56); error bars: standard error over evaluation items (Appendix~\ref{app:seeds}).}
    \label{fig:xdepth_think}
\end{figure}

\paragraph{Comparing the assignments.}
Figure~\ref{fig:assignment_comparison_1} compares the three assignments on Qwen3-8B/0.6B
from $\nu = 1$ to $\nu = 8$, and Figure~\ref{fig:assignment_comparison_2} does the same on
Qwen3-4B/1.7B. Table~\ref{tab:xdepth} compares relative-depth match and the $R^2$ assignment at $\nu = 1$,
both closed-form and self-distilled, on all four cross-layer cells.
Table~\ref{app:tab:xdepthruler} breaks their RULER recall down by context budget in both
reasoning modes, and Table~\ref{app:tab:xdepthlme} gives LongMemEval with reasoning on.
Figure~\ref{fig:xdepth_think} draws the reasoning-on mean score of the Qwen3-0.6B/8B pair.

\begin{table}[tbp]
\centering
\footnotesize
\setlength{\tabcolsep}{5pt}
\caption{\textbf{Cross-layer cache translation on two Qwen3 pairs}, $28$ against
$36$ layers, both directions. Each target layer reads a single source layer, chosen
either by relative depth or by the $R^2$ of its best single-layer fit; each assignment is
reported as its closed-form fit and after self-distillation, the latter at seed 42.
$\pm$: standard error over evaluation items (Appendix~\ref{app:seeds}). The two anchor columns are the same measurement for both directions of a pair, so
they are shown even where that direction's translator cells are not.
\underline{Underline}: the translated cache beats the small model, and in $S \to L$ by more
than one standard error. Five further $S \to L$ cells clear the small model on
the value, by $+0.009$ to $+0.026$, but by less than their own standard error, and are left
unmarked. \textbf{Bold}: the best of the four translators on that row. A dash
marks a LongBench-v2 cell left out because the decoding model is at the chance floor. Reasoning-off throughout; Table~\ref{app:tab:xdepthlme}
gives reasoning-on LongMemEval and Table~\ref{app:tab:xdepthruler} the full RULER ladder. MMLU-Pro is the macro-average
over its $14$ categories and ARC-C the micro-average, both lenient generate-and-parse;
RULER is recall averaged over $4$k--$16$k; LongBench-v2 is accuracy on its untruncated
subset; LongMemEval is judge accuracy pooled over its oracle, $8$k, $16$k and $32$k rungs.}
\label{tab:xdepth}
\begin{tabular}{lrrrrrr}
\toprule
 & \multicolumn{2}{c}{native} & \multicolumn{2}{c}{closed form} & \multicolumn{2}{c}{self-distilled} \\
\cmidrule(lr){2-3} \cmidrule(lr){4-5} \cmidrule(lr){6-7}
Benchmark & small & large & depth & $R^2$ & depth & $R^2$ \\
\midrule
\multicolumn{7}{l}{\emph{$L \to S$}, Qwen3-8B $\to$ Qwen3-0.6B} \\
MMLU-Pro & $0.234_{\pm 0.011}$ & $0.615_{\pm 0.013}$ & $0.011_{\pm 0.003}$ & $0.104_{\pm 0.008}$ & $0.205_{\pm 0.011}$ & $\mathbf{0.208}_{\pm 0.011}$ \\
ARC-C & $0.568_{\pm 0.014}$ & $0.937_{\pm 0.007}$ & $0.000_{\pm 0.000}$ & $0.028_{\pm 0.005}$ & $\underline{0.650}_{\pm 0.014}$ & $\underline{\mathbf{0.659}}_{\pm 0.014}$ \\
RULER & $0.879_{\pm 0.006}$ & $0.998_{\pm 0.001}$ & $0.000_{\pm 0.000}$ & $0.001_{\pm 0.001}$ & $0.323_{\pm 0.010}$ & $\mathbf{0.327}_{\pm 0.010}$ \\
LongBench-v2 & $0.276_{\pm 0.042}$ & $0.431_{\pm 0.046}$ & --- & --- & --- & --- \\
LongMemEval & $0.331_{\pm 0.019}$ & $0.643_{\pm 0.020}$ & $0.010_{\pm 0.003}$ & $0.017_{\pm 0.004}$ & $\underline{\mathbf{0.348}}_{\pm 0.020}$ & $0.289_{\pm 0.019}$ \\
\midrule
\multicolumn{7}{l}{\emph{$L \to S$}, Qwen3-4B $\to$ Qwen3-1.7B} \\
MMLU-Pro & $0.379_{\pm 0.013}$ & $0.564_{\pm 0.013}$ & $0.134_{\pm 0.009}$ & $0.213_{\pm 0.011}$ & $0.283_{\pm 0.011}$ & $\mathbf{0.286}_{\pm 0.011}$ \\
ARC-C & $0.770_{\pm 0.012}$ & $0.920_{\pm 0.008}$ & $0.349_{\pm 0.014}$ & $0.580_{\pm 0.014}$ & $\underline{\mathbf{0.817}}_{\pm 0.011}$ & $\underline{0.811}_{\pm 0.011}$ \\
RULER & $0.980_{\pm 0.004}$ & $0.999_{\pm 0.001}$ & $0.000_{\pm 0.000}$ & $0.000_{\pm 0.000}$ & $0.415_{\pm 0.010}$ & $\mathbf{0.460}_{\pm 0.011}$ \\
LongBench-v2 & $0.319_{\pm 0.043}$ & $0.388_{\pm 0.045}$ & --- & --- & --- & --- \\
LongMemEval & $0.470_{\pm 0.020}$ & $0.633_{\pm 0.019}$ & $0.006_{\pm 0.002}$ & $0.095_{\pm 0.010}$ & $0.385_{\pm 0.019}$ & $\mathbf{0.410}_{\pm 0.019}$ \\
\midrule
\multicolumn{7}{l}{\emph{$S \to L$}, Qwen3-0.6B $\to$ Qwen3-8B} \\
MMLU-Pro & $0.234_{\pm 0.011}$ & $0.615_{\pm 0.013}$ & $0.122_{\pm 0.009}$ & $0.149_{\pm 0.010}$ & $0.211_{\pm 0.011}$ & $\mathbf{0.245}_{\pm 0.011}$ \\
ARC-C & $0.568_{\pm 0.014}$ & $0.937_{\pm 0.007}$ & $0.282_{\pm 0.013}$ & $0.400_{\pm 0.014}$ & $0.445_{\pm 0.015}$ & $\mathbf{0.522}_{\pm 0.015}$ \\
RULER & $0.879_{\pm 0.006}$ & $0.998_{\pm 0.001}$ & $0.000_{\pm 0.000}$ & $0.002_{\pm 0.001}$ & $0.681_{\pm 0.010}$ & $\mathbf{0.704}_{\pm 0.009}$ \\
LongBench-v2 & $0.276_{\pm 0.042}$ & $0.431_{\pm 0.046}$ & $0.009_{\pm 0.009}$ & $0.181_{\pm 0.036}$ & $0.284_{\pm 0.042}$ & $\underline{\mathbf{0.362}}_{\pm 0.045}$ \\
LongMemEval & $0.331_{\pm 0.019}$ & $0.643_{\pm 0.020}$ & $0.092_{\pm 0.010}$ & $0.114_{\pm 0.011}$ & $\underline{\mathbf{0.361}}_{\pm 0.019}$ & $0.340_{\pm 0.020}$ \\
\midrule
\multicolumn{7}{l}{\emph{$S \to L$}, Qwen3-1.7B $\to$ Qwen3-4B} \\
MMLU-Pro & $0.379_{\pm 0.013}$ & $0.564_{\pm 0.013}$ & $0.016_{\pm 0.003}$ & $0.226_{\pm 0.011}$ & $\mathbf{0.324}_{\pm 0.012}$ & $0.297_{\pm 0.012}$ \\
ARC-C & $0.770_{\pm 0.012}$ & $0.920_{\pm 0.008}$ & $0.012_{\pm 0.003}$ & $0.595_{\pm 0.014}$ & $\underline{\mathbf{0.794}}_{\pm 0.012}$ & $0.717_{\pm 0.013}$ \\
RULER & $0.980_{\pm 0.004}$ & $0.999_{\pm 0.001}$ & $0.003_{\pm 0.001}$ & $0.007_{\pm 0.002}$ & $0.645_{\pm 0.009}$ & $\mathbf{0.651}_{\pm 0.010}$ \\
LongBench-v2 & $0.319_{\pm 0.043}$ & $0.388_{\pm 0.045}$ & $0.181_{\pm 0.036}$ & $0.284_{\pm 0.042}$ & $0.328_{\pm 0.044}$ & $\mathbf{0.345}_{\pm 0.044}$ \\
LongMemEval & $0.470_{\pm 0.020}$ & $0.633_{\pm 0.019}$ & $0.145_{\pm 0.012}$ & $0.141_{\pm 0.012}$ & $0.445_{\pm 0.020}$ & $\mathbf{0.465}_{\pm 0.020}$ \\
\bottomrule
\end{tabular}
\end{table}

\begin{table}[htbp]
\centering
\footnotesize
\setlength{\tabcolsep}{3pt}
\caption{\textbf{RULER recall by context length}, four cross-layer Qwen3 cells, both
reasoning modes. \emph{small} and \emph{large} are the two models of the pair decoding
their own caches; \emph{depth} and \emph{$R^2$} are the two self-distilled translators,
at seed 42. $\pm$: standard error over evaluation items (Appendix~\ref{app:seeds}).
\underline{Underline}: the translated cache beats the small model,
and in $S \to L$ by more than one standard error; no cell of this table
qualifies. \textbf{Bold}: the better of the two assignments on that row, within each
reasoning mode. The translators are fitted at up to $16$k tokens, so $32$k is
extrapolation. The
closed-form fits score $0.000$--$0.013$ everywhere in this table and are omitted.}
\label{app:tab:xdepthruler}
\resizebox{\ifdim\width>\linewidth\linewidth\else\width\fi}{!}{%
\begin{tabular}{lrrrrrrrr}
\toprule
 & \multicolumn{4}{c}{reasoning-off} & \multicolumn{4}{c}{reasoning-on} \\
\cmidrule(lr){2-5} \cmidrule(lr){6-9}
Context & small & large & depth & $R^2$ & small & large & depth & $R^2$ \\
\midrule
\multicolumn{9}{l}{\emph{$L \to S$}, Qwen3-8B $\to$ Qwen3-0.6B} \\
$4$k & $0.964_{\pm 0.008}$ & $0.998_{\pm 0.003}$ & $0.361_{\pm 0.017}$ & $\mathbf{0.395}_{\pm 0.017}$ & $0.819_{\pm 0.013}$ & $0.992_{\pm 0.004}$ & $\mathbf{0.372}_{\pm 0.018}$ & $0.335_{\pm 0.017}$ \\
$8$k & $0.855_{\pm 0.012}$ & $0.998_{\pm 0.003}$ & $\mathbf{0.352}_{\pm 0.017}$ & $0.333_{\pm 0.016}$ & $0.737_{\pm 0.008}$ & $0.966_{\pm 0.008}$ & $\mathbf{0.348}_{\pm 0.017}$ & $0.307_{\pm 0.017}$ \\
$16$k & $0.819_{\pm 0.012}$ & $0.999_{\pm 0.001}$ & $\mathbf{0.257}_{\pm 0.017}$ & $0.254_{\pm 0.016}$ & $0.708_{\pm 0.011}$ & $0.983_{\pm 0.006}$ & $0.246_{\pm 0.017}$ & $\mathbf{0.274}_{\pm 0.017}$ \\
$32$k & $0.717_{\pm 0.013}$ & $0.996_{\pm 0.003}$ & $0.150_{\pm 0.015}$ & $\mathbf{0.161}_{\pm 0.013}$ & $0.637_{\pm 0.014}$ & $0.986_{\pm 0.006}$ & $\mathbf{0.155}_{\pm 0.015}$ & $0.149_{\pm 0.014}$ \\
\midrule
\multicolumn{9}{l}{\emph{$L \to S$}, Qwen3-4B $\to$ Qwen3-1.7B} \\
$4$k & $0.994_{\pm 0.004}$ & $1.000_{\pm 0.000}$ & $0.506_{\pm 0.019}$ & $\mathbf{0.511}_{\pm 0.018}$ & $0.869_{\pm 0.013}$ & $0.960_{\pm 0.009}$ & $\mathbf{0.336}_{\pm 0.020}$ & $0.336_{\pm 0.020}$ \\
$8$k & $0.981_{\pm 0.006}$ & $1.000_{\pm 0.000}$ & $0.425_{\pm 0.019}$ & $\mathbf{0.483}_{\pm 0.018}$ & $0.865_{\pm 0.015}$ & $0.949_{\pm 0.010}$ & $0.250_{\pm 0.019}$ & $\mathbf{0.359}_{\pm 0.020}$ \\
$16$k & $0.963_{\pm 0.008}$ & $0.998_{\pm 0.003}$ & $0.314_{\pm 0.017}$ & $\mathbf{0.386}_{\pm 0.018}$ & $0.862_{\pm 0.015}$ & $0.955_{\pm 0.009}$ & $0.204_{\pm 0.017}$ & $\mathbf{0.320}_{\pm 0.019}$ \\
$32$k & $0.891_{\pm 0.013}$ & $0.995_{\pm 0.003}$ & $0.146_{\pm 0.014}$ & $\mathbf{0.237}_{\pm 0.017}$ & $0.820_{\pm 0.014}$ & $0.944_{\pm 0.010}$ & $0.107_{\pm 0.011}$ & $\mathbf{0.194}_{\pm 0.015}$ \\
\midrule
\multicolumn{9}{l}{\emph{$S \to L$}, Qwen3-0.6B $\to$ Qwen3-8B} \\
$4$k & $0.964_{\pm 0.008}$ & $0.998_{\pm 0.003}$ & $0.748_{\pm 0.015}$ & $\mathbf{0.754}_{\pm 0.014}$ & $0.819_{\pm 0.013}$ & $0.992_{\pm 0.004}$ & $0.419_{\pm 0.021}$ & $\mathbf{0.476}_{\pm 0.020}$ \\
$8$k & $0.855_{\pm 0.012}$ & $0.998_{\pm 0.003}$ & $0.668_{\pm 0.018}$ & $\mathbf{0.723}_{\pm 0.016}$ & $0.737_{\pm 0.008}$ & $0.966_{\pm 0.008}$ & $0.397_{\pm 0.020}$ & $\mathbf{0.534}_{\pm 0.019}$ \\
$16$k & $0.819_{\pm 0.012}$ & $0.999_{\pm 0.001}$ & $0.627_{\pm 0.018}$ & $\mathbf{0.636}_{\pm 0.018}$ & $0.708_{\pm 0.011}$ & $0.983_{\pm 0.006}$ & $0.447_{\pm 0.018}$ & $\mathbf{0.493}_{\pm 0.018}$ \\
$32$k & $0.717_{\pm 0.013}$ & $0.996_{\pm 0.003}$ & $0.411_{\pm 0.017}$ & $\mathbf{0.421}_{\pm 0.017}$ & $0.637_{\pm 0.014}$ & $0.986_{\pm 0.006}$ & $0.312_{\pm 0.018}$ & $\mathbf{0.388}_{\pm 0.017}$ \\
\midrule
\multicolumn{9}{l}{\emph{$S \to L$}, Qwen3-1.7B $\to$ Qwen3-4B} \\
$4$k & $0.994_{\pm 0.004}$ & $1.000_{\pm 0.000}$ & $0.697_{\pm 0.015}$ & $\mathbf{0.698}_{\pm 0.016}$ & $0.869_{\pm 0.013}$ & $0.960_{\pm 0.009}$ & $0.446_{\pm 0.020}$ & $\mathbf{0.457}_{\pm 0.022}$ \\
$8$k & $0.981_{\pm 0.006}$ & $1.000_{\pm 0.000}$ & $0.646_{\pm 0.016}$ & $\mathbf{0.664}_{\pm 0.017}$ & $0.865_{\pm 0.015}$ & $0.949_{\pm 0.010}$ & $0.366_{\pm 0.020}$ & $\mathbf{0.413}_{\pm 0.020}$ \\
$16$k & $0.963_{\pm 0.008}$ & $0.998_{\pm 0.003}$ & $\mathbf{0.593}_{\pm 0.017}$ & $0.593_{\pm 0.019}$ & $0.862_{\pm 0.015}$ & $0.955_{\pm 0.009}$ & $0.368_{\pm 0.020}$ & $\mathbf{0.391}_{\pm 0.020}$ \\
$32$k & $0.891_{\pm 0.013}$ & $0.995_{\pm 0.003}$ & $0.378_{\pm 0.016}$ & $\mathbf{0.422}_{\pm 0.018}$ & $0.820_{\pm 0.014}$ & $0.944_{\pm 0.010}$ & $0.319_{\pm 0.017}$ & $\mathbf{0.346}_{\pm 0.018}$ \\
\bottomrule
\end{tabular}}
\end{table}

\begin{table}[htbp]
\centering
\footnotesize
\setlength{\tabcolsep}{5pt}
\caption{\textbf{LongMemEval with reasoning on}, four cross-layer Qwen3 cells. Judge
accuracy pooled over the oracle, $8$k, $16$k and $32$k rungs, on the $389$ questions common
to all four ($n = 1556$ question-rung pairs). \underline{Underline}: the translated cache
beats the small model, and in $S \to L$ by more than one standard error.
\textbf{Bold}: the best of the four translators on that row. Self-distilled columns are
seed 42. $\pm$: standard error over evaluation items (Appendix~\ref{app:seeds}).}
\label{app:tab:xdepthlme}
\resizebox{\ifdim\width>\linewidth\linewidth\else\width\fi}{!}{%
\begin{tabular}{lrrrrrr}
\toprule
 & \multicolumn{2}{c}{native} & \multicolumn{2}{c}{closed form} & \multicolumn{2}{c}{self-distilled} \\
\cmidrule(lr){2-3} \cmidrule(lr){4-5} \cmidrule(lr){6-7}
Cell & small & large & depth & $R^2$ & depth & $R^2$ \\
\midrule
Qwen3-8B $\to$ Qwen3-0.6B & $0.383_{\pm 0.019}$ & $0.763_{\pm 0.017}$ & $0.000_{\pm 0.000}$ & $0.000_{\pm 0.000}$ & $\mathbf{0.373}_{\pm 0.019}$ & $0.343_{\pm 0.018}$ \\
Qwen3-4B $\to$ Qwen3-1.7B & $0.531_{\pm 0.019}$ & $0.723_{\pm 0.017}$ & $0.000_{\pm 0.000}$ & $0.000_{\pm 0.000}$ & $\mathbf{0.453}_{\pm 0.019}$ & $0.453_{\pm 0.018}$ \\
Qwen3-0.6B $\to$ Qwen3-8B & $0.383_{\pm 0.019}$ & $0.763_{\pm 0.017}$ & $0.006_{\pm 0.002}$ & $0.004_{\pm 0.002}$ & $\underline{0.506}_{\pm 0.020}$ & $\underline{\mathbf{0.531}}_{\pm 0.019}$ \\
Qwen3-1.7B $\to$ Qwen3-4B & $0.531_{\pm 0.019}$ & $0.723_{\pm 0.017}$ & $0.002_{\pm 0.001}$ & $0.003_{\pm 0.001}$ & $\underline{0.566}_{\pm 0.019}$ & $\underline{\mathbf{0.585}}_{\pm 0.019}$ \\
\bottomrule
\end{tabular}}
\end{table}

\paragraph{Head-mixing against head-wise translators.}
Figure~\ref{fig:head_wise_vs_head_mixing} sets the head-mixing KV-Lingo translator against its
head-wise counterpart on the Qwen3-8B/0.6B pair, at every $\nu$ of the sweep.

\begin{figure}[htbp]
    \centering
    \includegraphics[width=\linewidth]{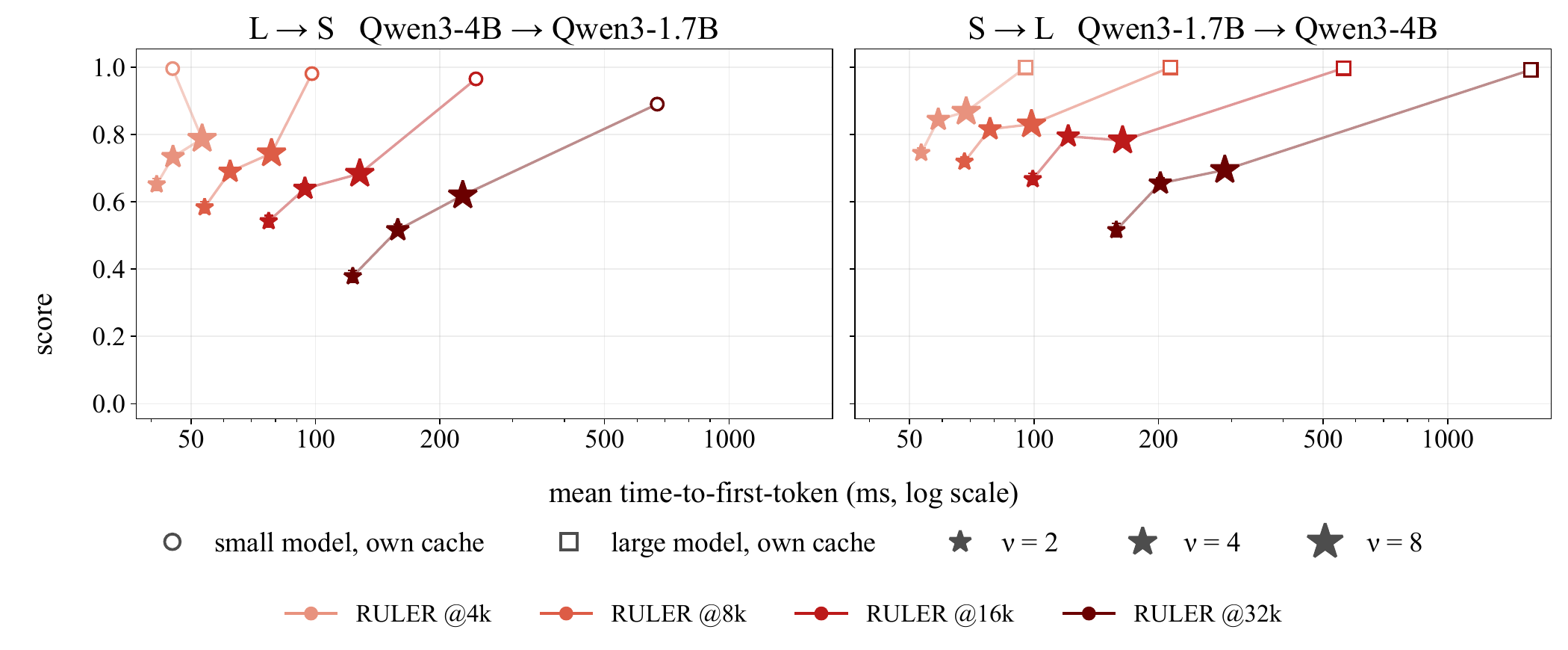}\\[1.5ex]
    \includegraphics[width=\linewidth]{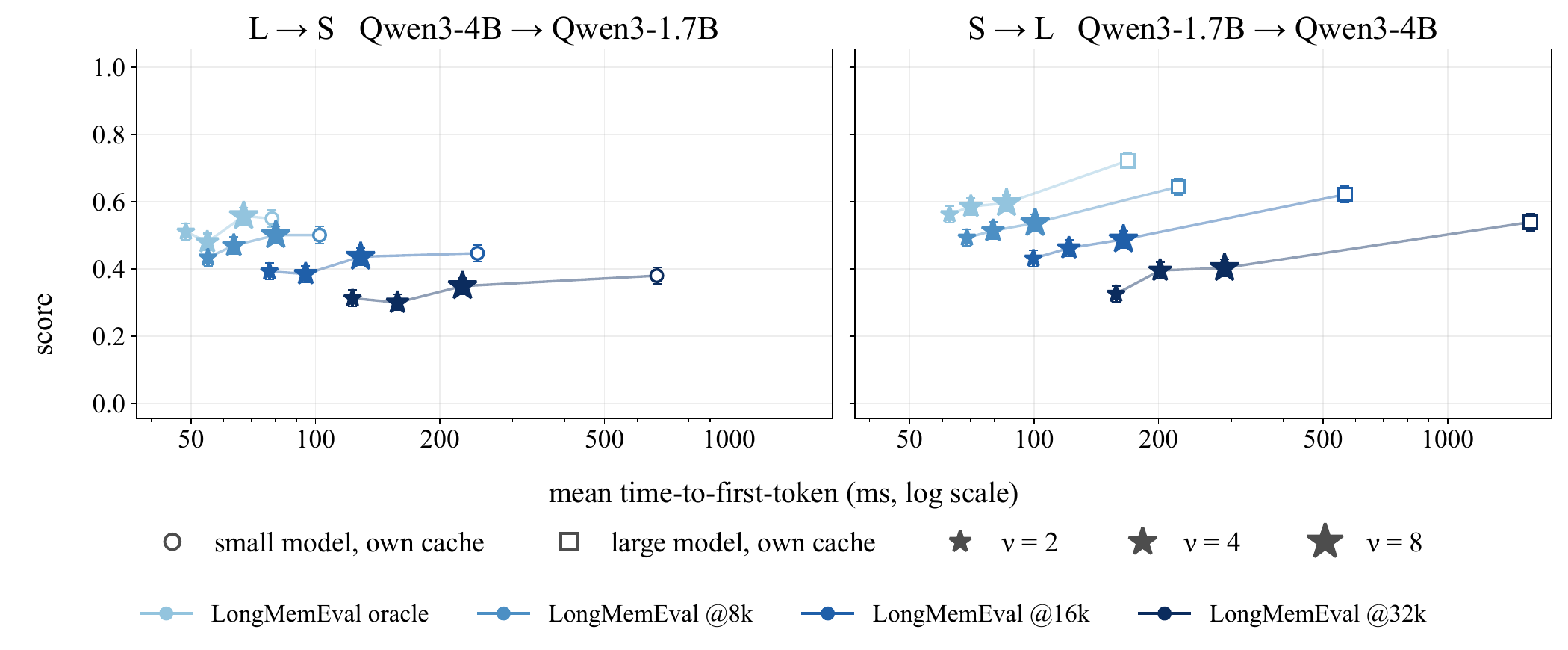}
    \caption{\textbf{KV-Lingo score against time to first token, Qwen3-$4$B/Qwen3-$1.7$B, $R^2$
    assignment, reasoning-off.} RULER above and LongMemEval below, sharing an $x$ window:
    $x$ is the mean time to first token over the eval's own prefix lengths, $y$ the eval's
    own metric, and colour the context budget, light at the shortest rung to dark at
    $32$k. Filled stars are the translated cache at $\nu = 2, 4, 8$ source layers per
    target layer, growing with $\nu$, each a single seed (seed 56), charged the translation and
    one decoding step (Appendix~\ref{app:cost}); open markers are the decoding
    model on its own cache, the only native baseline each panel carries. Error bars: standard
    error over evaluation items (Appendix~\ref{app:seeds}).}
    \label{fig:nsrc_ttft}
\end{figure}

\begin{figure}[tbp]
    \centering
    \includegraphics[width=0.9\linewidth]{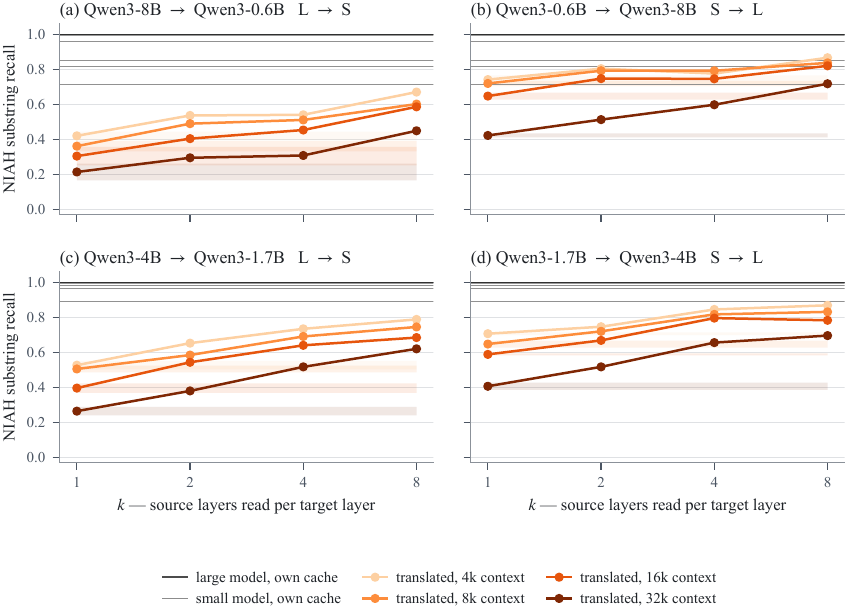}
    \caption{\textbf{RULER recall against the number of source layers per target layer}
    $\nu$, the four cross-layer Qwen3 cells, head-mixing
    KV-Lingo with the $R^2$ assignment, reasoning-off. Colour is the context budget, light at
    $4$k to dark at $32$k. The black and grey hairlines are the large and the small model on
    their own cache, one line per budget. Every point is a single seed, seed 42 at $\nu = 1$
    and seed 56 at $\nu \in \{2, 4, 8\}$, and the shaded band carries the standard error over
    evaluation items (Appendix~\ref{app:seeds}).}
    \label{fig:n_src_cross_depth}
\end{figure}

\paragraph{The number of source layers.}
Widening the assignment $\mathcal{E}_\ell$ of \S\ref{sec:method:family} so that every target
layer reads $\nu$ source layers instead of one costs $\nu$ matrices per target layer, to
store and to apply at every translation. Figure~\ref{fig:nsrc_ttft} reads the resulting
trade-off on the two long-context evaluations at once: translation gets more expensive as
$\nu$ grows from $2$ to $8$, so a cell's markers move right, and what the extra source
layers buy is the vertical distance they cover. RULER is where they buy the most, gaining
recall at every context budget in both directions; on LongMemEval the same widening moves
the score by little. How much cheaper than re-prefilling the widest map still is depends on
the context budget: at $32$k every translated marker sits far to the left of the decoding
model's own prefill. On the shortest RULER rung, the $\nu = 8$ marker of the 4B $\to$ 1.7B
direction sits level with the re-prefill, but more than half of its time there is the
timing harness's fixed decoding step of about $30$\,ms, which does not depend on $\nu$
(Appendix~\ref{app:cost}): the translation itself takes about $25$\,ms, against $49$\,ms for
the 1.7B's prefill.
Figure~\ref{fig:n_src_cross_depth} draws the RULER side of this sweep for all four
cross-layer cells, rung by rung, and Table~\ref{app:tab:nsrc} gives $\nu \in \{1, 2, 4, 8\}$
on MMLU-Pro, ARC-C, RULER and LongMemEval for the same four cells.

\begin{table}[tbp]
\centering
\footnotesize
\setlength{\tabcolsep}{5pt}
\caption{\textbf{Sweeping the number of source layers per target layer} on the four
cross-layer Qwen3 cells of Table~\ref{tab:xdepth}: $\nu \in \{1, 2, 4, 8\}$, the $R^2$
assignment, self-distilled, reasoning-off. The $\nu = 1$ column is the seed-42 run of
Table~\ref{tab:xdepth}; each wider map is a single seed, seed 56, the only seed trained.
$\pm$: standard error over evaluation items (Appendix~\ref{app:seeds}).
\underline{Underline}: the translated cache
beats the small model, and in $S \to L$ by more than one standard error. \textbf{Bold}: the best $\nu$ on that row.
MMLU-Pro is the macro-average over its $14$ categories and ARC-C the micro-average, both
lenient generate-and-parse; RULER is recall averaged over $4$k--$32$k; LongMemEval is
judge accuracy pooled over its oracle, $8$k, $16$k and $32$k rungs.
}
\label{app:tab:nsrc}
\resizebox{\ifdim\width>\linewidth\linewidth\else\width\fi}{!}{%
\begin{tabular}{lrrrrrr}
\toprule
 & \multicolumn{2}{c}{native} & \multicolumn{4}{c}{translated, $R^2$ assignment} \\
\cmidrule(lr){2-3} \cmidrule(lr){4-7}
Benchmark & small & large & $\nu = 1$ & $\nu = 2$ & $\nu = 4$ & $\nu = 8$ \\
\midrule
\multicolumn{7}{l}{\emph{$L \to S$}, Qwen3-8B $\to$ Qwen3-0.6B} \\
MMLU-Pro & $0.234_{\pm 0.011}$ & $0.616_{\pm 0.013}$ & $0.208_{\pm 0.011}$ & $\underline{\mathbf{0.367}}_{\pm 0.012}$ & $\underline{0.274}_{\pm 0.012}$ & $\underline{0.320}_{\pm 0.012}$ \\
ARC-C & $0.564_{\pm 0.014}$ & $0.939_{\pm 0.007}$ & $\underline{0.659}_{\pm 0.014}$ & $\underline{\mathbf{0.838}}_{\pm 0.011}$ & $\underline{0.826}_{\pm 0.011}$ & $\underline{0.832}_{\pm 0.011}$ \\
RULER & $0.834_{\pm 0.006}$ & $0.997_{\pm 0.001}$ & $0.286_{\pm 0.008}$ & $0.431_{\pm 0.009}$ & $0.453_{\pm 0.009}$ & $\mathbf{0.576}_{\pm 0.008}$ \\
LongMemEval & $0.333_{\pm 0.019}$ & $0.652_{\pm 0.019}$ & $0.289_{\pm 0.019}$ & $\underline{0.337}_{\pm 0.020}$ & $\underline{0.368}_{\pm 0.020}$ & $\underline{\mathbf{0.392}}_{\pm 0.021}$ \\
\midrule
\multicolumn{7}{l}{\emph{$L \to S$}, Qwen3-4B $\to$ Qwen3-1.7B} \\
MMLU-Pro & $0.386_{\pm 0.013}$ & $0.546_{\pm 0.013}$ & $\mathbf{0.286}_{\pm 0.011}$ & $0.211_{\pm 0.011}$ & $0.250_{\pm 0.011}$ & $0.281_{\pm 0.012}$ \\
ARC-C & $0.772_{\pm 0.012}$ & $0.919_{\pm 0.008}$ & $\underline{0.811}_{\pm 0.011}$ & $0.725_{\pm 0.013}$ & $\underline{0.784}_{\pm 0.012}$ & $\underline{\mathbf{0.823}}_{\pm 0.011}$ \\
RULER & $0.958_{\pm 0.004}$ & $0.998_{\pm 0.001}$ & $0.404_{\pm 0.009}$ & $0.539_{\pm 0.009}$ & $0.645_{\pm 0.007}$ & $\mathbf{0.709}_{\pm 0.007}$ \\
LongMemEval & $0.470_{\pm 0.021}$ & $0.632_{\pm 0.020}$ & $0.410_{\pm 0.019}$ & $0.413_{\pm 0.020}$ & $0.409_{\pm 0.019}$ & $\mathbf{0.461}_{\pm 0.020}$ \\
\midrule
\multicolumn{7}{l}{\emph{$S \to L$}, Qwen3-0.6B $\to$ Qwen3-8B} \\
MMLU-Pro & $0.234_{\pm 0.011}$ & $0.616_{\pm 0.013}$ & $\underline{0.245}_{\pm 0.011}$ & $\underline{0.261}_{\pm 0.012}$ & $\underline{0.250}_{\pm 0.011}$ & $\underline{\mathbf{0.287}}_{\pm 0.012}$ \\
ARC-C & $0.564_{\pm 0.014}$ & $0.939_{\pm 0.007}$ & $0.522_{\pm 0.015}$ & $0.561_{\pm 0.015}$ & $0.549_{\pm 0.015}$ & $\mathbf{0.562}_{\pm 0.014}$ \\
RULER & $0.834_{\pm 0.006}$ & $0.997_{\pm 0.001}$ & $0.633_{\pm 0.008}$ & $0.714_{\pm 0.007}$ & $0.728_{\pm 0.007}$ & $\mathbf{0.810}_{\pm 0.006}$ \\
LongMemEval & $0.333_{\pm 0.019}$ & $0.650_{\pm 0.020}$ & $0.340_{\pm 0.020}$ & $\underline{0.404}_{\pm 0.019}$ & $\underline{0.385}_{\pm 0.019}$ & $\underline{\mathbf{0.416}}_{\pm 0.020}$ \\
\midrule
\multicolumn{7}{l}{\emph{$S \to L$}, Qwen3-1.7B $\to$ Qwen3-4B} \\
MMLU-Pro & $0.386_{\pm 0.013}$ & $0.546_{\pm 0.013}$ & $0.297_{\pm 0.012}$ & $0.348_{\pm 0.013}$ & $0.346_{\pm 0.013}$ & $\underline{\mathbf{0.405}}_{\pm 0.013}$ \\
ARC-C & $0.772_{\pm 0.012}$ & $0.919_{\pm 0.008}$ & $0.717_{\pm 0.013}$ & $0.763_{\pm 0.012}$ & $0.746_{\pm 0.013}$ & $\mathbf{0.770}_{\pm 0.012}$ \\
RULER & $0.958_{\pm 0.004}$ & $0.998_{\pm 0.001}$ & $0.594_{\pm 0.009}$ & $0.662_{\pm 0.008}$ & $0.778_{\pm 0.006}$ & $\mathbf{0.794}_{\pm 0.006}$ \\
LongMemEval & $0.470_{\pm 0.021}$ & $0.632_{\pm 0.020}$ & $0.465_{\pm 0.020}$ & $0.454_{\pm 0.020}$ & $0.490_{\pm 0.020}$ & $\underline{\mathbf{0.506}}_{\pm 0.020}$ \\
\bottomrule
\end{tabular}}
\end{table}

\clearpage
\section{Gemma-3-4B/12B-it}
\label{app:gemma3}
\suppressfloats[t]

\begin{figure}[!htbp]
    \centering
    \includegraphics[width=0.9\linewidth]{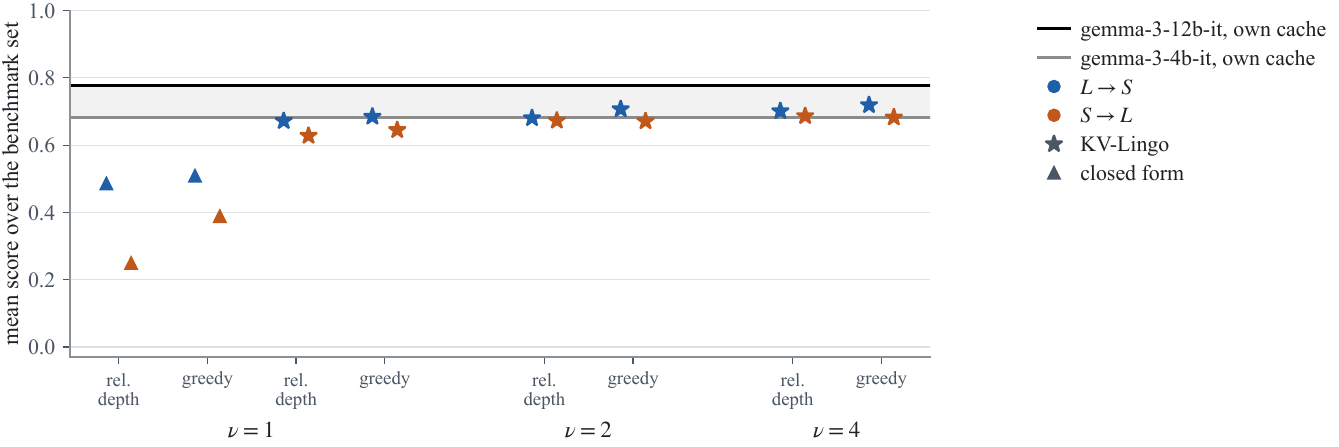}
    \caption{Cache translation with KV-Lingo (head-mixing) with the pair Gemma-3-4B/12B-it, both directions, with $\nu \in \{1, 2, 4\}$ source layers per target layer. Averaged score over MMLU-Pro, ARC-C, LongBench-v2, GSM8K, CoQA, IFEval, XQuAD, RULER at $4$k, $8$k and $16$k, and RepLiQA. MT-Bench-101 is left out of the average; Tables~\ref{tab:gemma3} and~\ref{tab:gemma3nu} report it. LongBench-v2 is kept in the average although Table~\ref{tab:gemma3} omits it, since its native anchors sit near chance at this pair's $16$k budget. The $\nu = 1$ closed-form fits and self-distilled depth map are those of Table~\ref{tab:gemma3}; every other self-distilled point is a single map from Table~\ref{tab:gemma3nu}. Every self-distilled point, the $\nu = 1$ depth map included, is seed 42; error bars: standard error over evaluation items (Appendix~\ref{app:seeds}).}
    \label{fig:gemma-3}
\end{figure}

Table~\ref{tab:gemma3} reports cache translation on the Gemma-3-4B/12B-it pair with one
source layer per target layer, Table~\ref{tab:gemma3nu} widens the maps to $\nu \in \{1, 2, 4\}$
source layers, and Figure~\ref{fig:gemma-3} draws the mean score of both.

\begin{table}[tbp]
\centering
\footnotesize
\setlength{\tabcolsep}{5pt}
\caption{\textbf{Cross-layer cache translation on the Gemma-3-4B/12B-it pair}, $34$ against
$48$ layers and a KV width of $1024$ against $2048$, both directions. One rectangular map
covers the stack; each target layer reads a single source layer chosen inside its own
attention group, either by relative depth or greedy match, and each assignment is reported as
its closed-form fit and after self-distillation, the latter seed 42. $\pm$: standard error
over evaluation items (Appendix~\ref{app:seeds}).
\underline{Underline}: the translated cache beats the small model, and in $S \to L$ by more
than one standard error. \textbf{Bold}: the best of the four translators on that
row. MMLU-Pro and ARC-C are lenient generate-and-parse accuracy, GSM8K flexible exact
match, CoQA and XQuAD token-F1, IFEval prompt-level strict accuracy, RepLiQA judged
accuracy on the answerable questions, MT-Bench-101 a $1$--$10$ judge score, and
RULER is recall at each of three prefix lengths. Gemma-3's chat template carries no
reasoning switch, so the suite has a single mode rather than the reasoning-on/off pair
reported elsewhere. Three points to note. The 4B outscores the 12B on RepLiQA, so on that
row the small model is the harder anchor. On the $S \to L$
MT-Bench-101 row both maps print $9.03$, but only the depth map clears the 4B's $8.883$ by one
standard error ($9.033 - 0.148 = 8.884$, against $9.031 - 0.153 = 8.878$).
And the two blocks' native anchors come from byte-identical generations that on IFEval alone
score differently in the third decimal ($0.715$ against $0.717$ for the 4B, $0.762$ against
$0.765$ for the 12B), because IFEval's language-detection check was unseeded; the block's own
pair is the one its rows are read against. LongBench-v2 is
omitted: at the $16384$-token budget this pair was evaluated at, $94\%$ of its items are
truncated and its $L \to S$ native anchor sits half a standard deviation above the
four-way chance floor, so no delta against native is readable from it.}
\label{tab:gemma3}
\begin{tabular}{lrrrrrr}
\toprule
 & \multicolumn{2}{c}{native} & \multicolumn{2}{c}{closed form} & \multicolumn{2}{c}{self-distilled} \\
\cmidrule(lr){2-3} \cmidrule(lr){4-5} \cmidrule(lr){6-7}
Benchmark & small & large & depth & greedy & depth & greedy \\
\midrule
\multicolumn{7}{l}{\emph{$L \to S$}, Gemma-3-4B/12B-it} \\
MMLU-Pro & $0.400_{\pm 0.013}$ & $0.584_{\pm 0.013}$ & $0.315_{\pm 0.012}$ & $0.198_{\pm 0.011}$ & $\underline{\mathbf{0.429}}_{\pm 0.013}$ & $0.391_{\pm 0.013}$ \\
ARC-C & $0.791_{\pm 0.012}$ & $0.933_{\pm 0.007}$ & $0.665_{\pm 0.014}$ & $0.682_{\pm 0.014}$ & $\underline{\mathbf{0.830}}_{\pm 0.011}$ & $0.773_{\pm 0.012}$ \\
GSM8K & $0.871_{\pm 0.021}$ & $0.883_{\pm 0.020}$ & $0.738_{\pm 0.028}$ & $0.680_{\pm 0.029}$ & $0.855_{\pm 0.022}$ & $\mathbf{0.871}_{\pm 0.021}$ \\
CoQA & $0.607_{\pm 0.027}$ & $0.657_{\pm 0.026}$ & $\underline{0.644}_{\pm 0.026}$ & $\underline{0.608}_{\pm 0.027}$ & $0.541_{\pm 0.027}$ & $\underline{\mathbf{0.706}}_{\pm 0.025}$ \\
IFEval & $0.715_{\pm 0.019}$ & $0.762_{\pm 0.018}$ & $0.396_{\pm 0.021}$ & $0.442_{\pm 0.021}$ & $0.590_{\pm 0.021}$ & $\mathbf{0.610}_{\pm 0.021}$ \\
XQuAD & $0.443_{\pm 0.007}$ & $0.629_{\pm 0.006}$ & $0.243_{\pm 0.007}$ & $0.210_{\pm 0.007}$ & $0.428_{\pm 0.007}$ & $\underline{\mathbf{0.522}}_{\pm 0.008}$ \\
RepLiQA & $0.908_{\pm 0.015}$ & $0.899_{\pm 0.016}$ & $0.570_{\pm 0.026}$ & $0.804_{\pm 0.021}$ & $0.858_{\pm 0.018}$ & $\underline{\mathbf{0.916}}_{\pm 0.015}$ \\
MT-Bench-101 & $8.88_{\pm 0.16}$ & $9.29_{\pm 0.12}$ & $8.57_{\pm 0.18}$ & $8.47_{\pm 0.17}$ & $8.64_{\pm 0.19}$ & $\underline{\mathbf{8.96}}_{\pm 0.17}$ \\
RULER, $4$k & $0.983_{\pm 0.006}$ & $1.000_{\pm 0.000}$ & $0.688_{\pm 0.018}$ & $0.710_{\pm 0.016}$ & $0.955_{\pm 0.009}$ & $\mathbf{0.981}_{\pm 0.007}$ \\
RULER, $8$k & $0.823_{\pm 0.014}$ & $0.964_{\pm 0.008}$ & $0.469_{\pm 0.020}$ & $0.578_{\pm 0.017}$ & $0.812_{\pm 0.015}$ & $\underline{\mathbf{0.832}}_{\pm 0.015}$ \\
RULER, $16$k & $0.700_{\pm 0.012}$ & $0.915_{\pm 0.012}$ & $0.341_{\pm 0.019}$ & $0.446_{\pm 0.017}$ & $0.682_{\pm 0.014}$ & $\underline{\mathbf{0.728}}_{\pm 0.014}$ \\
\midrule
\multicolumn{7}{l}{\emph{$S \to L$}, Gemma-3-4B/12B-it} \\
MMLU-Pro & $0.400_{\pm 0.013}$ & $0.584_{\pm 0.013}$ & $0.242_{\pm 0.011}$ & $0.256_{\pm 0.011}$ & $\mathbf{0.386}_{\pm 0.013}$ & $0.380_{\pm 0.013}$ \\
ARC-C & $0.791_{\pm 0.012}$ & $0.933_{\pm 0.007}$ & $0.484_{\pm 0.015}$ & $0.416_{\pm 0.014}$ & $\mathbf{0.750}_{\pm 0.013}$ & $0.747_{\pm 0.013}$ \\
GSM8K & $0.871_{\pm 0.021}$ & $0.883_{\pm 0.020}$ & $0.508_{\pm 0.031}$ & $0.523_{\pm 0.031}$ & $0.855_{\pm 0.022}$ & $\mathbf{0.879}_{\pm 0.020}$ \\
CoQA & $0.607_{\pm 0.027}$ & $0.657_{\pm 0.026}$ & $0.404_{\pm 0.026}$ & $0.401_{\pm 0.027}$ & $0.518_{\pm 0.028}$ & $\mathbf{0.538}_{\pm 0.028}$ \\
IFEval & $0.717_{\pm 0.019}$ & $0.765_{\pm 0.018}$ & $0.372_{\pm 0.021}$ & $0.399_{\pm 0.021}$ & $0.604_{\pm 0.021}$ & $\mathbf{0.634}_{\pm 0.021}$ \\
XQuAD & $0.443_{\pm 0.007}$ & $0.629_{\pm 0.006}$ & $0.071_{\pm 0.004}$ & $0.086_{\pm 0.004}$ & $\mathbf{0.447}_{\pm 0.008}$ & $0.427_{\pm 0.008}$ \\
RepLiQA & $0.908_{\pm 0.015}$ & $0.899_{\pm 0.016}$ & $0.176_{\pm 0.020}$ & $0.651_{\pm 0.025}$ & $0.821_{\pm 0.020}$ & $\mathbf{0.857}_{\pm 0.019}$ \\
MT-Bench-101 & $8.88_{\pm 0.16}$ & $9.29_{\pm 0.12}$ & $8.38_{\pm 0.22}$ & $8.41_{\pm 0.21}$ & $\underline{\mathbf{9.03}}_{\pm 0.15}$ & $9.03_{\pm 0.15}$ \\
RULER, $4$k & $0.983_{\pm 0.006}$ & $1.000_{\pm 0.000}$ & $0.095_{\pm 0.012}$ & $0.475_{\pm 0.020}$ & $0.945_{\pm 0.011}$ & $\mathbf{0.958}_{\pm 0.010}$ \\
RULER, $8$k & $0.823_{\pm 0.014}$ & $0.964_{\pm 0.008}$ & $0.080_{\pm 0.011}$ & $0.409_{\pm 0.020}$ & $0.733_{\pm 0.014}$ & $\mathbf{0.777}_{\pm 0.014}$ \\
RULER, $16$k & $0.700_{\pm 0.012}$ & $0.915_{\pm 0.012}$ & $0.113_{\pm 0.014}$ & $0.389_{\pm 0.017}$ & $0.661_{\pm 0.013}$ & $\mathbf{0.679}_{\pm 0.013}$ \\
\bottomrule
\end{tabular}
\end{table}

\begin{table}[tbp]
\centering
\footnotesize
\setlength{\tabcolsep}{2pt}
\caption{\textbf{Wider maps on the Gemma-3-4B/12B-it pair}: each target layer reads
$\nu$ source layers inside its own attention group, both directions, self-distilled.
\emph{depth} extends the relative-depth assignment of Table~\ref{tab:gemma3} to $\nu$
neighbouring layers; \emph{greedy} picks the $\nu$ source layers by orthogonal matching
pursuit on the cache reconstruction, and at $\nu = 1$ is therefore a different assignment
from the greedy match of Table~\ref{tab:gemma3}. Each cell is a single map (seed 42), trained at the
learning rate that won its own cell's sweep. $\pm$: standard error over evaluation items
(Appendix~\ref{app:seeds}). \underline{Underline}: the translated cache beats
the small model, and in $S \to L$ by more than one standard error. \textbf{Bold}: the best of the five maps on that row. Native anchors,
metrics and eval sets are those of Table~\ref{tab:gemma3}. No closed-form fits were evaluated at
$\nu > 1$.}
\label{tab:gemma3nu}
\resizebox{\ifdim\width>\linewidth\linewidth\else\width\fi}{!}{%
\begin{tabular}{lrrrrrrr}
\toprule
 & \multicolumn{2}{c}{native} & \multicolumn{2}{c}{depth} & \multicolumn{3}{c}{greedy} \\
\cmidrule(lr){2-3} \cmidrule(lr){4-5} \cmidrule(lr){6-8}
Benchmark & small & large & $\nu = 2$ & $\nu = 4$ & $\nu = 1$ & $\nu = 2$ & $\nu = 4$ \\
\midrule
\multicolumn{8}{l}{\emph{$L \to S$}, Gemma-3-4B/12B-it} \\
MMLU-Pro & $0.400_{\pm 0.013}$ & $0.584_{\pm 0.013}$ & $\underline{\mathbf{0.431}}_{\pm 0.013}$ & $\underline{0.429}_{\pm 0.013}$ & $\underline{0.416}_{\pm 0.013}$ & $\underline{0.412}_{\pm 0.013}$ & $\underline{0.429}_{\pm 0.013}$ \\
ARC-C & $0.791_{\pm 0.012}$ & $0.933_{\pm 0.007}$ & $\underline{\mathbf{0.846}}_{\pm 0.011}$ & $\underline{0.838}_{\pm 0.011}$ & $\underline{0.832}_{\pm 0.011}$ & $\underline{0.828}_{\pm 0.011}$ & $\underline{0.841}_{\pm 0.011}$ \\
GSM8K & $0.871_{\pm 0.021}$ & $0.883_{\pm 0.020}$ & $0.844_{\pm 0.023}$ & $0.867_{\pm 0.021}$ & $0.863_{\pm 0.022}$ & $0.863_{\pm 0.022}$ & $\underline{\mathbf{0.902}}_{\pm 0.019}$ \\
CoQA & $0.607_{\pm 0.027}$ & $0.657_{\pm 0.026}$ & $0.563_{\pm 0.028}$ & $0.556_{\pm 0.028}$ & $\underline{0.661}_{\pm 0.026}$ & $\underline{\mathbf{0.689}}_{\pm 0.025}$ & $\underline{0.686}_{\pm 0.025}$ \\
IFEval & $0.715_{\pm 0.019}$ & $0.762_{\pm 0.018}$ & $0.625_{\pm 0.021}$ & $\mathbf{0.654}_{\pm 0.021}$ & $0.591_{\pm 0.021}$ & $0.632_{\pm 0.021}$ & $0.643_{\pm 0.021}$ \\
XQuAD & $0.443_{\pm 0.007}$ & $0.629_{\pm 0.006}$ & $\underline{0.484}_{\pm 0.007}$ & $\underline{0.504}_{\pm 0.007}$ & $\underline{0.448}_{\pm 0.007}$ & $\underline{\mathbf{0.577}}_{\pm 0.007}$ & $\underline{0.527}_{\pm 0.007}$ \\
RepLiQA & $0.908_{\pm 0.015}$ & $0.899_{\pm 0.016}$ & $0.885_{\pm 0.017}$ & $\underline{0.913}_{\pm 0.015}$ & $0.891_{\pm 0.016}$ & $0.908_{\pm 0.015}$ & $\underline{\mathbf{0.922}}_{\pm 0.014}$ \\
MT-Bench-101 & $8.88_{\pm 0.16}$ & $9.29_{\pm 0.12}$ & $8.75_{\pm 0.18}$ & $8.80_{\pm 0.16}$ & $\mathbf{8.83}_{\pm 0.17}$ & $8.73_{\pm 0.19}$ & $8.73_{\pm 0.18}$ \\
RULER, $4$k & $0.983_{\pm 0.006}$ & $1.000_{\pm 0.000}$ & $\underline{0.991}_{\pm 0.004}$ & $\underline{0.998}_{\pm 0.001}$ & $0.971_{\pm 0.008}$ & $\underline{0.989}_{\pm 0.004}$ & $\underline{\mathbf{0.999}}_{\pm 0.001}$ \\
RULER, $8$k & $0.823_{\pm 0.014}$ & $0.964_{\pm 0.008}$ & $\underline{0.838}_{\pm 0.015}$ & $\underline{0.909}_{\pm 0.013}$ & $\underline{0.856}_{\pm 0.015}$ & $\underline{0.887}_{\pm 0.013}$ & $\underline{\mathbf{0.922}}_{\pm 0.012}$ \\
RULER, $16$k & $0.700_{\pm 0.012}$ & $0.915_{\pm 0.012}$ & $\underline{0.718}_{\pm 0.015}$ & $\underline{\mathbf{0.786}}_{\pm 0.014}$ & $\underline{0.729}_{\pm 0.014}$ & $\underline{0.749}_{\pm 0.014}$ & $\underline{0.772}_{\pm 0.014}$ \\
\midrule
\multicolumn{8}{l}{\emph{$S \to L$}, Gemma-3-4B/12B-it} \\
MMLU-Pro & $0.400_{\pm 0.013}$ & $0.584_{\pm 0.013}$ & $\underline{0.430}_{\pm 0.013}$ & $\underline{\mathbf{0.439}}_{\pm 0.013}$ & $0.404_{\pm 0.013}$ & $\underline{0.423}_{\pm 0.013}$ & $\underline{0.425}_{\pm 0.013}$ \\
ARC-C & $0.791_{\pm 0.012}$ & $0.933_{\pm 0.007}$ & $0.788_{\pm 0.012}$ & $\mathbf{0.794}_{\pm 0.012}$ & $0.758_{\pm 0.013}$ & $0.765_{\pm 0.012}$ & $0.760_{\pm 0.012}$ \\
GSM8K & $0.871_{\pm 0.021}$ & $0.883_{\pm 0.020}$ & $0.867_{\pm 0.021}$ & $0.887_{\pm 0.020}$ & $\underline{0.895}_{\pm 0.019}$ & $\underline{\mathbf{0.902}}_{\pm 0.019}$ & $0.891_{\pm 0.020}$ \\
CoQA & $0.607_{\pm 0.027}$ & $0.657_{\pm 0.026}$ & $0.543_{\pm 0.028}$ & $0.555_{\pm 0.028}$ & $0.508_{\pm 0.028}$ & $0.538_{\pm 0.028}$ & $\mathbf{0.569}_{\pm 0.028}$ \\
IFEval & $0.717_{\pm 0.019}$ & $0.765_{\pm 0.018}$ & $\mathbf{0.647}_{\pm 0.021}$ & $0.645_{\pm 0.021}$ & $0.636_{\pm 0.021}$ & $0.641_{\pm 0.021}$ & $0.640_{\pm 0.021}$ \\
XQuAD & $0.443_{\pm 0.007}$ & $0.629_{\pm 0.006}$ & $\underline{0.491}_{\pm 0.008}$ & $\underline{0.489}_{\pm 0.008}$ & $0.445_{\pm 0.008}$ & $\underline{0.475}_{\pm 0.008}$ & $\underline{\mathbf{0.499}}_{\pm 0.007}$ \\
RepLiQA & $0.908_{\pm 0.015}$ & $0.899_{\pm 0.016}$ & $0.841_{\pm 0.019}$ & $\mathbf{0.883}_{\pm 0.017}$ & $0.754_{\pm 0.023}$ & $0.872_{\pm 0.018}$ & $0.860_{\pm 0.018}$ \\
MT-Bench-101 & $8.88_{\pm 0.16}$ & $9.29_{\pm 0.12}$ & $8.94_{\pm 0.18}$ & $9.03_{\pm 0.16}$ & $8.86_{\pm 0.17}$ & $8.83_{\pm 0.17}$ & $\underline{\mathbf{9.13}}_{\pm 0.15}$ \\
RULER, $4$k & $0.983_{\pm 0.006}$ & $1.000_{\pm 0.000}$ & $0.970_{\pm 0.008}$ & $0.983_{\pm 0.006}$ & $0.951_{\pm 0.010}$ & $0.975_{\pm 0.008}$ & $\mathbf{0.984}_{\pm 0.006}$ \\
RULER, $8$k & $0.823_{\pm 0.014}$ & $0.964_{\pm 0.008}$ & $0.808_{\pm 0.013}$ & $\underline{0.841}_{\pm 0.013}$ & $0.770_{\pm 0.014}$ & $0.812_{\pm 0.014}$ & $\underline{\mathbf{0.844}}_{\pm 0.014}$ \\
RULER, $16$k & $0.700_{\pm 0.012}$ & $0.915_{\pm 0.012}$ & $\underline{0.723}_{\pm 0.013}$ & $\underline{\mathbf{0.747}}_{\pm 0.011}$ & $0.686_{\pm 0.013}$ & $\underline{0.713}_{\pm 0.012}$ & $\underline{0.741}_{\pm 0.012}$ \\
\bottomrule
\end{tabular}}
\end{table}

\clearpage
\section{Gemma-4-E2B/E4B-it}
\label{app:gemma4}
\suppressfloats[t]

\begin{figure}[!htbp]
    \centering
    \includegraphics[width=0.9\linewidth]{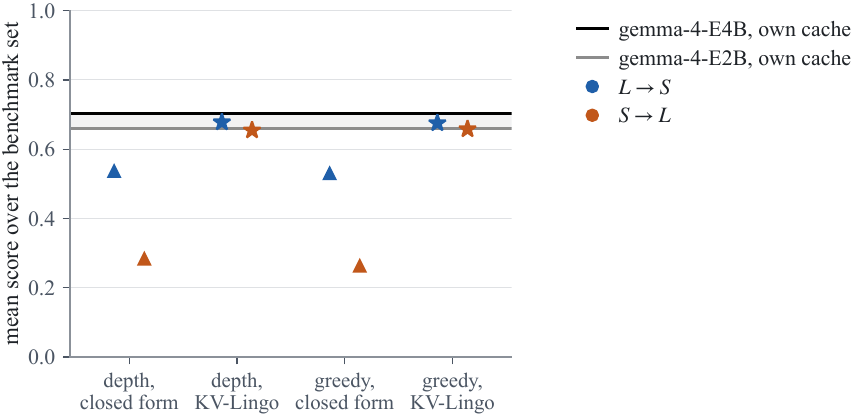}
    \caption{Cache translation with KV-Lingo on the pair Gemma-4-E2B/E4B-it, both directions, for the relative-depth (``depth'') and greedy within-group assignments, each as its closed-form fit (triangles) and after self-distillation (stars, seed 42); error bars: standard error over evaluation items (Appendix~\ref{app:seeds}). Averaged score over MMLU-Pro, ARC-C, LongBench-v2, XQuAD, RULER (mean over $4$k--$16$k) and MT-Bench-101, all reasoning-off, with judge scores rescaled to $[0, 1]$. Per-benchmark scores are in Table~\ref{tab:gemma4}, which reports MT-Bench-101 and LongBench-v2 reasoning-on instead.}
    \label{fig:gemma-4}
\end{figure}

Table~\ref{tab:gemma4} reports cache translation on the Gemma-4-E2B/E4B-it pair in both
directions, and Figure~\ref{fig:gemma-4} draws its mean scores.

\begin{table}[tbp]
\centering
\footnotesize
\setlength{\tabcolsep}{4pt}
\caption{\textbf{Grouped cache translation on the Gemma-4-E2B/E4B-it pair.} Both native
anchors and two within-group source-layer assignments (relative-depth and
greedy), each as its closed-form fit and after
self-distillation. Trained cells are seed 42. $\pm$: standard error over evaluation items
(Appendix~\ref{app:seeds}). \underline{Underline}: the
translated cache beats the small model, and in $S \to L$ by more than one standard
error. \textbf{Bold}: the best of the four translators on that row. Reasoning-off,
except MT-Bench-101 and LongBench-v2, which are reasoning-on (with reasoning off both
native anchors fall below LongBench-v2's $25\%$ chance floor). MMLU-Pro and ARC-C are lenient
generate-and-parse accuracy; XQuAD is token-F1 over the eleven languages; MT-Bench-101 is
a $1$--$10$ judge score; RULER is recall averaged over $4$k--$16$k; LongBench-v2 is
accuracy. Note that E2B outscores E4B on MT-Bench-101, so on that row the small model is
the harder anchor.}
\label{tab:gemma4}
\begin{tabular}{lrrrrrr}
\toprule
 & \multicolumn{2}{c}{native} & \multicolumn{2}{c}{closed form} & \multicolumn{2}{c}{self-distilled} \\
\cmidrule(lr){2-3} \cmidrule(lr){4-5} \cmidrule(lr){6-7}
Benchmark & small & large & depth & greedy & depth & greedy \\
\midrule
\multicolumn{7}{l}{\emph{$L \to S$}, Gemma-4-E2B/E4B-it} \\
MMLU-Pro & $0.468_{\pm 0.013}$ & $0.576_{\pm 0.013}$ & $0.346_{\pm 0.012}$ & $0.354_{\pm 0.013}$ & $\underline{\mathbf{0.481}}_{\pm 0.013}$ & $\underline{0.474}_{\pm 0.013}$ \\
ARC-C & $0.897_{\pm 0.009}$ & $0.939_{\pm 0.007}$ & $0.872_{\pm 0.010}$ & $0.879_{\pm 0.010}$ & $\underline{\mathbf{0.912}}_{\pm 0.008}$ & $\underline{0.905}_{\pm 0.009}$ \\
XQuAD & $0.470_{\pm 0.009}$ & $0.546_{\pm 0.009}$ & $0.298_{\pm 0.010}$ & $0.296_{\pm 0.010}$ & $\underline{0.488}_{\pm 0.010}$ & $\underline{\mathbf{0.544}}_{\pm 0.010}$ \\
MT-Bench-101 & $9.45_{\pm 0.13}$ & $9.35_{\pm 0.15}$ & $8.94_{\pm 0.19}$ & $8.74_{\pm 0.21}$ & $9.25_{\pm 0.16}$ & $\mathbf{9.36}_{\pm 0.12}$ \\
RULER & $0.996_{\pm 0.001}$ & $1.000_{\pm 0.000}$ & $0.755_{\pm 0.009}$ & $0.717_{\pm 0.009}$ & $\mathbf{0.964}_{\pm 0.004}$ & $0.931_{\pm 0.005}$ \\
LongBench-v2 & $0.304_{\pm 0.023}$ & $0.332_{\pm 0.024}$ & $0.175_{\pm 0.019}$ & $0.190_{\pm 0.020}$ & $\mathbf{0.284}_{\pm 0.023}$ & $0.266_{\pm 0.022}$ \\
\midrule
\multicolumn{7}{l}{\emph{$S \to L$}, Gemma-4-E2B/E4B-it} \\
MMLU-Pro & $0.468_{\pm 0.013}$ & $0.576_{\pm 0.013}$ & $0.086_{\pm 0.007}$ & $0.084_{\pm 0.007}$ & $0.446_{\pm 0.013}$ & $\mathbf{0.450}_{\pm 0.013}$ \\
ARC-C & $0.897_{\pm 0.009}$ & $0.939_{\pm 0.007}$ & $0.261_{\pm 0.013}$ & $0.116_{\pm 0.009}$ & $0.790_{\pm 0.012}$ & $\mathbf{0.828}_{\pm 0.011}$ \\
XQuAD & $0.470_{\pm 0.009}$ & $0.546_{\pm 0.009}$ & $0.012_{\pm 0.002}$ & $0.012_{\pm 0.002}$ & $\underline{\mathbf{0.509}}_{\pm 0.011}$ & $\underline{0.508}_{\pm 0.011}$ \\
MT-Bench-101 & $9.45_{\pm 0.13}$ & $9.35_{\pm 0.15}$ & $7.83_{\pm 0.24}$ & $8.00_{\pm 0.23}$ & $9.34_{\pm 0.13}$ & $\mathbf{9.35}_{\pm 0.13}$ \\
RULER & $0.996_{\pm 0.001}$ & $1.000_{\pm 0.000}$ & $0.471_{\pm 0.010}$ & $0.472_{\pm 0.010}$ & $\mathbf{0.972}_{\pm 0.004}$ & $0.971_{\pm 0.004}$ \\
LongBench-v2 & $0.304_{\pm 0.023}$ & $0.332_{\pm 0.024}$ & $0.205_{\pm 0.020}$ & $0.258_{\pm 0.022}$ & $0.284_{\pm 0.023}$ & $\mathbf{0.296}_{\pm 0.023}$ \\
\bottomrule
\end{tabular}
\end{table}

\clearpage
\section{Cross-tokenizer cache translation: Qwen3-8B/Mistral-7B-Instruct-v0.3}
\label{app:subsec:xtok}
\suppressfloats[t]

This section describes the procedure to apply KV-Lingo translators to the pair Qwen3-8B/Mistral-7B-Instruct-v0.3.

The source prefills its own rendering of the prefix and the translator is
applied to its pre-RoPE, pre-normalisation cache token by token, as in the matched case, so the translated cache holds
one entry per translated source token.
However, we do not apply RoPE canonically to the translated keys, which would induce a mismatch 
during training between the rotation of the last translated token of the prefix, and the first continuation token.
Instead, we assign a target token to every translated token, with the method explained below, and rotate the translated keys according to the position of
their assigned target token. That way, some rotations can be identical between translated tokens, and not all
target rotations are represented in the set of the translated tokens; but the translated cache has virtually the same
number of tokens as the target cache.

\begin{figure}[!htbp]
\centering
\includegraphics[width=\linewidth]{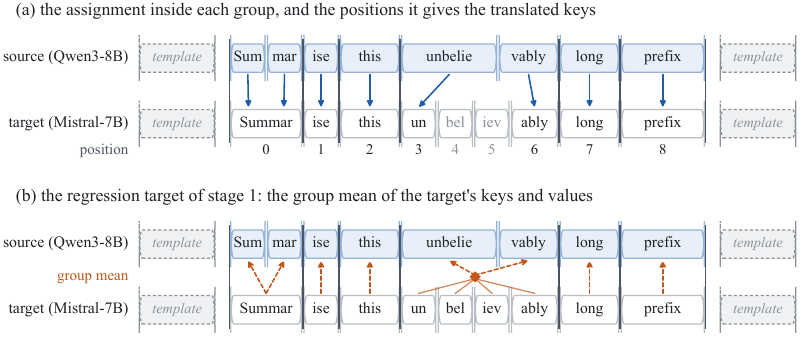}
\caption{\textbf{Cache translation across a tokenizer boundary, on one span of text.} The two
models segment \emph{Summarise this unbelievably long prefix} differently; the vertical bars stand
at the character offsets where both of them end a token, which delimit the minimal groups.
\textbf{(a)} Inside a group, the source tokens are spread evenly over the target ones, and the
translated keys of a source token are rotated at the position of the target token it is
assigned to: a position can carry two translated tokens, another none. \textbf{(b)} Stage 1
regresses the keys and values of a source token onto the average of those of the target tokens
of its group, which is the assigned token itself in a group holding one token on each side.
Boilerplate from the two chat templates is prefilled by the target. The source row is the
segmentation Qwen3 actually produces; the target row stands in for a SentencePiece one.}
\label{fig:xtok_assignment}
\end{figure}

\paragraph{Assigning each source token to a target token.}
Take a piece of text rendered by the source as the tokens $s_1, \dots, s_p$, by the target as
the tokens $t_1, \dots, t_q$, with $p \neq q$ in general. We assign one target token to every
source token, as follows.
Source tokens are spread evenly over the target ones: for $p \geq 2$, the source token $s_i$
is assigned to $t_{j(i)}$ with $j(i) = 1 + \mathrm{round}\bigl((i-1)(q-1)/(p-1)\bigr)$, while
a single source token is assigned to $t_1$; Figure~\ref{fig:xtok_assignment}(a) shows the
result on one span of text.
This assignment is non-decreasing. Its two ends are those of the
text: $s_1$ is sent to $t_1$, $s_p$ to $t_q$.
The position of the assigned target token is the position at which we rotate the translated
keys of a source token.
Note that some target tokens can be assigned to several source tokens, while others might be assigned to none.

\paragraph{Breaking down the prefix into minimal mismatching groups.}
We do not apply the rule above to the whole prefix at once. The prefix is first cut at every
token at which the source and the target both end at the same character. Two consecutive such
offsets delimit a group of tokens covering exactly the same characters on both sides, with at
least one token on either side. Cutting at every shared offset makes these groups minimal:
none of them can be split into two smaller parts that both cover the same characters on the
two sides. Each group is then treated as the text of the previous paragraph, independently of
the others. On our fitting corpus, $83\%$ of the groups hold a single token on each side,
where the two tokens are simply assigned to each other. 

\paragraph{Initialising the translators (stage 1).}
The closed-form solution
of stage 1 (\eqref{eq:hat}) takes the keys and values of a source token as input, and the average of
the target keys and values in the corresponding minimal mismatching group as regression
target (Figure~\ref{fig:xtok_assignment}(b)); the regression
target is the assigned token itself when
$p = q = 1$.

\paragraph{Boilerplate tokens are prefilled.}
Boilerplate tokens from the chat templates are left out of the
construction, turn delimiters and generation prompts having no counterpart across the two
families; the target prefills them itself.

\begin{figure}[t]
\centering
\includegraphics[width=0.6\linewidth]{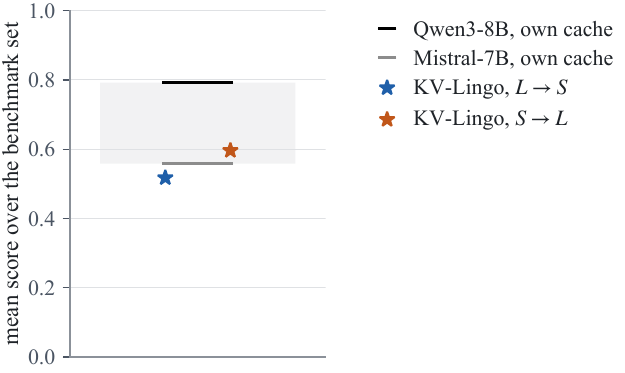}
\caption{\textbf{Cache translation across a tokenizer boundary, Qwen3-8B $\leftrightarrow$
Mistral-7B-Instruct-v0.3.} Mean over the seven benchmarks of Table~\ref{tab:xtok}, both
directions, a single map per direction. Error bars: standard error over evaluation items
(Appendix~\ref{app:seeds}).}
\label{fig:xtok_pair}
\end{figure}

\paragraph{Results.}
Table~\ref{tab:xtok} reports both directions on eight benchmarks, and
Figure~\ref{fig:xtok_pair} draws their mean.

\begin{table}[tbp]
\centering
\footnotesize
\setlength{\tabcolsep}{5pt}
\caption{\textbf{Cache translation across a tokenizer boundary}, Qwen3-8B
$\leftrightarrow$ Mistral-7B-Instruct-v0.3. The \emph{small} model is
Mistral-7B-Instruct-v0.3 and the \emph{large} one Qwen3-8B, each shown decoding from its
own cache; both blocks are scored on the same evaluation sets, so the two anchor columns
repeat. The translator combines the three assigned source layers with one rectangular map
per source layer ($201$M parameters), trained at a single seed. $\pm$: standard error over
evaluation items (Appendix~\ref{app:seeds}). \underline{Underline}: the
translated cache beats the small model, and in $S \to L$ by more than one standard error.
MMLU-Pro and ARC-C are lenient generate-and-parse accuracy at a $2048$-token
budget, GSM8K flexible exact match, IFEval prompt-level strict accuracy, and XQuAD token-F1
pooled over its eleven languages; RepLiQA is judged correctness over the whole $512$-item
set, answerable and unanswerable together; LongMemEval is judged accuracy averaged over its oracle, $8$k and $16$k rungs. The
\emph{mean} row weights the seven benchmarks equally.}
\label{tab:xtok}
\begin{tabular}{@{}cl cc c@{}}
\toprule
 & & \multicolumn{2}{c}{native} & translated \\
\cmidrule(lr){3-4} \cmidrule(lr){5-5}
 & Benchmark & small & large & self-distilled \\
\midrule
\multirow{8}{*}{\rotatebox[origin=c]{90}{$L \to S$}} & MMLU-Pro & $0.318_{\pm 0.012}$ & $0.629_{\pm 0.013}$ & $\underline{0.406}_{\pm 0.013}$ \\
 & ARC-C & $0.725_{\pm 0.013}$ & $0.937_{\pm 0.007}$ & $\underline{0.833}_{\pm 0.011}$ \\
 & RepLiQA & $0.754_{\pm 0.019}$ & $0.795_{\pm 0.018}$ & $0.740_{\pm 0.019}$ \\
 & LongMemEval & $0.542_{\pm 0.022}$ & $0.702_{\pm 0.021}$ & $0.455_{\pm 0.022}$ \\
 & GSM8K & $0.516_{\pm 0.031}$ & $0.922_{\pm 0.017}$ & $0.387_{\pm 0.030}$ \\
 & IFEval & $0.504_{\pm 0.021}$ & $0.804_{\pm 0.017}$ & $0.451_{\pm 0.021}$ \\
 & XQuAD & $0.549_{\pm 0.007}$ & $0.753_{\pm 0.006}$ & $0.351_{\pm 0.007}$ \\
\cmidrule(l){2-5}
 & mean & $0.558_{\pm 0.007}$ & $0.792_{\pm 0.006}$ & $0.517_{\pm 0.007}$ \\
\midrule
\multirow{8}{*}{\rotatebox[origin=c]{90}{$S \to L$}} & MMLU-Pro & $0.318_{\pm 0.012}$ & $0.629_{\pm 0.013}$ & $\underline{0.376}_{\pm 0.013}$ \\
 & ARC-C & $0.725_{\pm 0.013}$ & $0.937_{\pm 0.007}$ & $\underline{0.758}_{\pm 0.013}$ \\
 & RepLiQA & $0.754_{\pm 0.019}$ & $0.795_{\pm 0.018}$ & $0.666_{\pm 0.021}$ \\
 & LongMemEval & $0.542_{\pm 0.022}$ & $0.702_{\pm 0.021}$ & $0.506_{\pm 0.021}$ \\
 & GSM8K & $0.516_{\pm 0.031}$ & $0.922_{\pm 0.017}$ & $\underline{0.852}_{\pm 0.022}$ \\
 & IFEval & $0.504_{\pm 0.022}$ & $0.805_{\pm 0.017}$ & $\underline{0.595}_{\pm 0.021}$ \\
 & XQuAD & $0.549_{\pm 0.007}$ & $0.753_{\pm 0.006}$ & $0.422_{\pm 0.008}$ \\
\cmidrule(l){2-5}
 & mean & $0.558_{\pm 0.007}$ & $0.792_{\pm 0.006}$ & $\underline{0.596}_{\pm 0.007}$ \\
\bottomrule
\end{tabular}
\end{table}

\clearpage
\section{MLA cache translation}
\label{app:mla}
\suppressfloats[t]

\begin{figure}[!htbp]
\centering
\includegraphics[width=0.6\linewidth]{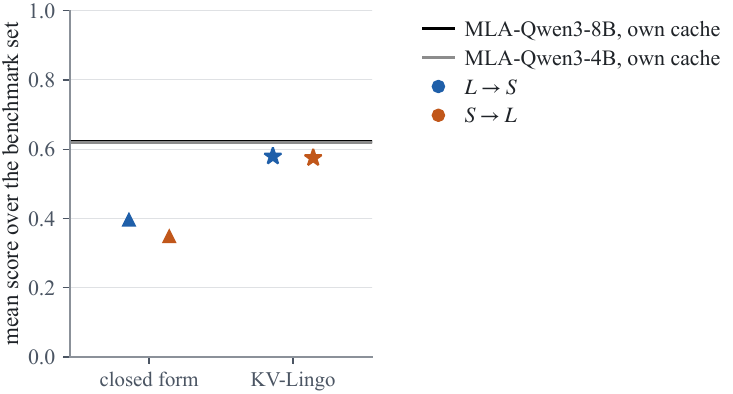}
\caption{\textbf{MLA latent translation, Qwen3-$4$B/$8$B converted with CARE.} Mean over
the benchmarks of Table~\ref{tab:mla} other than IFEval, both directions. Triangles are the closed-form
fit, stars the self-distilled KV-Lingo map, a single seed. Error bars: standard error over
evaluation items (Appendix~\ref{app:seeds}).}
\label{fig:mla_pair}
\end{figure}

This section complements \S\ref{sec:exp:mla}; Figure~\ref{fig:mla_pair} draws the suite
mean of Table~\ref{tab:mla}, and Table~\ref{app:tab:mla} compares translating the latents
with translating the keys and values of an unconverted pair.

\paragraph{The MLA models.}
The source and target models have caches of the same shape, holding one $1024$-dimensional compressed latent and a $64$-dimensional rotary key per token, per layer.
The models were converted into MLA models from Qwen3-4B and Qwen3-8B pretrained models, with CARE~\citep{zhou2026care}.
CARE is calibrated on $256$ sequences of $1024$ Dolci-Instruct-SFT tokens~\citep{olmo2025olmo3}, packed without padding, with a latent rank of $1024$, a rotary key of width $64$, and the query and key normalisation of Qwen3 kept.
Then, as CARE appears to hurt the performance of the native models, we fine-tune them for $120$k steps on Dolci-Instruct-SFT.
The loss is the cross-entropy of the final assistant turn plus, with weight $1$, the KL divergence $\mathrm{KL}(p_{\mathrm{MLA}} \,\|\, p_{\mathrm{GQA}})$ at temperature $1$ between the converted model's next-token distribution and that of its own frozen unconverted original, on the same positions.
We use AdamW at a learning rate of $10^{-5}$ with $2\%$ warmup and linear decay, weight decay $10^{-4}$, gradient clipping at $1.0$ and \texttt{bf16}, with one conversation of at most $4096$ tokens per GPU on $8$ H100 GPUs, and hold out $512$ Dolci conversations, selected by content hash, for validation.
We report that $120$k checkpoint throughout; its native accuracies are the anchor columns of Table~\ref{tab:mla}, $0.759$ and $0.819$ on ARC-C and $0.306$ and $0.351$ on MMLU-Pro for the $4$B and the $8$B. Most of the distance to stock Qwen3-Instruct ($0.920$ and $0.937$ on ARC-C, $0.564$ and $0.615$ on MMLU-Pro) comes from the recovery fine-tune rather than from the conversion: an unconverted GQA pair given the same fine-tune at the same budget scores $0.852$ and $0.900$ on ARC-C, $0.417$ and $0.486$ on MMLU-Pro. The converted pair is read against that matched control.

\begin{table}[]
\centering
\footnotesize
\setlength{\tabcolsep}{6pt}
\caption{Performance of KV-Lingo on an MLA pair (small: 4B, large: 8B), compared to the performance of each model equipped with its own cache. One map per cell (seed 42). $\pm$: standard error over evaluation items (Appendix~\ref{app:seeds}). \underline{Underline}: the translated cache beats the small model, and in $S \to L$ by more than one standard error. \textbf{Bold}: the better of the two translators on that row. IFEval is prompt-level strict accuracy, and the \emph{mean} row weights the six benchmarks equally. On the mean, KV-Lingo lands about $0.04$ below both native models, whose own means differ by only $0.004$, so in both directions the switch costs some quality. The closed-form fit beats the self-distilled translator on MMLU-Pro and ARC-C in $L \to S$ and on ARC-C in $S \to L$.}
\label{tab:mla}
\begin{tabular}{@{}cl cc cc@{}}
\toprule
 & & \multicolumn{2}{c}{native} & \multicolumn{2}{c}{translated} \\
\cmidrule(lr){3-4} \cmidrule(lr){5-6}
 & \multirow{2}{*}{Benchmark} & \multirow{2}{*}{small} & \multirow{2}{*}{large}
   & closed form & KV-Lingo \\
 & & & & head-mixing & head-mixing \\
\midrule
\multirow{7}{*}{\rotatebox[origin=c]{90}{$L \to S$}} & MMLU-Pro & $0.306_{\pm 0.012}$ & $0.351_{\pm 0.012}$ & $\mathbf{0.291}_{\pm 0.012}$ & $0.258_{\pm 0.011}$ \\
 & ARC-C & $0.759_{\pm 0.013}$ & $0.819_{\pm 0.011}$ & $\underline{\mathbf{0.785}}_{\pm 0.012}$ & $\underline{0.769}_{\pm 0.012}$ \\
 & XQuAD & $0.505_{\pm 0.008}$ & $0.508_{\pm 0.008}$ & $0.381_{\pm 0.008}$ & $\mathbf{0.440}_{\pm 0.008}$ \\
 & CoQA & $0.639_{\pm 0.027}$ & $0.567_{\pm 0.028}$ & $0.519_{\pm 0.028}$ & $\mathbf{0.577}_{\pm 0.028}$ \\
 & IFEval & $0.682_{\pm 0.020}$ & $0.691_{\pm 0.020}$ & $0.543_{\pm 0.021}$ & $\mathbf{0.630}_{\pm 0.021}$ \\
 & RepLiQA & $0.881_{\pm 0.017}$ & $0.860_{\pm 0.018}$ & $0.006_{\pm 0.004}$ & $\mathbf{0.854}_{\pm 0.019}$ \\
\cmidrule(l){2-6}
 & mean & $0.629_{\pm 0.007}$ & $0.633_{\pm 0.007}$ & $0.421_{\pm 0.007}$ & $\mathbf{0.588}_{\pm 0.007}$ \\
\midrule
\multirow{7}{*}{\rotatebox[origin=c]{90}{$S \to L$}} & MMLU-Pro & $0.306_{\pm 0.012}$ & $0.351_{\pm 0.012}$ & $0.268_{\pm 0.012}$ & $\mathbf{0.292}_{\pm 0.012}$ \\
 & ARC-C & $0.759_{\pm 0.013}$ & $0.819_{\pm 0.011}$ & $\mathbf{0.745}_{\pm 0.013}$ & $0.738_{\pm 0.013}$ \\
 & XQuAD & $0.505_{\pm 0.008}$ & $0.508_{\pm 0.008}$ & $0.340_{\pm 0.008}$ & $\mathbf{0.436}_{\pm 0.008}$ \\
 & CoQA & $0.639_{\pm 0.027}$ & $0.567_{\pm 0.028}$ & $0.226_{\pm 0.023}$ & $\mathbf{0.554}_{\pm 0.028}$ \\
 & IFEval & $0.682_{\pm 0.020}$ & $0.691_{\pm 0.020}$ & $0.429_{\pm 0.021}$ & $\mathbf{0.645}_{\pm 0.021}$ \\
 & RepLiQA & $0.881_{\pm 0.017}$ & $0.860_{\pm 0.018}$ & $0.163_{\pm 0.020}$ & $\mathbf{0.854}_{\pm 0.019}$ \\
\cmidrule(l){2-6}
 & mean & $0.629_{\pm 0.007}$ & $0.633_{\pm 0.007}$ & $0.362_{\pm 0.007}$ & $\mathbf{0.587}_{\pm 0.007}$ \\
\bottomrule
\end{tabular}
\end{table}

\begin{table}[tbp]
\centering
\footnotesize
\setlength{\tabcolsep}{3pt}
\caption{\textbf{Translating latents against translating the keys and values they
compress.} One pair, Qwen3-$4$B/$8$B, converted to MLA and, as the baseline,
left as it is; both versions given the same recovery fine-tune at two budgets, and their
translators fit by the same recipe on the same data, so architecture is the only
difference. \emph{Continuation KL} is the held-out KL to the target's own next-token
distribution, the quantity stage 2 minimises, reported for the closed-form fit and after
self-distillation. \emph{Accuracy $-$ own native} is the translated cache's
generate-and-parse accuracy minus the target's own, so $0$ is a lossless translation and
positive means the translation helped. One seed per cell (seed 42). $\pm$: standard error over
evaluation items, paired between the translated and the native run (Appendix~\ref{app:seeds});
the KLs carry none, as per-sequence KLs were not stored. Because the KL is measured against
each target's own cache and the target differs between the two directions, the columns
are commensurable in meaning rather than against a fixed reference, but the closed-form
pair, which involves no optimiser, separates the two versions the same way and by more.}
\label{app:tab:mla}
\resizebox{\ifdim\width>\linewidth\linewidth\else\width\fi}{!}{%
\begin{tabular}{llrrrrrrrr}
\toprule
 & & \multicolumn{4}{c}{continuation KL} & \multicolumn{4}{c}{accuracy $-$ own native} \\
\cmidrule(lr){3-6} \cmidrule(lr){7-10}
 & & \multicolumn{2}{c}{closed form} & \multicolumn{2}{c}{self-distilled} & \multicolumn{2}{c}{MMLU-Pro} & \multicolumn{2}{c}{ARC-C} \\
\cmidrule(lr){3-4} \cmidrule(lr){5-6} \cmidrule(lr){7-8} \cmidrule(lr){9-10}
Recovery & Dir. & latent & K/V & latent & K/V & latent & K/V & latent & K/V \\
\midrule
$60$k & $L \to S$ & $2.613$ & $0.117$ & $0.103$ & $0.023$ & $+0.011_{\pm 0.013}$ & $+0.004_{\pm 0.013}$ & $+0.023_{\pm 0.014}$ & $+0.045_{\pm 0.010}$ \\
$60$k & $S \to L$ & $0.977$ & $0.098$ & $0.136$ & $0.021$ & $-0.095_{\pm 0.012}$ & $-0.086_{\pm 0.013}$ & $-0.090_{\pm 0.014}$ & $-0.048_{\pm 0.010}$ \\
\midrule
$120$k & $L \to S$ & $3.033$ & $0.117$ & $0.152$ & $0.021$ & $-0.049_{\pm 0.013}$ & $+0.047_{\pm 0.013}$ & $+0.010_{\pm 0.013}$ & $+0.053_{\pm 0.010}$ \\
$120$k & $S \to L$ & $1.651$ & $0.103$ & $0.109$ & $0.021$ & $-0.059_{\pm 0.013}$ & $-0.087_{\pm 0.013}$ & $-0.081_{\pm 0.012}$ & $-0.061_{\pm 0.010}$ \\
\bottomrule
\end{tabular}}
\end{table}

\clearpage
\section{Comparison with DroidSpeak}
\label{app:droidspeak}

DroidSpeak~\citep{liu2024droidspeak} learns no map: the target recomputes a contiguous group
of its own layers from a hidden state of the source, and adopts the source's keys and values verbatim everywhere else, which requires the source and the target
caches to have the same shape, and the source and target hidden dimensions to be the same.
None of our pairs satisfies both conditions.
We therefore adapt DroidSpeak to Qwen3-0.6B/1.7B, whose caches have the same shape (28 layers, 8 key-value heads of
dimension 128) but whose hidden widths differ ($1024$ vs.\ $2048$). The source's embedding
cache therefore cannot be reused, so the recomputed group must start at layer $0$: the
target prefills layers $0$ to $j-1$ itself and reuses the source's cache for layers $j$ to
$27$. We pick $j$ with DroidSpeak's own offline profiler, which selects the cheapest group
closing all but $5\%$ of the output-KL gap between verbatim reuse and the target's own
cache. Scoring is our reasoning-off protocol on ARC-C and MMLU-Pro.

\begin{table}[htbp]
\centering
\footnotesize
\setlength{\tabcolsep}{3pt}
\caption{\textbf{DroidSpeak against KV-Lingo, Qwen3-0.6B/1.7B, reasoning-off.} ARC-C
($1172$ items) and MMLU-Pro ($1400$ items) accuracy. \emph{Compute} is the target-side
cost before decoding starts, in analytic MACs per token as a fraction of the target's own
prefill. In $L \to S$ the profiler selects all $28$ layers, so DroidSpeak coincides with
the target's own cache by construction. KV-Lingo is seed $42$. $\pm$: standard error over
evaluation items (Appendix~\ref{app:seeds}); the $L \to S$ verbatim cell is the $j = 0$
point of Table~\ref{tab:droidspeak_sweep}, on its subsample.}
\label{tab:droidspeak}
\begin{tabular}{lcccccc}
\toprule
& \multicolumn{3}{c}{$S \to L$ (target 1.7B)} & \multicolumn{3}{c}{$L \to S$ (target 0.6B)} \\
\cmidrule(lr){2-4} \cmidrule(lr){5-7}
& ARC-C & MMLU-Pro & compute & ARC-C & MMLU-Pro & compute \\
\midrule
Target, own cache              & $0.770_{\pm 0.012}$ & $0.379_{\pm 0.013}$ & $1.00$ & $0.568_{\pm 0.014}$ & $0.234_{\pm 0.011}$ & $1.00$ \\
Source alone                   & $0.568_{\pm 0.014}$ & $0.234_{\pm 0.011}$ & ---    & $0.770_{\pm 0.012}$ & $0.379_{\pm 0.013}$ & --- \\
\midrule
DroidSpeak, verbatim ($j=0$)   & $0.000_{\pm 0.000}$ & $0.000_{\pm 0.000}$ & $0.00$ & $0.000_{\pm 0.000}$ & $0.000_{\pm 0.000}$ & $0.00$ \\
DroidSpeak, profiler           & $0.770_{\pm 0.012}$ & $0.381_{\pm 0.013}$ & $0.93$ & $0.568_{\pm 0.014}$ & $0.234_{\pm 0.011}$ & $1.00$ \\
                               & \multicolumn{3}{c}{($j = 26$)} & \multicolumn{3}{c}{($j = 28$)} \\
\midrule
Closed-form                    & $0.497_{\pm 0.015}$ & $0.191_{\pm 0.011}$ & $0.03$ & $0.640_{\pm 0.014}$ & $0.210_{\pm 0.011}$ & $0.07$ \\
KV-Lingo                       & $0.467_{\pm 0.015}$ & $0.238_{\pm 0.011}$ & $0.03$ & $0.616_{\pm 0.014}$ & $0.219_{\pm 0.011}$ & $0.07$ \\
\bottomrule
\end{tabular}
\end{table}

\begin{table}[htbp]
\centering
\footnotesize
\setlength{\tabcolsep}{4pt}
\caption{\textbf{DroidSpeak accuracy against the recompute boundary $j$.} The target
recomputes layers $0$ to $j-1$ and reuses the source's cache above. Subsampled to $400$
ARC-C and $420$ MMLU-Pro items; the last row is the target's own cache on the same
items. $\pm$: standard error over evaluation items (Appendix~\ref{app:seeds}).}
\label{tab:droidspeak_sweep}
\begin{tabular}{@{}lcccc@{}}
\toprule
 & \multicolumn{2}{c}{$S \to L$} & \multicolumn{2}{c}{$L \to S$} \\
\cmidrule(lr){2-3} \cmidrule(lr){4-5}
$j$ & ARC-C & MMLU-Pro & ARC-C & MMLU-Pro \\
\midrule
$0$   & $.000_{\pm .000}$ & $.000_{\pm .000}$ & $.000_{\pm .000}$ & $.000_{\pm .000}$ \\
$4$   & $.000_{\pm .000}$ & $.000_{\pm .000}$ & $.000_{\pm .000}$ & $.000_{\pm .000}$ \\
$8$   & $.000_{\pm .000}$ & $.000_{\pm .000}$ & $.000_{\pm .000}$ & $.000_{\pm .000}$ \\
$12$  & $.018_{\pm .007}$ & $.005_{\pm .003}$ & $.000_{\pm .000}$ & $.002_{\pm .002}$ \\
$16$  & $.075_{\pm .013}$ & $.021_{\pm .007}$ & $.000_{\pm .000}$ & $.002_{\pm .002}$ \\
$18$  & $.170_{\pm .019}$ & $.050_{\pm .011}$ & $.003_{\pm .003}$ & $.005_{\pm .003}$ \\
$20$  & $.410_{\pm .025}$ & $.136_{\pm .017}$ & $.500_{\pm .025}$ & $.124_{\pm .016}$ \\
$22$  & $.768_{\pm .021}$ & $.360_{\pm .023}$ & $.548_{\pm .025}$ & $.221_{\pm .020}$ \\
$24$  & $.760_{\pm .021}$ & $.419_{\pm .024}$ & $.543_{\pm .025}$ & $.233_{\pm .021}$ \\
$26$  & $.775_{\pm .021}$ & $.393_{\pm .024}$ & $.555_{\pm .025}$ & $.257_{\pm .021}$ \\
\midrule
own   & $.758_{\pm .021}$ & $.410_{\pm .024}$ & $.565_{\pm .025}$ & $.257_{\pm .021}$ \\
\bottomrule
\end{tabular}
\end{table}

Table~\ref{tab:droidspeak} gives the headline comparison and
Table~\ref{tab:droidspeak_sweep} the full range of recompute boundaries. Three
observations stand out.
\emph{Verbatim reuse fails completely.} At $j = 0$ the target almost never emits a
parseable answer, and accuracy stays near zero, below chance, up to $j = 18$ in both
directions. The target's own accuracy is first recovered at $j = 22$, that is, after
recomputing $79\%$ of its layers, so the pair offers no cheap operating point.
\emph{DroidSpeak's accuracy is capped by the target's.} Everything it does not reuse, it
recomputes with the target's own stack, so it cannot exceed the target; in $L \to S$ its
profiler recomputes the whole stack and saves nothing. KV-Lingo instead brings part of the
source's advantage along with the cache, and outscores the 0.6B target on ARC-C ($0.616$
vs.\ $0.568$).
\emph{In $S \to L$, the two methods sit at opposite ends of the cost range.} DroidSpeak
matches the target's accuracy at $93\%$ of its prefill, whereas KV-Lingo costs $3\%$ but
loses accuracy. DroidSpeak does win on transfer size, since it ships only the $2$ layers it
reuses where a translator ships all $28$.
DroidSpeak was designed for fine-tuned variants of one base model, whose caches are
nearly interchangeable; these results show that the premise does not hold across model
sizes.

\clearpage
\section{Dense-to-MoE cache translation}
\label{app:moe}
\suppressfloats[t]

This section details the experiment of \S\ref{sec:exp:moe}, where Qwen3-4B prefills and
Qwen3-30B-A3B decodes from the translated cache, and the reverse translator, from the MoE's
cache to the 4B's.

\paragraph{The pair.}
Qwen3-30B-A3B~\citep{yang2025qwen3} replaces the feed-forward block of each of its $48$
layers by $128$ experts, of which each token is routed to $8$: it activates $3.3$B of its
$30.5$B parameters per token. Its attention has $4$ key-value heads of dimension $128$, so
$d_\tgt = 512$, against $8$ heads of the same dimension in each of the $36$ layers of
Qwen3-4B, so $d_\src = 1024$; the two models share a tokenizer. For the KV-Lingo translator, each target layer reads
$\nu = 3$ source layers, assigned by relative depth (\S\ref{sec:method:family}), through one
$512 \times 3072$ matrix per side, for $151$M parameters in all. The reverse translator
assigns three of the MoE's layers to each layer of the 4B the same way, through one
$1024 \times 1536$ matrix per side, for $113$M parameters.

\paragraph{Training.}
The translator is trained as in Appendix~\ref{app:protocol}, on the long-context mixture of
Appendix~\ref{app:data}, with one change: stage 1 fits the closed-form map under the
attention-weighted norm~\eqref{eq:attn-norm}, the \emph{attention mass} norm of
Appendix~\ref{app:norms}, rather than the Frobenius norm. That fit alone reaches a held-out
KL of $0.207$. Stage 2 then runs $5000$ steps of~\eqref{eq:kl} on $8$ B200 GPUs, one
conversation per GPU and step. Of the learning rates
$\{10^{-6}, 3{\times}10^{-6}, 10^{-5}\}$, $3{\times}10^{-6}$ ends at the lowest held-out KL,
$0.0634$ (Table~\ref{app:tab:moemlp}), and is the translator evaluated below. The reverse
translator follows the same recipe: its closed-form fit starts at $0.304$, and of the
learning rates $\{3{\times}10^{-5}, 10^{-4}, 3{\times}10^{-4}\}$, $10^{-4}$ ends lowest, at
$0.0858$.

\begin{figure}
    \centering
    \includegraphics[width=0.7\linewidth]{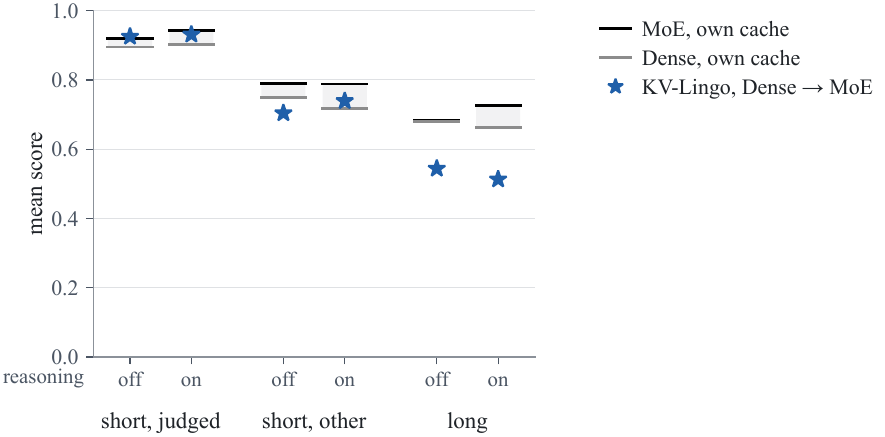}
    \caption{\textbf{Dense$\to$MoE mean scores, by task group and reasoning mode.} Qwen3-4B
  (\emph{Dense}, grey) and Qwen3-30B-A3B (\emph{MoE}, black) decode from their own caches;
  the stars are the MoE decoding from the KV-Lingo translation of the dense model's cache.
  \emph{Short, judged} averages the two judge-scored short-context benchmarks, RepLiQA and
  MT-Bench-101, with the $1$--$10$ judge ratings rescaled to $0$--$1$; \emph{short,
  other} averages the six others, MMLU-Pro, ARC-C, GSM8K, IFEval, CoQA and XQuAD; \emph{long}
  averages RULER, LongMemEval and LongBench-v2, with RULER and LongMemEval first averaged
  over their context lengths. Per-benchmark scores are in Table~\ref{app:tab:moe}.}
    \label{app:fig:moe}
\end{figure}

\paragraph{Evaluation.}
The translated caches are scored by the same harness and on the same items as the
Qwen3-4B/8B pair (Table~\ref{tab:evalsets}), with every benchmark run in both reasoning modes. Table~\ref{app:tab:moe} reports the whole suite, and Figure~\ref{app:fig:moe} averages it by task group. The reverse translator falls short
of the 4B on its own cache in all but four of its $36$ cells, and its RULER recall collapses
to $0.07$--$0.28$.

\begin{table}[htbp]
\centering
\footnotesize
\setlength{\tabcolsep}{3.5pt}
\caption{\textbf{Qwen3-4B and Qwen3-30B-A3B: the whole evaluation suite, both directions
and both reasoning on and off.} \emph{4B} and \emph{MoE} decode from their own caches;
\emph{4B$\to$MoE} is the MoE decoding from the translated cache of the 4B, and
\emph{MoE$\to$4B} the 4B decoding from the translated cache of the MoE.
\underline{Underline}: the translated cache beats the 4B on its own cache. Metrics as in
Table~\ref{tab:qwen3ls}, with GSM8K scored by flexible exact match, and RULER and
LongMemEval given per context length rather than averaged. The eight benchmarks above the
middle rule have prefixes of at most $3$k tokens. The translators are trained on prefixes
of up to $16$k tokens, so $32$k is extrapolation.}
\label{app:tab:moe}
\begin{tabular}{lrrrrrrrr}
\toprule
 & \multicolumn{4}{c}{reasoning-off} & \multicolumn{4}{c}{reasoning-on} \\
\cmidrule(lr){2-5} \cmidrule(lr){6-9}
 & \multicolumn{2}{c}{own cache} & \multicolumn{2}{c}{translated}
 & \multicolumn{2}{c}{own cache} & \multicolumn{2}{c}{translated} \\
\cmidrule(lr){2-3} \cmidrule(lr){4-5} \cmidrule(lr){6-7} \cmidrule(lr){8-9}
Benchmark & 4B & MoE & 4B$\to$MoE & MoE$\to$4B & 4B & MoE & 4B$\to$MoE & MoE$\to$4B \\
\midrule
MMLU-Pro & $0.548$ & $0.674$ & $0.525$ & $0.368$ & $0.607$ & $0.736$ & $0.606$ & $0.437$ \\
ARC-C & $0.918$ & $0.948$ & $0.916$ & $0.849$ & $0.938$ & $0.970$ & $\underline{0.959}$ & $0.905$ \\
GSM8K & $0.914$ & $0.930$ & $0.898$ & $0.863$ & $0.820$ & $0.859$ & $\underline{0.863}$ & $\underline{0.836}$ \\
IFEval & $0.771$ & $0.839$ & $0.745$ & $0.632$ & $0.758$ & $0.845$ & $\underline{0.795}$ & $0.706$ \\
CoQA & $0.722$ & $0.664$ & $0.675$ & $0.702$ & $0.612$ & $0.625$ & $\underline{0.614}$ & $\underline{0.613}$ \\
RepLiQA & $0.915$ & $0.905$ & $\underline{0.934}$ & $0.795$ & $0.911$ & $0.938$ & $\underline{0.939}$ & $0.900$ \\
XQuAD & $0.613$ & $0.688$ & $0.464$ & $0.444$ & $0.564$ & $0.693$ & $\underline{0.598}$ & $0.427$ \\
MT-Bench-101 & $8.87$ & $9.40$ & $\underline{9.25}$ & $\underline{8.91}$ & $9.06$ & $9.51$ & $\underline{9.30}$ & $8.93$ \\
\midrule
RULER $4$k & $1.000$ & $1.000$ & $0.766$ & $0.278$ & $0.968$ & $0.950$ & $0.544$ & $0.174$ \\
RULER $8$k & $1.000$ & $1.000$ & $0.704$ & $0.249$ & $0.953$ & $0.972$ & $0.548$ & $0.171$ \\
RULER $16$k & $0.998$ & $1.000$ & $0.638$ & $0.211$ & $0.927$ & $0.972$ & $0.524$ & $0.150$ \\
RULER $32$k & $0.994$ & $0.998$ & $0.531$ & $0.078$ & $0.937$ & $0.987$ & $0.476$ & $0.070$ \\
LongMemEval oracle & $0.722$ & $0.784$ & $0.674$ & $0.622$ & $0.830$ & $0.851$ & $0.797$ & $0.720$ \\
LongMemEval $8$k & $0.648$ & $0.717$ & $0.612$ & $0.563$ & $0.817$ & $0.810$ & $0.707$ & $0.643$ \\
LongMemEval $16$k & $0.617$ & $0.658$ & $0.576$ & $0.486$ & $0.699$ & $0.774$ & $0.645$ & $0.512$ \\
LongMemEval $32$k & $0.548$ & $0.576$ & $0.440$ & $0.409$ & $0.586$ & $0.717$ & $0.535$ & $0.419$ \\
LongBench-v2 & $0.405$ & $0.362$ & $0.397$ & $0.388$ & $0.310$ & $0.422$ & $\underline{0.345}$ & $0.250$ \\
\bottomrule
\end{tabular}
\end{table}

\paragraph{A non-linear translator.}
To test whether linearity costs this pair anything, we add a two-layer MLP branch to the
linear map of each target layer and side,
$\widehat{K}^\ell_i = T_K^\ell k_i + W_2\,\mathrm{GELU}(W_1 k_i + b_1) + b_2$, where
$k_i \in \R^{3 d_\src}$ stacks $K_{\src,i}^{\ell'}$ over $\ell' \in \mathcal{E}_\ell$ and the
hidden width is $4096$, and likewise for values. $W_2$ and $b_2$ start at zero, so stage 2 starts from the
same closed-form fit, and the recipe is otherwise unchanged, the learning rate being shared by
both parts (Table~\ref{app:tab:moemlp}). From the 4B to the MoE, the MLP brings the
translator to $1.56$B parameters, $10\times$ the linear map, and buys nothing: at its best
learning rate it ends at $0.0644$, against $0.0634$ for the linear map, and takes $1.8\times$
as long to train. In the reverse direction, at $0.87$B parameters ($8\times$), it ends $2\%$
below the best linear map, at $0.0842$ against $0.0858$, and takes $1.5\times$ as long to
train.

\begin{table}[htbp]
\centering
\footnotesize
\setlength{\tabcolsep}{5pt}
\caption{\textbf{Linear against non-linear translators}, Qwen3-4B and Qwen3-30B-A3B, both
directions. Held-out KL of~\eqref{eq:kl} during stage 2, one seed per run; step $0$ is the
closed-form fit, shared by every run of a direction. \emph{Time} is the wall-clock of stage
2 on $8$ B200 GPUs. The 4B $\to$ MoE MLP at $3{\times}10^{-5}$ was stopped at step $2000$,
at $0.150$ against $0.073$ for the linear map at $3{\times}10^{-6}$. In the reverse
direction, the MLP's best learning rate is the larger of the two it was run at, so its
optimum is not bracketed. \textbf{Bold}: the lowest final KL of each direction.}
\label{app:tab:moemlp}
\begin{tabular}{lrrrrrrr}
\toprule
 & & & \multicolumn{4}{c}{held-out KL at step} & \\
\cmidrule(lr){4-7}
Translator & Parameters & Learning rate & $0$ & $1000$ & $2500$ & $5000$ & Time \\
\midrule
\multicolumn{8}{l}{\emph{Qwen3-4B $\to$ Qwen3-30B-A3B}} \\
linear & $151$M & $10^{-6}$ & $0.207$ & $0.0749$ & $0.0689$ & $0.0659$ & $3.8$\,h \\
 & & $3{\times}10^{-6}$ & $0.207$ & $0.0846$ & $0.0706$ & $\mathbf{0.0634}$ & $3.8$\,h \\
 & & $10^{-5}$ & $0.207$ & $0.1154$ & $0.0894$ & $0.0742$ & $3.8$\,h \\
linear $+$ MLP & $1.56$B & $10^{-6}$ & $0.207$ & $0.0768$ & $0.0690$ & $0.0654$ & $6.9$\,h \\
 & & $3{\times}10^{-6}$ & $0.207$ & $0.0871$ & $0.0729$ & $0.0644$ & $6.9$\,h \\
 & & $3{\times}10^{-5}$ & $0.207$ & $0.1818$ & --- & --- & --- \\
\midrule
\multicolumn{8}{l}{\emph{Qwen3-30B-A3B $\to$ Qwen3-4B}} \\
linear & $113$M & $3{\times}10^{-5}$ & $0.304$ & $0.1078$ & $0.0965$ & $0.0921$ & $3.2$\,h \\
 & & $10^{-4}$ & $0.304$ & $0.1027$ & $0.0933$ & $0.0858$ & $3.2$\,h \\
 & & $3{\times}10^{-4}$ & $0.304$ & $0.1217$ & $0.1097$ & $0.0891$ & $3.2$\,h \\
linear $+$ MLP & $0.87$B & $3{\times}10^{-6}$ & $0.304$ & $0.1212$ & $0.1029$ & $0.0989$ & $4.9$\,h \\
 & & $3{\times}10^{-5}$ & $0.304$ & $0.0997$ & $0.0892$ & $\mathbf{0.0842}$ & $4.9$\,h \\
\bottomrule
\end{tabular}
\end{table}

\paragraph{Timing.}
Table~\ref{app:tab:moespeed} times both models and the translation on an Apple M3 Ultra with
MLX, in \texttt{bf16}, at batch size $1$. The MoE's prefill takes $2.2\times$ as long as the 4B's at $64$
tokens and $1.07\times$ at $1024$, while it decodes $1$--$2\%$ faster, and the translator adds
$2.6$--$4.4\%$ to the 4B's prefill in \texttt{fp32}, $2.5$--$3.4\%$ in \texttt{bf16}. As in
Appendix~\ref{app:cost}, the translated path's time to first token also counts the MoE's forward
pass on the last prompt token, one decoding step, timed as the inverse of its decoding speed.
The table prices the two uses of the translator. In a switch, the 4B has already served the
conversation and holds the cache, so the MoE's first token comes after the translation and
that decoding step: $15.8$\,ms at $64$ tokens and $31.7$\,ms at $1024$, $9.6$ to $13.3\times$
sooner than after its own prefill. As a prefill accelerator, the 4B first prefills the prompt
itself, so its prefill joins the path: $1.8\times$ sooner at $64$ tokens, and no sooner at
$1024$, where the 4B prefills about as slowly as the MoE. Letting the 4B emit the first token
instead, since its prefill predicts it, would remove the MoE's step and bring the first token
to the 4B's prefill time, $2.2\times$ sooner than the MoE's at $64$ tokens. Our evaluation does
not cover that protocol, in which the first token of every answer is the 4B's.

\begin{table}[htbp]
\centering
\footnotesize
\setlength{\tabcolsep}{5pt}
\caption{\textbf{Latency on an Apple M3 Ultra}, using MLX in \texttt{bf16}, batch size $1$.
Decoding speed is averaged over $256$ tokens generated after the prompt. In a switch, the
MoE's first token comes after the \texttt{fp32} translation of the 4B's cache and one decoding
step of the MoE; as a prefill accelerator, after the 4B's prefill as well. \emph{Speedup}
divides the MoE's prefill by that time.}
\label{app:tab:moespeed}
\begin{tabular}{lrrrrr}
\toprule
Prompt tokens & $64$ & $128$ & $256$ & $512$ & $1024$ \\
\midrule
Qwen3-4B prefill (ms) & $67.9$ & $94.9$ & $140.4$ & $248.7$ & $393.7$ \\
Qwen3-30B-A3B prefill (ms) & $151.8$ & $188.6$ & $215.8$ & $287.9$ & $421.4$ \\
translation, \texttt{fp32} (ms) & $1.8$ & $3.1$ & $5.7$ & $11.0$ & $17.2$ \\
translation, \texttt{bf16} (ms) & $1.7$ & $2.7$ & $4.7$ & $7.3$ & $13.1$ \\
\midrule
Qwen3-4B decoding (tokens/s) & $70.1$ & $70.0$ & $69.6$ & $68.6$ & $67.9$ \\
Qwen3-30B-A3B decoding (tokens/s) & $71.5$ & $71.3$ & $70.7$ & $69.6$ & $68.9$ \\
\midrule
switch, first token (ms) & $15.8$ & $17.1$ & $19.8$ & $25.3$ & $31.7$ \\
speedup & $9.63\times$ & $11.02\times$ & $10.89\times$ & $11.36\times$ & $13.29\times$ \\
\midrule
prefill accelerator, first token (ms) & $83.6$ & $112.0$ & $160.2$ & $274.0$ & $425.4$ \\
speedup & $1.81\times$ & $1.68\times$ & $1.35\times$ & $1.05\times$ & $0.99\times$ \\
\bottomrule
\end{tabular}
\end{table}

\clearpage
\section{Measuring the distance between caches}
\label{app:norms}

\iclronly{}

Table~\ref{app:tab:cachedist} shows that a better cache fit does not guarantee better
downstream accuracy: post-$k$-norm translators have the higher $R^2$ but decode worse.

\begin{table}[htbp]
\centering
\small
\setlength{\tabcolsep}{4pt}
\caption{\textbf{The MSE objective is not always aligned with downstream performance: the example of pre- versus post-$k$-norm.} Closed-form translators fitted before and after the target's $k$-norm, on both
Qwen3 pairs in both directions. $R^2$ is measured against the target's own cache on
held-out prefixes of Nemotron-SFT-Instruction-Following-Chat-v2.
Accuracy is that of the target decoding from that
cache. Better capture point in bold. Post-norm translators have a better $R^2$ fit, but worse downstream performance: the MSE objective is not always aligned with downstream performance.}
\label{app:tab:cachedist}
\begin{tabular}{@{}l rrr @{\hspace{2em}} rrr@{}}
\toprule
 & \multicolumn{3}{c}{pre-$k$-norm} & \multicolumn{3}{c}{post-$k$-norm} \\
\cmidrule(lr){2-4} \cmidrule(lr){5-7}
Source $\to$ target & $R^2$ & MMLU-Pro & ARC-C & $R^2$ & MMLU-Pro & ARC-C \\
\midrule
Qwen3-1.7B $\to$ 0.6B & $0.805$ & $\mathbf{0.257}$ & $\mathbf{0.671}$ & $\mathbf{0.852}$ & $0.202$ & $0.591$ \\
Qwen3-0.6B $\to$ 1.7B & $0.797$ & $\mathbf{0.218}$ & $\mathbf{0.544}$ & $\mathbf{0.801}$ & $0.079$ & $0.262$ \\
Qwen3-8B $\to$ 4B     & $0.849$ & $\mathbf{0.541}$ & $\mathbf{0.930}$ & $\mathbf{0.882}$ & $0.469$ & $0.859$ \\
Qwen3-4B $\to$ 8B     & $0.847$ & $\mathbf{0.526}$ & $0.912$ & $\mathbf{0.867}$ & $0.497$ & $\mathbf{0.914}$ \\
\bottomrule
\end{tabular}
\end{table}

\paragraph{The family, and why it is small.} 
Stage~1 of~\eqref{eq:hat} charges every position equally: the Frobenius residual is
$\sum_i \|\delta_i\|_2^2$ over the columns $\delta_i$ of $M K_\src^\ell - K_\tgt^\ell$.
Replacing each term by a quadratic form $\delta_i^\top \Omega_i \delta_i$ with $\Omega_i \succeq 0$ gives one generalised least-squares translator per choice of $\{\Omega_i\}_i$, of which the attention-weighted norm of~\eqref{eq:attn-norm} is the case $\Omega_i = \bar a_i I$. The obvious choices all collapse.
If $\Omega_t \equiv \Omega$ is the same at
every token and full rank, it factors out of the normal equations and the minimiser is
\emph{exactly} the Frobenius solution, whatever $\Omega$ is: whitening the target's
cache, or any fixed re-weighting of its coordinates, is a no-op. Only a metric that
varies across tokens moves the map.
Among those we take the form of~\eqref{eq:attn-norm}, $\Omega_t = w_t I$, one scalar
weight per position, which turns~\eqref{eq:hat} into a weighted least-squares problem
solved by the same second-moment statistics with $w_t$ inserted in both sums. 
We compare four families: \emph{Uniform} ($w_t = 1$) is our default.
\emph{Sink} follows StreamingLLM~\citep{xiao2024efficient} in up-weighting the first
four positions by a factor of $s \in \{2, 4, 8, 16, 32\}$.
\emph{Attention mass} takes $w_i = \bar a_i$ of~\eqref{eq:attn-norm}, the attention the target's own heads pay to position $i$, averaged over the continuation's queries and over heads on the fitting corpus.
Its \emph{attention-Fisher} refinement weights by $\bar a_i(1-\bar a_i)$, the diagonal of the softmax Jacobian, which discounts a position whose attention is already saturated and therefore cannot respond to a change in its key.

\begin{table}[htbp]
\centering
\caption{
\textbf{Fitting norms on the Qwen3-1.7B-Base/0.6B-Base pair}, $L \to S$, short prefixes. Left:
the closed-form translator on its own, next-token KL against native decoding. Right: the
same maps after stage 2, run to convergence ($15$k steps, one seed, norms paired on a
common validation batch). 
The two columns are measured on different continuation lengths and should be read down, not across.}
\label{app:tab:norms}
\begin{tabular}{lrr}
\toprule
fitting norm & off-the-shelf KL & after stage 2 \\
\midrule
Frobenius (uniform)      & $0.175$ & $0.0195$ \\
sink $\times 2$           & $0.134$ & $0.0196$ \\
sink $\times 4$           & $0.103$ & $0.0194$ \\
sink $\times 8$           & $0.085$ & $0.0195$ \\
sink $\times 16$          & $0.078$ & $0.0196$ \\
sink $\times 32$          & $0.076$ & $0.0198$ \\
attention mass           & $0.064$ & $\mathbf{0.0186}$ \\
attention Fisher         & $0.064$ & $\mathbf{0.0187}$ \\
\bottomrule
\end{tabular}
\end{table}

\paragraph{Off the shelf, the norm matters a great deal.} The norms rank strictly,
attention $<$ sink $<$ Frobenius
(Table~\ref{app:tab:norms}, left): the next-token KL
against native decoding falls from $0.175$ to $0.064$, a $63\%$ reduction, and the
continuation perplexity gap to native decoding from $1.77$ to $0.25$ (native $12.87$).
The Fisher correction buys nothing over raw attention mass, and both beat the sink family at its plateau, so what matters is up-weighting every heavily read key rather than
the first few. Uniform cache MSE is essentially unchanged across norms ($0.102$ against
$0.103$ on keys): the norm is invisible to reconstruction error and visible only in
decoding, one more instance of the mismatch that motivates stage 2.

\paragraph{The effect grows with context.} Refitted under the long-context Instruct
recipe of our main results ($S \to L$, prefixes packed to $12$--$16$k), the
attention-weighted translator cuts the validation KL of~\eqref{eq:kl} from $1.49$ to
$0.26$, putting the continuation cross-entropy within $0.04$ nats of the target
prefilling the context itself, against $1.56$ nats for the Frobenius fit. Downstream that
separates a usable training-free translator from an unusable one: on RULER the Frobenius
translator essentially fails to retrieve ($0.10$--$0.35$ recall from $4$ to $32$k) where
the attention-weighted one reaches $0.60$--$0.84$, against $0.72$--$0.96$ for the source
decoding on its own cache; on MMLU-Pro and ARC-C it moves from $0.191$ and $0.497$,
below the source's own accuracy, to $0.235$ and $0.549$, level with it ($0.234$ and
$0.568$; the target on its own cache scores $0.379$ and $0.770$).

\begin{figure}[htbp]
\centering
\includegraphics[width=\linewidth]{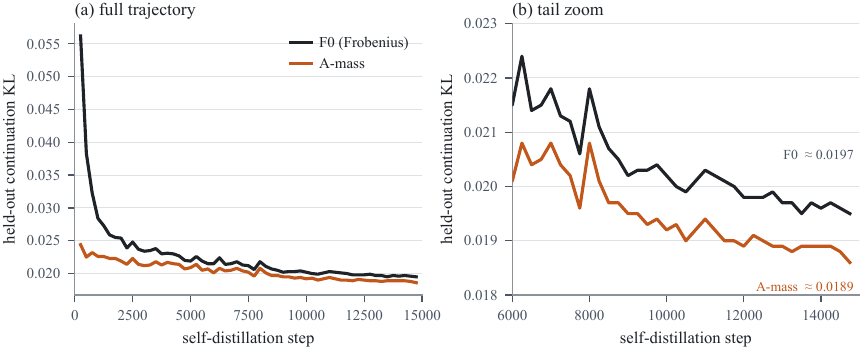}
\caption{\textbf{Stage 2 from two of the eight closed-form initialisations.} Held-out
continuation KL against native decoding over $15$k self-distillation steps;
Qwen3-1.7B-Base $\to$ Qwen3-0.6B-Base, $L \to S$, one seed, both runs paired on a common
validation batch. F0 is the Frobenius row of Table~\ref{app:tab:norms} and A-mass the
attention-mass one; the five sink initialisations track F0 and the attention-Fisher one tracks
A-mass, so only the two endpoints are drawn and the table carries the rest.
(a) shows the full run, (b) zooms the tail, where the attention-weighted initialisation holds about $0.0008$ below the level the Frobenius run settles into.}
\label{app:fig:norms}
\end{figure}

\paragraph{Stage 2 absorbs most of the difference.} Fine-tuned to convergence, and again
at twice the budget so that a head start is not read as a floor, the sink family lands on
the Frobenius level while the attention norms sit $3$--$4\%$ below
it (Table~\ref{app:tab:norms}, right; $0.0183$ against $0.0188$ at the doubled budget, a
gap flat over the second half of training); Figure~\ref{app:fig:norms} draws the two
endpoint runs.
What survives unambiguously is speed: the
attention-weighted initialisation reaches in $250$ steps the validation KL the Frobenius
one needs $5000$ steps for. The long-context pair agrees: after stage 2 the two are
within $0.03$ recall at every RULER length, and the attention norm keeps a small
multiple-choice edge ($+0.01$ on MMLU-Pro, $+0.10$ on ARC-C, with the
Frobenius-initialised translator failing to terminate on $20\%$ of ARC items against
$4\%$). The attention weights cost one extra attention pass over the fitting corpus,
computed by a query-blocked softmax reduction to stay feasible at $16$k, and nothing at
inference: the translator is the same object.

\clearpage
\clearpage
\section{The impact of reasoning on cache translation}
\label{app:reasoningboost}

\begin{figure}[htbp]
\centering
\includegraphics[width=0.8\linewidth]{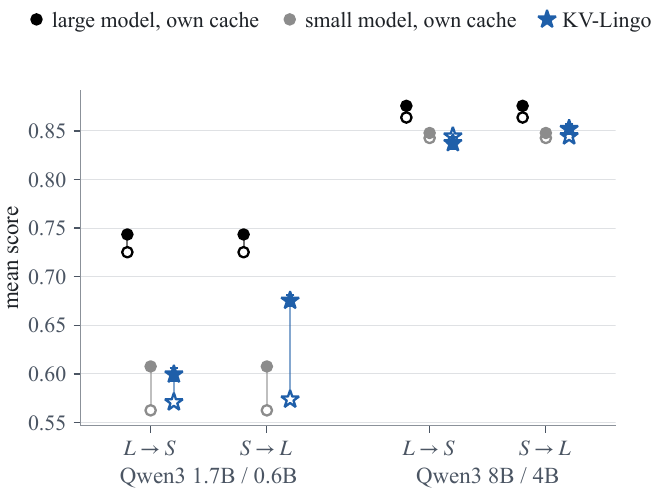}
\caption{\textbf{Reasoning off against reasoning on, both shape-preserving Qwen3 pairs.} Mean
score over four equally weighted benchmark families (MMLU-Pro, ARC-C, MT-Bench-101
rescaled from its $1$--$10$ rating, and RULER over $4$/$8$/$16$k), with reasoning off
(open markers) and on (filled), so the segment joining a pair of markers is that
condition's reasoning boost. A translated cache is decoded by the small model in
$L \to S$ and by the large one in $S \to L$, and is therefore read against grey and
black respectively. Translated caches are the mean over three seeds. Error bars, in both
reasoning modes and on the native models too, are the standard error over evaluation items
(Appendix~\ref{app:seeds}).}
\label{app:fig:reasoningboost}
\end{figure}

Figure~\ref{app:fig:reasoningboost} gives the reasoning-off and reasoning-on scores, and
so the reasoning boost, for each model on its own cache and for the translated cache. On
the 4B/8B pair the translated cache tracks the target's own boost in both directions. The
0.6B/1.7B pair splits: translating into the larger model amplifies the boost ($+0.102$
against $+0.018$), translating into the smaller one dampens it ($+0.029$ against
$+0.045$).

\clearpage
\clearpage
\section{Multi-turn cache translation}
\label{app:multiturn}
\suppressfloats[t]

\begin{figure}[!htbp]
\centering
\includegraphics[width=0.7\linewidth]{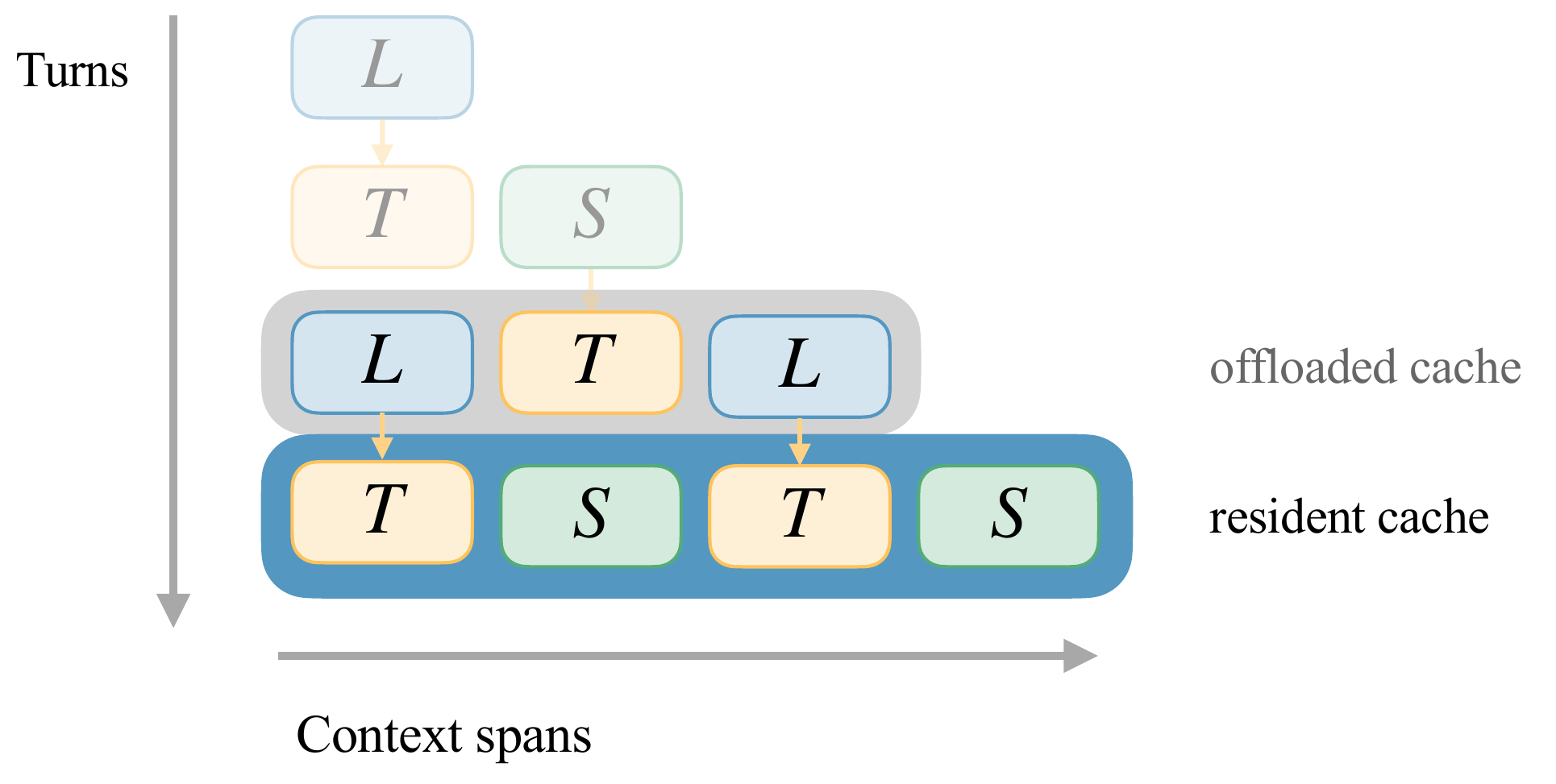}
\caption{\textbf{KV-Lingo multi-turn protocol.} 
$L$ and $S$ alternate over a shared context. Each row is the cache
line of the model that decodes at that turn, and each block one turn's span of it: a block
labelled $L$ or $S$ holds the cache that model wrote itself, a block labelled $T$ the
translation of a span the other model wrote. At each switch, the outgoing model's line is
offloaded, and the incoming model's line is extended by translating only the span written
since it last decoded; here, at turn~$4$, $S$ decodes from its resident line while $L$'s
line waits off the accelerator. No span passes through more than one translator.
}
\label{fig:multiturn}
\end{figure}

Figure~\ref{fig:multiturn} illustrates the two strategies compared in
\S\ref{sec:exp:multiturn}: the naive one, which translates the whole prefix again at every
switch, and KV-Lingo, where each model retains the spans it decoded and the translated spans of the other model. At a switch, the
incoming model rebuilds its cache by translating the other model's newly generated cache since the last switch, so each span is translated at most once.

Keeping one full cache line per model doubles the cache memory of a single model for an
equal-shape pair such as Qwen3-4B/8B, since each line covers the whole prefix. Only the
active line needs to sit on the accelerator, however. The idle line is not read at all until
the next switch, so it gains nothing from being resident, and it can be offloaded. It then
moves once per switch, an $O(n)$ copy of a cache that already
exists, whereas full re-prefill recomputes that cache with a forward pass whose attention costs
$O(n^2)$; the copy should stay far cheaper at the long contexts where memory matters. Both
directions of the copy can also be hidden: the outgoing line is offloaded while the incoming
model decodes, and the incoming line is reloaded while the new span is being translated. The
accelerator therefore holds the same cache as when serving a single model.

\begin{figure}[!htbp]
\centering
\includegraphics[width=\linewidth]{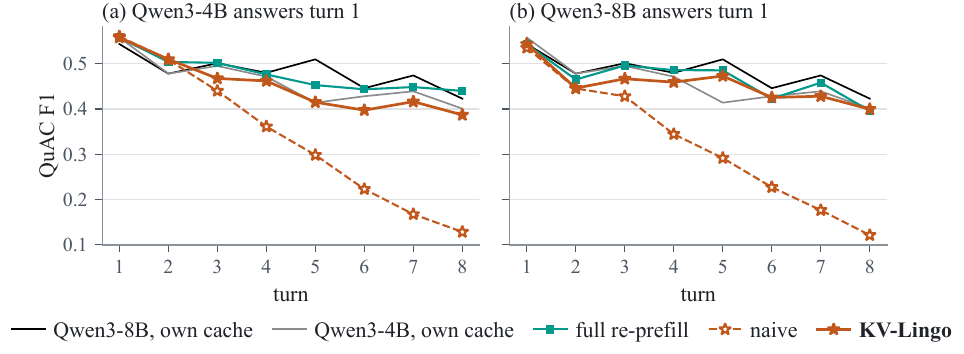}
\caption{\textbf{Multi-turn model switching on QuAC.} Span F1 against turn index, same pair
and protocol as Figure~\ref{fig:multiturn_coqa}, $256$ frozen conversations truncated to
depth $8$. Full re-prefill is exact and costs a native prefill over the whole history at
every switch; KV-Lingo translates only the span the target's cache lacks. Translated conditions are the mean over three seeds; error bars, on
every condition including re-prefill, are the standard error over evaluation items
(Appendix~\ref{app:seeds}).}
\label{app:fig:multiturn_quac}
\end{figure}

\begin{figure}[!htbp]
\centering
\includegraphics[width=\linewidth]{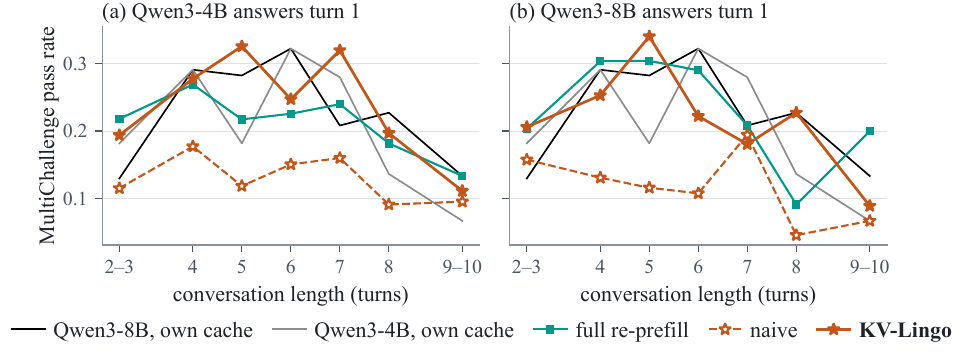}
\caption{\textbf{Multi-turn model switching on MultiChallenge.} Judged pass
rate against conversation length, same pair and protocol as
Figure~\ref{fig:multiturn_coqa}. The history is pre-written and fixed, so every
condition reaches the final prompt over a byte-identical sequence and the two models' caches
split the tokens roughly evenly, unlike CoQA and QuAC, where the passage sits in the turn-$1$
block. Bins are disjoint item sets of the size annotated at the foot of each panel, so the
depth axis varies across items rather than within one. Translated conditions are the mean
over three seeds; error bars, on every condition including re-prefill, are the standard
error over evaluation items (Appendix~\ref{app:seeds}).}
\label{app:fig:multiturn_multichallenge}
\end{figure}

Figures~\ref{app:fig:multiturn_quac} and~\ref{app:fig:multiturn_multichallenge} repeat the
CoQA comparison of Figure~\ref{fig:multiturn_coqa} on QuAC and MultiChallenge, with the
same pair, protocol and translators. Both agree with CoQA: the naive strategy falls below
full re-prefill within a few turns (from turn~$3$ on QuAC, and in most bins on
MultiChallenge), while KV-Lingo stays with full re-prefill to the end of each sweep:
turn~$8$ on QuAC, the $9$--$10$ bin on MultiChallenge.

\clearpage

\clearpage
\section{Pre-$k$-norm vs.\ post-$k$-norm translation}
\label{app:knorm}

Keys can be captured before key normalisation (pre-$k$-norm, pre-norm in
\S\ref{sec:method:family}) or after it and before RoPE (post-$k$-norm, post-norm in
\S\ref{sec:method:family}). We fit the same translator at both capture points, holding
everything else fixed (both Qwen3 pairs, both directions, one training recipe, the same
evaluation items), and compare them on the whole downstream suite. Only
keys are affected: values are captured at the same point either way.
Table~\ref{app:tab:knorm} summarises the difference per evaluation.

\begin{table}[htbp]
\centering
\footnotesize
\caption{\textbf{Post-$k$-norm minus pre-$k$-norm capture, per evaluation.} Self-distilled
translator, over the two Qwen3 pairs in both directions. A cell is one pair and direction
and, for long-context evaluations, one context length, scored on that evaluation's own
metric. $\pm$: standard
error of the mean difference over evaluation items, accounting for the items the cells of a
row share (Appendix~\ref{app:seeds}).}
\label{app:tab:knorm}
\begin{tabular}{@{}lccc@{}}
\toprule
evaluation & cells & post-$k$-norm worse in & mean $\Delta$ \\
\midrule
RULER, reasoning-off ($4$--$32$k)   & $16$ & $14$ & $-0.027_{\pm 0.002}$ \\
RULER, reasoning-on ($4$--$32$k)    & $16$ & $12$ & $-0.026_{\pm 0.005}$ \\
LongMemEval, reasoning-on, four rungs & $16$ & $12$ & $-0.010_{\pm 0.005}$ \\
LongMemEval, reasoning-on, multi-turn at $16$k & $4$  & $4$  & $-0.027_{\pm 0.010}$ \\
LongBench-v2, reasoning-off         & $4$  & $3$  & $-0.010_{\pm 0.007}$ \\
RepLiQA, reasoning-on, judged accuracy and token-F1 & $8$  & $4$  & $-0.004_{\pm 0.004}$ \\
MMLU-Pro and ARC-C                  & $8$  & $4$  & $+0.021_{\pm 0.003}$ \\
\bottomrule
\end{tabular}
\end{table}

\paragraph{Post-$k$-norm loses on long context and gains on multiple choice.} Pooling the
long-context evaluations (RULER and LongBench-v2 scored reasoning-off, LongMemEval and its
multi-turn variant reasoning-on), post-$k$-norm is the worse capture point in $33$ of $40$ cells,
mean $-0.019_{\pm 0.003}$ (sign test $p < 10^{-4}$), and no single cell is catastrophic: the worst is
$-0.063_{\pm 0.012}$, on RULER at $16$k. Taken one cell at a time, the difference exceeds two
standard errors in $12$ of the $40$ cells, $11$ worse and $1$ better. On MMLU-Pro and
ARC-C the sign flips: post-$k$-norm is better by $0.021_{\pm 0.003}$ on average, although it
wins in only four of the eight cells.

\clearpage
\section{Pre-$k$-norm versus post-RoPE translation}
\label{app:rope}

\begin{figure}[!htbp]
    \centering
    \includegraphics[width=\linewidth]{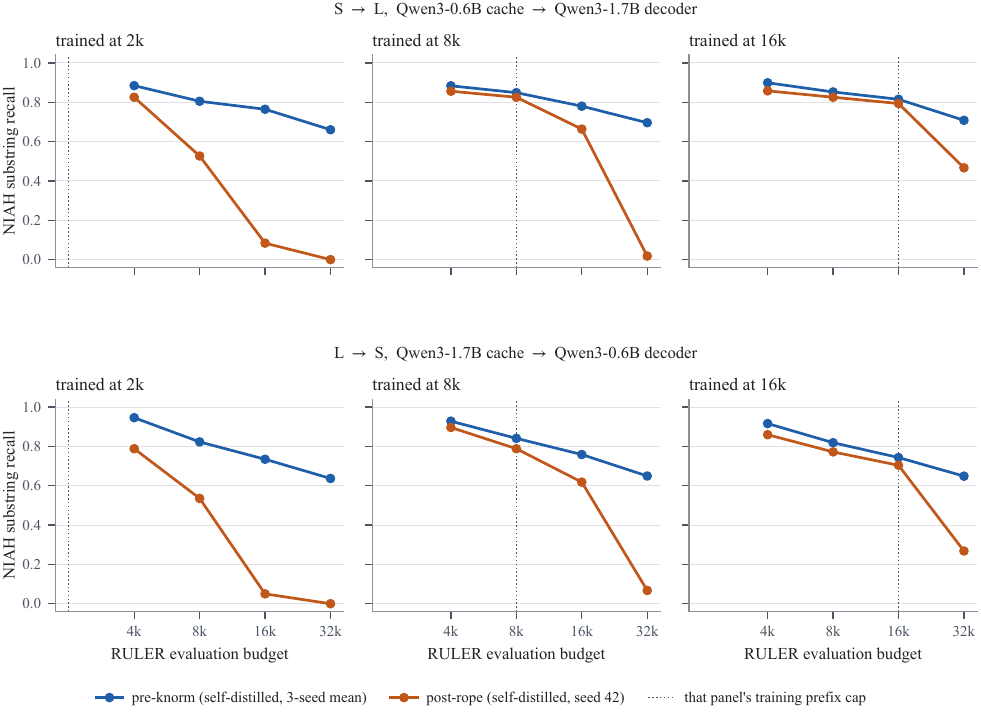}
    \caption{\textbf{Where the two capture points diverge: extrapolation past the training
    prefix cap.} RULER NIAH substring recall against evaluation context budget on the
    Qwen3-$0.6$B/$1.7$B pair, both directions. One panel per training prefix cap; the
    dotted line marks that cap, so everything to its right is extrapolation. Both curves are
    the same self-distilled translator fitted at the two capture points, pre-$k$-norm
    averaged over three seeds and post-RoPE at seed $42$, its only seed; error bars are
    the standard error over evaluation items (Appendix~\ref{app:seeds}). Post-RoPE matches pre-$k$-norm
    within the cap and falls away beyond it, to near-zero recall at $32$k for the $2$k and
    $8$k caps, while pre-$k$-norm degrades smoothly across the whole range.}
    \label{fig:pre_vs_post_rope_extrapolation}
\end{figure}

Keys can also be captured after the rotary embedding has been applied (post-RoPE), rather
than before key normalisation (pre-$k$-norm, called pre-norm in \S\ref{sec:method:family}). We fit the same translator at both capture
points, holding the model pairs, directions, objective and training recipe fixed, and
compare them on the downstream suite (Table~\ref{tab:pre_vs_post_rope}) and on RULER, at
four evaluation context budgets and for translators fitted with three different training
prefix caps (Figure~\ref{fig:pre_vs_post_rope_extrapolation}). Short-context scores are
mixed, but at long context pre-$k$-norm is clearly ahead: on RULER at $32$k it reaches
$0.65$--$0.98$ against $0.27$--$0.85$ for post-RoPE.
Figure~\ref{fig:pre_vs_post_rope_extrapolation} shows why: post-RoPE collapses beyond its
training prefix cap.

\begin{table}[htbp]
\centering
\caption{\textbf{Pre-$k$-norm against post-RoPE capture on the evaluation suite},
translators fitted at a $16$k training prefix cap, seed $42$ (post-RoPE's only seed),
reasoning-off except GSM8K. $\pm$: standard error over evaluation items
(Appendix~\ref{app:seeds}). \underline{Underline}: beats the small model, in $S \to L$ by
more than one standard error. \textbf{Bold}: the better capture point.}
\footnotesize
\setlength{\tabcolsep}{5pt}
\label{tab:pre_vs_post_rope}
\begin{tabular}{@{}l rr rr@{}}
\toprule
 & \multicolumn{2}{c}{native} & \multicolumn{2}{c}{translated} \\
\cmidrule(lr){2-3} \cmidrule(lr){4-5}
Benchmark & small & large & pre-$k$-norm & post-RoPE \\
\midrule
\multicolumn{5}{@{}l}{\emph{$L \to S$}, Qwen3-0.6B/1.7B} \\
MMLU-Pro & $0.234_{\pm 0.011}$ & $0.379_{\pm 0.013}$ & $0.219_{\pm 0.011}$ & $\underline{\mathbf{0.249}}_{\pm 0.011}$ \\
ARC-C & $0.568_{\pm 0.014}$ & $0.770_{\pm 0.012}$ & $\underline{0.616}_{\pm 0.014}$ & $\underline{\mathbf{0.660}}_{\pm 0.014}$ \\
GSM8K & $0.641_{\pm 0.030}$ & $0.746_{\pm 0.027}$ & $\underline{\mathbf{0.695}}_{\pm 0.029}$ & $0.641_{\pm 0.030}$ \\
CoQA & $0.475_{\pm 0.026}$ & $0.672_{\pm 0.026}$ & $\mathbf{0.469}_{\pm 0.026}$ & $0.386_{\pm 0.026}$ \\
IFEval & $0.579_{\pm 0.021}$ & $0.662_{\pm 0.020}$ & $\mathbf{0.529}_{\pm 0.021}$ & $0.497_{\pm 0.022}$ \\
RepLiQA & $0.766_{\pm 0.023}$ & $0.921_{\pm 0.014}$ & $\underline{\mathbf{0.811}}_{\pm 0.021}$ & $\underline{0.790}_{\pm 0.022}$ \\
MT-Bench-101 & $6.11_{\pm 0.26}$ & $7.95_{\pm 0.22}$ & $\mathbf{6.05}_{\pm 0.26}$ & $5.90_{\pm 0.28}$ \\
LongMemEval, oracle & $0.396_{\pm 0.025}$ & $0.558_{\pm 0.025}$ & $\underline{\mathbf{0.486}}_{\pm 0.025}$ & $\underline{0.468}_{\pm 0.025}$ \\
LongMemEval, $32$k & $0.247_{\pm 0.022}$ & $0.365_{\pm 0.024}$ & $\underline{\mathbf{0.342}}_{\pm 0.024}$ & $0.082_{\pm 0.014}$ \\
RULER, $4$k & $0.964_{\pm 0.008}$ & $0.994_{\pm 0.004}$ & $\mathbf{0.928}_{\pm 0.011}$ & $0.859_{\pm 0.012}$ \\
RULER, $32$k & $0.717_{\pm 0.013}$ & $0.891_{\pm 0.013}$ & $\mathbf{0.651}_{\pm 0.011}$ & $0.268_{\pm 0.015}$ \\
\midrule
\multicolumn{5}{@{}l}{\emph{$L \to S$}, Qwen3-4B/8B} \\
MMLU-Pro & $0.564_{\pm 0.013}$ & $0.615_{\pm 0.013}$ & $\underline{\mathbf{0.569}}_{\pm 0.013}$ & $0.561_{\pm 0.013}$ \\
ARC-C & $0.920_{\pm 0.008}$ & $0.937_{\pm 0.007}$ & $\underline{\mathbf{0.925}}_{\pm 0.008}$ & $0.918_{\pm 0.008}$ \\
GSM8K & $0.824_{\pm 0.024}$ & $0.824_{\pm 0.024}$ & $\underline{\mathbf{0.863}}_{\pm 0.022}$ & $0.824_{\pm 0.024}$ \\
CoQA & $0.721_{\pm 0.024}$ & $0.712_{\pm 0.023}$ & $0.721_{\pm 0.024}$ & $\underline{\mathbf{0.722}}_{\pm 0.023}$ \\
IFEval & $0.787_{\pm 0.018}$ & $0.802_{\pm 0.017}$ & $0.769_{\pm 0.018}$ & $\mathbf{0.771}_{\pm 0.018}$ \\
RepLiQA & $0.916_{\pm 0.015}$ & $0.935_{\pm 0.013}$ & $\underline{\mathbf{0.935}}_{\pm 0.013}$ & $\underline{0.929}_{\pm 0.014}$ \\
MT-Bench-101 & $9.00_{\pm 0.19}$ & $9.15_{\pm 0.16}$ & $\mathbf{8.97}_{\pm 0.19}$ & $8.96_{\pm 0.18}$ \\
LongBench-v2 & $0.388_{\pm 0.045}$ & $0.431_{\pm 0.046}$ & $\underline{\mathbf{0.448}}_{\pm 0.046}$ & $\underline{\mathbf{0.448}}_{\pm 0.046}$ \\
LongMemEval, oracle & $0.715_{\pm 0.023}$ & $0.725_{\pm 0.023}$ & $0.704_{\pm 0.023}$ & $\mathbf{0.712}_{\pm 0.023}$ \\
LongMemEval, $32$k & $0.542_{\pm 0.025}$ & $0.535_{\pm 0.025}$ & $\underline{\mathbf{0.548}}_{\pm 0.025}$ & $0.465_{\pm 0.025}$ \\
RULER, $4$k & $1.000_{\pm 0.000}$ & $0.998_{\pm 0.003}$ & $0.998_{\pm 0.003}$ & $\mathbf{1.000}_{\pm 0.000}$ \\
RULER, $32$k & $0.995_{\pm 0.003}$ & $0.996_{\pm 0.003}$ & $\mathbf{0.968}_{\pm 0.008}$ & $0.792_{\pm 0.014}$ \\
\midrule
\multicolumn{5}{@{}l}{\emph{$S \to L$}, Qwen3-0.6B/1.7B} \\
MMLU-Pro & $0.234_{\pm 0.011}$ & $0.379_{\pm 0.013}$ & $\mathbf{0.238}_{\pm 0.011}$ & $0.216_{\pm 0.011}$ \\
ARC-C & $0.568_{\pm 0.014}$ & $0.770_{\pm 0.012}$ & $0.467_{\pm 0.015}$ & $\mathbf{0.509}_{\pm 0.015}$ \\
GSM8K & $0.641_{\pm 0.030}$ & $0.746_{\pm 0.027}$ & $\underline{0.711}_{\pm 0.028}$ & $\underline{\mathbf{0.738}}_{\pm 0.028}$ \\
CoQA & $0.475_{\pm 0.026}$ & $0.672_{\pm 0.026}$ & $\underline{0.535}_{\pm 0.027}$ & $\underline{\mathbf{0.607}}_{\pm 0.027}$ \\
IFEval & $0.580_{\pm 0.021}$ & $0.662_{\pm 0.020}$ & $\mathbf{0.538}_{\pm 0.021}$ & $0.505_{\pm 0.022}$ \\
RepLiQA & $0.766_{\pm 0.023}$ & $0.921_{\pm 0.014}$ & $\underline{0.859}_{\pm 0.018}$ & $\underline{\mathbf{0.878}}_{\pm 0.017}$ \\
MT-Bench-101 & $6.11_{\pm 0.26}$ & $7.95_{\pm 0.22}$ & $\underline{7.19}_{\pm 0.23}$ & $\underline{\mathbf{7.48}}_{\pm 0.22}$ \\
LongMemEval, oracle & $0.396_{\pm 0.025}$ & $0.558_{\pm 0.025}$ & $0.378_{\pm 0.025}$ & $\underline{\mathbf{0.442}}_{\pm 0.025}$ \\
LongMemEval, $32$k & $0.247_{\pm 0.022}$ & $0.365_{\pm 0.024}$ & $\mathbf{0.219}_{\pm 0.021}$ & $0.170_{\pm 0.019}$ \\
RULER, $4$k & $0.964_{\pm 0.008}$ & $0.994_{\pm 0.004}$ & $\mathbf{0.901}_{\pm 0.009}$ & $0.858_{\pm 0.007}$ \\
RULER, $32$k & $0.717_{\pm 0.013}$ & $0.891_{\pm 0.013}$ & $\mathbf{0.707}_{\pm 0.011}$ & $0.467_{\pm 0.016}$ \\
\midrule
\multicolumn{5}{@{}l}{\emph{$S \to L$}, Qwen3-4B/8B} \\
MMLU-Pro & $0.564_{\pm 0.013}$ & $0.615_{\pm 0.013}$ & $\mathbf{0.571}_{\pm 0.013}$ & $0.542_{\pm 0.013}$ \\
ARC-C & $0.920_{\pm 0.008}$ & $0.937_{\pm 0.007}$ & $0.915_{\pm 0.008}$ & $\mathbf{0.917}_{\pm 0.008}$ \\
GSM8K & $0.824_{\pm 0.024}$ & $0.824_{\pm 0.024}$ & $\mathbf{0.840}_{\pm 0.023}$ & $0.777_{\pm 0.026}$ \\
CoQA & $0.721_{\pm 0.024}$ & $0.712_{\pm 0.023}$ & $\mathbf{0.738}_{\pm 0.024}$ & $0.733_{\pm 0.023}$ \\
IFEval & $0.787_{\pm 0.018}$ & $0.800_{\pm 0.017}$ & $0.784_{\pm 0.018}$ & $\mathbf{0.786}_{\pm 0.018}$ \\
RepLiQA & $0.916_{\pm 0.015}$ & $0.935_{\pm 0.013}$ & $\underline{\mathbf{0.949}}_{\pm 0.012}$ & $\underline{0.944}_{\pm 0.012}$ \\
MT-Bench-101 & $9.00_{\pm 0.19}$ & $9.15_{\pm 0.16}$ & $9.00_{\pm 0.18}$ & $\underline{\mathbf{9.19}}_{\pm 0.16}$ \\
LongBench-v2 & $0.388_{\pm 0.045}$ & $0.431_{\pm 0.046}$ & $0.405_{\pm 0.046}$ & $\mathbf{0.414}_{\pm 0.046}$ \\
LongMemEval, oracle & $0.715_{\pm 0.023}$ & $0.725_{\pm 0.023}$ & $\mathbf{0.699}_{\pm 0.023}$ & $0.694_{\pm 0.023}$ \\
LongMemEval, $32$k & $0.542_{\pm 0.025}$ & $0.535_{\pm 0.025}$ & $\mathbf{0.527}_{\pm 0.025}$ & $0.458_{\pm 0.025}$ \\
RULER, $4$k & $1.000_{\pm 0.000}$ & $0.998_{\pm 0.003}$ & $0.998_{\pm 0.003}$ & $\mathbf{1.000}_{\pm 0.000}$ \\
RULER, $32$k & $0.995_{\pm 0.003}$ & $0.996_{\pm 0.003}$ & $\mathbf{0.979}_{\pm 0.007}$ & $0.854_{\pm 0.011}$ \\
\bottomrule
\end{tabular}
\end{table}

\clearpage
\clearpage
\section{Moving the switch point}
\label{app:handoffpoint}
\suppressfloats[t]

We move the point at which the cache changes hands. Table~\ref{tab:ctxonly} translates only
the document of each long-context task and lets the target prefill the question itself:
self-distilled translators stay close to the whole-prompt protocol, the largest drops being
$0.07$ on LongBench-v2 in $L \to S$ and $0.05$ on RULER at $32$k in $S \to L$, while the
closed-form fit collapses in $L \to S$. Table~\ref{tab:cothandoff} instead switches after
the source's reasoning trace, so the target only writes the answer: the $0.6$B target then
matches the $1.7$B source on MMLU-Pro and ARC-C and exceeds it on GSM8K.

Two readings of Table~\ref{tab:ctxonly} are worth noting. LongMemEval is scored there on
$256$ questions per rung, so its absolute values are not comparable with those of
Table~\ref{tab:qwen3ls}; the difference between translating the whole prefix and the context only is, since it is paired per question.
And the large movements in the closed-form columns are empty-answer failures rather than
retrieval failures: in every cell that collapses, under either protocol, the closed-form
fit leaves a large fraction of the items (up to $0.96$) with no parseable answer at all,
whereas the trained maps answer on every item of every cell.

\begin{table}[htbp]
\centering
\footnotesize
\setlength{\tabcolsep}{2pt}
\caption{\textbf{Translating the context alone instead of the whole prompt.} The four
evaluation sets that carry a document followed by a question about it, on the
Qwen3-$4$B/$8$B pair, reasoning-off, both directions. \emph{whole} is the protocol used
everywhere else in the paper: the source prefills the entire rendered prompt and the map
translates all of it. \emph{context} translates the document alone, and the target
prefills the question, the assistant header and the empty reasoning block on top of the
translated cache before answering. The two protocols are scored on the same items, so
they pair per item. Translated cells are the mean over three seeds; $\pm$: standard
error over evaluation items (Appendix~\ref{app:seeds}). \emph{mixed}
adds a cross-entropy term to the self-distillation loss (Appendix~\ref{app:objective}).
\underline{Underline}: the translated cache beats the small model, and in $S \to L$ by
more than one standard error. \textbf{Bold}: the best translator on that row, across
both protocols. RULER is
substring recall, LongBench-v2 accuracy on the untruncated subset ($n = 116$),
LongMemEval and RepLiQA judged accuracy.}
\label{tab:ctxonly}
\resizebox{\ifdim\width>\linewidth\linewidth\else\width\fi}{!}{%
\begin{tabular}{@{}lrrrrrrrr@{}}
\toprule
 & \multicolumn{2}{c}{native} & \multicolumn{2}{c}{closed form} & \multicolumn{2}{c}{KV-Lingo} & \multicolumn{2}{c}{mixed} \\
\cmidrule(lr){2-3} \cmidrule(lr){4-5} \cmidrule(lr){6-7} \cmidrule(lr){8-9}
Benchmark & small & large & whole & context & whole & context & whole & context \\
\midrule
\multicolumn{9}{l}{\emph{$L \to S$}, Qwen3-$4$B/$8$B} \\
RULER, $4$k & $1.000_{\pm 0.000}$ & $0.998_{\pm 0.003}$ & $0.994_{\pm 0.004}$ & $0.823_{\pm 0.011}$ & $\mathbf{0.998}_{\pm 0.003}$ & $\mathbf{0.998}_{\pm 0.001}$ & $\mathbf{0.998}_{\pm 0.002}$ & $0.996_{\pm 0.002}$ \\
RULER, $8$k & $1.000_{\pm 0.000}$ & $0.998_{\pm 0.003}$ & $0.989_{\pm 0.005}$ & $0.706_{\pm 0.016}$ & $\mathbf{1.000}_{\pm 0.000}$ & $\mathbf{1.000}_{\pm 0.000}$ & $\mathbf{1.000}_{\pm 0.000}$ & $0.999_{\pm 0.001}$ \\
RULER, $16$k & $0.998_{\pm 0.003}$ & $0.999_{\pm 0.001}$ & $0.960_{\pm 0.009}$ & $0.350_{\pm 0.017}$ & $\mathbf{0.997}_{\pm 0.001}$ & $0.992_{\pm 0.004}$ & $0.995_{\pm 0.002}$ & $0.983_{\pm 0.004}$ \\
RULER, $32$k & $0.995_{\pm 0.003}$ & $0.996_{\pm 0.003}$ & $0.897_{\pm 0.010}$ & $0.168_{\pm 0.011}$ & $0.972_{\pm 0.007}$ & $0.949_{\pm 0.009}$ & $\mathbf{0.977}_{\pm 0.007}$ & $0.966_{\pm 0.007}$ \\
LongMemEval oracle & $0.582_{\pm 0.031}$ & $0.633_{\pm 0.030}$ & $0.504_{\pm 0.031}$ & $0.129_{\pm 0.021}$ & $0.559_{\pm 0.028}$ & $\mathbf{0.570}_{\pm 0.029}$ & $\mathbf{0.570}_{\pm 0.027}$ & $0.549_{\pm 0.029}$ \\
LongMemEval @8k & $0.594_{\pm 0.031}$ & $0.621_{\pm 0.030}$ & $0.543_{\pm 0.031}$ & $0.066_{\pm 0.016}$ & $\underline{\mathbf{0.599}}_{\pm 0.027}$ & $0.572_{\pm 0.028}$ & $0.587_{\pm 0.026}$ & $0.572_{\pm 0.028}$ \\
LongMemEval @16k & $0.465_{\pm 0.031}$ & $0.547_{\pm 0.031}$ & $0.434_{\pm 0.031}$ & $0.000_{\pm 0.000}$ & $\underline{\mathbf{0.466}}_{\pm 0.027}$ & $0.449_{\pm 0.028}$ & $0.464_{\pm 0.027}$ & $0.452_{\pm 0.028}$ \\
LongMemEval @32k & $0.410_{\pm 0.031}$ & $0.387_{\pm 0.030}$ & $0.324_{\pm 0.029}$ & $0.012_{\pm 0.007}$ & $0.379_{\pm 0.027}$ & $0.375_{\pm 0.028}$ & $0.372_{\pm 0.026}$ & $\mathbf{0.382}_{\pm 0.027}$ \\
LongBench-v2 & $0.388_{\pm 0.045}$ & $0.431_{\pm 0.046}$ & $0.388_{\pm 0.045}$ & $0.138_{\pm 0.032}$ & $\underline{0.445}_{\pm 0.044}$ & $0.376_{\pm 0.043}$ & $\underline{\mathbf{0.448}}_{\pm 0.043}$ & $0.362_{\pm 0.042}$ \\
RepLiQA & $0.916_{\pm 0.015}$ & $0.935_{\pm 0.013}$ & $\underline{\mathbf{0.944}}_{\pm 0.012}$ & $0.629_{\pm 0.026}$ & $\underline{0.934}_{\pm 0.011}$ & $\underline{0.924}_{\pm 0.012}$ & $\underline{0.942}_{\pm 0.011}$ & $\underline{0.936}_{\pm 0.011}$ \\
\midrule
\multicolumn{9}{l}{\emph{$S \to L$}, Qwen3-$4$B/$8$B} \\
RULER, $4$k & $1.000_{\pm 0.000}$ & $0.998_{\pm 0.003}$ & $0.999_{\pm 0.001}$ & $0.917_{\pm 0.012}$ & $0.999_{\pm 0.001}$ & $\mathbf{1.000}_{\pm 0.000}$ & $\mathbf{1.000}_{\pm 0.000}$ & $\mathbf{1.000}_{\pm 0.000}$ \\
RULER, $8$k & $1.000_{\pm 0.000}$ & $0.998_{\pm 0.003}$ & $0.998_{\pm 0.003}$ & $0.943_{\pm 0.010}$ & $\mathbf{0.999}_{\pm 0.001}$ & $\mathbf{0.999}_{\pm 0.001}$ & $\mathbf{0.999}_{\pm 0.001}$ & $\mathbf{0.999}_{\pm 0.001}$ \\
RULER, $16$k & $0.998_{\pm 0.003}$ & $0.999_{\pm 0.001}$ & $0.978_{\pm 0.007}$ & $0.875_{\pm 0.011}$ & $\mathbf{0.997}_{\pm 0.001}$ & $0.981_{\pm 0.006}$ & $\mathbf{0.997}_{\pm 0.002}$ & $0.987_{\pm 0.004}$ \\
RULER, $32$k & $0.995_{\pm 0.003}$ & $0.996_{\pm 0.003}$ & $0.864_{\pm 0.012}$ & $0.801_{\pm 0.012}$ & $\mathbf{0.980}_{\pm 0.006}$ & $0.936_{\pm 0.010}$ & $0.973_{\pm 0.007}$ & $0.920_{\pm 0.010}$ \\
LongMemEval oracle & $0.582_{\pm 0.031}$ & $0.633_{\pm 0.030}$ & $0.527_{\pm 0.031}$ & $0.484_{\pm 0.031}$ & $0.564_{\pm 0.028}$ & $\underline{\mathbf{0.611}}_{\pm 0.028}$ & $0.572_{\pm 0.028}$ & $0.579_{\pm 0.028}$ \\
LongMemEval @8k & $0.594_{\pm 0.031}$ & $0.621_{\pm 0.030}$ & $0.520_{\pm 0.031}$ & $0.500_{\pm 0.031}$ & $0.555_{\pm 0.027}$ & $0.561_{\pm 0.028}$ & $0.572_{\pm 0.027}$ & $\mathbf{0.599}_{\pm 0.028}$ \\
LongMemEval @16k & $0.465_{\pm 0.031}$ & $0.547_{\pm 0.031}$ & $0.395_{\pm 0.031}$ & $0.391_{\pm 0.031}$ & $0.409_{\pm 0.027}$ & $\mathbf{0.487}_{\pm 0.028}$ & $0.443_{\pm 0.027}$ & $0.461_{\pm 0.028}$ \\
LongMemEval @32k & $0.410_{\pm 0.031}$ & $0.387_{\pm 0.030}$ & $0.297_{\pm 0.029}$ & $0.266_{\pm 0.028}$ & $0.333_{\pm 0.026}$ & $0.365_{\pm 0.027}$ & $0.372_{\pm 0.027}$ & $\mathbf{0.410}_{\pm 0.027}$ \\
LongBench-v2 & $0.388_{\pm 0.045}$ & $0.431_{\pm 0.046}$ & $0.319_{\pm 0.043}$ & $0.371_{\pm 0.045}$ & $0.385_{\pm 0.041}$ & $\mathbf{0.399}_{\pm 0.043}$ & $\mathbf{0.399}_{\pm 0.043}$ & $0.376_{\pm 0.044}$ \\
RepLiQA & $0.916_{\pm 0.015}$ & $0.935_{\pm 0.013}$ & $0.274_{\pm 0.024}$ & $\underline{0.930}_{\pm 0.014}$ & $\underline{0.936}_{\pm 0.012}$ & $\underline{0.938}_{\pm 0.011}$ & $\underline{0.937}_{\pm 0.011}$ & $\underline{\mathbf{0.940}}_{\pm 0.011}$ \\
\bottomrule
\end{tabular}}
\end{table}

\begin{table}[htbp]
\centering
\footnotesize
\setlength{\tabcolsep}{2.8pt}
\caption{\textbf{Handing the cache over after the reasoning trace rather than at the
prompt.} Qwen3-$1.7$B $\to$ Qwen3-$0.6$B, reasoning-on. Under \emph{prompt} the source
prefills the prompt, the map translates it, and the target does its own thinking and
answering, the protocol used everywhere else in the paper. Under \emph{trace} the
source also generates its chain of thought on its own cache, the map translates the
prompt and the trace together, and the target only writes the answer that follows the
closing reasoning marker. \emph{text} is the reference for the second protocol: the same
target answering from its \emph{own} prefill of prompt $+$ trace, which is what an
errorless map would hand it. Translated cells are the mean over three seeds; $\pm$:
standard error over evaluation items (Appendix~\ref{app:seeds}).
\underline{Underline}: the translated cache beats the small model; this pair is read
$L \to S$ throughout, so no standard-error margin applies. \textbf{Bold}: the best translator
on that row, across both switch points. MMLU-Pro and ARC-C are
letter accuracy with a parse failure counted incorrect, GSM8K flexible exact match. On
MMLU-Pro, the \emph{trace} variants are helped by a format effect worth
about $6$ accuracy points (asked only to commit an answer, the target emits a parseable
one about twice as reliably as either model reasoning and concluding in one turn), so
matching the source there is not a sign of better reasoning.}
\label{tab:cothandoff}
\begin{tabular}{@{}lrrrrrrr@{}}
\toprule
 & \multicolumn{3}{c}{native} & \multicolumn{2}{c}{closed form} & \multicolumn{2}{c}{KV-Lingo} \\
\cmidrule(lr){2-4} \cmidrule(lr){5-6} \cmidrule(lr){7-8}
Benchmark & small & large & text & prompt & trace & prompt & trace \\
\midrule
MMLU-Pro & $0.319_{\pm 0.012}$ & $0.473_{\pm 0.013}$ & $0.508_{\pm 0.013}$ & $0.001_{\pm 0.001}$ & $\underline{0.335}_{\pm 0.012}$ & $0.260_{\pm 0.009}$ & $\underline{\mathbf{0.473}}_{\pm 0.013}$ \\
ARC-C & $0.690_{\pm 0.014}$ & $0.873_{\pm 0.010}$ & $0.875_{\pm 0.010}$ & $0.009_{\pm 0.003}$ & $0.666_{\pm 0.014}$ & $\underline{0.726}_{\pm 0.011}$ & $\underline{\mathbf{0.870}}_{\pm 0.010}$ \\
GSM8K & $0.641_{\pm 0.030}$ & $0.746_{\pm 0.027}$ & $0.848_{\pm 0.023}$ & $0.031_{\pm 0.011}$ & $0.297_{\pm 0.029}$ & $\underline{0.686}_{\pm 0.025}$ & $\underline{\mathbf{0.862}}_{\pm 0.020}$ \\
\bottomrule
\end{tabular}
\end{table}

\clearpage
\section{Objective ablation: cross-entropy and mixed objectives}
\label{app:objective}

We compare the self-distillation loss~\eqref{eq:kl} with two alternatives.

\paragraph{Cross-entropy.}
The cross-entropy (CE) objective fits the translator to the ground-truth continuation
rather than to the target's own predictions:
\begin{equation}
  \label{eq:ce}
  \mathcal{L}_{\mathrm{CE}}(T) \;=\; -\,\E_{(x_{1:n},\,y_{1:m})}\ \frac{1}{m}\sum_{j=1}^{m}
  \log p_\tgt\bigl(y_j \mid T(\mathcal{C}_\src(x_{1:n})),\, y_{<j}\bigr).
\end{equation}
CE improves strongly on the closed-form fit, but self-distillation beats it even on the
perplexity of the corpus both are trained on: on the from-scratch pair, CE leaves a gap of
$0.048$--$0.138$ nats to native decoding, against $0.031$--$0.101$ for self-distillation
(Table~\ref{tab:scratchppl}). CE is also fragile: at long context or reasoning-on, it
drives the decoder off its own manifold (unclosed reasoning blocks, repetition loops),
collapsing RULER recall and MT-Bench scores where self-distillation translators keep
working.

\begin{remark}
    \citet{chen2026see} train their translators with cross-entropy on continuations generated by the target model, while we use ground-truth continuations, typically generated by more capable language models. Training on target continuations might mitigate the brittleness we report.
\end{remark}

\paragraph{Mixed objective.}
The mixed objective adds the CE term to the self-distillation loss,
$\mathcal{L}_{\mathrm{mix}} = \mathcal{L} + \lambda\,\mathcal{L}_{\mathrm{CE}}$ with
$\lambda = 1$. Mixed translators are reported in the ``mixed'' columns of
Tables~\ref{tab:qwen3ls} and~\ref{tab:qwen3sl} (both Qwen3 pairs, both directions, the
whole evaluation suite) and Table~\ref{tab:ctxonly} (long-context benchmarks). They perform
about as well as self-distillation: neither dominates, and most gaps between the two are
within one standard error of their difference over evaluation items.

\clearpage
\section{Cache translation with Qwen3-like models pretrained from scratch}
\label{app:scratch}
\suppressfloats[t]

The Qwen3 checkpoints we translate between come from a heavily engineered pretraining
pipeline whose data and recipe are only partly disclosed~\citep{yang2025qwen3}, so the
success of a linear translator could stem from that pipeline rather than from the
architecture. To rule this out, we pretrain a $0.6$B/$1.7$B pair from scratch with the
Qwen3 architectures, under a recipe we fully control: $35$B tokens each, independent
initialisations and no distillation. We train the same translator with the same recipe as
for the released pair. The closed-form translator suffers on the from-scratch pair, leaving
$1.8$--$4.7\times$ the released pair's perplexity gap, whether because of the recipe or of
the much smaller training budget. Self-distillation closes most of that difference, so
translatability does not depend on the Qwen3 pretraining pipeline.

\paragraph{Setup.} The two models have exactly the Qwen3-$0.6$B and Qwen3-$1.7$B
architectures ($N=28$ layers, $H=8$ key-value heads of width $d_h=128$, so $d=1024$), and
therefore the same translator shape as the released pair. We pretrain each on $35$B tokens
of Nemotron-CC~\citep{su2025nemotron}, Chinchilla-optimal for the larger
model~\citep{hoffmann2022training}, with the cl100k\_base tokenizer of tiktoken rather than
Qwen3's; the two share this corpus and tokenizer but no checkpoint or initialisation. Both
follow one recipe: sequences of $4096$ tokens in batches of $256$ (${\approx}1.05$M tokens,
${\approx}33$k steps), AdamW with $(\beta_1, \beta_2) = (0.9, 0.95)$, $2000$ warmup steps, a
constant learning rate decayed to zero over the last $30\%$ of the steps, gradient clipping
at $1.0$, a z-loss of weight $10^{-4}$ and \texttt{bf16} mixed precision. The peak learning
rate, $3.79{\times}10^{-3}$, and the weight decay, $0.2$, are set for a base model of width
$512$ and depth $32$ and transferred to each model by CompleteP
scaling~\citep{dey2026dontlazycompletepenables}. We cannot replicate the Qwen3
post-training, so we compare against the released Qwen3-$0.6$B/$1.7$B-Base pair rather than
the Instruct checkpoints of our main results, which leaves pretraining as the only
difference between the two pairs. Base models have no chat template, so both pairs'
translators are trained on Nemotron-CC, with samples of a $256$-token prefix and a
$256$-token continuation: the split translator with head mixing, the identity layer
assignment, a closed-form stage 1 and $5000$ self-distillation steps. The released pair uses
the learning rate selected per direction, the from-scratch pair a single rate. We measure
continuation perplexity on a $64$-token forced continuation after a $256$-token prefix, over
$128$ documents of Nemotron-CC and $128$ of WikiText~\citep{merity2016pointer}, and zero-shot
accuracy on BoolQ~\citep{clark2019boolq}, HellaSwag~\citep{zellers2019hellaswag},
PIQA~\citep{bisk2020piqa}, the two ARC splits~\citep{clark2018think} and
WinoGrande~\citep{sakaguchi2021winogrande}, at most $2000$ examples each. The from-scratch
models are much weaker than the released ones, so each translator is compared with its own
pair's native decoding rather than on raw scores.

\begin{figure}[tbp]
\centering
\includegraphics[width=\linewidth]{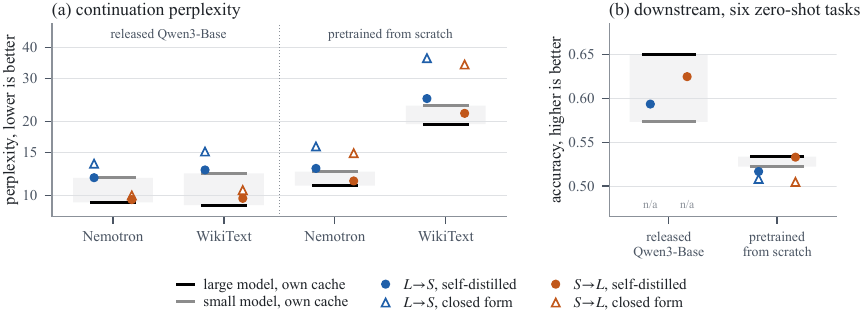}
\caption{\textbf{The same translator and recipe on the released Qwen3-$0.6$B/$1.7$B-Base
pair and on a $0.6$B/$1.7$B pair pretrained from scratch.} Both panels report absolute
scores. Each pair's two native models are short horizontal lines (black the larger, grey the
smaller) with the band between them shaded, and the cost of translation is the distance
from a marker to \emph{its own} target's line (grey for $L \to S$, black for $S \to L$).
Triangles are the closed-form translator (stage 1), circles the self-distilled one
(stage 2). \emph{(a)} Continuation perplexity on Nemotron-CC, which the translators are
fitted on, and on WikiText; the axis is logarithmic, so equal
vertical distances are equal log-perplexity gaps in both pairs. \emph{(b)} Accuracy
averaged over the six zero-shot tasks; the released pair's closed-form translator was not
run on this suite. Every self-distilled translator is a single run (seed $42$). Error bars in \emph{(b)}
are the standard error over evaluation items (Appendix~\ref{app:seeds}), binomial from the
aggregate accuracies since only aggregate scores were stored; \emph{(a)} has none, as
per-document perplexities were not stored.}
\label{app:fig:scratch}
\end{figure}

\paragraph{Results.} We compare the log-perplexity of the target's continuation under the
translated and the native cache. Over $L \to S$ and $S \to L$, each on Nemotron-CC and
WikiText (Figure~\ref{app:fig:scratch}a), the closed-form translator leaves a gap of
$0.24$, $0.44$, $0.30$ and $0.56$ nats on the from-scratch pair against $0.13$, $0.20$,
$0.06$ and $0.14$ on the released pair, $1.8$--$4.7\times$ larger. Self-distillation cuts the from-scratch gap by $5$--$9\times$, to $0.031$, $0.067$,
$0.043$ and $0.101$, which leaves $0.017$ to $0.037$ nats between the two pairs. Downstream
(Figure~\ref{app:fig:scratch}b), the from-scratch translator costs $0.014$ and $0.029$
average accuracy in the two directions in closed form, and $0.006$ and $0.001$ after
self-distillation, against $-0.020$ and $0.026$ for the released pair's self-distilled
translator (a negative value means the translated cache beats the target's own). After
self-distillation, neither pair is clearly easier to translate. The measurements of
Table~\ref{tab:linearity} are consistent with this: the from-scratch pair is barely harder
to fit linearly ($R^2_{\mathrm{lin}}$ of $0.752$ and $0.651$ on keys and values, against $0.766$ and $0.658$)
but has more nonlinear headroom ($\rho$ of $0.935$ and $0.903$ against $0.965$ and
$0.945$), which a least-squares fit pays for and an objective on the target's outputs can
work around.

\paragraph{Head mixing.} The two from-scratch models start from independent
initialisations, so their heads have no reason to correspond. Table~\ref{tab:scratch} uses
this pair to compare KV-Lingo, which mixes heads, with the head-wise MLP translator
of~\citet{chen2026see}, at matched parameters and FLOPs and at $4\times$ the parameters, all
trained with the same recipe. KV-Lingo is best in all twelve rows, and on perplexity
(Table~\ref{tab:scratchppl}) the head-wise MLP leaves $1.7$--$3.4\times$ our gap at matched
parameters and still $1.5$--$2.3\times$ at $4\times$. On released Qwen3 pairs, whose heads
are aligned, head-wise translators instead match head-mixing ones
(Appendix~\ref{app:head_alignment}).

\begin{table}[tbp]
\centering
\footnotesize
\setlength{\tabcolsep}{5pt}
\caption{\textbf{KV-Lingo against the head-wise MLP translator on a pair with unaligned
heads.} The from-scratch $0.6$B/$1.7$B pair of Appendix~\ref{app:scratch}, whose two models
start from independent initialisations. KV-Lingo against the head-wise
gated MLP of~\citet{chen2026see} at matched parameters and FLOPs ($59.0$M against our
$58.7$M) and at $4\times$ that budget ($235.9$M), all trained with the same two-stage recipe
and $5000$ self-distillation steps. \underline{Underline}: the translated cache beats the
small model, and in $S \to L$ by more than one standard error. \textbf{Bold}: the best
translator on that row. BoolQ and WinoGrande are
accuracy, HellaSwag, PIQA and the two ARC splits length-normalised accuracy; zero-shot,
$2000$ examples per task, one map per cell. $\pm$: standard error over evaluation items
(Appendix~\ref{app:seeds}), binomial from the aggregate accuracy, since per-item outcomes
were not stored for these runs. On BoolQ the large model scores below the small
one.}
\label{tab:scratch}
\begin{tabular}{@{}cl cc ccc@{}}
\toprule
 & & \multicolumn{2}{c}{native} & \multicolumn{3}{c}{translated} \\
\cmidrule(lr){3-4} \cmidrule(lr){5-7}
 & \multirow{2}{*}{Benchmark} & \multirow{2}{*}{small} & \multirow{2}{*}{large}
   & \multirow{2}{*}{KV-Lingo} & head-wise MLP & head-wise MLP \\
 & & & & & $1.0\times$ & $4.0\times$ \\
\midrule
\multirow{6}{*}{\rotatebox[origin=c]{90}{$L \to S$}} 
 & BoolQ & $0.523_{\pm 0.011}$ & $0.496_{\pm 0.011}$ & $\underline{\mathbf{0.547}}_{\pm 0.011}$ & $0.510_{\pm 0.011}$ & $0.519_{\pm 0.011}$ \\
 & HellaSwag & $0.457_{\pm 0.011}$ & $0.484_{\pm 0.011}$ & $\mathbf{0.448}_{\pm 0.011}$ & $0.439_{\pm 0.011}$ & $0.443_{\pm 0.011}$ \\
 & PIQA & $0.691_{\pm 0.011}$ & $0.708_{\pm 0.011}$ & $\mathbf{0.685}_{\pm 0.011}$ & $0.677_{\pm 0.011}$ & $0.683_{\pm 0.011}$ \\
 & ARC-E & $0.578_{\pm 0.011}$ & $0.636_{\pm 0.011}$ & $\mathbf{0.557}_{\pm 0.011}$ & $0.509_{\pm 0.011}$ & $0.529_{\pm 0.011}$ \\
 & ARC-C & $0.306_{\pm 0.013}$ & $0.338_{\pm 0.014}$ & $\mathbf{0.293}_{\pm 0.013}$ & $0.262_{\pm 0.013}$ & $0.276_{\pm 0.013}$ \\
 & WinoGrande & $0.529_{\pm 0.014}$ & $0.557_{\pm 0.014}$ & $\underline{\mathbf{0.534}}_{\pm 0.014}$ & $0.523_{\pm 0.014}$ & $0.525_{\pm 0.014}$ \\
\midrule
\multirow{6}{*}{\rotatebox[origin=c]{90}{$S \to L$}} 
 & BoolQ & $0.523_{\pm 0.011}$ & $0.496_{\pm 0.011}$ & $\mathbf{0.511}_{\pm 0.011}$ & $0.419_{\pm 0.011}$ & $0.458_{\pm 0.011}$ \\
 & HellaSwag & $0.457_{\pm 0.011}$ & $0.484_{\pm 0.011}$ & $\mathbf{0.465}_{\pm 0.011}$ & $0.462_{\pm 0.011}$ & $0.460_{\pm 0.011}$ \\
 & PIQA & $0.691_{\pm 0.011}$ & $0.708_{\pm 0.011}$ & $\underline{\mathbf{0.706}}_{\pm 0.011}$ & $0.698_{\pm 0.011}$ & $\underline{0.702}_{\pm 0.011}$ \\
 & ARC-E & $0.578_{\pm 0.011}$ & $0.636_{\pm 0.011}$ & $\underline{\mathbf{0.612}}_{\pm 0.011}$ & $0.561_{\pm 0.011}$ & $0.569_{\pm 0.011}$ \\
 & ARC-C & $0.306_{\pm 0.013}$ & $0.338_{\pm 0.014}$ & $\underline{\mathbf{0.328}}_{\pm 0.014}$ & $0.309_{\pm 0.013}$ & $0.306_{\pm 0.013}$ \\
 & WinoGrande & $0.529_{\pm 0.014}$ & $0.557_{\pm 0.014}$ & $\mathbf{0.541}_{\pm 0.014}$ & $0.538_{\pm 0.014}$ & $0.533_{\pm 0.014}$ \\
\bottomrule
\end{tabular}
\end{table}

\begin{table}[tbp]
\centering
\footnotesize
\setlength{\tabcolsep}{6pt}
\caption{\textbf{Continuation perplexity on the from-scratch pair}, for the translators of
Table~\ref{tab:scratch}, plus KV-Lingo's closed-form initialisation and KV-Lingo trained
with cross-entropy (CE) instead of self-distillation\iclronly{ (Appendix~\ref{app:objective})}. The last five columns are the gap
$\ln(\text{translated PPL}) - \ln(\text{native PPL})$ in nats, so zero is native and lower
is better; \emph{native PPL} is the target's own continuation perplexity on that corpus.
A $64$-token continuation after a $256$-token prefix, $128$ documents, one map per cell.
No standard errors are shown: per-document perplexities were not stored.
\textbf{Bold}: the lowest gap on that row.}
\label{tab:scratchppl}
\begin{tabular}{@{}ll c ccc cc@{}}
\toprule
 & & & \multicolumn{3}{c}{KV-Lingo (ours)} & \multicolumn{2}{c}{head-wise MLP} \\
\cmidrule(lr){4-6} \cmidrule(lr){7-8}
 & Corpus & native PPL & closed form & CE & self-distilled & $1.0\times$ & $4.0\times$ \\
\midrule
\multirow{2}{*}{$L \to S$} & Nemotron & $12.51$ & $0.236$ & $0.048$ & $\mathbf{0.031}$ & $0.085$ & $0.069$ \\
 & WikiText & $23.20$ & $0.444$ & $0.106$ & $\mathbf{0.067}$ & $0.229$ & $0.155$ \\
\midrule
\multirow{2}{*}{$S \to L$} & Nemotron & $10.99$ & $0.303$ & $0.060$ & $\mathbf{0.043}$ & $0.074$ & $0.065$ \\
 & WikiText & $19.51$ & $0.558$ & $0.138$ & $\mathbf{0.101}$ & $0.181$ & $0.170$ \\
\bottomrule
\end{tabular}
\end{table}

\paragraph{Released Base pairs.} Tables~\ref{tab:base} and~\ref{tab:baseppl} report the released Base pairs on their
own, at three seeds: downstream accuracy for both Qwen3-Base pairs in both directions, and
the cost of the same translators on corpora they were not fitted on.
Averaged over the four tasks, the $S \to L$ translated cache lands within $0.005$ of the
small model that prefilled it, not near the large target.

\begin{table}[tbp]
\centering
\footnotesize
\setlength{\tabcolsep}{4pt}
\caption{\textbf{Cache translation on two Qwen3-Base pairs.} The released
Qwen3-$0.6$B/$1.7$B-Base and Qwen3-$4$B/$8$B-Base pairs, both directions, with KV-Lingo, the
identity layer assignment and the recipe of \S\ref{sec:exp:setup}. KV-Lingo cells are
the mean over three seeds; the closed-form fit is a single solve. $\pm$: standard error
over evaluation items (Appendix~\ref{app:seeds}), binomial from the aggregate accuracy
since only aggregate scores were stored for these runs (for KV-Lingo, taken at the
three-seed mean); SQuAD-v1 token-F1 is not binary, so its cells carry none. Base models
have no chat template, so both pairs are fitted on plain text and every
task is run $5$-shot on at most $2000$ examples. \underline{Underline}: the translated
cache beats the small model, in $S \to L$ by more than one standard error.
\textbf{Bold}: the better translator on that row. ARC-C is length-normalised accuracy,
SQuAD-v1 token-F1, BoolQ and WinoGrande accuracy. The native columns repeat within a pair.
On the $4$B/$8$B $L \to S$ SQuAD-v1 row, the underlined $0.906$ is $0.9063$ against the small
model's $0.9057$.}
\label{tab:base}
\begin{tabular}{@{}ccl cccc@{}}
\toprule
 & & & \multicolumn{2}{c}{native} & \multicolumn{2}{c}{translated} \\
\cmidrule(lr){4-5} \cmidrule(lr){6-7}
 & & Benchmark & small & large & closed form & KV-Lingo \\
\midrule
\multirow{8}{*}{\rotatebox[origin=c]{90}{Qwen3 $0.6$B/$1.7$B}} & \multirow{4}{*}{\rotatebox[origin=c]{90}{$L \to S$}} & ARC-C & $0.441_{\pm 0.015}$ & $0.550_{\pm 0.015}$ & $\mathbf{0.429}_{\pm 0.014}$ & $0.423_{\pm 0.014}$ \\
 &  & SQuAD-v1 & $0.826$ & $0.881$ & $0.785$ & $\mathbf{0.821}$ \\
 &  & BoolQ & $0.720_{\pm 0.010}$ & $0.799_{\pm 0.009}$ & $\underline{\mathbf{0.783}}_{\pm 0.009}$ & $\underline{0.772}_{\pm 0.009}$ \\
 &  & WinoGrande & $0.597_{\pm 0.014}$ & $0.658_{\pm 0.013}$ & $0.597_{\pm 0.014}$ & $\underline{\mathbf{0.611}}_{\pm 0.014}$ \\
\cmidrule(l){2-7}
 & \multirow{4}{*}{\rotatebox[origin=c]{90}{$S \to L$}} & ARC-C & $0.441_{\pm 0.015}$ & $0.550_{\pm 0.015}$ & $\underline{0.472}_{\pm 0.015}$ & $\underline{\mathbf{0.493}}_{\pm 0.015}$ \\
 &  & SQuAD-v1 & $0.826$ & $0.881$ & $0.746$ & $\mathbf{0.794}$ \\
 &  & BoolQ & $0.720_{\pm 0.010}$ & $0.799_{\pm 0.009}$ & $0.696_{\pm 0.010}$ & $\mathbf{0.705}_{\pm 0.010}$ \\
 &  & WinoGrande & $0.597_{\pm 0.014}$ & $0.658_{\pm 0.013}$ & $0.597_{\pm 0.014}$ & $\mathbf{0.610}_{\pm 0.014}$ \\
\midrule
\multirow{8}{*}{\rotatebox[origin=c]{90}{Qwen3 $4$B/$8$B}} & \multirow{4}{*}{\rotatebox[origin=c]{90}{$L \to S$}} & ARC-C & $0.645_{\pm 0.014}$ & $0.683_{\pm 0.014}$ & $0.624_{\pm 0.014}$ & $\mathbf{0.636}_{\pm 0.014}$ \\
 &  & SQuAD-v1 & $0.906$ & $0.915$ & $0.884$ & $\underline{\mathbf{0.906}}$ \\
 &  & BoolQ & $0.860_{\pm 0.008}$ & $0.877_{\pm 0.007}$ & $\underline{0.870}_{\pm 0.008}$ & $\underline{\mathbf{0.871}}_{\pm 0.008}$ \\
 &  & WinoGrande & $0.713_{\pm 0.013}$ & $0.777_{\pm 0.012}$ & $\underline{0.735}_{\pm 0.012}$ & $\underline{\mathbf{0.737}}_{\pm 0.012}$ \\
\cmidrule(l){2-7}
 & \multirow{4}{*}{\rotatebox[origin=c]{90}{$S \to L$}} & ARC-C & $0.645_{\pm 0.014}$ & $0.683_{\pm 0.014}$ & $0.613_{\pm 0.014}$ & $\mathbf{0.646}_{\pm 0.014}$ \\
 &  & SQuAD-v1 & $0.906$ & $0.915$ & $0.861$ & $\mathbf{0.896}$ \\
 &  & BoolQ & $0.860_{\pm 0.008}$ & $0.877_{\pm 0.007}$ & $\mathbf{0.857}_{\pm 0.008}$ & $0.854_{\pm 0.008}$ \\
 &  & WinoGrande & $0.713_{\pm 0.013}$ & $0.777_{\pm 0.012}$ & $\underline{0.725}_{\pm 0.013}$ & $\underline{\mathbf{0.741}}_{\pm 0.012}$ \\
\bottomrule
\end{tabular}
\end{table}

\begin{table}[tbp]
\centering
\footnotesize
\setlength{\tabcolsep}{6pt}
\caption{\textbf{The Base-pair translators off their fitting distribution.} Relative
continuation-perplexity gap $(\text{translated} - \text{native})/\text{native}$, in
percent, for the self-distilled translators of Table~\ref{tab:base} on Nemotron-CC, which
they are fitted on, and four corpora they never see. Mean over three seeds; a $256$-token
continuation after a $256$-token prefix, $128$ documents per corpus; zero is native and
lower is better. No standard errors are shown: per-document perplexities were not stored. The gap is relative because absolute perplexity varies six-fold across
corpora ($2.4$ on code to $14.7$ on books). The worst of the twenty cells costs $6.5\%$ and
the median $2.1\%$, so the translators do not rely on their fitting corpus. The closed-form
fit was not run on this suite.}
\label{tab:baseppl}
\begin{tabular}{@{}ll ccccc@{}}
\toprule
 & & Nemotron & WikiText & Code & Books & arXiv \\
\midrule
\multirow{2}{*}{Qwen3 $0.6$B/$1.7$B} & $L \to S$ & $+0.03$ & $+3.23$ & $+4.51$ & $+2.18$ & $+2.74$ \\
 & $S \to L$ & $+2.58$ & $+6.50$ & $+5.92$ & $+5.50$ & $+5.07$ \\
\midrule
\multirow{2}{*}{Qwen3 $4$B/$8$B} & $L \to S$ & $+0.28$ & $+1.41$ & $+0.62$ & $+1.20$ & $+1.08$ \\
 & $S \to L$ & $+1.04$ & $+0.36$ & $+1.51$ & $+2.57$ & $+2.03$ \\
\bottomrule
\end{tabular}
\end{table}

\clearpage
\section{Split vs.\ joint maps}
\label{app:splitjoint}
\suppressfloats[t]
A joint map translates the stacked $[K;V]$ with a single
$2d_\tgt \times 2d_\src$ matrix. KV-Lingo's split map is the joint map with its
key-to-value and value-to-key blocks set to zero, leaving $T_K^\ell$ and $T_V^\ell$, which
halves the parameters. We compare the two on the Qwen3-0.6B/1.7B-Base and 4B/8B-Base pairs
in both directions, with the recipe of \S\ref{sec:method:training} on Nemotron text and one
seed per map (Table~\ref{tab:splitjoint}). The split map loses almost nothing: its
validation KL is higher by only $0.0008$--$0.0027$ nats per token, below two seed standard
deviations of the split map itself ($0.0010$--$0.0024$ over three seeds); its continuation
log-perplexity is higher by at most $0.008$ nats on Nemotron and WikiText; and its BoolQ
accuracy differs by at most $0.013$, within the sampling error of the $2000$-item
evaluation (two-proportion $z$-test, $p \geq 0.28$). KV-Lingo therefore translates keys and
values independently throughout.

\begin{table}[htbp]
\centering
\footnotesize
\setlength{\tabcolsep}{4pt}
\caption{\textbf{Split against joint maps on the Qwen3 Base pairs}, after
self-distillation, one seed per map. Validation KL is the forward KL from the target
decoding from its own cache to the target decoding from the translated cache, in nats per
token over $2048$ Nemotron continuation tokens. $\Delta\ln\mathrm{PPL}$ is the split map's
continuation log-perplexity minus the joint map's, in nats, over $128$ documents per corpus
with a $256$-token prefix and a $256$-token continuation. BoolQ is $0$-shot accuracy on
$2000$ items.}
\label{tab:splitjoint}
\begin{tabular}{@{}ll cc cc cc@{}}
\toprule
 & & \multicolumn{2}{c}{validation KL} & \multicolumn{2}{c}{$\Delta\ln\mathrm{PPL}$} & \multicolumn{2}{c}{BoolQ} \\
\cmidrule(lr){3-4} \cmidrule(lr){5-6} \cmidrule(lr){7-8}
Pair & Direction & split & joint & Nemotron & WikiText & split & joint \\
\midrule
$0.6$B/$1.7$B & $L \to S$ & $0.0202$ & $0.0175$ & $+0.0037$ & $+0.0076$ & $0.740$ & $0.729$ \\
              & $S \to L$ & $0.0265$ & $0.0257$ & $+0.0012$ & $+0.0039$ & $0.716$ & $0.709$ \\
$4$B/$8$B     & $L \to S$ & $0.0143$ & $0.0120$ & $+0.0013$ & $+0.0045$ & $0.818$ & $0.807$ \\
              & $S \to L$ & $0.0175$ & $0.0164$ & $+0.0018$ & $+0.0009$ & $0.831$ & $0.844$ \\
\bottomrule
\end{tabular}
\end{table}

\clearpage
\section{Cache translation across languages}
\label{app:xquad}

\begin{figure}[htbp]
\centering
\includegraphics[width=0.62\linewidth]{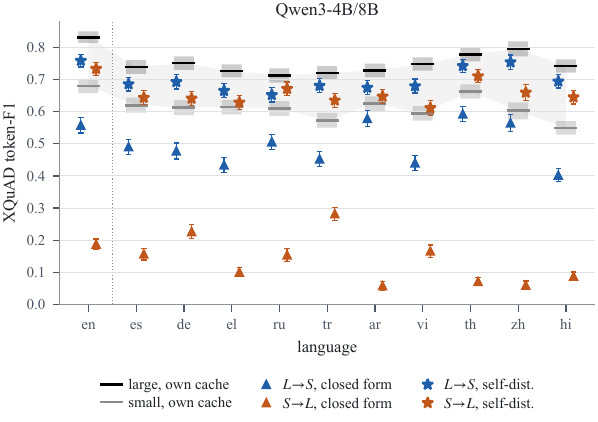}
\caption{\textbf{XQuAD token-F1 by language, both translation directions, Qwen3-4B/8B.}
Horizontal bars indicate the small model's and large model's scores, with the band between them shaded.
Both directions are then overlaid: $L \to S$ in blue, $S \to L$ in orange. Stars
are the score of KV-Lingo translators, averaged over three seeds;
triangles are the closed-form baseline. Error bars, on the native scores and the closed-form
fits too, are the standard error over evaluation items (Appendix~\ref{app:seeds}). These translators have been fitted on English-only data (from Nemotron-SFT-Instruction-Following-Chat-v2), so
English, left of the dotted line, is their training language and the other ten are
not. Reasoning-off; $256$ questions per language. The distilled translators outperform the small model in most of the languages, showing striking OOD generalisation.}
\label{fig:xquad}
\end{figure}

Figure~\ref{fig:xquad} reports the per-language XQuAD scores behind the multilingual
paragraph of \S\ref{sec:exp:shape}. The translators are fitted on an English-only stream
(the English subset of Nemotron-SFT-Instruction-Following-Chat-v2), which makes the ten
other languages unseen at training time.

\clearpage
\section{Computational gains of cache translation}
\label{app:cost}

We measure time to first token, decoding throughput and peak memory in each
scenario of \S\ref{sec:setting:metrics}.

\textbf{Time to first token.}
Re-prefilling runs the target over the $n$ tokens of the prefix, and its last position
predicts the first new token. Translation replaces that forward pass with two operations.
The translator applies, at each token and for each target layer, one matrix of size
$d_\tgt \times \nu d_\src$ to the keys and one to the values: $4\nu\, d_\src d_\tgt$ FLOPs
per token and target layer, and no attention. On the shape-preserving Qwen3 pairs
($d_\src = d_\tgt = 1024$, $\nu = 1$), that is $117$\,MFLOP per token into the $28$-layer
models and $151$\,MFLOP into the $36$-layer ones, against $0.88$, $2.8$, $7.3$ and
$13.9$\,GFLOP for the weight matrices of Qwen3-0.6B, 1.7B, 4B and 8B: $7.5$, $24$, $48$ and
$92$ times fewer, before counting the prefill's attention. A head-wise translator is
$H = 8$ times cheaper still. The translated cache holds keys and values but no prediction,
so the target then runs one forward pass over the current token, that is, one decoding
step, whose cost grows with $n$ only through reading the cache. The two paths reach the
first token after $t_{\mathrm{prefill}}(n)$ and after $t_T(n) + t_{\mathrm{step}}(n)$.

Table~\ref{app:tab:moespeed} times these terms on an Apple M3 Ultra (MLX, \texttt{bf16},
batch size $1$), for Qwen3-4B and Qwen3-30B-A3B (Figure~\ref{fig:cost}c). A decoding step
of the MoE, the inverse of its decoding speed, takes $14.0$\,ms at $64$ tokens and
$14.5$\,ms at $1024$, against $152$ and $421$\,ms for its prefill, and the translation takes
$1.8$ to $17.2$\,ms. A switch from the 4B, which already holds the cache, therefore reaches
the MoE's first token $9.6\times$ sooner than re-prefilling at $64$ tokens, $10.9$ to
$11.4\times$ sooner at $128$ to $512$, and $13.3\times$ sooner at $1024$. Used instead as a
prefill accelerator (\S\ref{sec:exp:moe}), the 4B first prefills the prompt itself, which
puts its own prefill on the path: the first token then comes $1.8\times$ sooner at $64$
tokens, and no sooner at $1024$, where the 4B prefills about as slowly as the MoE.

On an H100, we serve the targets with vLLM~\citep{kwon2023efficient} and divide the
target's time to first token by the translation time plus one vLLM decoding step, in the
four directions of the two shape-preserving Qwen3 pairs (Tables~\ref{app:tab:vllm}
and~\ref{tab:cost}). The translation is timed outside vLLM, as matrix products in PyTorch.
The switch then reaches the first token $4.8$ to $5.6\times$ sooner already at $2$k tokens,
and $12.0\times$, $14.8\times$, $23.2\times$ and $28.9\times$ sooner at $32$k into the $0.6$B,
$1.7$B, $4$B and $8$B (Figure~\ref{fig:cost}a,b). As a same-stack check, we also time
both paths end to end with Hugging Face Transformers (Table~\ref{tab:cost}). That harness
spends a fixed $29$ to $40$\,ms on every decoding step, against $1.8$ to $8.5$\,ms under
vLLM, an overhead that hides the gain at short prefixes; past a few thousand tokens the
prefill's arithmetic takes over, and the gap grows with $n$ as under vLLM.
Reading $\nu$
source layers per target layer multiplies the translation cost by $\nu$
(Figure~\ref{fig:nsrc_ttft}).

\textbf{Decoding throughput.}
When the cache is translated before starting to decode, after the switch the target decodes
at its native speed, as the translated cache has the target's own shape: on the harness of
Table~\ref{tab:cost}, the $0.6$B/$1.7$B pair decodes at the same $34$ tokens/s with either
cache, at every length.

\textbf{Peak memory.}
Translation also skips the prefill's activations: at $32$k tokens on the $0.6$B/$1.7$B
pair, the path with KV-Lingo peaks at $8.7$\,GB against $9.5$\,GB (into $0.6$B) and $10.3$\,GB (into
$1.7$B) for re-prefill.

\begin{table}[htbp]
\centering
\footnotesize
\setlength{\tabcolsep}{4pt}
\caption{\textbf{Components of the switch cost on an H100}, the two shape-preserving Qwen3
pairs, Hugging Face Transformers (\texttt{bf16}, batch $1$, both paths on the same stack):
translation, one decoding step of the target, and its re-prefill, in ms, every term measured
at every length. The translation times enter the speedups of Figure~\ref{fig:cost}a.
This harness spends a fixed $29$ to $40$\,ms on every decoding step, against $1.8$ to
$8.5$\,ms with vLLM (Table~\ref{app:tab:vllm}), so ratios taken within it understate the
gain.}
\label{tab:cost}
\begin{tabular}{@{}rrrrrrr@{}}
\toprule
& \multicolumn{3}{c}{into the smaller model} & \multicolumn{3}{c}{into the larger model} \\
\cmidrule(lr){2-4} \cmidrule(lr){5-7}
$n$ & translate & step & re-prefill & translate & step & re-prefill \\
\midrule
\multicolumn{7}{l}{\emph{Qwen3-0.6B/1.7B}} \\
$2048$ & $3.5$ & $29.2$ & $29.9$ & $3.5$ & $29.0$ & $30.0$ \\
$4096$ & $4.3$ & $29.4$ & $34.0$ & $4.3$ & $29.0$ & $49.6$ \\
$8192$ & $7.9$ & $29.4$ & $75.7$ & $7.9$ & $29.6$ & $107.0$ \\
$16384$ & $15.2$ & $29.4$ & $189.2$ & $15.5$ & $29.7$ & $250.9$ \\
$32768$ & $30.8$ & $29.6$ & $547.3$ & $30.7$ & $29.6$ & $671.6$ \\
\midrule
\multicolumn{7}{l}{\emph{Qwen3-4B/8B}} \\
$2048$ & $4.5$ & $38.6$ & $51.6$ & $4.5$ & $39.2$ & $74.1$ \\
$4096$ & $5.5$ & $38.8$ & $104.7$ & $5.5$ & $40.4$ & $149.9$ \\
$8192$ & $10.1$ & $38.5$ & $230.7$ & $10.1$ & $40.4$ & $325.3$ \\
$16384$ & $19.9$ & $38.9$ & $569.0$ & $19.9$ & $39.7$ & $759.0$ \\
$32768$ & $39.9$ & $39.3$ & $1577.6$ & $39.4$ & $39.7$ & $1981.8$ \\
\bottomrule
\end{tabular}
\end{table}

\begin{table}[htbp]
\centering
\footnotesize
\setlength{\tabcolsep}{5pt}
\caption{\textbf{Serving with vLLM on an H100}, \texttt{bf16}, one request at a time: median
time to first token / median time per output token, in ms, over $16$ requests per prompt
length, each generating $128$ tokens. Time to first token is measured at the client. The
last row is Gemma-3-12B given the $4$ KV heads of Gemma-3-4B, with random weights, which
times its decoding step against a cache of the 4B's width under a memory budget.}
\label{app:tab:vllm}
\begin{tabular}{lrrrrr}
\toprule
Prompt tokens & $2$k & $4$k & $8$k & $16$k & $32$k \\
\midrule
Qwen3-0.6B & $30$ / $1.8$ & $38$ / $1.9$ & $58$ / $2.1$ & $123$ / $2.4$ & $405$ / $2.9$ \\
Qwen3-1.7B & $31$ / $2.6$ & $37$ / $2.6$ & $67$ / $2.8$ & $145$ / $3.1$ & $506$ / $3.6$ \\
Qwen3-4B & $44$ / $4.6$ & $54$ / $4.7$ & $111$ / $4.9$ & $275$ / $5.3$ & $1065$ / $6.0$ \\
Qwen3-8B & $61$ / $7.0$ & $72$ / $7.1$ & $150$ / $7.3$ & $543$ / $7.7$ & $1384$ / $8.5$ \\
\midrule
Gemma-3-4B & $41$ / $4.7$ & $51$ / $4.7$ & $92$ / $4.7$ & $179$ / $4.7$ & $561$ / $4.7$ \\
Gemma-3-12B & $90$ / $11.1$ & $101$ / $11.1$ & $345$ / $10.9$ & $681$ / $11.1$ & $1451$ / $11.4$ \\
Gemma-3-12B, $4$ KV heads & $84$ / $10.4$ & $97$ / $10.4$ & $190$ / $10.4$ & $656$ / $10.4$ & $1361$ / $10.5$ \\
\bottomrule
\end{tabular}
\end{table}

\clearpage
\section{Decoding under a memory budget without materialising the target cache}
\label{app:memory}

In the memory-budget scenario of \S\ref{sec:setting:metrics}, the context is kept as the
cache of the small model $S$ and translators are merged to queries, so that $L$ can directly decode from the small cache.
This requires a linear and post-RoPE translation map. In that case,
linearity carries the map across the inner product:
$\bigl\langle q,\, T_K^{\ell} \bar K_{\src,i}^{\ell} \bigr\rangle =
\bigl\langle T_K^{\ell\top} q,\, \bar K_{\src,i}^{\ell} \bigr\rangle$, $i \in [n]$, where $q$
is a query of the target at the current step. The same holds for values: the attention
weights of a target head pool the values of $S$ first, and the map is applied to the pooled
vector, folding into the target's output projection. The map is then applied once per layer and decoding step rather than
once per cached token. 

To get throughput acceleration, however, the translators have to be learned head-wise.
A head-wise map pulls each target query back onto the source heads it reads from, so the
scores are computed against keys of the width of one head of $S$; a head-mixing map pulls it
back onto the whole cache width of $S$, and every target head then pays for all heads of $S$
at every cached token.

Our experiments show that head-wise translators land significantly below the score of the large model, when small and large have different cache shapes — which is required in this setting.
Hence, Table~\ref{app:tab:budget} times this decoding path as a proof-of-concept on Gemma-3-12B, against a cache narrowed
to the width of Gemma-3-4B, using head-wise post-RoPE translators with random weights.

\begin{table}[htbp]
\centering
\footnotesize
\setlength{\tabcolsep}{5pt}
\caption{\textbf{Decoding under a memory budget}, Gemma-3-12B in $4$-bit on Apple M3 Ultra
(MLX), random weights and cache. The full cache has $8$ KV heads, the narrowed one $4$, the
width of Gemma-3-4B, at unchanged depth; the translator is a random per-head map applied to
the query and to the attention output. Each cell generates $128$ tokens, best of $3$; ranges
span two machines. Throughput changes are relative to the full cache. Peak memory is that of
the whole process during decoding (weights, cache and temporaries).}
\label{app:tab:budget}
\begin{tabular}{rrrrrr}
\toprule
& & \multicolumn{3}{c}{decoding throughput} & peak memory (GB) \\
\cmidrule(lr){3-5} \cmidrule(lr){6-6}
context & batch & full (tokens/s) & narrowed & narrowed $+$ translator & full $\to$ narrowed $+$ translator \\
\midrule
$8$k  & $1$ & $62$--$65$   & $+9$--$10\%$  & $-5\%$         & $9.9 \to 9.1$ \\
$8$k  & $8$ & $160$--$171$ & $+33$--$37\%$ & $+22$--$25\%$  & $21.6 \to 14.9$ \\
$32$k & $1$ & $53$         & $+16$--$20\%$ & $+1$--$5\%$    & $13.9 \to 11.0$ \\
$32$k & $8$ & $110$--$115$ & $+44$--$49\%$ & $+36$--$40\%$  & $50.2 \to 29.4$ \\
\bottomrule
\end{tabular}
\end{table}

\applefootnote{ \textcolor{textgray}{\sffamily Apple and the Apple logo are trademarks of Apple Inc., registered in the U.S. and other countries and regions.}}

\end{document}